\documentclass{article}
\usepackage{authblk}
\usepackage{fullpage}
\usepackage{amsthm}
\usepackage{hyperref}
\usepackage{url}
\usepackage[noend]{algorithmic}
\usepackage[algoruled,nofillcomment,algo2e]{algorithm2e}
\newtheorem{definition}{Definition}
\usepackage{amsmath,bm}
\usepackage{graphicx}
\usepackage{subcaption}
\usepackage{todonotes}
\usepackage{xspace}
\usepackage{colortbl}
\usepackage{multirow}
\usepackage{booktabs}
\usepackage{cleveref}
\usepackage{enumitem}
\usepackage{natbib}

\def\sgsc{\textsc{SeqGenSeqClass}\xspace}
\def\sg{\textsc{SeqGen}\xspace}
\def\seqc{\textsc{SeqClass}\xspace}
\def\sgnsc{\textsc{SeqGenNonSeqClass}\xspace}
\def\algname{\textsc{STAR}\xspace}
\def\claalgname{\textsc{cla-STAR}\xspace}
\def\genalgname{\textsc{gen-STAR}\xspace}
\def\distrname{\textsc{strata}\xspace}

\usepackage{amsmath,amsfonts,bm}

\def\eqref#1{equation~\ref{#1}}

\def\1{\bm{1}}

\DeclareMathAlphabet{\mathsfit}{\encodingdefault}{\sfdefault}{m}{sl}
\SetMathAlphabet{\mathsfit}{bold}{\encodingdefault}{\sfdefault}{bx}{n}

\renewcommand{\paragraph}[1]{\noindent{\bf #1}}

\usepackage{xcolor}

\renewcommand{\thefootnote}{\fnsymbol{footnote}}

\begin{document}
\title{Fairness Feedback Loops:\\Training on Synthetic Data Amplifies Bias}

\author[1]{Sierra Wyllie \footnote{Corresponding author: sierra@wyllie.net}
}
\author[2]{Ilia Shumailov}
\author[1]{Nicolas Papernot}

\affil[1]{University of Toronto and Vector Institute}
\affil[2]{University of Oxford}

\date{February 5, 2024}
\maketitle
\renewcommand{\thefootnote}{\arabic{footnote}}

\begin{abstract}
Model-induced distribution shifts (MIDS) occur as previous model outputs pollute new model training sets over generations of models. 
This is known as \textit{model collapse} in the case of generative models, and \textit{performative prediction} or \textit{unfairness feedback loops} for supervised models. When a model induces a distribution shift, it also encodes its mistakes, biases, and unfairnesses into the ground truth of its data ecosystem. We introduce a framework that allows us to track multiple MIDS over many generations, finding that they can lead to loss in performance, fairness, and minoritized group representation, even in initially unbiased datasets. Despite these negative consequences, we identify how models might be used for positive, intentional, interventions in their data ecosystems, providing redress for historical discrimination through a framework called algorithmic reparation (AR). We simulate AR interventions by curating representative training batches for stochastic gradient descent to demonstrate how AR can improve upon the unfairnesses of models and data ecosystems subject to other MIDS. Our work takes an important step towards identifying, mitigating, and taking accountability for the unfair feedback loops enabled by the idea that ML systems are inherently neutral and objective. 
\end{abstract}

\section{Introduction}
Fairness feedback loops have posed problems for both machine learning practitioners' models and societies' policies for some time. 
One example are the 1930s Home Owner Loan Corporation Security Maps rediscovered by historian Kenneth T. Jackson in the 1980s. These depict `redlining,' where minoritized communities (especially Black and Jewish people) were discriminated against in housing in the United States~\citep{HOLC1938, Nelson2020Mapping, jackson1985crabgrass}.
These maps were likely used by government and banks to determine which neighborhoods should be provided programs and loans, feeding a feedback loop of segregation, limited Black home ownership, environmental racism, and increasing median household income gap between Black and white families~\citep{Shkembi2022environmental, Einhorn2021detroit}. 
More recently, automated systems used for policies such as loan eligibility and approval prediction~\citep{wu2022machine} risk further entrenchment of inequitable feedback loops~\citep{Davis2021reparation}. 

In the machine learning fairness community, this effect is known as performative prediction~\citep{perdomo2020performative} or fairness feedback loops~\citep{lum2016predict}, where the errors and behaviors of a model influence its future inputs, causing runaway unfairness~\citep{Ensign2017runaway}. 
In economics this is known as the performativity thesis, where economic theories attempting to describe markets instead shape them~\citep{callon1998economic}.
Increased attention to these effects and the proliferation of generated content on the internet, has created terminology for a dataset `ecosystem.' 
These ecosystems may suffer from `synthetic data spills,' such as unrealistic AI-generated images of baby peacocks polluting and dominating the image search results for real peachicks~\citep{bender2023IAS}. 
Despite the harms that models might cause their data ecosystems, practitioners lack understanding of the mechanics of these distribution shifts. This can lead to unawareness of the distribution shifts, especially when multiple models participate in the data ecosystem, and a lack of understanding of the fairness and equity harms that may arise.

In this work we introduce \textit{model-induced distribution shift} (MIDS) to describe model-induced changes to the data ecosystem. There are several phenomena in existing literature that we re-specify as MIDS; a subset of distribution shifts which are caused by past generations of model (mis)behaviours (in)advertently impacting successive generations. Each MIDS entails a model causing a gradual change in the data ecosystem; such as when synthetic data is published to the web and re-scraped to form new training sets. Once re-scraped, this polluted data becomes the ground truth for future generations of models, and the MIDS continues. We first unify several existing MIDS into a common framework, allowing a more nuanced understanding of their common causes and enabling analysis even where multiple MIDS occur in the same data ecosystem. We analyze the model behavior and fairness impacts of MIDS continuing over generations of models; including supervised classification models trained with labels sourced from their predecessors' predictions, and generative models trained from synthetic data created by their predecessors' outputs. 
We evaluate the performance and a variety of properties that may indicate the fairness of these models in each generation; finding disproportionately negative impacts on minoritized groups.
We find that chains of generative models eventually converge to the majority and amplify model mistakes that eventually come to dominate and degrade the data until little information from the original distribution remains, causing representational disparity between sensitive groups. 
We identify similar trends in chains of supervised classification models, showing that harms can arise even if distribution shift occurs through the labels alone. These harms are also present where MIDS co-occur; a chain of generative and supervised models can interact when the generators provide training data for the classifiers. %

In contrast to these harms, we study a conceptual framework introduced in~\citet{Davis2021reparation} called \textit{algorithmic reparation} (AR). 
While not strictly limited to algorithmic changes, AR uses ML models to provide redress for past harms to people with marginalized intersectional identities. 
For example, if attempting reparative predictive policing for Black communities, AR interventions in models could include re-weighting marginalized peoples' records in a dataset to compensate for over-representation in policing, increasing the model threshold for the detrimental prediction, or perhaps removing the predictive system entirely~\citep{Humerick2019reprogramming}. 
Because AR operates in settings where the `ground truth' is of questionable validity (due to current and historical discrimination), it provides a valuable avenue to provide reparation, and also to counter the injustices of MIDS. 
Of course, AR interventions also create model-induced distribution shifts; AR co-opts the mechanics of other MIDS to promote equity. 
To that effect, we simulate the effects of AR interventions through progressive intersectional categorical sampling, showing how prioritising representation lessens the unfair impacts of coexisting MIDS and their discriminative data ecosystems.

In summary, we make the following contributions:
\begin{itemize}
    \item We define a new term, model-induced distribution shift (MIDS), to unify several distribution shifts under one concept, and explore empirical settings to illustrate their impact. This unification draws attention to the common causes of MIDS and enables analysis even where MIDS co-occur.
    \item We use our settings to evaluate the impact of the fairness feedback loop and model collapse MIDS in several datasets, including face \texttt{CelebA} and \texttt{FairFace} datasets. We find that MIDS can lead to poor performance within a few generations of models, causing class imbalance, a lack of minoritized group representation, and unfairness. For example, our experiments on \texttt{CelebA} undergoing model collapse and performative prediction leads to a 15\% drop in accuracy and complete erasure of the minoritized group. %
    \item We position algorithmic reparation as an intentional MIDS with the goal of using model impact to promote equity and justice in the broader data setting. We create an algorithm, STratified AR (\algname) to simulate AR interventions by making training representative of intersectional identities. These simulations demonstrate how AR interventions can lessen disparate impact between sensitive groups and combat the unfair effects of other MIDS. 
\end{itemize}

\section{Background}\label{sec:related_work}
Several terms in existing literature describe distribution shifts perpetuated by models. 
We provide an overview of these MIDS, their enablers, and their relationships to Fairness in ML (FML, acronym from~\citet{Davis2021reparation}), then also connect algorithmic reparation to MIDS. 

\subsection{What are MIDS?}

\begin{table}[ht]
\centering
\begin{tabular}{lcr}
\toprule
    \textbf{MIDS}                    & \textbf{Model action/property}                         & $\Delta$ \textbf{Data Ecosystem}                    \\ \midrule
    Fairness Feedback Loops &  \multirow{2}{*}{Model predictions}           & Predictions become outcomes  \\
    Performative Prediction &                                               & and future labels            \\ \arrayrulecolor{black!30}\midrule
    \multirow{2}{*}{Model Collapse} & \multirow{2}{*}{Generated outputs}    & Synthetic data in ecosystem  \\ 
                                    &                                       & become new inputs            \\ \arrayrulecolor{black!30}\midrule
    Disparity Amplification & Poor utility for marginalized groups          & Marginalized groups leave                 \\  
        \arrayrulecolor{black}\bottomrule 
\end{tabular}
\caption{MIDS in the existing literature as organized by the model action that induces the MIDS and the effect on the data ecosystem. }
\label{tab:taxonomy}
\end{table}

In \Cref{tab:taxonomy}, we organize three phenomena from the literature into MIDS by determining how the model changes the data ecosystem: 
1) \textit{Performative prediction} occurs when a model's predictions influence outcomes, such as when recommender models influence and change a person's preferences~\citep{perdomo2020performative, Tong2023feedback}. This is also known as \textit{fairness feedback loops} when the outcomes of model predictions entrench bias or discrimination, as in redlining~\citep{green2020false}. 
2) \textit{Model collapse} may occur due to a similar phenomenon for generative models. If synthetic outputs are used to train a new generative model, over the course of several generations of models, the data distribution loses its tails and converges to a point estimate~\citep{shumailov2023curse, alemohammad2023selfconsuming}. 
3) \textit{Disparity amplification} occurs due to poor performance on a group of users. These negatively impacted users disengage from the data ecosystem, causing representational disparity. If trained upon, the altered data ecosystem could lead to even worse performance disparity~\citep{Hashimoto2018fairness}. 
While all of these effects cause distribution shift after deploying one model, the changes to the data ecosystem become entrenched as the ground truth if used to train the next generation of models. Throughout the remainder of the paper, we refer to generations, lineages, or sequences of generative and classifier models to indicate the teacher--student (similar to knowledge distillation, ~\citep{hinton2015distilling}) model chains underlying these MIDS.

There are other effects, which we refer to as enablers, that provide signal to data ecosystems undergoing MIDS. If the enabler misrepresents the training distribution to a model, this may bias its behavior and outputs. Enablers are \textit{not} innately MIDS, and can include sampling, data annotation, generative feedback, and pseudo-labelling methods (for background material on these concepts, see \Cref{ssec:mids_enablers}).
These enablers can also permit MIDS to co-occur: a pseudo-labeling model may annotate synthetic data created from generative models to use for supervised training, allowing model collapse and fairness feedback loops to co-occur. Furthermore, if the classifier resulting from the supervised training then impacts humans (or the non-synthetic portion of the data ecosystem), the next generations of any of these models may also be subject to disparity amplification. 
We model these MIDS and their interactions in \Cref{sec:methods}. For a review of MIDS and enablers with examples, see \Cref{app:mids_in_lit}.

\subsection{Algorithmic Reparation}
\textit{Algorithmic reparation} (AR), introduced in~\citet{Davis2021reparation}, proposes to substitute traditional frameworks of fairness in ML with a reparative approach to the design, development, and evaluation of machine learning systems for social interventions. AR is primarily inspired by Intersectionality theories, and seeks to promote justice in the broader data ecosystem through interventions from carefully-trained models. 
These actions are not restricted to algorithmic changes; a truly reparative approach requires transdisciplinary collaboration and a shift of economic, legal, and societal incentives. 
While AR specifically operates in machine learning, it encourages reflection on whether use of ML or computation in general may be inappropriate; and if so, advocates for eliminating these systems. 

AR is set as an alternative framework to FML, which generally attempts to equate model properties such as accuracy or positive prediction rate over sensitive groups. 
Instead, an AR approach focuses not on an equal distribution of resources and benefits, but on a highly task-dependent and potentially uneven allocation targeted to benefit marginalized intersectional identities in consideration of historical discrimination. 
This arises from AR's basis in Intersectionality theories, which acknowledges that harms compound at intersecting marginalized identities (see~\citep{romero2017introducing} for an overview and~\citep{combahee1977statement} for a prominent example). 
AR rejects the notion that equality necessarily begets equity and rejects that technology, including machine learning, can be neutral and objective (see ~\citet{representationalist_crit} for a detailed discussion of representational thinking, algorithmic idealism, and algorithmic objectivity).
Therefore, AR inherently questions the validity of the `ground truth' data used when training an ML system; this makes it a critical framework for addressing model-induced changes to the data ecosystem. Further background on the foundational assumptions of FML and their critiques which motivate AR may be found in~\Cref{sec:fairness_background}. 

In this paper, we empirically simulate how intersectional interventions at each model generation may constitute AR, harm reduction, and better representation. In this perspective, where AR functions as a MIDS, AR provides data ecosystem maintenance with a focus on reparative justice.

\subsection{Related Work}\label{ssec:related_work}
We overview several pertinent related works that study MIDS and how they connect to our work. 
Performative prediction, from~\citet{perdomo2020performative}, is detected by comparing the data generating distribution before and after a distribution shift caused by a function of the model's parameters. A performatively optimal model minimizes risk on the data distribution that manifests after its own deployment, and can be approached by methods such as repeated risk minimization and stochastic gradient updates~\citep{perdomo2020performative, hardt2023performative}. These discussions are usually constrained to the impacts of a model after one generation, which we extend over several generations and consider alongside other MIDS.

Another work, ~\citet{pmlr-v202-taori23a}, observe data feedback loops caused as model predictions contaminate datasets. They provide an upper bound for bias amplification depending on the amount of synthetic predictions and on whether the model has the same label bias as the original dataset. This second criteria is met in classifiers that have high uncertainty over the true labels, which follows from distributional generalization. We build on this work by considering bias amplification due to a changing data ecosystem subject to model collapse and performative prediction, as well as considering fairness impacts beyond remaining faithful to dataset label bias. While in our results some of our metrics converge and stabilize, we do not intentionally aim for distributional generalization.

Discussions of model collapse and the impact of generative models on future training sets are frequent in the natural language processing (NLP) literature, which concludes that removing this data ensures better future performance as NLP models improve~\citep{Rarrick2011MTDI}. More recent work answers questions on how synthetic data impacts downstream tasks. ~\citep{Hataya2022WillLG} finds worse downstream classifier performance when training from a synthetic dataset instead of the original. Evaluation settings with multiple, connected generative models have since been investigated in ~\citep{shumailov2023curse, alemohammad2023selfconsuming, martínez2023understanding}. Each of these works finds negative impact to utility if there is a sufficient lack of non-synthetic data. We depart from all of these works by considering the impacts of model collapse on fairness and equity; we combine the downstream performance task of ~\citep{Hataya2022WillLG} with the model collapse evaluation scheme of ~\citep{martínez2023understanding} and add fairness considerations, as well as co-occuring MIDS.

The work that introduces disparity amplification,~\citet{Hashimoto2018fairness}, considers fairness cases where sensitive information is unavailable. To minimize the risk that the minoritized group incurs high loss and disengages from the dataset ecosystem, they use distributionally robust optimization (DRO). As mentioned in their discussions, DRO might not protect minoritized groups so much as some worse-off group (as in Rawlsian justice), which for our focus on algorithmic reparation and intersectionality is inappropriate.  

\section{Methodology}\label{sec:methods}
In this section we introduce two settings, sequences of classifiers and sequences of generators, to allow for the observation and evaluation of MIDS. In these settings, each new model in the sequence is (at least partially) trained using the outputs of its predecessor(s) as inputs and/or labels. As that model is deployed and used, it propagates MIDS through its own properties and outputs, potentially affecting both the synthetic and non-synthetic portions of the data ecosystem. We also position AR as an intentional MIDS aiming for justice for historical discrimination and oppression. 
These settings provide an understanding of model impact over many generations, enabling informed maintenance of model and data ecosystem `health,' and reinforcing accountability for model impacts. 

\subsection{Modeling Assumptions}
Measuring distribution shift requires comparison between the current and the reference distribution, which represents the data ecosystem before the presence of any MIDS. The original reference distribution is given by $\mathcal{H}=\mathcal{X}\times\mathcal{L}\times\mathcal{S}$, where $\mathcal{X}$ represents the inputs, and $\mathcal{L}$ and $\mathcal{S}$ are annotations for the labels and sensitive attribute(s). 
Sampling from $\mathcal{H}$ gives dataset $D=X\times L\times S$.
(\textit{i.e.,} via disparity amplification). To compare the current data ecosystem against the original, we would need access to the original input distribution $\mathcal{X}$ and oracles for $\mathcal{X}\rightarrow\mathcal{L}$ and $\mathcal{X}\rightarrow\mathcal{S}$. Instead, we approximate these with a generative model $G_0$ and classifiers $A_L$ and $A_S$, all trained from $D$. We use these to approximate data from the original distribution and to annotate the class and group of generated data. 

These oracles provide an infinite data source that may be used to train all of the downstream models in our settings. Therefore, if the oracles misrepresent the distribution, the models trained from their outputs will experience MIDS relative to the original training distribution. These oracle approximations are not strongly limiting as we are primarily interested in the effects of MIDS relative to some distribution, be it the original or its approximation. 
Using classifiers to annotate generated samples has been used for the fair training of generative models, though labeling oracle $A_L$ and sensitive attribute oracle $A_S$ could also represent human annotators conducting manual data annotation~\citep{fairGAN, grover2019bias, grover2019fair, Hataya2022WillLG}. The initial generator, $G_0$, is also relevant for scenarios where synthetic data is preferred over human-generated data for a downstream task, which may sometimes arise in FML and privacy~\citep{zemelFair2013, ganev2022robin, NIST2018privacy, stadler2022synthetic}. Sampling from $G_0$, as opposed to the training set, also allows a chance at sampling from groups that might not otherwise be well-represented in the dataset, as in~\citep{zemelFair2013}.

\subsection{Sequential classifiers}\label{ssec:seq_cla}
\begin{figure}[ht]
    \centering
    \includegraphics[width=.6\textwidth]{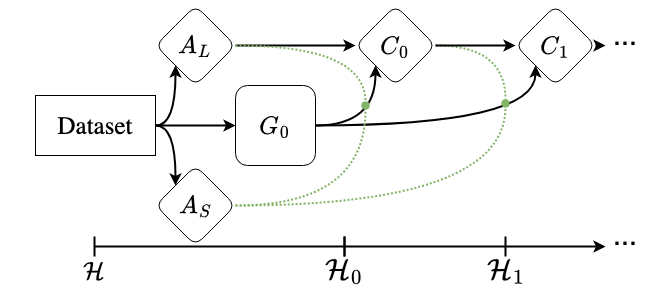}
    \caption{A high-level depiction of sequentially training classifiers (\seqc setting) for MIDS such as performative prediction and runaway feedback loops. The oracle models, $A_L$, $A_S$, and $G_0$ provide an infinite source of labels, sensitive group annotations, and inputs. We use these to train classifiers $C_i$, where $C_{i}$ is trained using labels from $C_{i-1}$. To alleviate the harms caused from sequentially training, \claalgname may be used to incorporate sensitive attribute data from $A_S$, as shown by the narrowly-dashed green lines.}
    \label{fig:flow_nomc}
\end{figure}

The sequential classifier (\seqc) setting permits us to pursue the study of MIDS such as fairness feedback loops and performative prediction, where distribution shift is mediated by classifier predictions becoming the ground truth of the next generation, as shown in~\Cref{fig:flow_nomc}. In the first generation, we train a classifier $C_0$ by sampling inputs from $G_0$ and labeling these with oracle $A_L$. In subsequent generations $i=1\dots n$, the classifier $C_i$ is trained on data sampled from $G_0$ but labeled by the preceding classifier $C_{i-1}$. 

\looseness=-1
Disparity amplification can be modeled by taking non-synthetic samples $h_{i}\sim\mathcal{H}_{i}$ to train $C_i$, where $\mathcal{H}_{i}$ is the non-synthetic data distribution after models from generation $i$ were deployed. We assume that disparity amplification has already influenced the label and sensitive group balances of $\mathcal{H}_{i}$ via $C_{i-1}$. Therefore, to get $h_{i}$ in practice, we inference $C_{i-1}$ on a held-out subset of $D$, recording label prediction frequencies over the categories formed from the Cartesian product of the sensitive attribute values and the possible labels. We use this to define a categorical distribution which we use to perform quota sampling on $D$. Quota sampling refers to partitioning a population into strata (in our case defined by label and group intersections) and selecting from each partition until we reach its quota, which is given by the categorical distribution multiplied by the total number of samples we wish to select. Henceforth we refer to this categorical distribution as a \distrname.   
In a nutshell, if $C_{i-1}$ often assigns negative predictions to a minoritized group, then $h_{i}$ will contain a proportional number of minoritized group samples with the negative label. Therefore, $C_i$ may be influenced by $C_{i-1}$ in two ways: 1) through data labeled by $C_{i-1}$ and 2) through non-synthetic data undergoing disparity amplification due to prediction disparity in $C_{i-1}$. When training $C_0$, we sample $h_{0}\sim D$ as disparity amplification has not yet occurred.

The formulations are shown below, where $\mathcal{T_C}(\cdot; \cdot)$ is the classifier training algorithm trained from the data in its first argument(s) as labeled by its second argument(s), $\mathcal{T_G}$ is the generator training algorithm, and $\texttt{Sample}(\cdot; \cdot)$ samples from its first argument according to a property (such as group representation) of its second argument(s). The terms causing performative prediction and disparity amplification are in \textcolor{red}{\textbf{red bold}} and \textcolor{teal}{\textbf{teal bold}} face: 
\begin{align}\label{eqn:nomc}
    C_i=\mathcal{T_C}(g_i, \textcolor{teal}{\bm{h_{i}}}; \textcolor{red}{\bm{C_{i-1}}}) \text{, where } C_0=\mathcal{T_C}(g_0, h_0; A_L) \text{, }  G_0=\mathcal{T_G}(X), \text{ and }  g_i\sim G_0 \\
    \text{ and } h_{i}=\texttt{Sample}(D; \textcolor{teal}{\bm{C_{i-1}}}) \text{, where } h_{0}=\texttt{Sample}(D).
\end{align} 

\subsection{Sequential generators and classifiers}\label{ssec:seq_gen}
\begin{figure}[ht]
    \centering
    \begin{minipage}{0.5\textwidth}
        \centering
        \includegraphics[width=\textwidth]{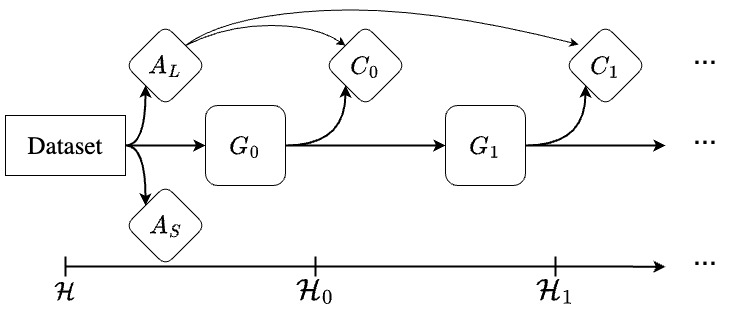}
    \end{minipage}\hfill
    \begin{minipage}{0.5\textwidth}
        \centering
        \includegraphics[width=\textwidth]{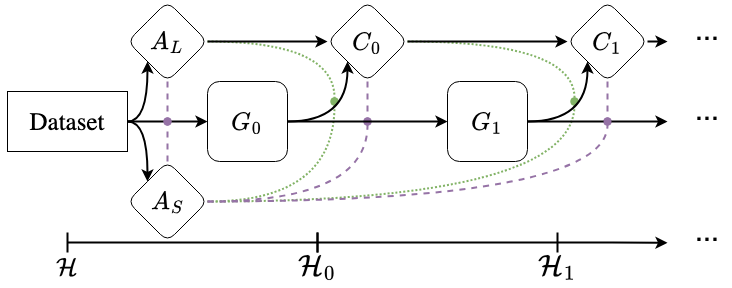} 
    \end{minipage}
    \caption{A high-level depiction of sequentially training generators with and without sequential classifiers (\textit{left:} \sgnsc, \textit{right:} \sgsc). The oracle models, $A_L$, $A_S$, and $G_0$ provide an infinite source of labels, sensitive group annotations, and inputs. We train a lineage of generative models ($G_i$) and train classifiers $C_i$ from these, where $C_{i}$ is trained using labels from $A_L$ (\textit{left}) or $C_{i-1}$ (\textit{right}). To alleviate the harms caused from sequentially training, \claalgname and \genalgname may be used to incorporate sensitive attribute data from $A_S$, as shown by the narrowly-dashed green lines and the broad-dashed purple lines.}
    \label{fig:flow_mc}
\end{figure}

The sequential generator (\sg) setting primarily investigates the model collapse MIDS, where distribution shift occurs as synthetic data is used for training new models, as shown in~\Cref{fig:flow_mc}. For this setting, we train generators $G_i$ sequentially from the samples of the preceding generator $G_{i-1}$, where the first generator $G_0$ is trained from the original dataset. 
This chain of generators is the same setting as used by~\citet{shumailov2023curse}. 
Departing from them, we also train a downstream classifier $C_i$ by sampling inputs from $G_i$ and labels from either the labeling oracle $A_L$ or from the preceding classifier $C_{i-1}$. The former case is the sequential generator and non-sequential classifier setting (henceforth \sgnsc), and the latter the sequential generator sequential classifier setting (\sgsc). In \sgsc, in addition to the generators being chained together, the classifiers are chained together and suffer the MIDS described in the \seqc setting. 
These downstream classifiers allow us to initiate the study of downstream classifier performance and FML fairness metrics while also tracking the devolution of minoritized group representation and model collapse. 

Additionally, disparity amplification due to $C_{i-1}$ and $G_{i-1}$ may impact the non-synthetic data distribution, $\mathcal{H}_i$, that may be used when training $G_i$ and/or $C_i$. For example, if $C_{i-1}$ has poor performance on a minoritized group of users, they may choose to disengage from the data ecosystem, meaning $\mathcal{H}_i$ will be less representative, likely harming the next generation of models. Similarly to the \seqc setting, we calculate the categorical distribution of $C_{i-1}$ over the sensitive attributes and labels and then quota sample proportionally from $D$ to form $h_i$. While this directly impacts $G_i$, it also impacts $C_i$ since it trains from $G_i$. 

The formulation for \sg with classifiers is shown below, with a substitute model $C$ that may stand for oracle $A_L$ or $C_{i-1}$ depending on whether the classifiers are sequential. The terms for performative prediction, model collapse, and disparity amplification are bolded in \textcolor{red}{\textbf{red}}, \textcolor{blue}{\textbf{blue}}, and \textcolor{teal}{\textbf{teal}} face.
\begin{align}\label{eqn:mc}
    &G_i=\mathcal{T_G}(\textcolor{blue}{\bm{g_{i-1}}}, \textcolor{teal}{\bm{h_i}}) \text{, where } G_0=\mathcal{T_G}(X) \text{ and } g_i\sim G_i \\
    &C_i=\mathcal{T_C}(g_i, \textcolor{teal}{\bm{h_i}}; \textcolor{red}{\bm{C}}) \text{, where } C_0=\mathcal{T_C}(g_0, h_0; A_L) \\
    & h_i=\texttt{Sample}(D; \textcolor{teal}{\bm{C_{i-1}}}, \textcolor{teal}{\bm{g_{i-1}}}) \text{, where } h_0=\texttt{Sample}(D).
\end{align}
\subsection{Simulating Algorithmic Reparation}\label{ssec:methods_ar}
In our experiments, we measure and simulate equity-oriented interventions as a change in the discrete distribution (\distrname) formed from the Cartesian product of the label and sensitive attribute values. For example, the \distrname of the current data ecosystem may be formed from $L$ and $S$, and the \distrname of a classifier $C_i$ may be formed from its predictions (on data sampled from $G_i$ or taken from $X$) and sensitive group annotations (from oracle $A_S$ or data $S$). 
We use \distrname to characterize the intersectional and label distributions of training sets, where these \distrname can change over the generations due to MIDS. 

To simulate the effects of AR interventions, we introduce an algorithm called STratified sampling AR (\algname; see \Cref{alg:AR_batches}). \algname creates model training batches by taking a biased sample according to these \distrname (within a resampling budget). For our simulations, we use a uniform distribution over these categories to give a quota for the number of samples from each category that should be present in the batch. 
As we are not experts in AR and do not provide a case study, we use a uniform distribution as the target for the biased sampling. Using an `ideal' (finite) distribution may be inappropriate as a distribution cannot neutrally or objectively determine the `best' mixture of demographics for a task. 
Due to this concession, we will refer to these measurements as fairness/unfairness (in the FML sense), as we cannot make claims of equity or justice without considering the myriad sources of bias in the ML life cycle~\citep{sources_ml_lifecycle} and the specific data ecosystem. 

In the \seqc setting, we use the name Classifier-\algname, or \claalgname, to refer to creating more intersectionally  representative batches for the classifiers in the lineage. This is not the only avenue for AR, but is inspired by work done by the FML community for performative prediction~\citep{Ensign2017runaway}. 
To train $C_i$, \claalgname labels generated outputs from $G_0$ using $C_{i-1}$ and sensitive attribute oracle $A_S$, then selects a subset of these samples for each training batch such that each label and group category in the batch meets the quota set by the fairness ideal. We update $\mathcal{T_{C}}$ to $\mathcal{T_{A,C}}$ and show the additional labeling and sensitive group annotations from $C_{i-1}$ and $A_S$ in \textcolor{green}{\textbf{green}}:
\begin{equation}\label{eqn:arnomc}
    C_i=\mathcal{T_{A,C}}(g_i, h_i; \textcolor{green}{\bm{C_{i-1}}}, \textcolor{green}{\bm{A_S}}) \text{, where } C_0=\mathcal{T_{A,C}}(g_0, h_0; A_L, \textcolor{green}{\bm{A_S}}).
\end{equation}

\algname in \sgsc may occur at all the same points described above, with the addition of interventions taken while training the generators. We examine both classifier-side \algname (\claalgname, taken while training classifiers, shown in \textcolor{green}{\textbf{green}}), and generator-side \algname (\genalgname, while training generators, shown in \textcolor{purple}{\textbf{purple}}). \genalgname also uses annotations from $C_{i-1}$ and sensitive attribute oracle $A_S$ to fill the label and group category quotas set by the fairness ideal. Both \claalgname and \genalgname are described in~\Cref{eqn:arcla} and~\Cref{eqn:argen}, respectively:
\begin{equation}\label{eqn:arcla}
    C_i=\mathcal{T_{A,C}}(g_i, h_i; \textcolor{green}{\bm{C}}, \textcolor{green}{\bm{A_S}}) \text{, where } C_0=\mathcal{T_{A,C}}(g_0, h_0; A_L, \textcolor{green}{\bm{A_S}})
\end{equation}
\begin{equation}\label{eqn:argen}
    G_i=\mathcal{T_{A,G}}(g_{i-1}, h_i; \textcolor{purple}{\bm{C}}, \textcolor{purple}{\bm{A_S}}) \text{, where } G_0=\mathcal{T_{A,G}}(X;\textcolor{purple}{\bm{L}},\textcolor{purple}{\bm{S}}).
\end{equation}

\subsubsection{\algname Implementation} We simulate algorithmic reparation using the two variants of \algname introduced in \Cref{ssec:methods_ar}; \claalgname and \genalgname. The algorithm uses sampling and pseudo-labelling to create training batches of size $b$ that meet the fairness ideal by having a prescribed number of samples to fill a quota in each category. However, the closeness between this fairness ideal and the resulting \distrname of the batch is bounded by the reparation budget $r$, which simulates costs to conducting reparation. For our experiments, we use a uniform distribution as the fairness ideal.

\algname creates a pool of $b+r$ samples from either the previous generator or the original dataset, which is then annotated by $A_S$ and either $A_L$ or $C_{i-1}$. The fairness ideal, multiplied by $b$, gives a quota for the number of samples ideally belonging to each category. \algname then attempts to populate each category to its quota from the pool of samples. If after this initial populating, some of the categories did not meet their quota, the remainder of the batch is populated with randomly selected samples from the remaining pool. This process (henceforth re-sampling) will most likely add samples representative of the majority group and class. 
There are therefore two barriers to meaningful reparation: 1) we cap the number of samples that may be drawn to form the batch yet attempt to create equal categories from an unequal dataset; and 2) the effects of MIDS. If \algname increases the representation of a minoritized group, then these samples may be re-selected more often than majoritized group peers, increasing their exposure to mislabeling. Additionally, the higher number of generations this data is subjected to may accelerate the model collapse for these groups. \algname is shown for binary labels and binary sensitive attribute (4 categories in the \distrname) in \Cref{alg:AR_batches}.

\section{Evaluation}\label{sec:results}
We conduct two main sets of experiments to illustrate MIDS in the \seqc and \sgsc settings.
Our results seek to answer several questions which we formalize and answer in brief:

\paragraph{Q1) What are the effects of MIDS on performance, representation, and fairness?}
In both \seqc and \sgsc, we find that the performative prediction, model collapse, and disparity amplification MIDS lead to a loss of accuracy, fairness, and representation in classes and/or groups. These effects are more pronounced in \sgsc, likely because model collapse sometimes results in the beneficial class and majoritized group dominating the generated samples. These effects are more severe in data ecosystems with higher proportions of synthetic data, which we ablate in \Cref{ssec:synthetic}.

\paragraph{Q2) Why is it important to be aware of MIDS?}
We find that unawareness of MIDS in either setting results in overstating the accuracy and fairness, which can be observed by measuring the relative performance of classifiers (comparing $C_i$ against labels provided by $C_{i-1}$) instead of the original data distribution. In \seqc, relative results show nearly 100\% accuracy and near-perfect fairness (using equalized odds difference), the same holds for \sgsc with the addition of mis-reported class and group balances (see \Cref{app:rel_mids}).

\paragraph{Q3) How do MIDS interact?}
We compare \sgsc and \sgnsc, revealing that the fairness feedback loop in the former allows the classifiers to adapt to distribution shift in the inputs caused by model collapse. This co-operation lessens the rate and degree of classifier performance decline. When training with a mixture of synthetic and non-synthetic data, we observe that the non-synthetic data greatly slows the degree of model collapse, though also enables disparity amplification amongst groups in the non-synthetic data ecosystem. 

\paragraph{Q4) Can AR interventions alleviate the harms of MIDS?}
Our AR interventions using \algname lessen these unfair behaviors and achieve better downstream classifier fairness, especially in \seqc. 
\claalgname reduces harms in \seqc and our experiments on disparity amplification, where we train with a mixture of synthetic and non-synthetic data. For 100\% synthetic training in \sgsc, \genalgname usually performs better than \claalgname, likely due to the strength of the model collapse MIDS in deteriorating the data ecosystem.

\subsection{Modeling MIDS}\label{ssec:modeling_mids}
\subsubsection{Experimental Setup} 
We provide computer vision experiments for four datasets. 
We modify \texttt{MNIST} and \texttt{SVHN} into \texttt{ColoredMNIST} and \texttt{ColoredSVHN} by adding color to create binary sensitive groups and by forming two classes for digits $<5$ and $\geq 5$.
We choose the beneficial class as the class converged to by model collapse, and bias the majoritized group towards it. These arbitrary choices simplify our presentation; we vary the class and group balance in \Cref{ssec:class_group_imbal}, finding little impact. 
We also use \texttt{FairFace} and \texttt{CelebA} for a more complicated and real-world task, but also contrast the fairness of MIDS on datasets with and without group balance. 
For \texttt{CelebA}, our task is to predict attractiveness with gender as the sensitive attribute; these attributes have well-documented errors and disparities~\citep{Lingenfelter2022issues}. For \texttt{FairFace}, we attempt to predict gender (2 values) with sensitive attributes race (7 values) and age (which we binarize at $<30$, $\geq 30$). Note that our \texttt{FairFace} experiments are intersectional, and for $A_S$ we use a different classifier for each sensitive attribute. For \texttt{CelebA} and \texttt{FairFace}, we provide between 5-10 generations,\footnote{The time to repeatedly train \texttt{CelebA} and \texttt{FairFace} from scratch took around a week, and due to the cost and \textit{CO$_2$} footprint, we elected to terminate these experiments upon realization of the MIDS.} for \texttt{ColoredMNIST} and \texttt{ColoredSVHN} we train for 40 generations. When training each generation, all synthetic data is sampled from the prior generator and/or classifier/annotator, while non-synthetic data is taken from the training distribution. 

Further details on the datasets (including their class and sensitive attribute distributions) and model architectures, hyperparameters, and compute specifics are in \Cref{sec:experimental_details}.\footnote{Our code is hosted anonymously here: \url{https://anonymous.4open.science/r/FairFeedbackLoops-1053/README.md} and will be released to GitHub if accepted for publication.} Loss values for generators may be found in \Cref{fig:gen_losses}, and accuracies and fairnesses for $A_L$ and $A_S$ are in \Cref{tab:annotator_perfs}. These performances reflect baseline results of training without fairness optimization. 
For \algname, we set the reparation budget $r$ at 25\% of $b$, the batch size, for all datasets aside from \texttt{ColoredSVHN}, which is set to 33\% of $b$. This budget was tuned via grid search by finding the smallest reparation budget that results in a decrease in the proportion of each batch resampled over generations, indicating that \algname is changing the data ecosystem towards its ideal. 

In both settings, we also experiment with training models from a 50-50 mixture of synthetic and non-synthetic data. This allows us to observe the effects of disparity amplification as it co-occurs with performative prediction and model collapse. We use this data mixture to train the classifiers in \seqc, and the generators in \sgsc where we observe downstream impacts in the classifiers.

\paragraph{MIDS Metrics.}
To measure MIDS and their fairness impacts, we inspect the original dataset, generated outputs and annotations, and model performances. We use a held-out evaluation set \textit{i.i.d.} from the original training set $D$. 
In the classifiers, we measure fairness using demographic parity difference (DP), equalized odds difference (EOdds), and group accuracy gaps, and measure utility with accuracy. DP difference (\Cref{def:dp}) compares the positive prediction rates between groups. EOdds difference (\Cref{def:eodds}) is the maximum between two values: the difference between the groups' true positive rates, or between their false positive rates. For performance disparity, we report the maximum accuracy gap when comparing all sensitive groups. For these metrics, a lower value (less difference between groups) indicates more fairness. Note that accuracy and EOdds require a ground truth label which may be taken from a biased original distribution. Therefore, to meet other fairness objectives, such as in \algname, EOdds and group accuracy differences may worsen as the label distribution changes. We measure these metrics on the evaluation set, but also between successive classifiers using images from $G_0$ or $G_i$ with sensitive annotations from $A_S$ and labels from $C_{i-1}$. 
The difference between the former (measuring with respect to $D$) and the latter (measuring with respect to the preceding models) shows how model performances can be misreported if the evaluator is unaware of MIDS.
To track the class and group representation of the generators, we generate 1000 samples and annotate class and group with $A_L$ and $A_S$.

\looseness=-1
\paragraph{MIDS \distrname.} We also observe the \distrname of the original dataset (using $X$, $L$, and $S$), the model training batch \distrname (using $G_0$ or $G_i$, $A_L$ or $C_{i-1}$, and $A_S$), and the model output \distrname (classifiers using $X$, $C_i$, and $A_S$, generators using ($G_i$, $A_L$, and $A_S$). 
We also record the Kullback–Leibler (KL) Divergence between these \distrname and the fairness ideal used in \algname. These KL-Divergence results provide a simple way to measure and visualize change in intersectional representation for our experiments; a `fairness ideal' is otherwise inappropriate (see \Cref{ssec:methods_ar} for clarity on representational thinking).
We also measure the progression of \algname through the \distrname it achieves during batch curation, the amount of resampling required when categories fail to meet their quotas, and the KL-Divergence between the \distrname and the fairness ideal.

\subsection{Results}\label{ssec:results}

\subsubsection{Sequential classifier setting}\label{ssec:results_cla}
\begin{figure}[ht]
\centering
\begin{tabular}{ccc}
\includegraphics[width=0.30\textwidth]{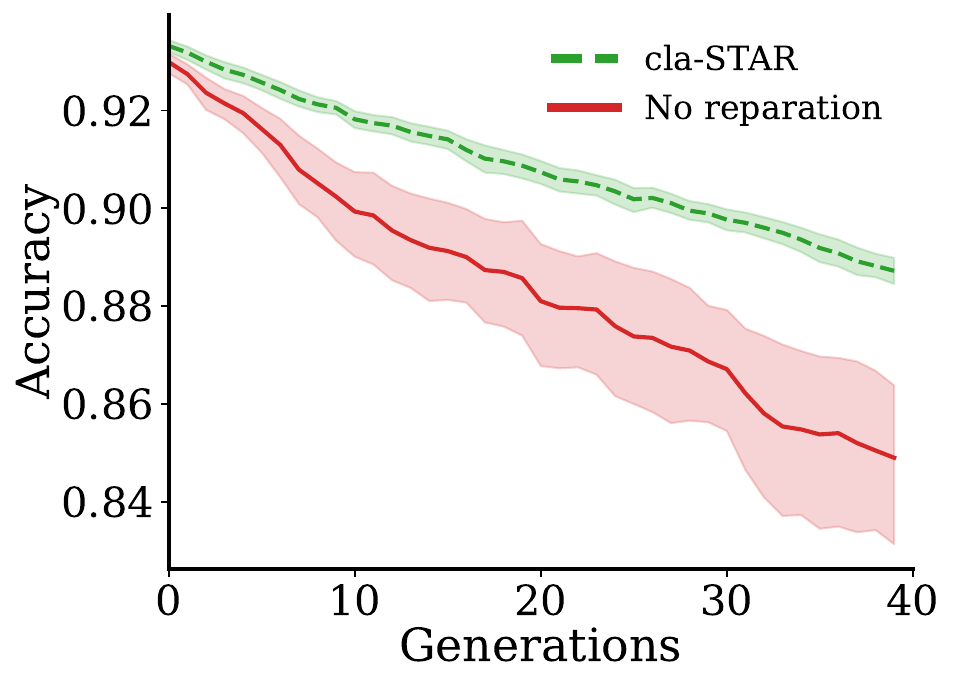} &
\includegraphics[width=0.30\textwidth]{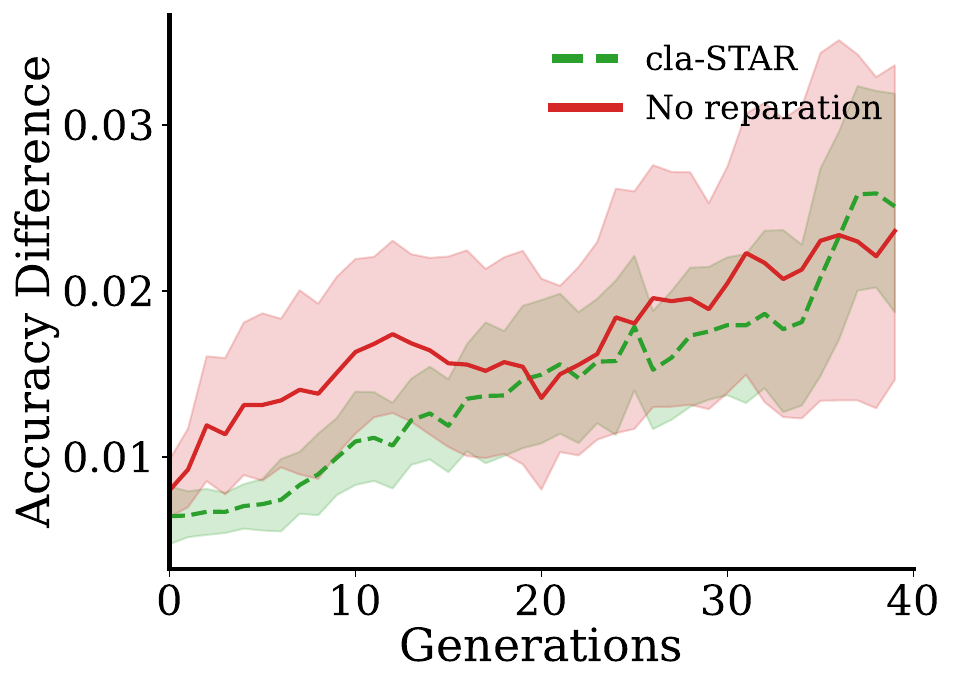} &
\includegraphics[width=0.30\textwidth]{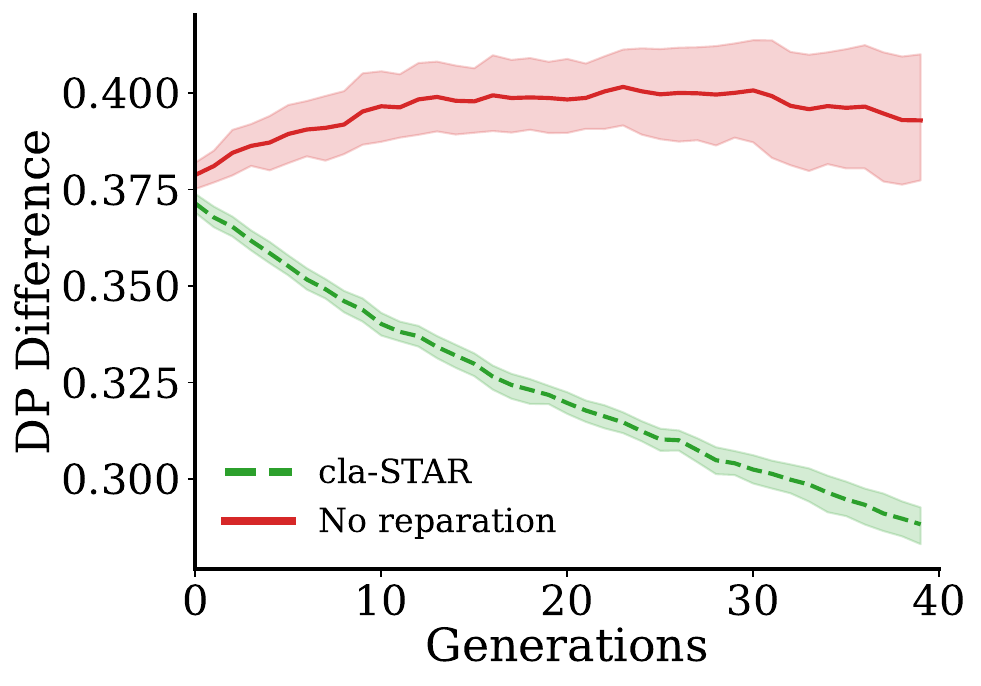} \\
\end{tabular}
\begin{tabular}{ccc}
\includegraphics[width=0.30\textwidth]{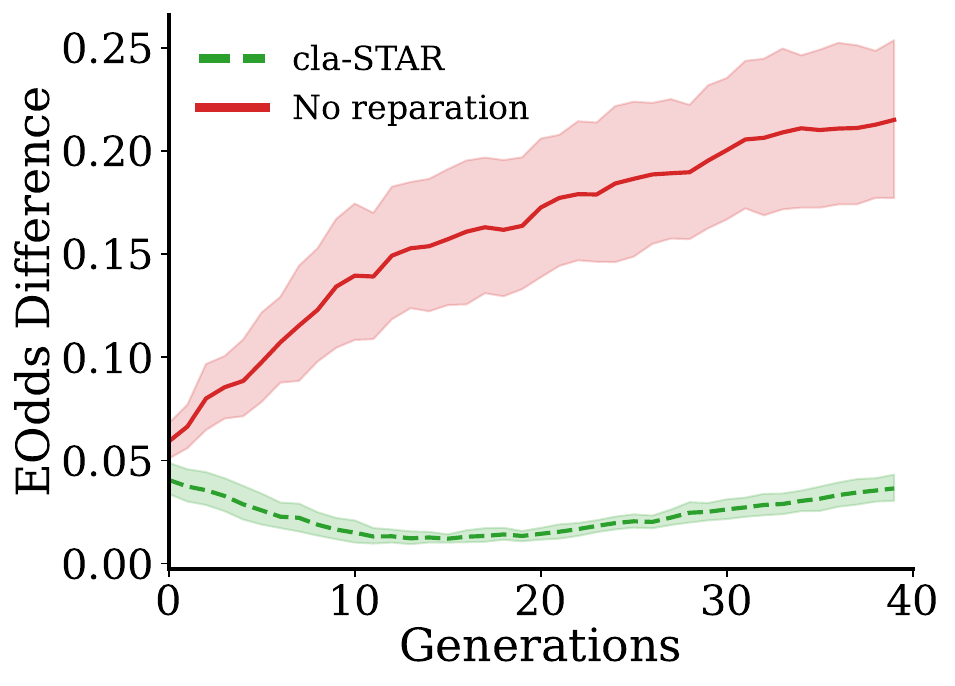}
\includegraphics[width=0.30\textwidth]{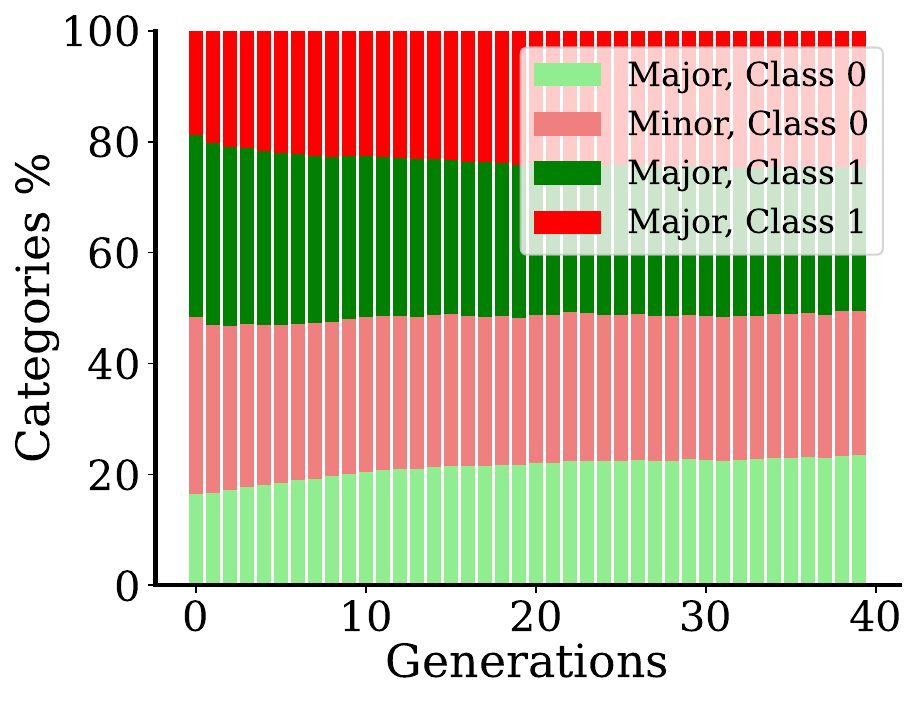} &
\includegraphics[width=0.30\textwidth]{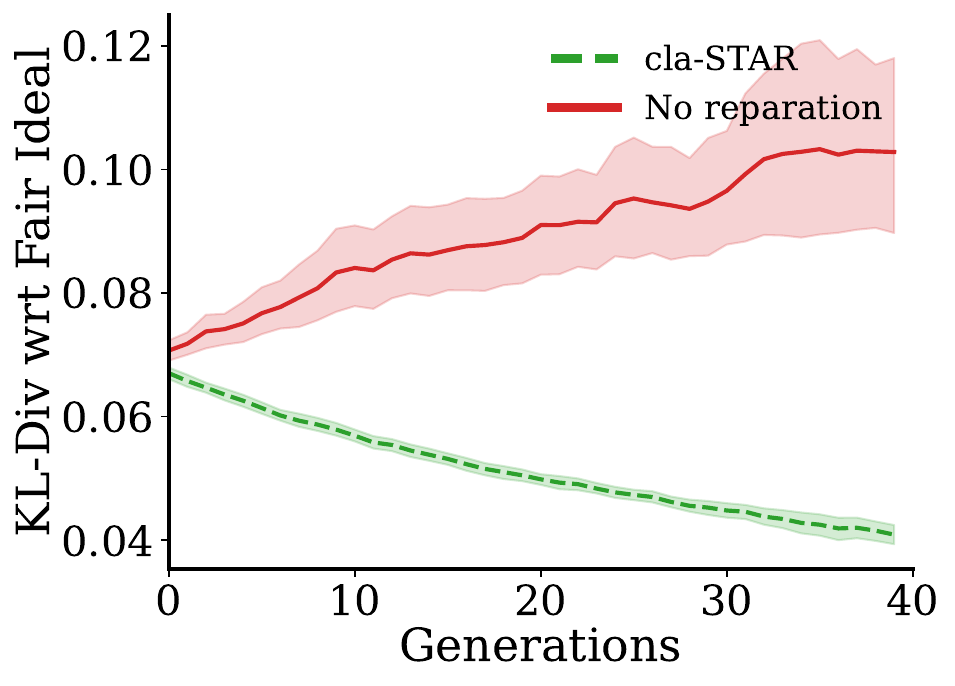}\\
\end{tabular}
\caption{\texttt{ColoredMNIST} results for \seqc on the evaluation set. \textit{Top:} accuracy, accuracy difference, and demographic parity difference. 
Better fairness (lower fairness difference) and higher accuracy is achieved with \claalgname. \textit{Bottom:} equalized odds difference, the \distrname created during \claalgname, and the KL-divergence between \claalgname fairness ideal and classifier \distrname. The KL-Divergence decreases with \claalgname, indicating more fairness, and the batches become more balanced across group and class. The accuracy difference and EOdds difference between groups is small, but increases during reparation due to metric tension with \claalgname.}
\label{fig:ColoredMNIST_nomc}
\end{figure}

Our first experiment suite uses \seqc as described in \Cref{ssec:seq_cla}; MIDS occur as a classifier's predictions are used to label the next generation's classifier. 
In \texttt{ColoredMNIST} and \texttt{ColoredSVHN} (Figures~\ref{fig:ColoredMNIST_nomc} and ~\ref{fig:SVHN_nomc}), we observe an accuracy drop of 10-15\% over 40 generations, with an increase in both DP and EOdds unfairnesses (in the case of \texttt{ColoredSVHN}, both metrics increased by roughly 0.2, where the maximum unfairness gap is 1). \texttt{CelebA} immediately suffers near-random classifier performance as $G_0$ misrepresents $D$ by incurring significant class imbalance towards the detrimental class (\Cref{fig:celeba_nomc}). \texttt{FairFace} classifier accuracies drop from 57\% to random accuracy within 10 generations, and also incur an accuracy difference increase of .1, with a .2 jump in EOdds unfairness (\Cref{fig:FairFace_nomc}). Note that these performances would likely worsen with additional generations.

\paragraph{\algname reduces performance degradation from MIDS.}
In \texttt{ColoredMNIST}, \texttt{ColoredSVHN}, and \texttt{CelebA} \claalgname lead to a significant reduction in DP and EOdds unfairness, and lessened the gap between \claalgname's fairness ideal and the data ecosystem \distrname (Figures~\ref{fig:ColoredMNIST_nomc}, ~\ref{fig:SVHN_nomc}, and ~\ref{fig:celeba_nomc}). Across most datasets, there is far less variance compared to results without reparation, where variance grows with the number of generations (all figures report the 95\% confidence interval).
For \texttt{FairFace}, classifier \distrname without reparation are constituted primarily of older white males (where younger white males are the plurality of the dataset, see \Cref{fig:fairface}). With reparation, the representation of young white non-males increases, but as $G_0$ fails to adequately generate samples from the other races, there are still large performance disparities and high unfairnesses (\Cref{fig:FairFace_nomc}). 
See \Cref{ssec:nomc_strata} for detailed figures on the representation of classes and groups in training batches.
Due to the high representational disparity between the white race and the other 6 races, \claalgname did not lead to better fairness.   
We also observe tension between FML metrics: for \texttt{ColoredMNIST} (\Cref{fig:ColoredMNIST_nomc}), the EOdds and accuracy difference increase after generation 15, while both DP and the KL-divergence continue to decrease. As EOdds is satisfied when error rates (relative to a potentially biased dataset) are similar across groups, meeting the \algname fairness ideal leads to an `unfair' allocation of beneficial labels to the minoritized class, and of detrimental labels to the majoritized class.  

\paragraph{Non-synthetic data slows MIDS, including \claalgname.}%
We also trained classifiers on an even mixture of synthetic and non-synthetic data as described in \Cref{ssec:seq_cla}, see \Cref{fig:nomc_50}. Unsurprisingly, adding non-synthetic data greatly improved the performance of classifiers compared to results with 100\% synthetic data and no reparation (see a full ablation of the amount of synthetic data in \Cref{app:ablation_studies}). While we were able to further increase this fairness by using \claalgname, the impact was far less than on the 100\% synthetic results, with FML unfairness metrics converging to higher values at around the $25^{\text{th}}$ generation. Because the non-synthetic data lessens the impact of the pseudo-labelling enabler in the fairness feedback loop, it likewise lessens the impact of \claalgname.

\subsubsection{Sequential generator and classifier setting}\label{ssec:results_gen}
\begin{figure}
\centering
\begin{tabular}{cccc}
\includegraphics[width=0.22\textwidth]{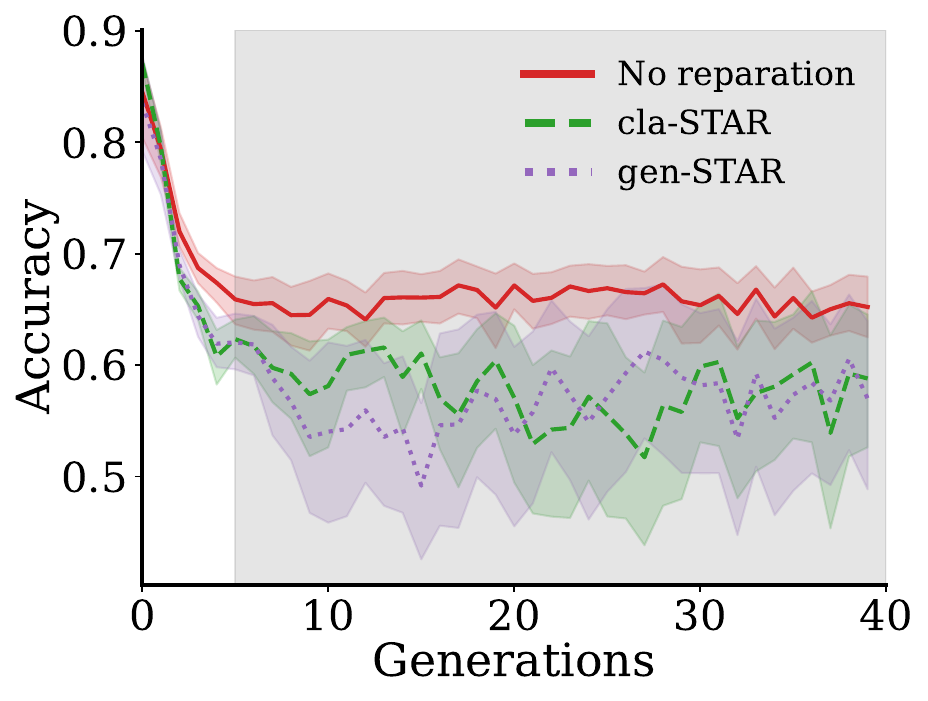} &
\includegraphics[width=0.22\textwidth]{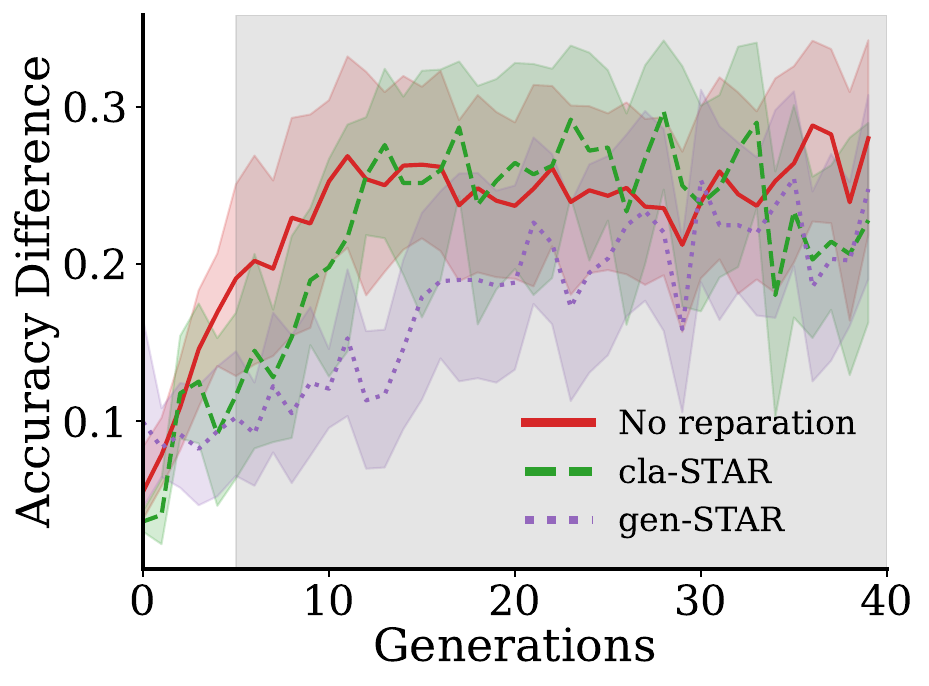} &
\includegraphics[width=0.22\textwidth]{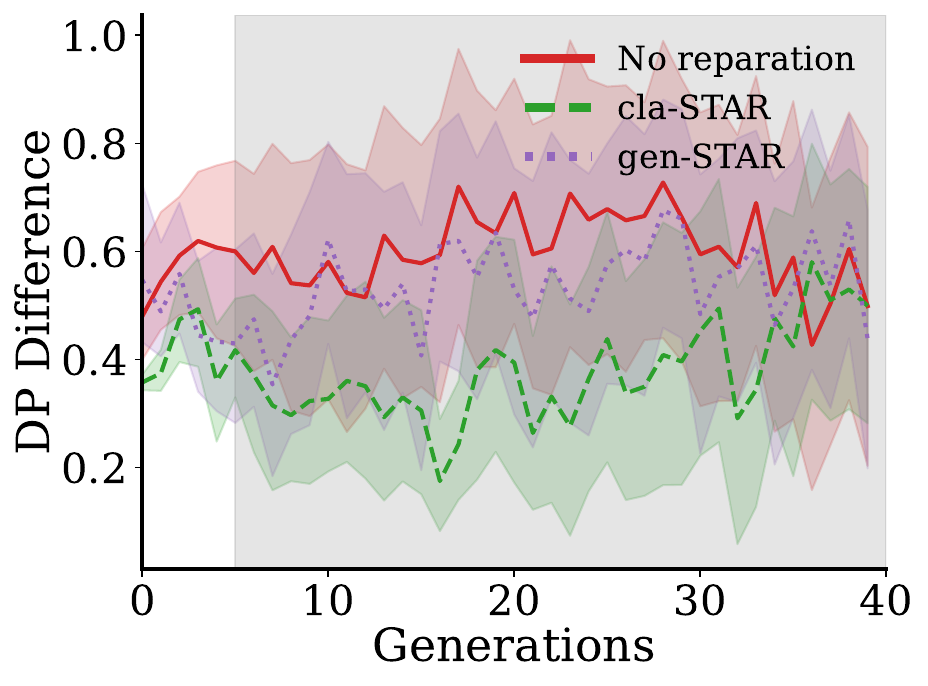} & 
\includegraphics[width=0.22\textwidth]{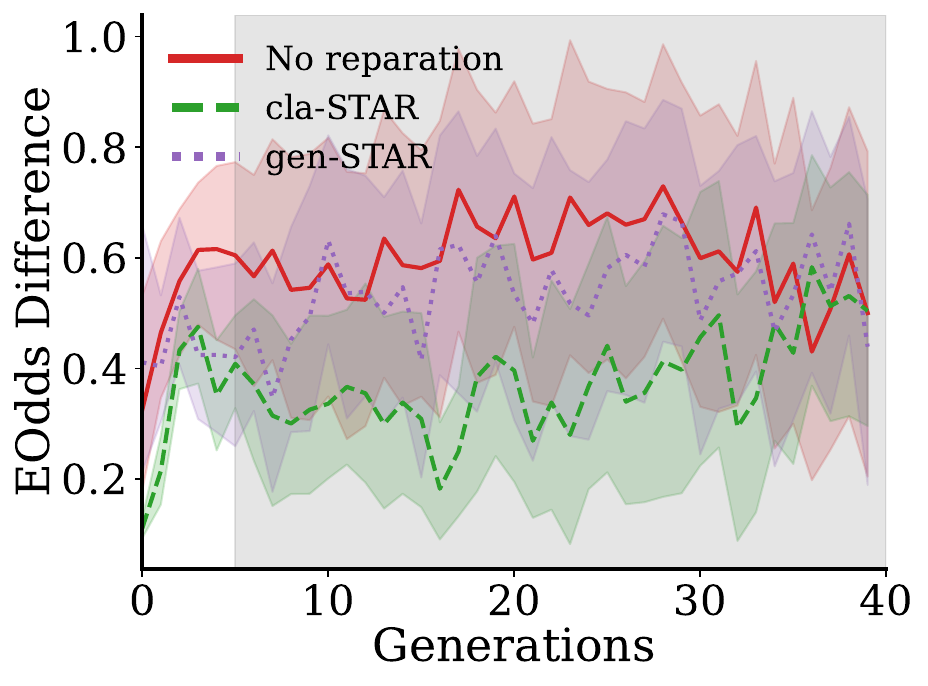} \\
\includegraphics[width=0.22\textwidth]{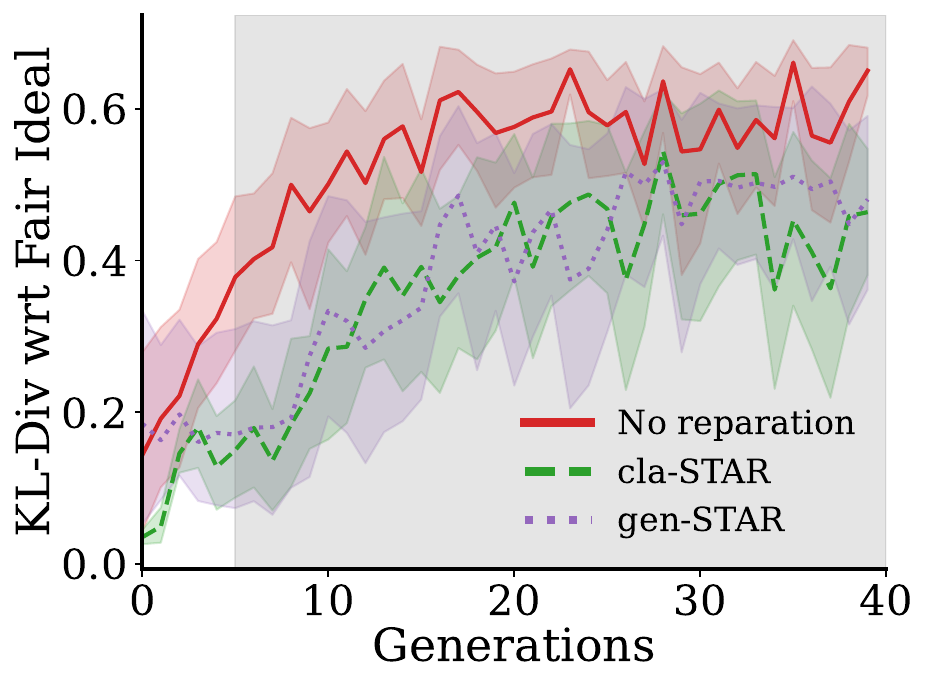} &
\includegraphics[width=0.22\textwidth]{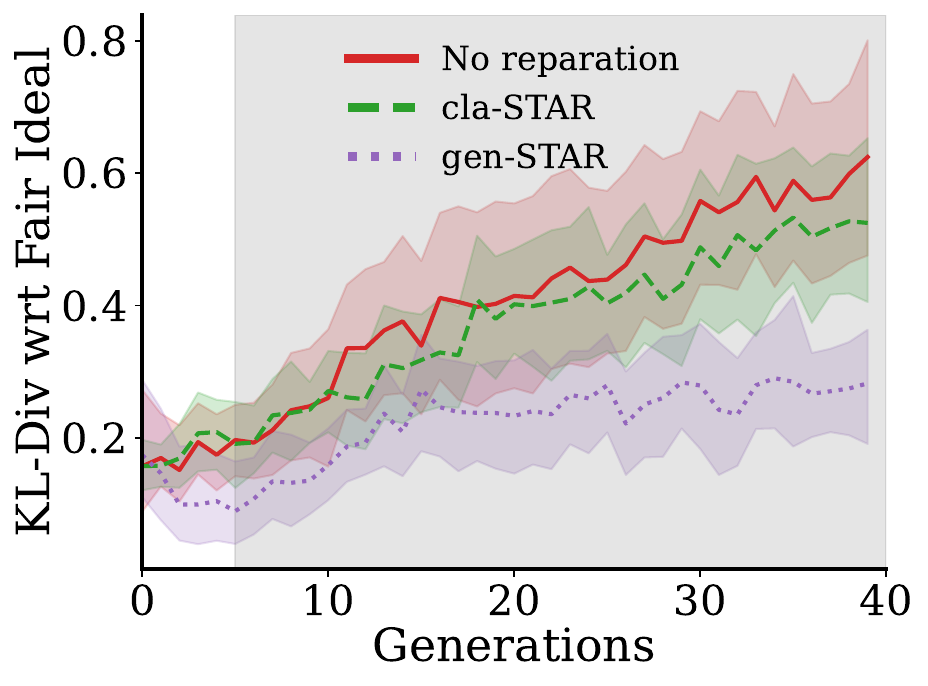} &
\includegraphics[width=0.22\textwidth]{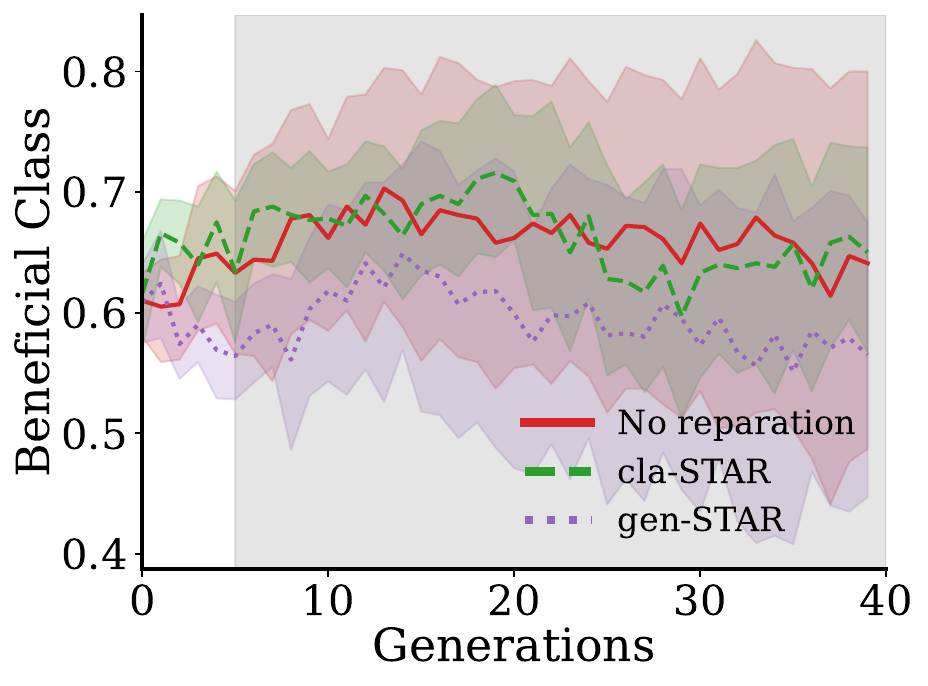} &
\includegraphics[width=0.22\textwidth]{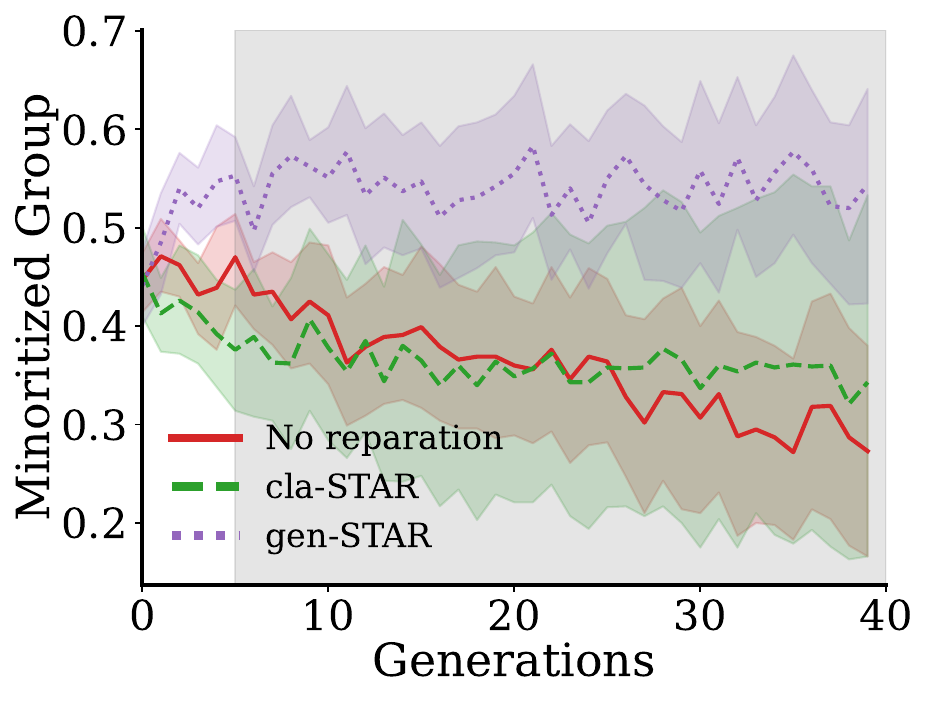} \\
\end{tabular}
\caption{\texttt{ColoredSVHN} results for \sgsc. \textit{Top:} shows accuracy, accuracy difference, demographic parity difference, and equalized odds difference. For the latter three, lower values are better. \textit{Bottom:} KL-Divergence between fairness ideal and classifiers, and between fairness ideal and generator \distrname, the class balance, and group balance. Shading shows collapsed generations.
We observe that \genalgname provides more minoritized group representation. While model collapse causes outputs to eventually resemble a `3,' which moves class balance towards the beneficial class, \genalgname also maintains the original dataset imbalance of 60\%. } 
\label{fig:SVHN_mc}
\end{figure}

\begin{figure}
\centering
\begin{tabular}{cccc}
\includegraphics[width=0.22\textwidth]{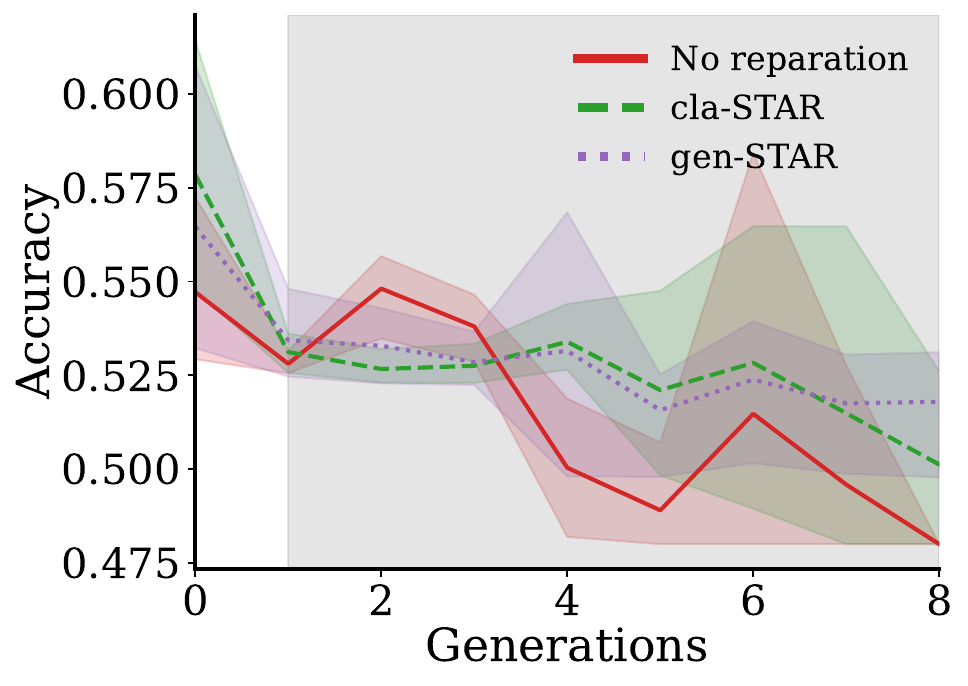} &
\includegraphics[width=0.22\textwidth]{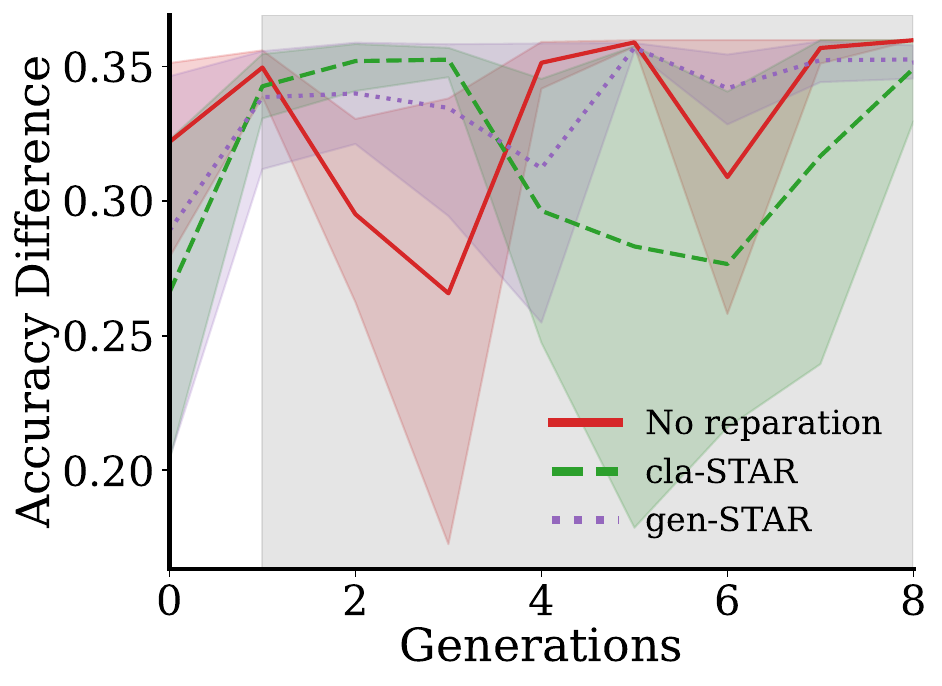} &
\includegraphics[width=0.22\textwidth]{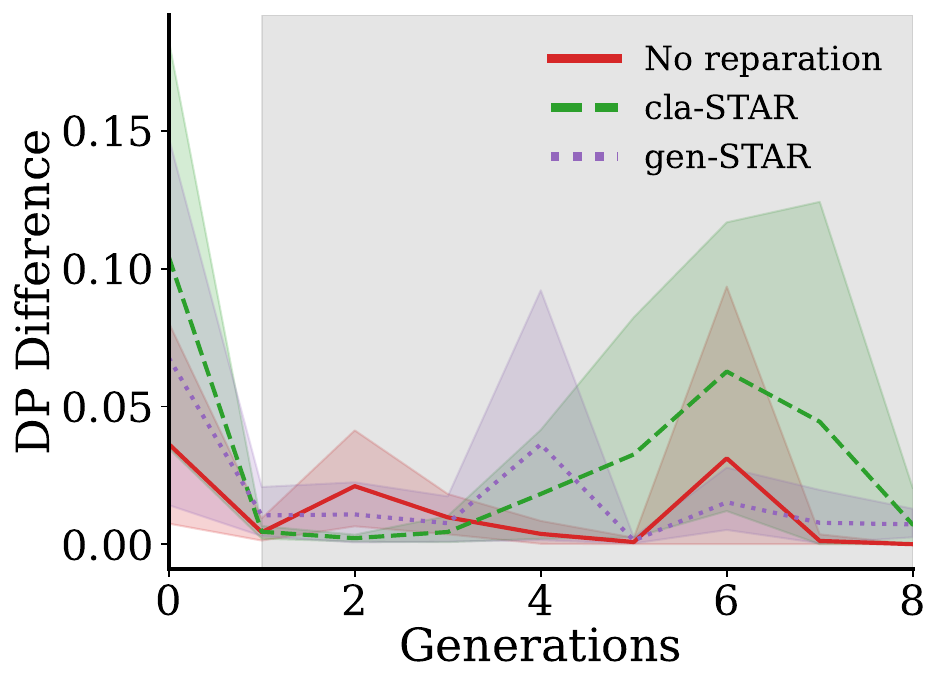} &
\includegraphics[width=0.22\textwidth]{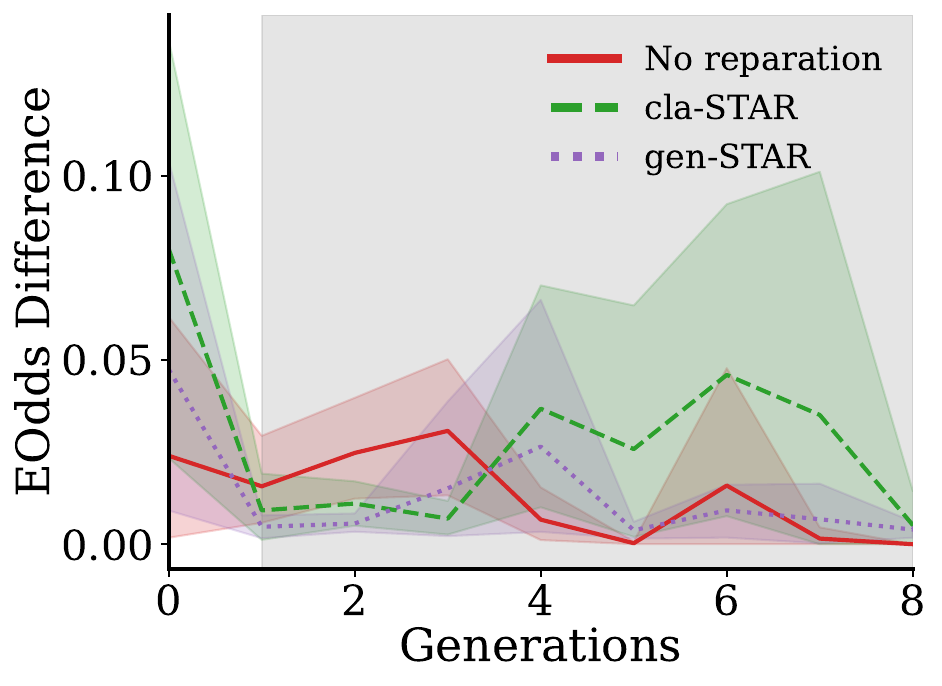} \\
\end{tabular}
\begin{tabular}{cc}
\includegraphics[width=0.22\textwidth]{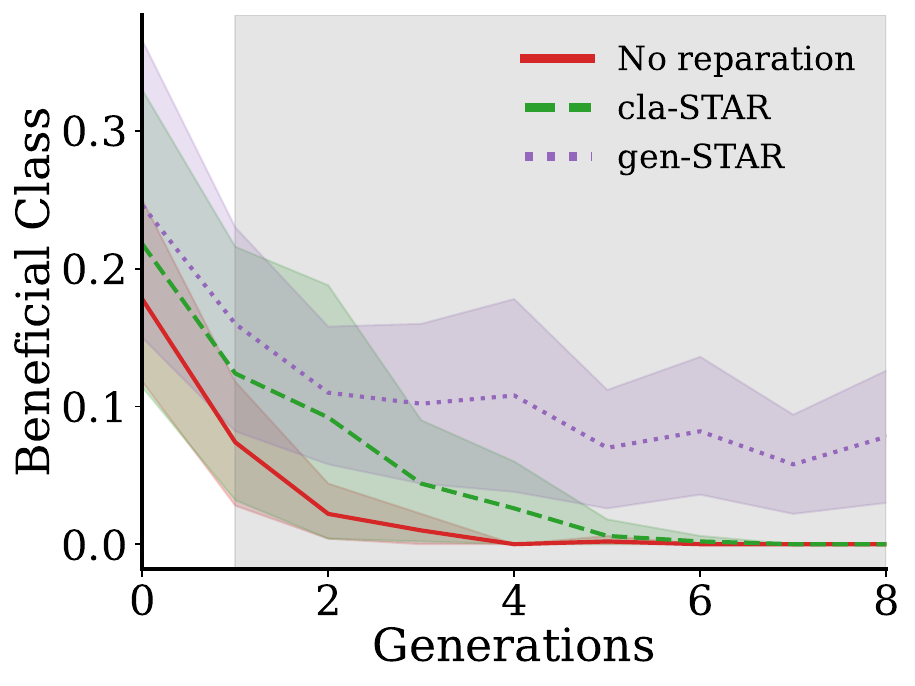} &
\includegraphics[width=0.22\textwidth]{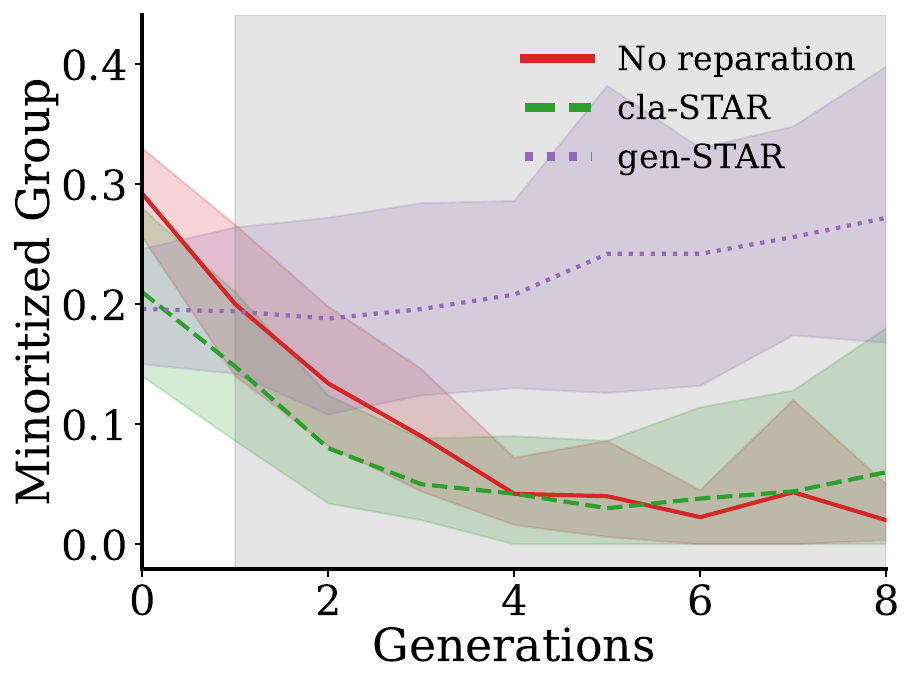} \\
\end{tabular}
\caption{\texttt{CelebA} results for \sgsc. \textit{Top:} shows accuracy, accuracy difference, demographic parity difference, and equalized odds difference. For the latter three, lower values are better. \textit{Bottom:} shows the class balance the and group balance. Shading shows collapsed generations. The performance of the classifiers was initially low, though \algname is moderately better in later generations. \genalgname provides better class and group balance compared to \claalgname and results without reparation, but is still unable to achieve uniform representation due to the strength of model collapse. } 
\label{fig:celeba_mc}
\end{figure}

These experiments refer to the \sgsc setting described in \Cref{ssec:seq_gen} and depicted in \Cref{fig:flow_mc}, which we use to depict model collapse and performative prediction, with additional results including the effects of disparity amplification. 
In our 100\% synthetic training experiments, we observe model collapse deteriorates the data, leading to class imbalance (\texttt{ColoredMNIST} \Cref{fig:ColoredMNIST_mc}, \texttt{CelebA} \Cref{fig:celeba_mc}) and/or to group imbalance (\texttt{ColoredSVHN} \Cref{fig:SVHN_mc}, \texttt{CelebA} \Cref{fig:FairFace_mc}). 
As model collapse progresses, the downstream classifiers either perform with random accuracy or constantly predict the beneficial class label, leading to poor fairness. 
Refer to \Cref{app:gen_loss} to see generated samples undergoing model collapse.

If we judge model collapse at the point when the generated data ceases to have any downstream utility, model collapse occurs at generation 15 for \texttt{ColoredMNIST}, 5 for \texttt{ColoredSVHN}, and between generations 1-5 for \texttt{CelebA} and \texttt{FairFace}.
These values correspond to the increasing difficulty of the datasets' tasks, which is correlated with heavy-tailedness in their distributions~\citep{Meng2023difficulty}. A small sample size may be able to represent a concentrated distribution, but finite samples of a heavy-tailed distribution will likely be biased, leading to faster distribution shift.
The steep decline to random accuracy is likely due in part to the amount of synthetic data used to train each generation. As the amount of synthetic training data decreases, so too does the rate of accuracy decline and beneficial class dominance, as shown in \Cref{ssec:synthetic}. 

\paragraph{Unpredictable convergence of model collapse.}
As model collapse progresses, it is difficult to predict the category of the \distrname in the generators that dominates. 
For example, in generation 40 of \texttt{ColoredMNIST}, both majoritized and minoritized groups of the beneficial class each constitute around 40\% of each batch, whereas in \texttt{ColoredSVHN}, the majoritized and beneficial category alone constitutes 60\% of each batch (Figures~\ref{fig:ColoredMNIST_mc} and~\ref{fig:SVHN_mc}). 
This is in part due to the initial class and label balances (see \Cref{tab:datasets}), but also due to how the model collapse manifests. 
For example, \texttt{ColoredMNIST} eventually converges to samples that resemble an `8', or all the digits superimposed (see samples from model collapse in \Cref{app:gen_loss}). As this happens to fall in the advantaged class, it becomes dominant in the generators, without strongly impacting the group balance. 
Meanwhile, in \texttt{CelebA}, model outputs become dominated by the majoritized group, yet these same samples are also classified into the detrimental class by $A_S$, which is counter to the original label balance in the dataset. Eventually, minoritized group representation in \texttt{CelebA} falls to 0\% (see \Cref{fig:celeba_mc}). 
\texttt{FairFace} has much deterioration in the data but maintains both class and group balance. 
These observations support a growing consensus that the features of the original data preserved by generative models in synthetic data is difficult to predict, or highly data dependent (for private synthetic data~\citep{stadler2022synthetic}, and at the intersection of fairness and privacy in synthetic data~\citep{cheng2021fake}).

\paragraph{Performative prediction adapts to model collapse.}
We also uncover co-operation between MIDS by evaluating the role of sequential classifiers in \sgsc and \sgnsc (see \Cref{ssec:seq_v_noseq}). 
In \texttt{ColoredMNIST} and \texttt{ColoredSVHN} (Figures \ref{fig:abl_noseq_cmnist} and \ref{fig:abl_noseq_svhn}): the non-sequential classifiers converge to accuracies 10-20 percentage points lower than the sequential classifiers. However, the sequential classifiers have considerably more unfairness (in the case of \texttt{ColoredMNIST}, by 0.6), likely due to their participation in fairness feedback loops. 
Performative prediction among sequential classifiers allows $C_{i-1}$ to provide meaningful labels for training $C_i$ from $G_i$. For the non-sequential classifiers, $A_L$ cannot adequately support the distribution represented by $G_i$ once the $i^{th}$ distribution substantially differs from the original. The inherited knowledge of $\mathbb{P}(Y|X)$ passed through the sequential classifiers allows them to preserve a more accurate map from the changing distribution to the classes.

\paragraph{\genalgname improves fairness and minoritized representation.}
Between \genalgname and \claalgname, the former leads to more preservation of the group and label balance in all four datasets. 
This result fits intuitively as the biased sampling enables these generators to maintain more balanced representations across the categories. For example, \genalgname leads to better fairness than \claalgname in \texttt{ColoredMNIST} and \texttt{CelebA}, though with cost to accuracy.
However, because \genalgname results in oversampling minority (in terms of population) categories relative to the original dataset, it may also expose these areas of the data distribution more to model collapse. In \texttt{ColoredSVHN}, for example, \genalgname results in more balanced \distrname in both the generators and classifiers (compared to \claalgname), but the classifier \distrname are still dominated by the (minoritized, detrimental) and (majoritized, beneficial) categories, leading to worse DP and EOdds fairness (see \Cref{ssec:mc_strata}).
In the case of \texttt{FairFace}, a combination of oracle model bias and unrepresentative generators causes large disparities between races as \genalgname cannot adequately sample from the smallest intersectional minorities. As \texttt{FairFace} has relatively balanced races and genders, these disparities indicate that algorithmic reparation should be considered when collecting data, and might require action beyond collecting balanced quotas of data from various groups. 
Overall, \claalgname did not show consistent performance across datasets, achieving worse or equivalent performance to the non-reparative results, likely due to the strength of the model collapse MIDS.  

\paragraph{Disparity amplification reduced with \claalgname.} 
Recall that we model disparity amplification by sampling non-synthetic data using the \distrname of the classifiers, which we use for half the training data for the generators (the other half is sampled from $G_{i-1}$). We evaluate this setting for \texttt{ColoredMNIST}. Similarly to \seqc, the non-synthetic data slows MIDS caused by synthetic data spills, including model collapse. We evaluate the fairness performance of \genalgname and \claalgname, finding substantially better performance and fairness with \claalgname (subject to an increase in accuracy disparity due to increased false negatives) than with \genalgname (see \Cref{app:fancy}). 
The \genalgname generator \distrname never achieve the fairness ideal as randomly sampling from initially biased generators leads to unfair classifiers which propagates disparity amplification in the non-synthetic data, which incidentally only protects the majority group and class categories from model collapse. Meanwhile, \claalgname trains classifiers with more ideal \distrname that reverses disparity amplification, providing balanced non-synthetic data to the generators and protecting all classes and groups equally from model collapse. We may see this by comparing the \algname \distrname for both algorithms, see \Cref{fig:50_stratas}.

\subsection{Limitations}\label{ssec:limitations}
We discuss three main limitations of our work. Firstly, we do not provide a specific use case and cannot fully evaluate algorithmic reparation, nor make any claims that our achievements in fairness lead to equity or justice. Secondly, when collecting the synthetic data for training a new model, we operate in a worst case where all of it is sampled only from the immediately preceding model(s); without provenance information the synthetic data could be sourced from any number of other models, including multiple predecessor models. Thirdly, in \texttt{CelebA} and \texttt{FairFace}, we rely on race annotations which might fail to represent the various skin tones within groups, an important consideration in computer vision tasks, and for better representation (as discussed in~\citep{Buolamwini2018GenderSI}). Additionally, racial categorizations are not universally consistent, and so these datasets provide a simplification that may be inappropriate. 

\section{Concluding Remarks}\label{sec:conclusions}
In this paper, we introduced model-induced distribution shifts (MIDS) and created empirical settings enabling the evaluation of their harms. With these settings we found that MIDS, both on their own and co-occurring with enablers such as data annotation, lead to major degradation in utility, fairness, and minoritized group representation. While MIDS can be intentional or unintentional, unawareness of their existence can lead to grossly overstating model utility and fairness. Based on these harms, we discussed how algorithmic reparation (from the literature of critical theory in ML) may act as an intentional MIDS with goals of equity and justice. By simulating the impacts of algorithmic reparation at various points in our settings, we saw a lessening in harms.

We would also like to acknowledge that throughout this work, we have implied that models cause model-induced distribution shift. This is not the case; agency over MIDS rests primarily on model owners, data publishers and collectors, and model users. Several related works, including ~\citet{shumailov2023curse} and ~\citet{hardt2023performative}, have also considered the power or advantage given to an entity that has more control over the amount of synthetic data spillage or has more access to non-synthetic data. As we have found that MIDS are of imminent concern in data ecosystems undergoing synthetic data spills; we now turn to solutions to MIDS and methods for taking accountability of them. 
One possible solution, also mentioned in~\citet{Davis2021reparation}, is an archival perspective on data curation as introduced in~\citet{JoGebru2020archives}. Specifically, adopting the tenets of archival description codes could enable gathering of high-quality provenance information, and adopting the moral obligations underlying many an archives' raison d'être could help to identify and repair structural and historical bias~\citep{DACS, RAD, JoGebru2020archives}. Another solution motivated by our results is the importance of non-synthetic data and human data annotation to prevent or slow the rate of MIDS. We therefore advocate for more attention to the often-unseen and underappreciated labor of human data workers. We end with a call for safer conditions for data workers given their current and increasing importance in our data ecosystems.

\clearpage
\section*{Statements}

\subsection*{Ethical Considerations}
We recognize that technical solutions are never disjoint from their societal impacts, and have striven towards a more sociotechnical framing for this work. We navigate several definitions and frameworks for algorithmic fairness and equity by considering multiple definitions of fairness and their contrasts with algorithmic reparation. However, we primarily focus on group-based fairness metrics, including when those groups are formed intersectionally, which we acknowledge can reinforce the ideologies behind them.  

\subsection*{Positionality}
We are researchers usually operating within the more technical areas of machine learning. Throughout our time working in this area, we have turned to \textit{Data Feminism} by ~\citet{dataFeminism} to inform our discussions around fairness and equity, and to inform some of our language choices. We also rely heavily on ~\citet{Davis2021reparation} and ~\citet{representationalist_crit} for their critiques of `representationalist thinking,' which has been admittedly ubiquitous in our education, and which seemingly appears commonly in ML research (including ML fairness research). 

\subsection*{Adverse Impact}
We note that this work might be used to fuel despair over the `long-term' existential harms of models, especially generative models. We advise readers to think critically about the systems of power behind machine learning and consider the current harms these permit, continue, and worsen. 
We also acknowledge that due to our lack of a specific use case to fully evaluate algorithmic reparation, we risk representing it as a mathematical or technical definition to be satisfied or optimized for. This runs counter to the tenets of AR (see~\citet{Davis2021reparation}) and we have been careful with our language around this area (for example in Sections \ref{ssec:methods_ar} and \ref{ssec:limitations}, and in \Cref{sec:fairness_background}).

\section*{Acknowledgements}
We would like to acknowledge our sponsors, who support our research with financial and in-kind contributions: CIFAR through the Canada CIFAR AI Chair program and the Catalyst grant program, Microsoft, and NSERC through the Discovery Grant and COHESA Strategic Alliance. Resources used in preparing this research were provided, in part, by the Province of Ontario, the Government of Canada through CIFAR, and companies sponsoring the Vector Institute. 
We would like to thank members of the CleverHans Lab for their feedback. We additionally thank David Glukhov, Syed Ishtiaque Ahmed, and Ramaravind K. Mothilal for feedback on earlier versions of this work.

\bibliographystyle{abbrvnat}
\bibliography{aaai24.bib}

\appendix

\section{Fairness in Machine Learning (FML)}\label{sec:fairness_background}
\citet{sources_ml_lifecycle} describe several sources of bias and oppression that may be encoded into a model as a result of the processes for gathering and encoding data, training, evaluating, and deploying the model. These sytems may then commit several types of harms, including allocative (where resources are withheld from certain groups, such as in redlining) and representational (where groups are stigmatized and stereotyped). Likewise, there are many different biases that may co-occur such as historical, representational, measurement, aggregation, learning, evaluation, and deployment biases~\citep{suresh2021framework}. These often arise from misrepresenting a complex feature (e.g., treating gender or sex as a binary), mis-measuring features, stripping data of its context (e.g., regional or dialectal language heteroglossia), and from historical oppression influencing the data modelling processes. There are several frameworks for defining and addressing issues of fairness; calibration (used when the sensitive identities have impact on the decision task), anti-classification (used when sensitive data is unavailable or illegal to use), individual fairness (``similar individuals should be treated similarly''), and classification parity. 

In group fairness, protected attributes are often chosen from legally-protected attributes such as race or gender, and encoded into categorical features to determine \textit{sensitive groups}. In this paper we use the terms \textit{majoritized} and \textit{minoritized} as in~\citet{dataFeminism} to emphasize the impact of a model's behavior on a group. Note that the majority population might not correspond with the majoritized (benefit-receiving) group; for example the Black population is a minoritized majority in the COMPAS dataset~\citep{Northpointe2016COMPAS}. This grouping often splits the dataset into two groups, delineated by one attribute with two possible values (e.g., `male' vs `female' or `white' vs `people of color') which may misrepresent or inappropriately group populations and ignores the compounding impact of possessing multiple marginalized identities. 

In classification parity, there are a variety of metrics that aim for some equality of rates between these groups, such as accuracy, positive selection rate, or error rates. In this work, we use accuracy difference, demographic parity difference, and equalized odds difference to cover a multitude of differing priorities model owners may value. It is often impossible to satisfy multiple fairness metrics simultaneously, so they are ideally chosen based upon the task~\citep{chouldechova2016fair, Kleinberg2018inherent}. The binary classification and binary grouping versions of these metrics are presented below. 

\begin{definition}[Demographic Parity (DP)~\citep{calders2009building}]\label{def:dp}
A classifier $\hat{Y}$ satisfies Demographic Parity with respect to the sensitive attribute $s$ if: 
\[\mathbb{P}(\hat{Y} = 1 \vert s = 0) = \mathbb{P}(\hat{Y} = 1 \vert s = 1) \qquad \forall 0, 1 \in s.\]
\end{definition}

In this work we consider demographic parity difference, which is the absolute value of the difference between the two terms equated above. Each term is also the selection rate, or rate of positive prediction, for the group.

\begin{definition}[Equalized Odds (EOdds)~\citep{hardt2016equality}]\label{def:eodds}
A classifier $\hat{Y}$ satisfies Equalized Odds with respect to the sensitive attribute $s$ if for ground truth $L$: 
\[\mathbb{P}(\hat{Y} = 1 \vert L = l, s = 0) = \mathbb{P}(\hat{Y} = 1 \vert L = l, s = 1) \qquad \forall l\in\{0,1\}, \forall 0, 1 \in s.\]
\end{definition}

We also use equalized odds difference. This is formulated as $\text{max}[|\mathbb{P}(\hat{Y} = 1 \vert L = 0, s = 0) - \mathbb{P}(\hat{Y} = 1 \vert L = 0, s = 1)|, |\mathbb{P}(\hat{Y} = 1 \vert L = 1, s = 0) - \mathbb{P}(\hat{Y} = 1 \vert L = 1, s = 1)|]$, or the larger of the absolute value differences between the false and true positive rates for the groups. These metrics may be used for multiple groups by taking each difference between every pair of groups and reporting the maximal disparity. Similarly, these groups may be formed by intersecting multiple sensitive attributes. 

Due to the biases that may exist in data collection, training, evaluation, and deployment, adherence to or achievement of any of these fairness metrics does not guarantee fairness or equity. For example, these fairness metrics assume that the cost of an error borne by a person of any group is the same, when in practice the costs and benefits may differ greatly depending on identity~\citep{makhlouf2021applicability}. There are several works that focus on the trade-offs between meeting a decision maker's FML criteria and the potentially inequitable social outcomes which cast doubt on the suitability of FML metrics for societal welfare~\citep{kasy2021fairness, Hu2020welfare, corbettdavies2017cost}.
This is in part a byproduct from FML's reliance upon algorithmic idealism, where computation assumes a meritocratic society whereby equalizing demographic disparities must therefore lead to fairness at a societal level~\citep{Davis2021reparation, green2020false, kasy2021fairness}. 
Additionally, FML may also engage in or reinforce two main biases: 1) automation bias, that machines are objective and are less biased than humans, and 2) that automation invites justice without regard for the objective and purpose of the models~\citep{Davis2021reparation, green2020false}.  

\section{MIDS in Literature}\label{app:mids_in_lit}
This section provides a more detailed review of the MIDS and enablers described in \Cref{sec:related_work}.

\subsection{MIDS}
\paragraph{Performative prediction.}
\textit{Performative prediction} is a distribution shift that occurs when a model's predictions impact the outcome. For example, when economists publish forecasts, they may influence the behavior of others in the market, causing the market to fit the forecast in a self-fulfilling prophesy~\citep{perdomo2020performative, hardt2023performative}. In this case, the model's predictions leak into the data ecosystem as they become outcomes. If this data ecosystem is trained upon, these outcomes are treated as the ground truth, and the MIDS continues into another generation of models. 

\begin{figure}[ht]
    \begin{tabular}{cc}
         \includegraphics[width=.4\textwidth]{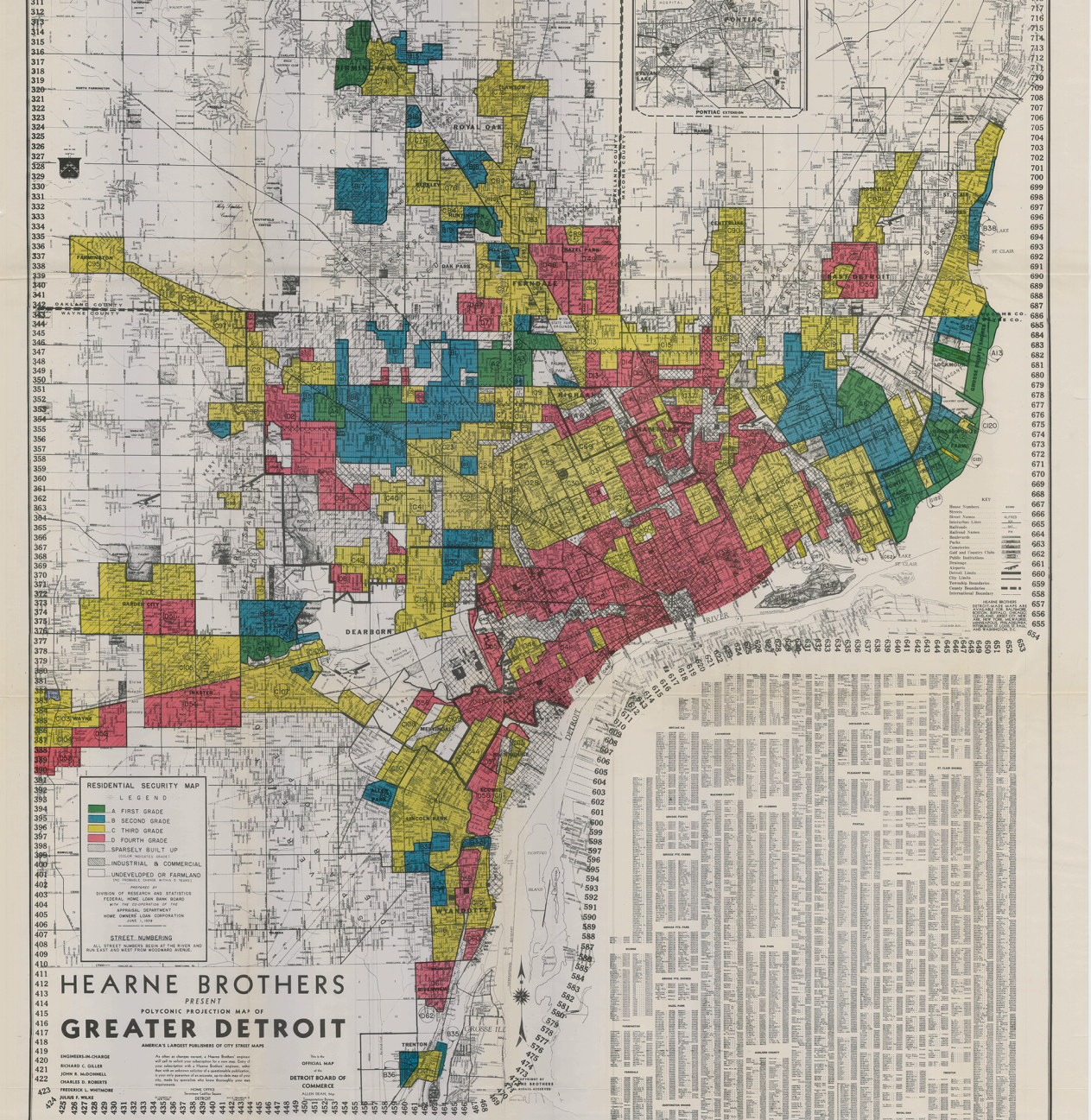} &  
         \includegraphics[width=.35\textwidth]{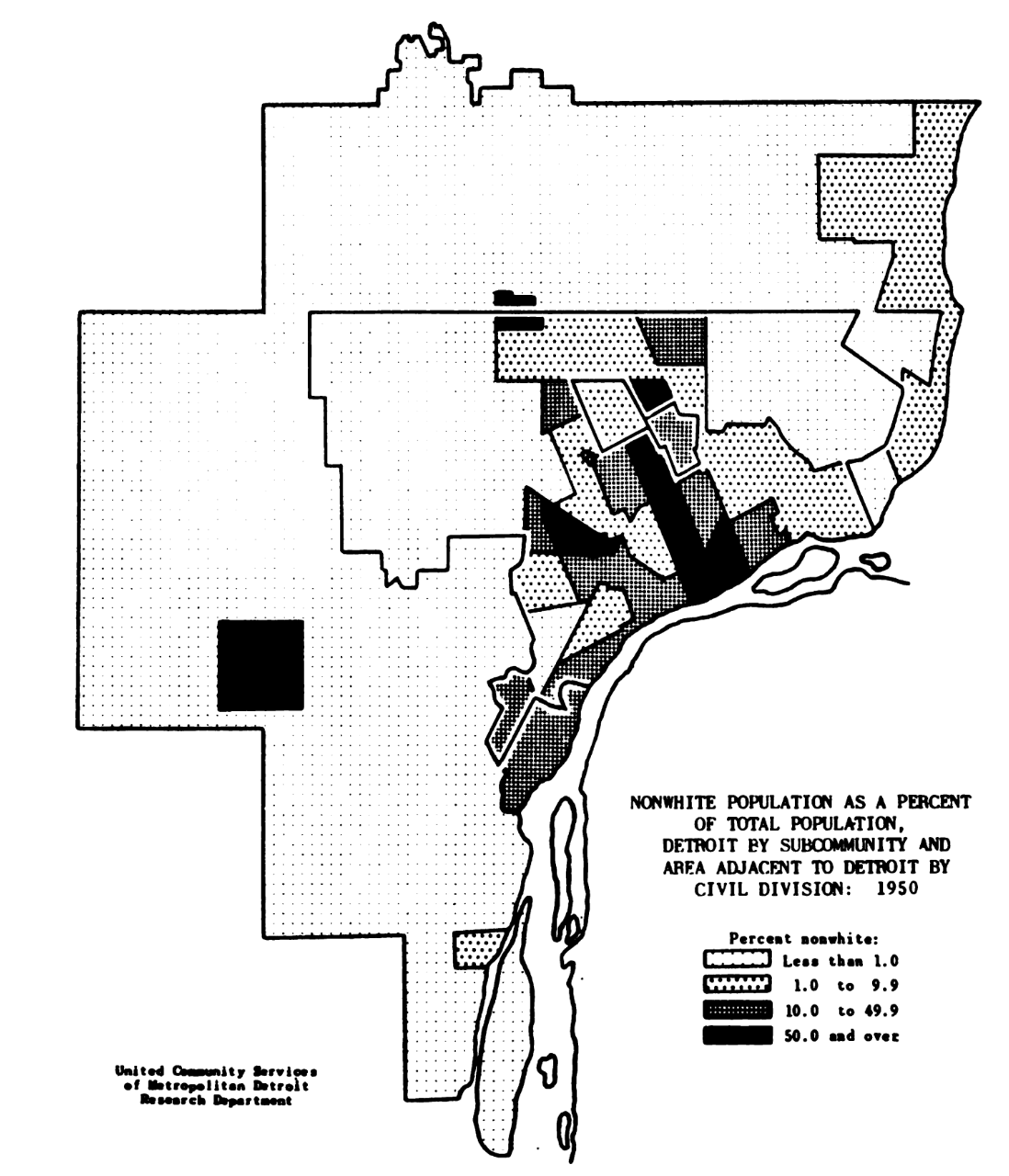}\\
    \end{tabular}
    \centering
    \caption{\textit{Left:} 1939 Home Owner Loan Corporation Security Map for Detroit. Red areas are Grade D, or `hazardous' locations due to the presence of racial and religious minorities~\citep{Nelson2020Mapping}. Banks of the time were likely privy to these otherwise secret maps and in surveys stated that only high-graded neighborhoods would be offered loans~\citep{jackson1985crabgrass, Nelson2020Mapping}. Redlining continued \textit{de jure} until the Fair Housing Act in 1968~\citep{fairHousingAct}. \textit{Right:} Census data on the non-white population in metropolitan Detroit from 1955~\citep{detroitDemographics}.}
    \label{fig:redlining}
\end{figure}

\paragraph{Fairness feedback loops.} The fairness community has studied performative prediction and fairness feedback loops in the context of risk assessment systems, including mortgaging and predictive policing~\citep{green2020false}. We provide two notable examples:

1) \Cref{fig:redlining} shows the 1939 HOLC Residential Security Maps for Detroit alongside 1955 demographic information of non-white communities. This map is a historical multiclass classification model that reflects the values and priorities of the individuals and institutions responsible for its creation; specifically the white male gaze of Depression-era professional realtors~\citep{So2022housing, dataFeminism, jackson1985crabgrass}. These values encoded into this model and the MIDS from the model itself has contributed to housing discrimination in Detroit, as seen in the right of \Cref{fig:redlining}.

2) Work in predictive policing found that policing locations may converge towards over-policing low-income non-white communities~\citep{lum2016predict, richardson2019dirty}. Theoretical follow-ups find that the degree of runaway feedback may be moderated with careful training set weighting, but cannot be negated entirely~\citep{Ensign2017runaway}.

\paragraph{Model Collapse.}
Where performative prediction and runaway feedback loops generally refer to classification models, model collapse describes the same effect for generative models. \textit{Model collapse} occurs when new generative models are trained on samples created by their predecessor over many generations, as introduced in~\citet{shumailov2023curse} and concurrently in~\citet{alemohammad2023selfconsuming}. This leads to new models forgetting the original data distribution as they recreate and amplify the failures of their ancestors. There are two error sources that contribute: 1) functional approximation error due to an inadequately expressive generator, and 2) statistical approximation error from finite sampling. 
Model collapse begins with a loss of information from the tails of the data distribution. In late-stage model collapse, the model mixes the modes of the original distribution, converging to a point estimate of some mean betwixt them. There are also concerns over the effects of model collapse on fairness, as model failure on ``low-probability events" may have negative effects on minoritized groups when datasets have poor representation~\citep{suresh2021framework}.

\paragraph{Disparity amplification.} Unlike the aforementioned MIDS, disparity amplification arises from human-model interaction. If a model suffers from problems derived from representational bias, it may have overall high performance but low performance on minoritized groups (performance disparity). This can lead to \textit{disparity amplification}, where minoritized users who suffer high error rates may choose to disengage from the model, shifting the future  dataset towards the majoritized group, and increasing the representational disparity of the data ecosystem~\citep{Hashimoto2018fairness}.

\subsection{MIDS Enablers}\label{ssec:mids_enablers}
We describe several enablers in more detail here. Note that even sampling is an enabler, and indeed it informs our approach to algorithmic reparation in our experiments. 

\paragraph{Pseudo-labelling.} \textit{Pseudo-labelling} generally refers to using a model to assign labels to unlabeled samples in a dataset, so that this data may also be used for supervised or semi-supervised training~\citep{Bonilla2020curriculum}. This may occur just once~\citep{Bonilla2020curriculum} or iteratively~\citep{McLachlan1975iterative}. Incentives for pseudo-labeling may arise in cases where manual labeling and/or annotation is too expensive for vast swathes of data. The fairness impact of using pseudo-labelling for self-supervised learning was discussed in~\citet{zhu2022rich}, finding that groups with high initial accuracy benefit whereas groups with low initial accuracy may see a degradation in performance.

\paragraph{Feedback and Data Annotation.}
Similarly to pseudo-labelling, feedback (whether human or model-based) is often used for labeling and annotating data for supervised training or for providing feedback on generative outputs. Reliance on human annotation can lead to unfairness arising from individual annotator bias and instructions for annotating~\citep{suresh2021framework}. Recent work has found indicators that some human data annotators use LLMs or other models, which may lead to MIDS if these models are updated and retrained on the data they labeled~\citep{veselovsky2023artificial}. AI feedback is also used in methods such as Constitutional AI, which uses a succession of fine-tuned supervised models to provide RLAIF (reinforcement learning from AI feedback) for training `harmless' AI assistants~\citep{bai2022constitutional}.

\section{Experimental Details}\label{sec:experimental_details}
\subsection{Datasets}\label{ssec:datasets}
We evaluate the fairness effects of model collapse and algorithmic reparation on several datasets; adapted versions of \texttt{MNIST} and \texttt{SVHN}, as well as \texttt{CelebA} and \texttt{FairFace}. 

\begin{table}
    \begin{subtable}[t]{.45\linewidth}
    \centering
    \begin{tabular}{lcc}
        \toprule
                \textbf{\texttt{ColoredMNIST}} & \multicolumn{2}{c}{Class} \\ \cmidrule(lr){2-3} 
                Group & Beneficial & Detrimental         \\ \midrule
                Majoritized & 0.350 & 0.150 \\ 
                Minoritized & 0.150 & 0.350 \\ \bottomrule 
        \end{tabular}
    \end{subtable} 
    \begin{subtable}[t]{.45\linewidth}
    \centering
    \begin{tabular}{lcc}
        \toprule
                \textbf{\texttt{ColoredSVHN}} & \multicolumn{2}{c}{Class} \\ \cmidrule(lr){2-3} 
                Group & Beneficial & Detrimental         \\ \midrule
                Majoritized & 0.424 & 0.118 \\ 
                Minoritized & 0.183 & 0.275 \\ \bottomrule 
        \end{tabular}
    \end{subtable} \par\bigskip
        \begin{subtable}[t]{\linewidth}
        \centering
            \label{tab:celeba}
            \begin{tabular}{lcc}
            \toprule
                    \textbf{\texttt{CelebA}} & \multicolumn{2}{c}{Class} \\ \cmidrule(lr){2-3} 
                    Group & Beneficial (Attractive) & Detrimental        \\ \midrule
                    Majoritized (Not Male) & 0.396 & 0.184 \\ 
                    Minoritized &            0.117 & 0.302 \\ \bottomrule 
            \end{tabular}
        \end{subtable}
        \caption{Class and group demographics of training datasets. Values show the proportion of that group--class category in the training dataset (and therefore sum to 1). \textit{Top:} \texttt{ColoredMNIST} and \texttt{ColoredSVHN} the majoritized is group skewed towards the beneficial class with probability 0.7, and to the detrimental class with probability 0.3. \textit{Bottom:} \texttt{CelebA}.}
        \label{tab:datasets}
\end{table}

\begin{figure}
    \centering
    \begin{minipage}{0.40\textwidth}
        \centering
        \includegraphics[width=\textwidth]{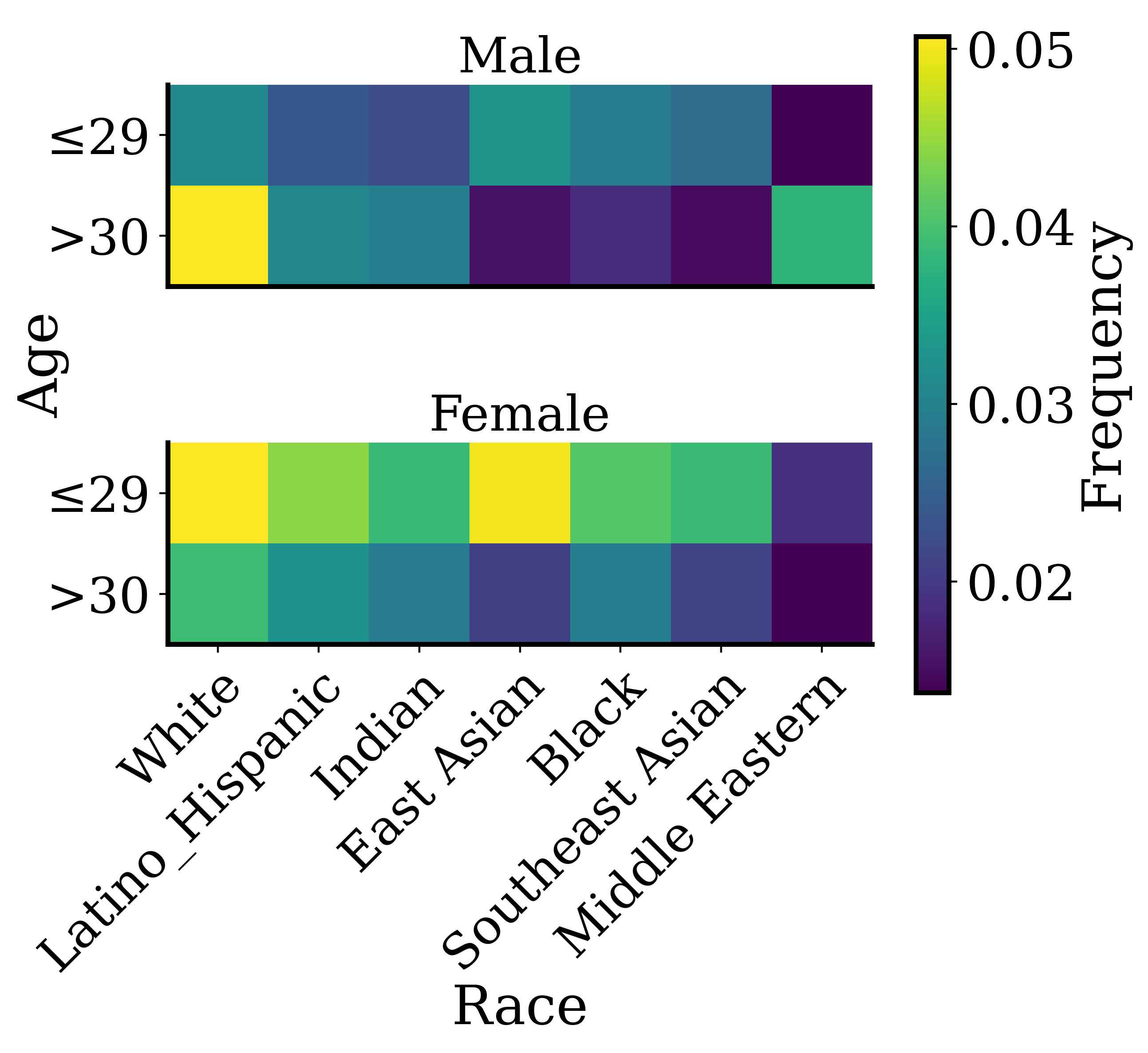} %
    \end{minipage}\hfill
    \begin{minipage}{0.55\textwidth}
        \centering
        \includegraphics[width=\textwidth]{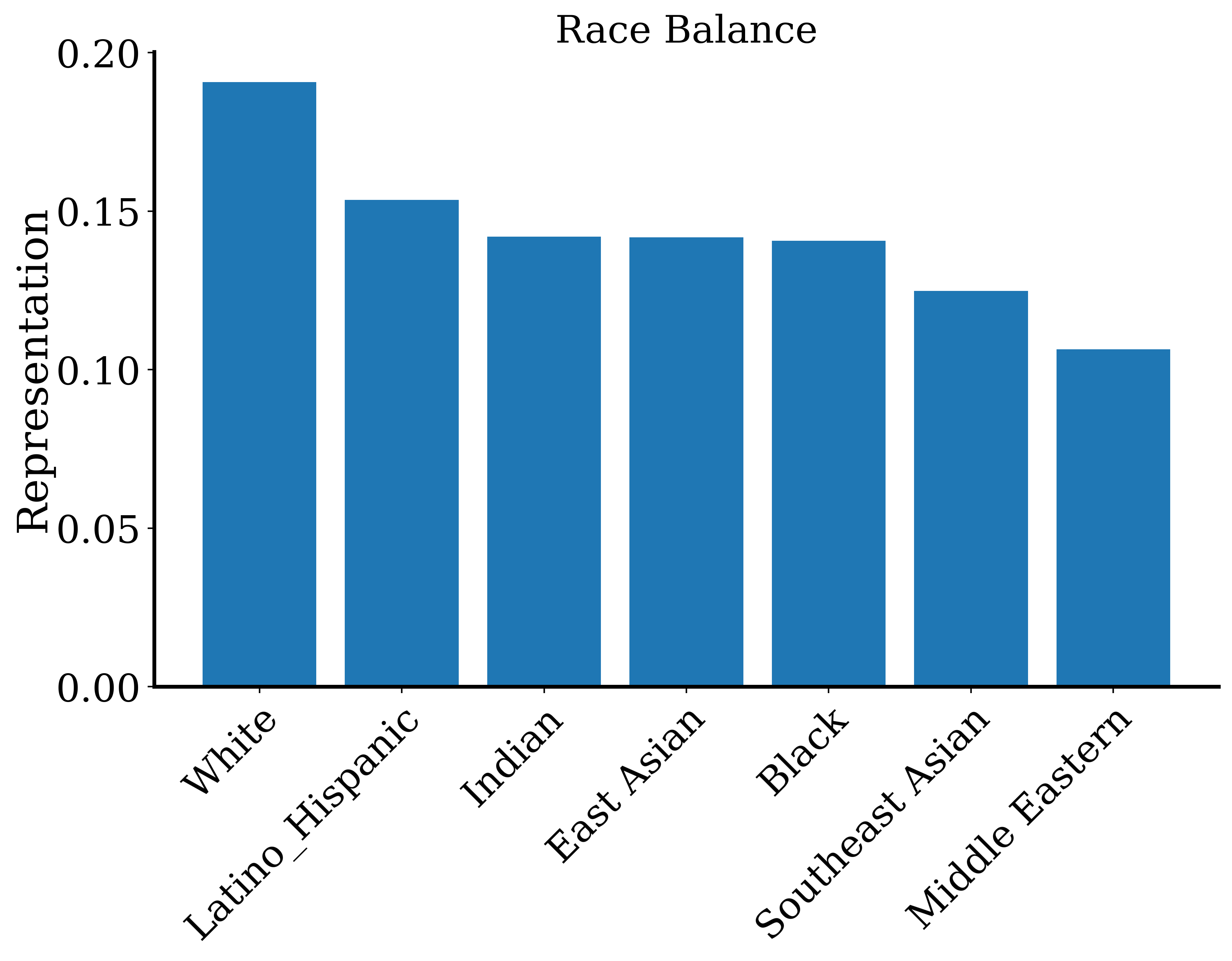} %
    \end{minipage}
    \begin{tabular}{lcc}
            \toprule
                     \multirow{2}{*}{Age}    & $\leq 29$ & $>30$ \\ \cmidrule(lr){2-3}
                         & 0.541  & 0.459 \\
                     \midrule 
                     \multirow{2}{*}{Gender} & Male      & Female \\ \cmidrule(lr){2-3}
                      & 0.531  & 0.469 \\
                    \bottomrule 
        \end{tabular}
    \caption{Class (gender) and group (race and binarized age) balance for the \texttt{FairFace} training set. }
    \label{fig:fairface}
\end{figure}

\paragraph{\texttt{ColoredMNIST}} \texttt{MNIST} is a single-channel handwritten digit recognition dataset~\citep{deng2012mnist}. We use 50000 images for training, 10000 for validation, and another 10000 for testing. We adapt \texttt{MNIST} to a binary classification scheme (determining if a digit is in [0..4] for class 0 or [5..9] for class 1). The class label is switched with a uniform probability of 5\% to add label noise, as in~\citet{arjovsky2020invariant}. We also adapt \texttt{MNIST} to have binary groups by coloring the sample either red or green, where green is treated as the `majoritized' group. We skew the majoritized group to the beneficial class, such that $\mathbb{P}(S=\textrm{majoritized}|L=\textrm{beneficial})=0.7$, $\mathbb{P}(S=\textrm{majoritized}|L=\textrm{detrimental})=0.3$. In this case, both classes and groups are balanced, as seen in the dataset composition matrix in Table~\ref{tab:datasets}. 
Ablations for this skew and class and label balances are in Appendix~\ref{app:ablation_studies}.

\paragraph{\texttt{ColoredSVHN}} \texttt{SVHN} (Street View House Numbers) is a digit recognition dataset composed of house numbers sourced from Google Street View~\citep{netzer2011svhn}. We use 52327 images for training, 20930 for validation, and another 26032 for testing. For \texttt{SVHN}, we adapt to a binary task similarly as in MNIST. We binarize the classification task to determining if a digit is in [0..4] for class 1 or [5..9] for class 0, where class 1 is the beneficial class as converged to by model collapse. The class label is swapped with a uniform probability of 5\% to add label noise, as in~\citet{arjovsky2020invariant}. Unlike in \texttt{ColoredMNIST}, class 1 is the lower numbers as \texttt{SVHN} converges to small numbers over the course of model collapse, as seen in Figure~\ref{fig:mc_images}. This causes class imbalance; class 1 composes 60.7\% of the data. We also add sensitive groups by converting the images to grayscale and then coloring the samples either red or green as in \texttt{ColoredMNIST}. The green group again serves as the majoritized group, and is skewed towards the beneficial class at rates $\mathbb{P}(S=\textrm{majoritized}|L=\textrm{beneficial})=0.7$, and $\mathbb{P}(S=\textrm{majoritized}|L=\textrm{detrimental})=0.3$, leading to group imbalance with the majoritized group as 54.3\%. A matrix showing the composition of the \texttt{ColoredSVHN} training distribution is shown in Table~\ref{tab:datasets}.

\paragraph{\texttt{CelebA}} \texttt{CelebA} is a dataset of celebrity faces~\citep{liu2015celeba}. We use an 80/10/10 train/validation/test split of the 202599 cropped and aligned images. The binary classification task is to predict attractiveness, where sensitive groups are given from gender (`Male,' `not Male'). The group and class balance is 58.\% `not male' and 51.3\% `attractive.' The composition of the dataset is shown in Table~\ref{tab:datasets}. This dataset is criticized for the inclusion of subjective features such as `attractive,' and there are many instances of incorrect labeling and annotation~\citep{Lingenfelter2022issues}. In the other datasets, we chose the beneficial class as the class most converged to by model collapse. Interestingly, against the class imbalance, model collapse converges to `unattractive,' which would benefit the `Male' group more. However, model collapse also converges to `Not Male.' We therefore use `attractive' as the beneficial class and `Not Male' as the majoritized group.

\paragraph{\texttt{FairFace}} \texttt{FairFace} is a face dataset that is balanced by both gender (two values) and race (seven values)~\citep{karkkainen2021fairface}. The composition, including the intersections of age, race, and gender, are shown in \Cref{fig:fairface}. We use an 80/10/10 train/validation/test split of the  108501 images. For binary classification, we predict gender (`Male,' `not Male'). For sensitive features, we use $S_1$ as race, which has seven potential values (`White', `Latino Hispanic', `Indian', `East Asian', `Black', `Southeast Asian', `Middle Eastern'), and $S_2$ as age, which we binarize as above and below 30 years (see \Cref{fig:fairface} for the class and group balances). For the group annotation oracles, we have two separate classifiers corresponding to these features. The beneficial class is `not Male,' as model collapse increases the representation of this class, even through they are a minority in the original dataset class balance. As there are 14 categories created from the intersection of age and race, we instead track the representations of both of these attributes alone and the largest and smallest categories among these over time. Note that these choices for the beneficial and majoritized annotations are arbitrary for these experiments as we do not make claims of justice (recall \Cref{ssec:methods_ar}). 

\paragraph{Justification of dataset choice. }We choose \texttt{ColoredMNIST} and \texttt{ColoredSVHN} due to their similarity. Both detect and classify digits, and may be easily adapted into a binary classification and binary fairness grouping task. They differ in their complexity, \texttt{SVHN} is usually the harder dataset to learn, and also in their class balance once binarized (see \Cref{tab:datasets}). These two datasets, despite their similarities, show very different points of model collapse and even opposite behavior when observing the accuracy difference between groups, as in \Cref{fig:celeba_mc} and \Cref{fig:FairFace_mc}. \texttt{CelebA} is chosen to represent a more complex and real-world dataset with well-documented disparities between the class and grouping we use for its task~\citep{Lingenfelter2022issues}. On the other hand, \texttt{FairFace} is chosen for its balance in both race and gender; alongside other metrics, it is simple to measure sensitive group intersections.

\subsection{Compute and codebase}\label{ssec:compute}
Experiments were performed on Ubuntu 18.04.6 LTS using 4 Intel Xeon CPU cores per GPU. We use the following GPUs per random seed: for \texttt{ColoredMNIST} we use 1 NVIDIA T4 GPU with 16 gigabytes of memory; for \texttt{ColoredSVHN} we use 2 NVIDIA RTX6000s with 40 gigabytes each; for \texttt{CelebA} we use 2 NVIDIA A100s with 40 gigabytes each (split GPUs); for \texttt{FairFace} we use 2 NVIDIA A40s with 48 gigabytes each. 
Our codebase is in Python 3.9 with PyTorch and fairness metrics from Fairlearn~\citep{bird2020fairlearn}. Existing code was adapted for our experiments: VAE models from~\citet{Subramanian2020vae}, \texttt{ColoredMNIST} from~\citet{arjovsky2020invariant}, SVHN with fairness from~\citet{kenfack2021fairness}, and ResNets from~\citet{Idelbayev18aresnet}.

\subsection{Models}\label{ssec:models}
The architectures and hyperparameters differ based on dataset. The performance of the annotator models $A_L$ and $A_S$ are shown in Table~\ref{tab:annotator_perfs}. All generative models were trained with ADAM (weight decal $1\times 10^{-5}$), and all classifiers with SGD and cross-entropy loss.

\textbf{\texttt{ColoredMNIST}}
\begin{itemize}
    \item Classifiers and annotators are 6 layer CNNs with 2 convolution layers and ReLU activations. Learning rate 0.1, batch size 256, for 30 epochs. 
    \item Generators (VAEs) are mirrored encoder and decoder CNNs. Each is 2 convolution layers with ReLU activations. Uses BCE Loss with KL-divergence term, a latent space dimension of 20, and a variational beta of 1. Learning rate 0.001, batch size 256, for 30 epochs.
\end{itemize}

\textbf{\texttt{ColoredSVHN}}
\begin{itemize}
    \item Classifiers and annotators are 32-layer ResNets adapted from~\citet{Idelbayev18aresnet}. Learning rate 0.001, batch size 32, for 30 epochs. 
    \item Generators are the deep convolutional VAE adapted from~\citet{sujitVAE}, using MSE loss with KL-divergence term, a latent space dimension of 32, and a variational beta of 1. Learning rate 0.0005, batch size 128, for 30 epochs. 
\end{itemize}
    
\textbf{\texttt{CelebA}}
\begin{itemize}
    \item Classifiers and annotators are 110-layer ResNets adapted from~\cite{Idelbayev18aresnet}. Learning rate 0.001, batch size 128, for 15 epochs. 
    \item Generator VAEs are composed of a 5-layer CNN encoder and 6-layer upsampling CNN decoder with LeakyReLU activations. Loss is BCE with KL-divergence term, a latent space dimension of 500, with a variational beta of $5\times 10^{-6}$. Learning rate 0.005, batch size 64, for 30 epochs.  
\end{itemize}

\textbf{\texttt{FairFace}}
\begin{itemize}
    \item Classifiers and annotators are pretrained 50-layer ResNets adapted from~\cite{Idelbayev18aresnet}. Pretraining is on Imagenet, using the version 1 weights from PyTorch. Learning rate 0.001, batch size 256, for 30 epochs. 
    \item Generator VAEs are composed of a 5-layer CNN encoder and 6-layer upsampling CNN decoder with LeakyReLU activations. Loss is MSE with KL-divergence term, a latent space dimension of 500, with a variational beta of $1\times 10^{-6}$. Learning rate 0.0001, batch size 256, for 50 epochs.  
\end{itemize}

\begin{table}[ht]
\centering
\caption{Performances (with standard deviations) of the label and sensitive attribute annotator models $A_L$ and $A_S$. Performances are shown for each dataset. The performance of $A_S$ is high for \texttt{ColoredMNIST} and \texttt{ColoredSVHN} as determining sample color is an easy task. The fairness metrics for $A_S$ should be close to 1, as these models should assign class based on the sensitive attribute alone. Reported accuracies are all macro-averaged. Note the high accuracy disparity for \texttt{FairFace} $A_{S_1}$, although racial groups are roughly balanced, we observed far higher accuracy for `white' than any other group, perhaps due to simplicity bias~\citep{BellSimplicityBias}.}
\begin{tabular}{@{}lllllll@{}}
\toprule
                  & \multicolumn{2}{c}{\texttt{ColoredMNIST}} & \multicolumn{2}{c}{\texttt{ColoredSVHN}} & \multicolumn{2}{c}{\texttt{CelebA}}    \\ \cmidrule(lr){2-3}\cmidrule(lr){4-5}\cmidrule(lr){6-7}
                  & \multicolumn{1}{c}{$A_L$} & \multicolumn{1}{c}{$A_S$} & \multicolumn{1}{c}{$A_L$} & \multicolumn{1}{c}{$A_S$} & \multicolumn{1}{c}{$A_L$} & \multicolumn{1}{c}{$A_S$} \\ \midrule
Accuracy          & 0.928 $\pm$ 0.003 & 1 $\pm$ 0     & 0.849 $\pm$ 0.080 & 1 $\pm$ 0     & 0.816 $\pm$ 0.005 & 0.976 $\pm$ 0.002 \\
$\Delta$ Accuracy & 0.009 $\pm$ 0.005 & 0 $\pm$ 0     & 0.052 $\pm$ 0.111 & 0 $\pm$ 0     & 0.029 $\pm$ 0.008 & 0.015 $\pm$ 0.008 \\
$\Delta$ DP       & 0.367 $\pm$ 0.008 & 1 $\pm$ 0     & 0.151 $\pm$ 0.193 & 1 $\pm$ 0     & 0.440 $\pm$ 0.022 & 0.951 $\pm$ 0.004 \\
$\Delta$ EOdds    & 0.032 $\pm$ 0.022 & 1 $\pm$ 0     & 0.163 $\pm$ 0.240 & 1 $\pm$ 0     & 0.271 $\pm$ 0.037 & 0.971 $\pm$ 0.008 \\ \bottomrule
\end{tabular}
\label{tab:annotator_perfs}
\end{table}

\begin{table}[ht]
\centering
\begin{tabular}{@{}llll@{}}
\toprule
                  & \multicolumn{3}{c}{\texttt{FairFace}} \\ 
                  \cmidrule(lr){2-4}
                  & \multicolumn{1}{c}{$A_L$} & \multicolumn{1}{c}{$A_{S_1}$} & \multicolumn{1}{c}{$A_{S_2}$} \\ \midrule
Accuracy          & 0.884 $\pm$ 0.013 &     0.617 $\pm$ 0.010     & 0.789 $\pm$ 0.017  \\
$\Delta$ Accuracy & 0.104 $\pm$ 0.011 &     0.428 $\pm$ 0.088     & 0.251 $\pm$ 0.094  \\
$\Delta$ DP       & 0.360 $\pm$ 0.023 &     0.443 $\pm$ 0.057     & 0.683 $\pm$ 0.031  \\
$\Delta$ EOdds    & 0.308 $\pm$ 0.014 &     0.438 $\pm$ 0.139     & 0.254 $\pm$ 0.075  \\ \bottomrule
\end{tabular}
\label{tab:annotator_perfs_cont}
\end{table}

\subsection{\algname Algorithm}

\Cref{alg:AR_batches} shows an example of \algname for binary group and binary sensitive attributes.

\begin{algorithm2e}[ht!]
\algsetup{linenosize=\tiny}
\DontPrintSemicolon
\SetKwComment{Comment}{{\scriptsize$\triangleright$\ }}{}
\caption{Training with algorithmic reparation batches.}
        \KwIn{Sample-providing generator $G$, batch size $b$, reparation budget $r$, label annotator $C$ (either $C_{i-1}$ or $A_L$), sensitive attribute annotator $A_S$.} 
        \KwOut{Reparation batch}
\BlankLine
    \begin{algorithmic}[1]
    \FOR{{batch in number batches}}
        \STATE Ideal $=[b/4, b/4, b/4, b/4]$\Comment*[r]{\scriptsize Ideal category sizes}
        \STATE Batch $= [b_{L=0,S=0}, b_{L=0,S=1}, b_{L=1,S=0}, b_{L=1,S=1}]= [0,0,0,0]$ \Comment*[r]{{\scriptsize Initialize batch categories}} 
        \STATE Temporary batch $=$ Sample $b+r$ times from $G$ \Comment*[r]{\scriptsize Initial batch from uniform sampling}
        \STATE Annotate temporary batch using $C$ and $A_S$
        \STATE Categorize batch depending on $L$ and $S$ values from annotations
        \STATE Populate Batch until Ideal$_i =$Batch$_i$ 
        \STATE To\_resample $=$sum($\text{Ideal} - \text{Batch}$) \Comment*[r]{\scriptsize Get amount to sample to fill deficient categories}
        \STATE Batch.append(Sample To\_resample times from $G$, annotate with $C$ and $A_S$) \Comment*[r]{\scriptsize Refill batch}
    \ENDFOR
    \STATE Update model on Batch.
    \STATE \textbf{return} Batch
\end{algorithmic}
\label{alg:AR_batches}
\end{algorithm2e}

\section{Appendix: Ablation Studies}\label{app:ablation_studies}

In this appendix we provide experiments to demonstrate the effects of MIDS over several ablated variables: the sensitive group imbalance, class imbalance, and amount of synthetic training data. We provide results for both \texttt{ColoredMNIST} and \texttt{ColoredSVHN}. 

\subsection{Class and group imbalance}\label{ssec:class_group_imbal}
\begin{figure}
\centering
\begin{tabular}{cc}
\includegraphics[width=0.4\textwidth]{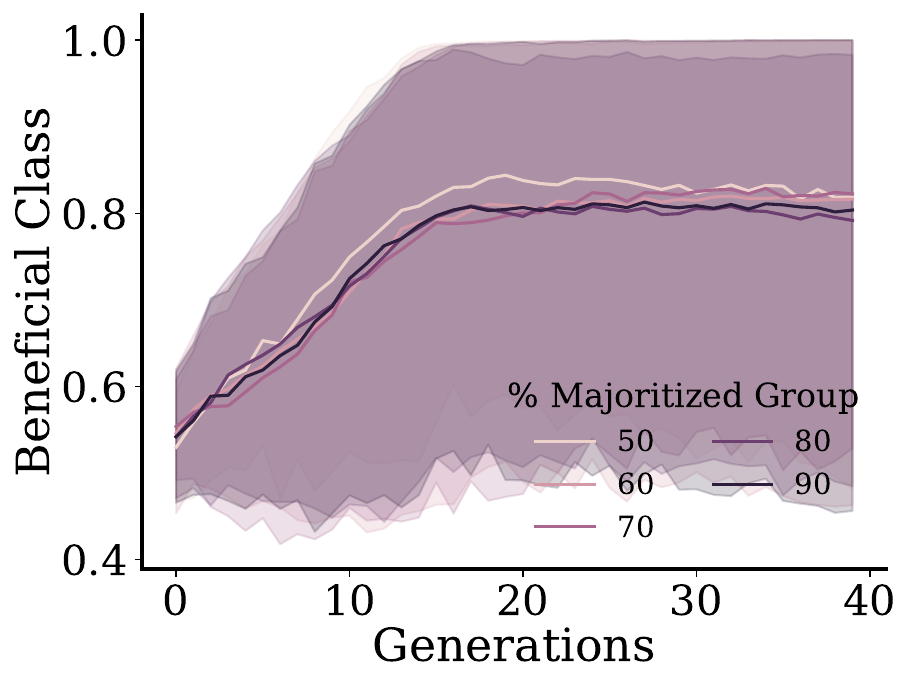} &
\includegraphics[width=0.4\textwidth]{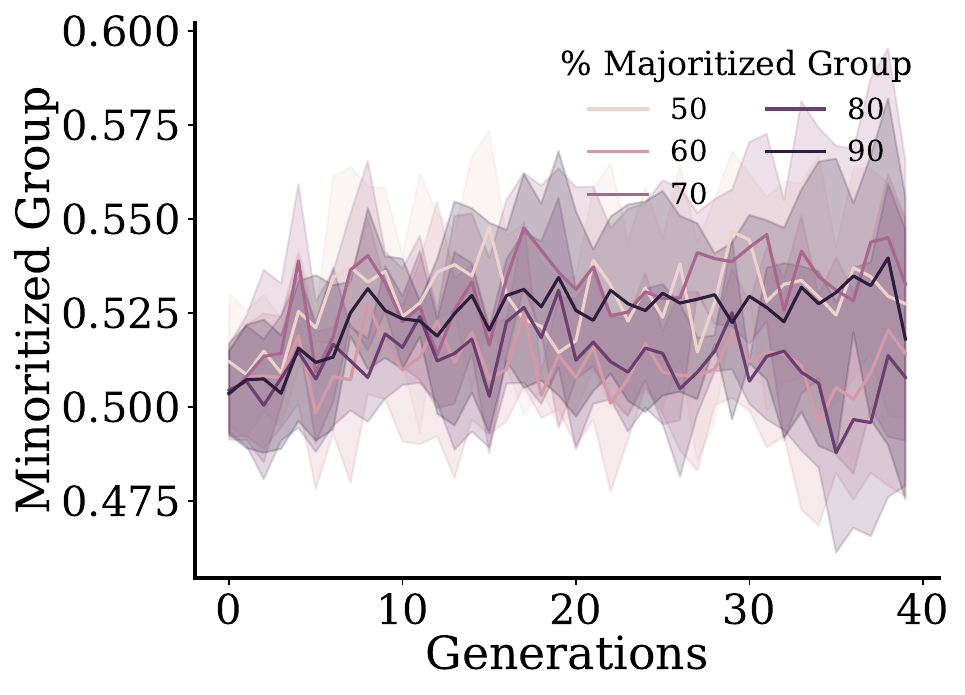}
\\
\end{tabular}
\caption{\texttt{ColoredMNIST} class and group balance while varying the group balance in \sgsc. }
\label{fig:abl_grp_bal}
\end{figure}

\begin{figure}
\centering
\begin{tabular}{cc}
\includegraphics[width=0.4\textwidth]{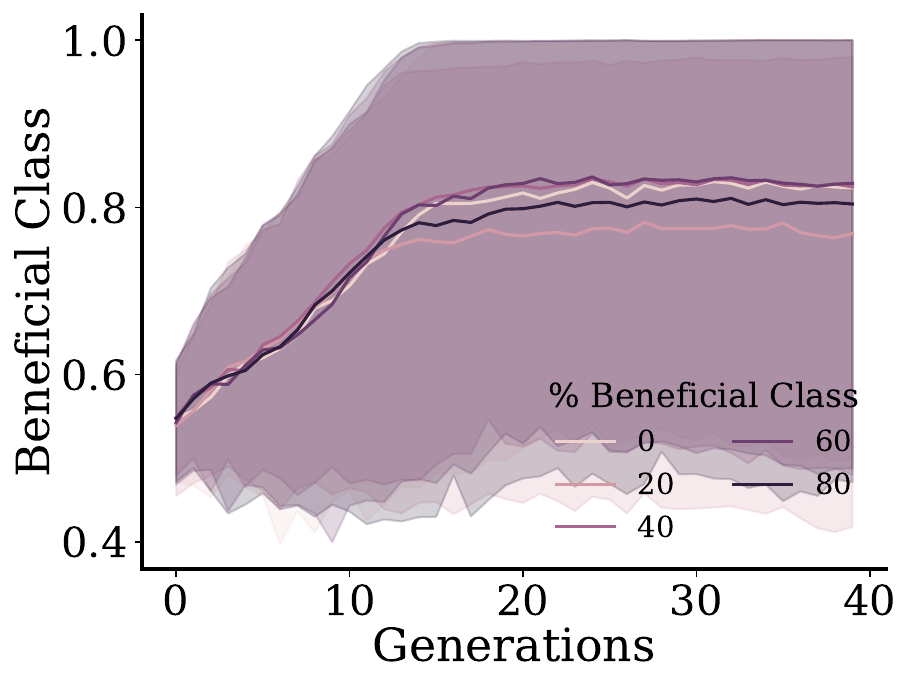} &
\includegraphics[width=0.4\textwidth]{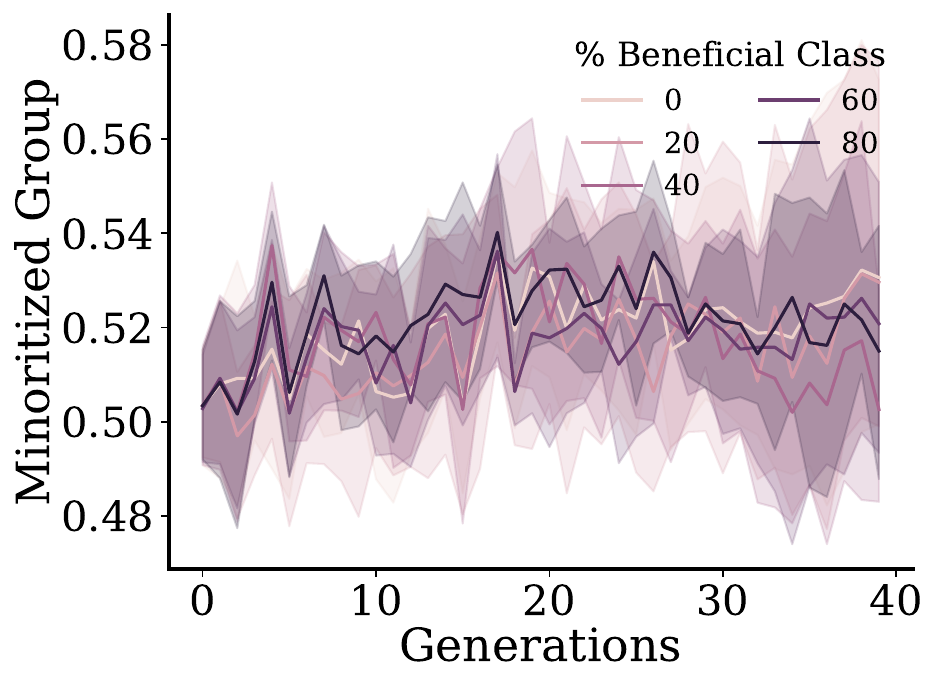}
\\
\end{tabular}
\caption{\texttt{ColoredMNIST} class and group balance while varying the class balance in \sgsc. }
\label{fig:abl_lab_bal}
\end{figure}

In these studies we varied the class balance or group balance. The study was carried out on \texttt{ColoredMNIST} with 5 seeds in the sequential generator and classifier setting. For group imbalance, the groups were equally likely to belong to the beneficial class, though their populations were varied. For class imbalance, the majoritized group was skewed towards the beneficial class in the same manner as discussed in \Cref{ssec:datasets}, and the class population varied. For this task, the variations in balance did not strongly effect the generated population or downstream classifier performance. The generator class and group balances are shown for varied group balance in \Cref{fig:abl_grp_bal} and for varied class balance in \Cref{fig:abl_lab_bal}. The results in \Cref{ssec:results_gen} use datasets with a mixture of class and group imbalance which better elucidate the effects of MIDS. 

\subsection{Amount of synthetic data}\label{ssec:synthetic}
\begin{figure}
\centering
\begin{tabular}{cc}
\includegraphics[width=0.4\textwidth]{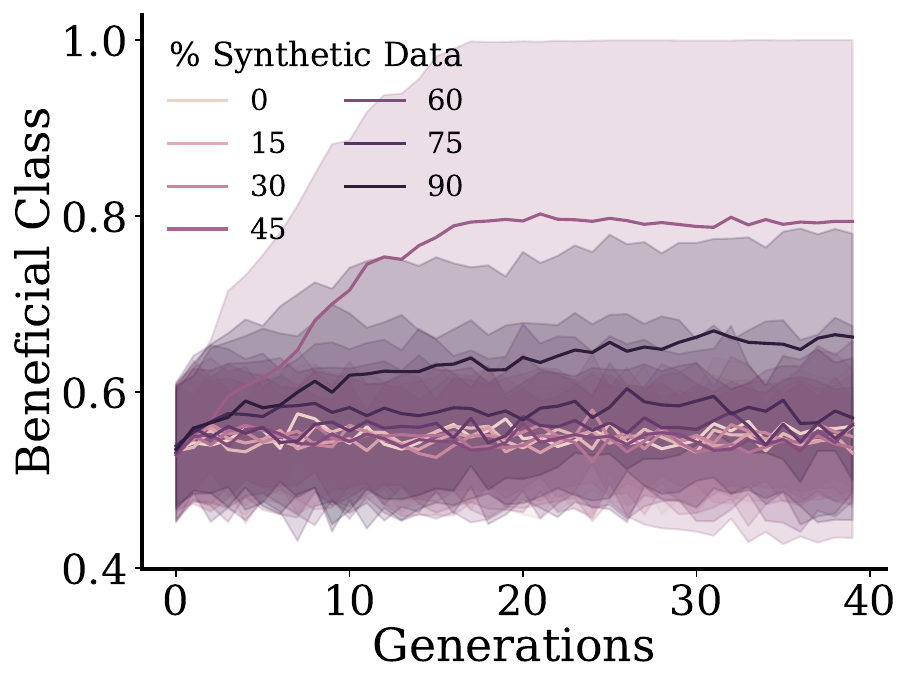} &
\includegraphics[width=0.4\textwidth]{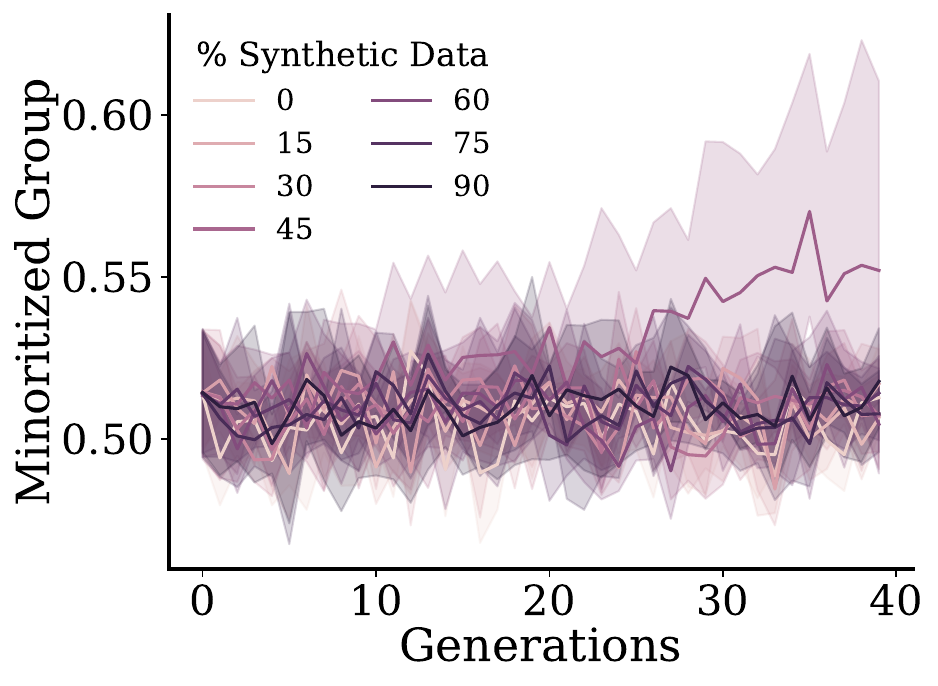}
\\
\includegraphics[width=0.4\textwidth]{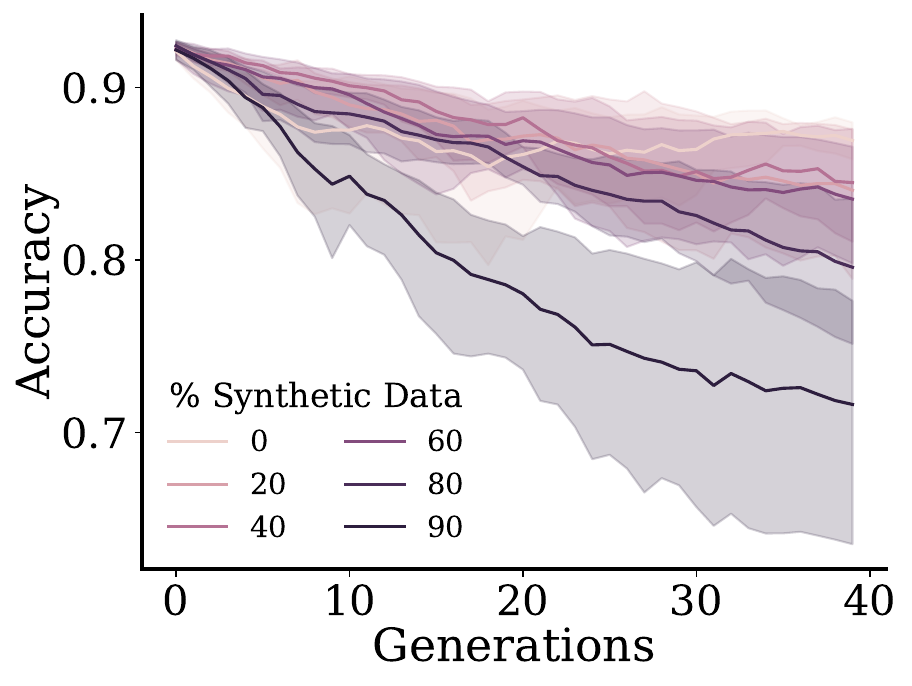} &
\includegraphics[width=0.4\textwidth]{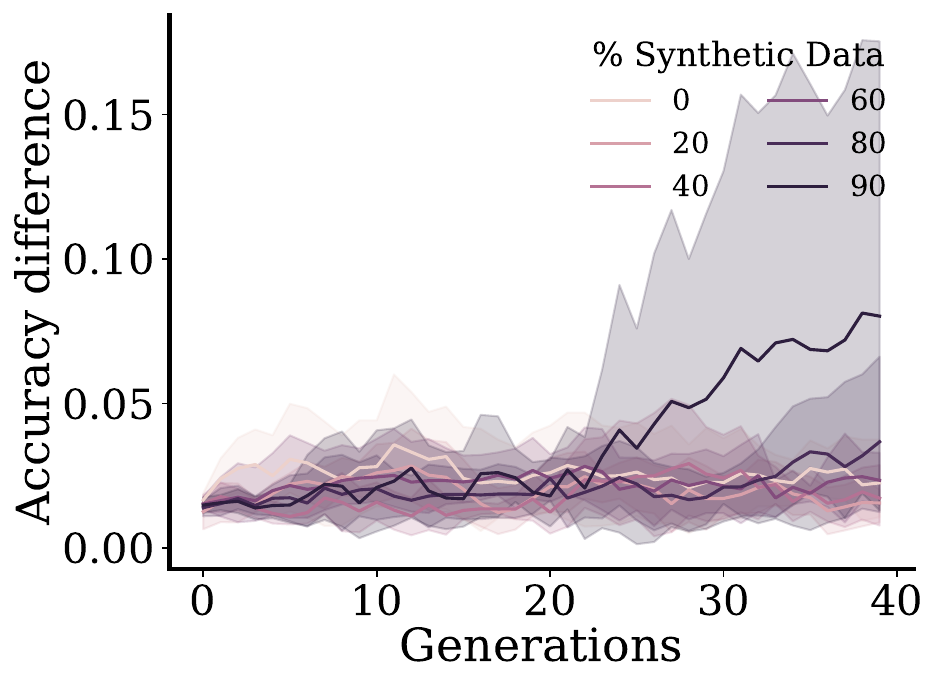}
\\
\includegraphics[width=0.4\textwidth]{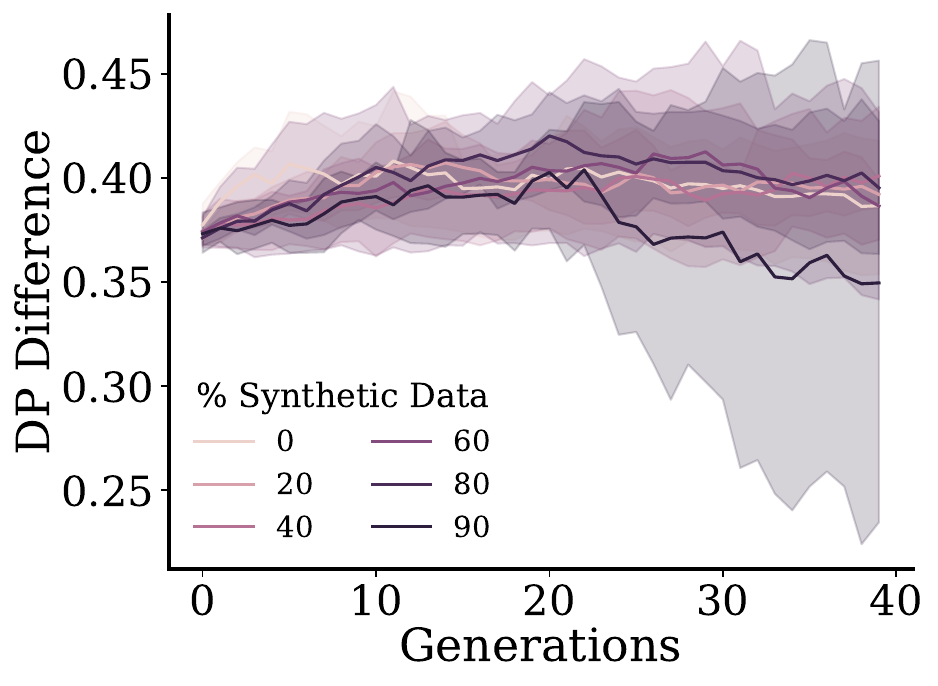} &
\includegraphics[width=0.4\textwidth]{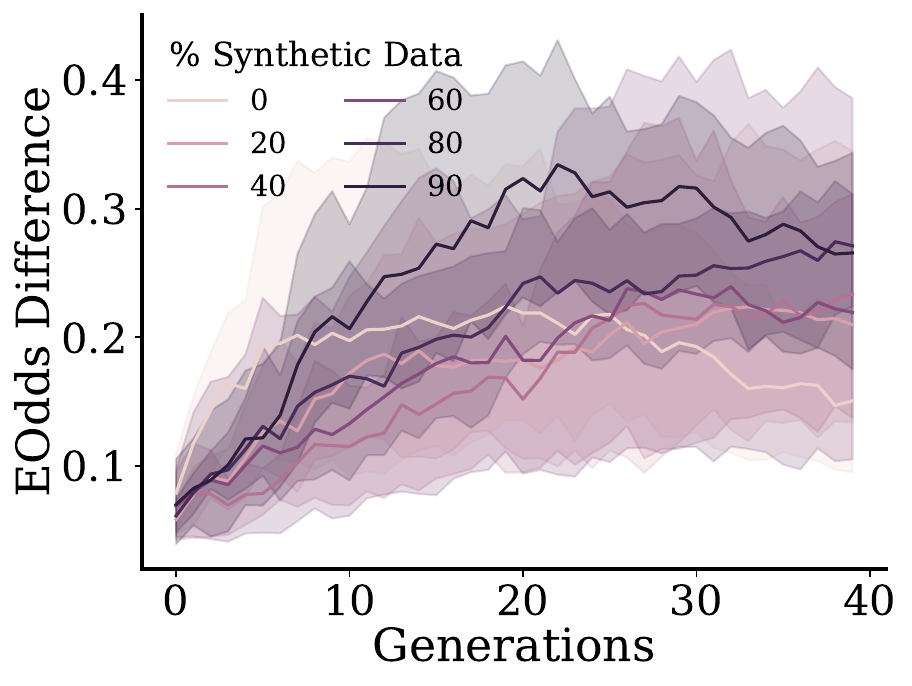}
\\
\end{tabular}
\caption{\texttt{ColoredMNIST} metrics while varying the amount of synthetic data in \sgsc. \textit{Top:} beneficial class balance and group balance. \textit{Center:} accuracy, and accuracy difference between groups. \textit{Bottom:} DP and EOdds difference. These generally show worse performance and fairness with more synthetic data, with larger variances. }
\label{fig:abl_syn_accs}
\end{figure}

In this study we varied the amount of original training data (drawn randomly from the training set) in each batch for training generators in the sequential generator and classifier setting. These experiments were carried out on \texttt{ColoredMNIST} for 5 seeds. There is a substantially higher accuracy cost and accuracy disparity between groups, as shown in \Cref{fig:abl_syn_accs}. Note that even with 0\% synthetic data (i.e., training each generator from the original training set) there is still an accuracy loss over time due to the effects of the sequential classifiers. While the group balance is not hugely effected (as in the other \texttt{ColoredMNIST} results in \Cref{ssec:results}), the class balance skews towards the beneficial class over the generations, fueling an increase of equalized odds difference with more synthetic data, see \Cref{fig:abl_syn_accs}.  

In practice, there may be several generations of synthetic data present when drawing from a corpus of polluted data. For example, when training $G_2$, samples from $G_0$ and $G_1$ might also be present. In this case, the compounded artefacts of model collapse will be lesser in these early generations. In this study, the synthetic data is only pooled from the most recent generator, and so these results may overstate the effect of model collapse in the aforementioned case.

\subsection{Sequential versus non-sequential classifiers in \sgsc and \sgnsc}\label{ssec:seq_v_noseq}
In this study we demonstrate the impact of sequential classifiers in \sgsc. These experiments were conducted for \texttt{ColoredMNIST} and \texttt{ColoredSVHN} for 25 and 10 seeds, respectively.

\begin{figure}
\centering
\begin{tabular}{cc}
\includegraphics[width=0.3\textwidth]{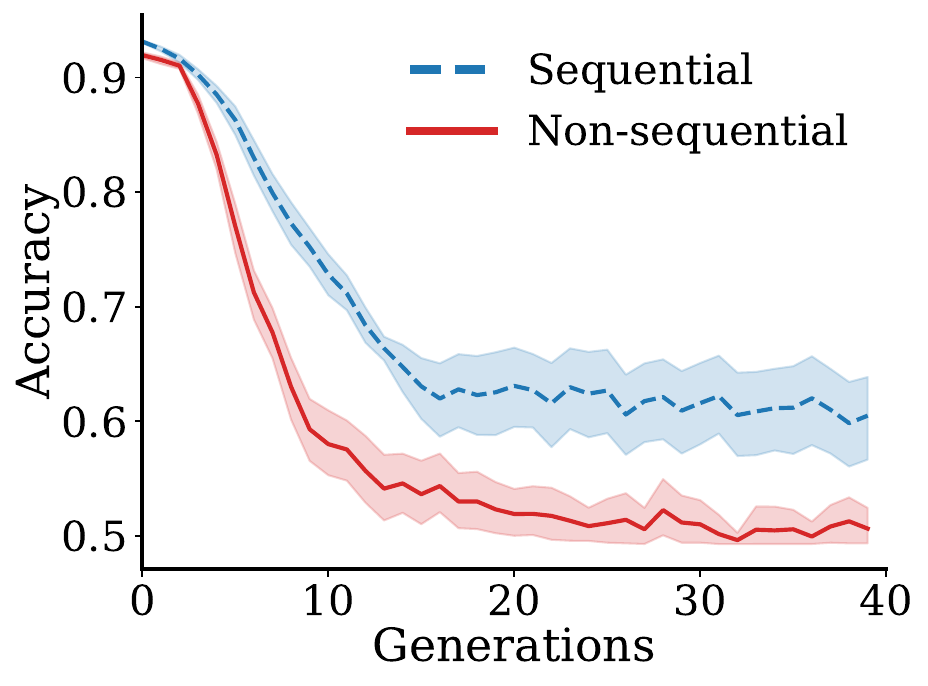} &
\includegraphics[width=0.3\textwidth]{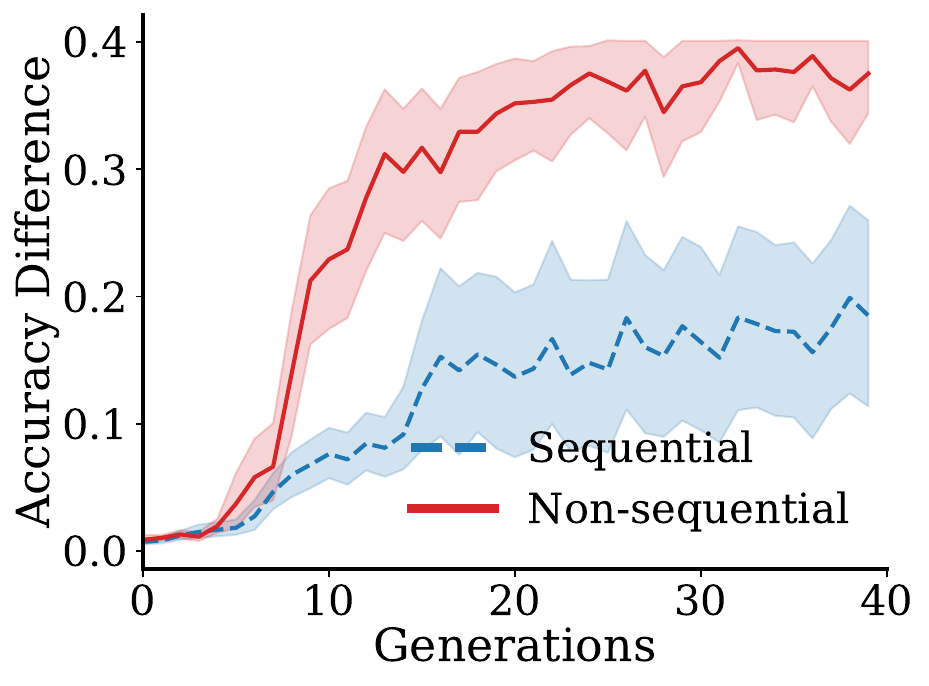}
\\
\end{tabular}
\begin{tabular}{ccc}
\includegraphics[width=0.3\textwidth]{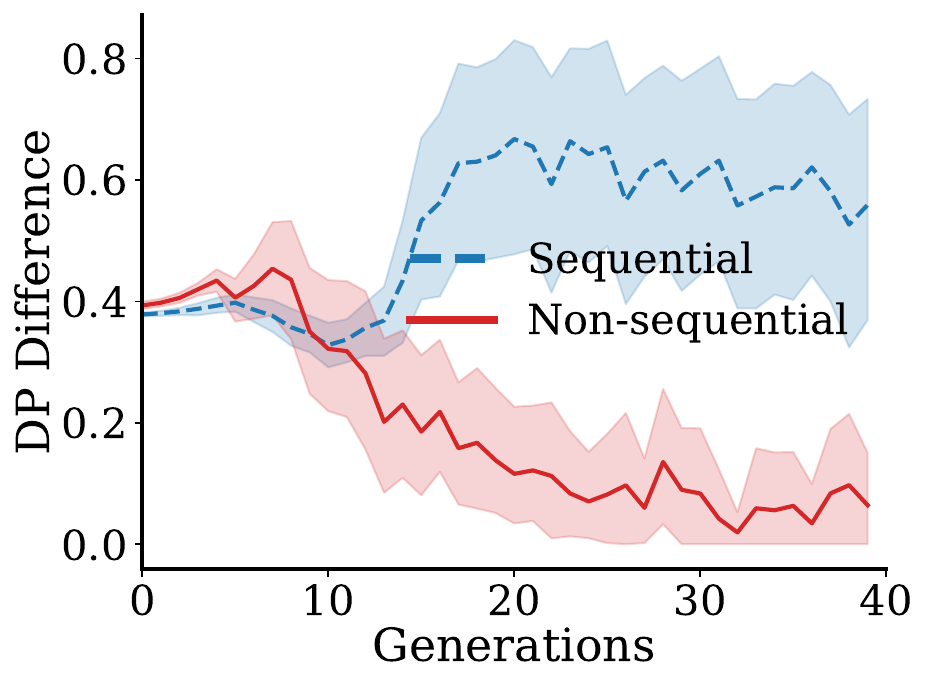} & 
\includegraphics[width=0.3\textwidth]{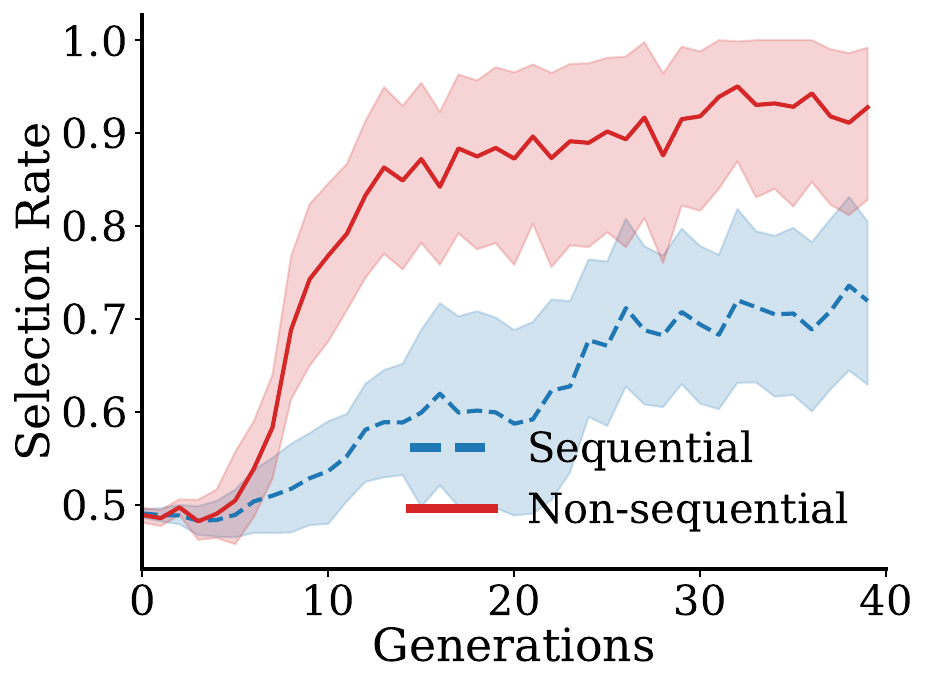} &
\includegraphics[width=0.3\textwidth]{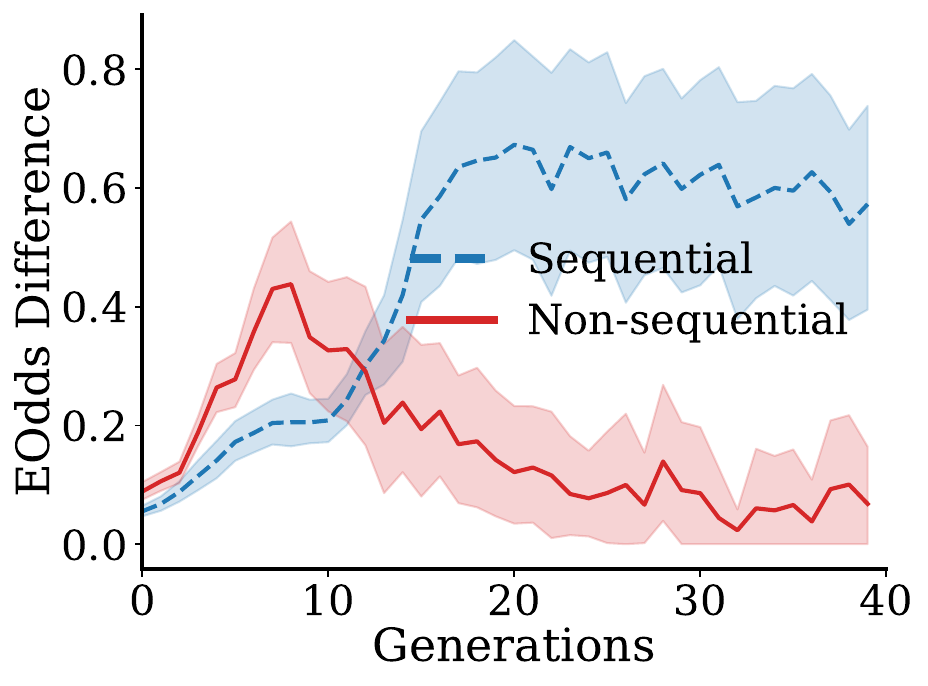}
\\
\end{tabular}
\caption{\texttt{ColoredMNIST} results for sequential versus non-sequential classifiers in \sgsc and \sgnsc. \textit{Top:} accuracy and accuracy difference between groups. \textit{Bottom:} demographic parity difference, selection rate, and equalized odds difference. }
\label{fig:abl_noseq_cmnist}
\end{figure}

\begin{figure}
\centering
\begin{tabular}{cc}
\includegraphics[width=0.3\textwidth]{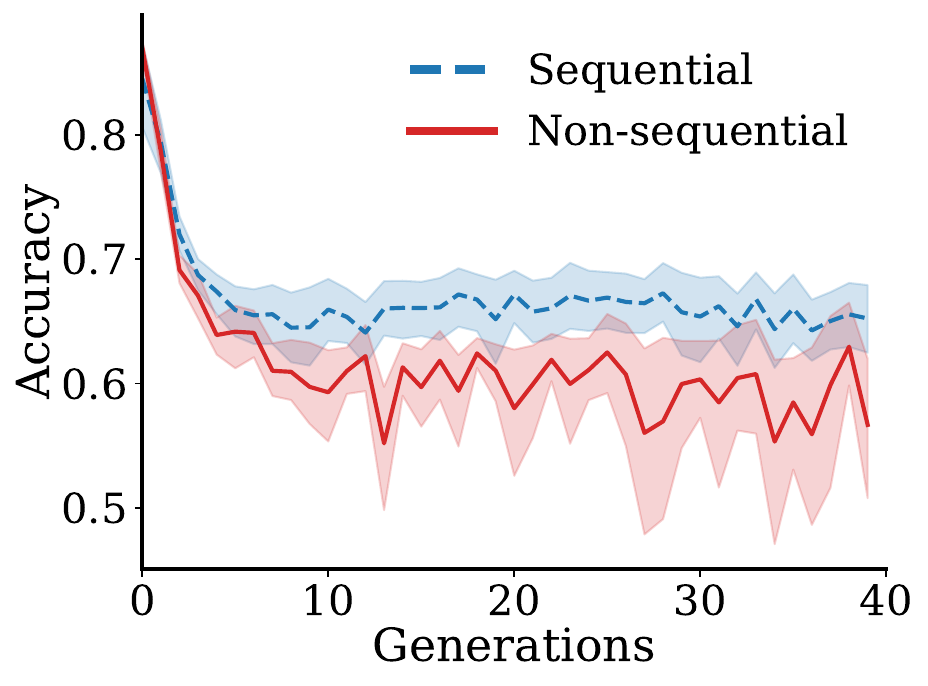} &
\includegraphics[width=0.3\textwidth]{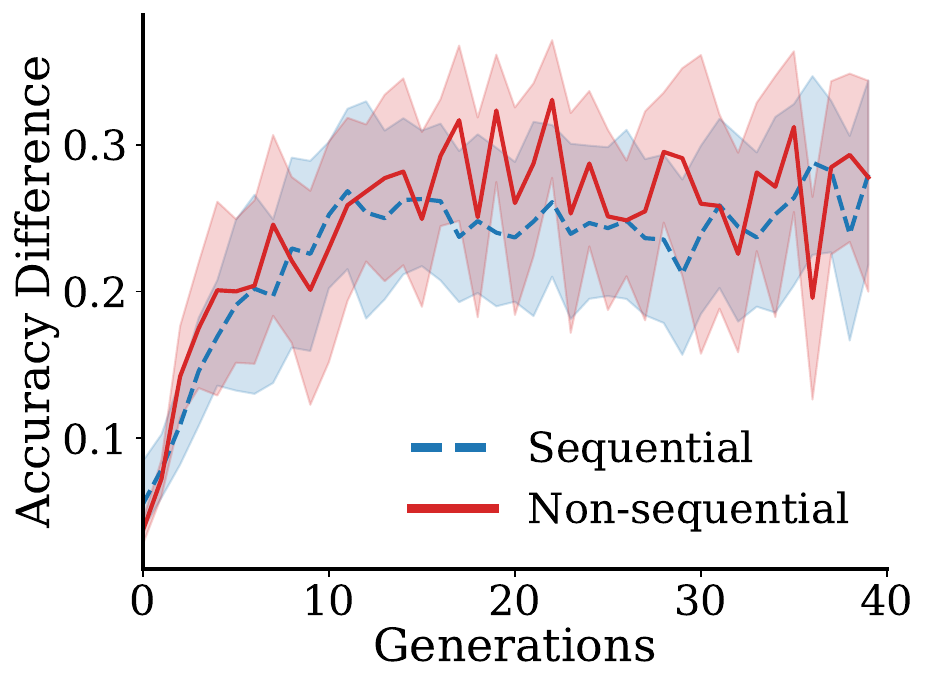}
\\
\end{tabular}
\begin{tabular}{ccc}
\includegraphics[width=0.3\textwidth]{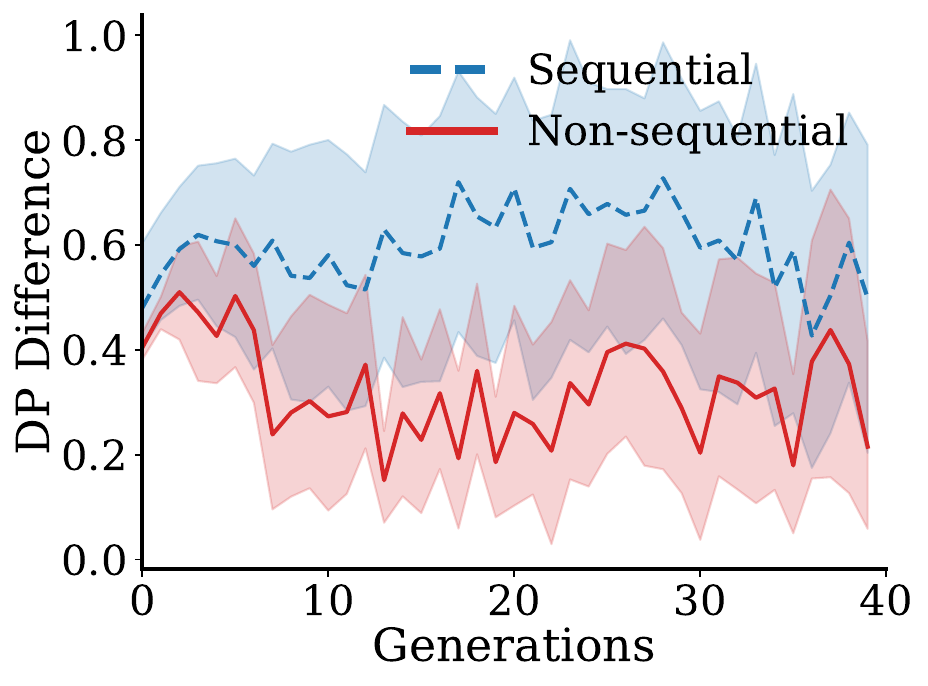} & 
\includegraphics[width=0.3\textwidth]{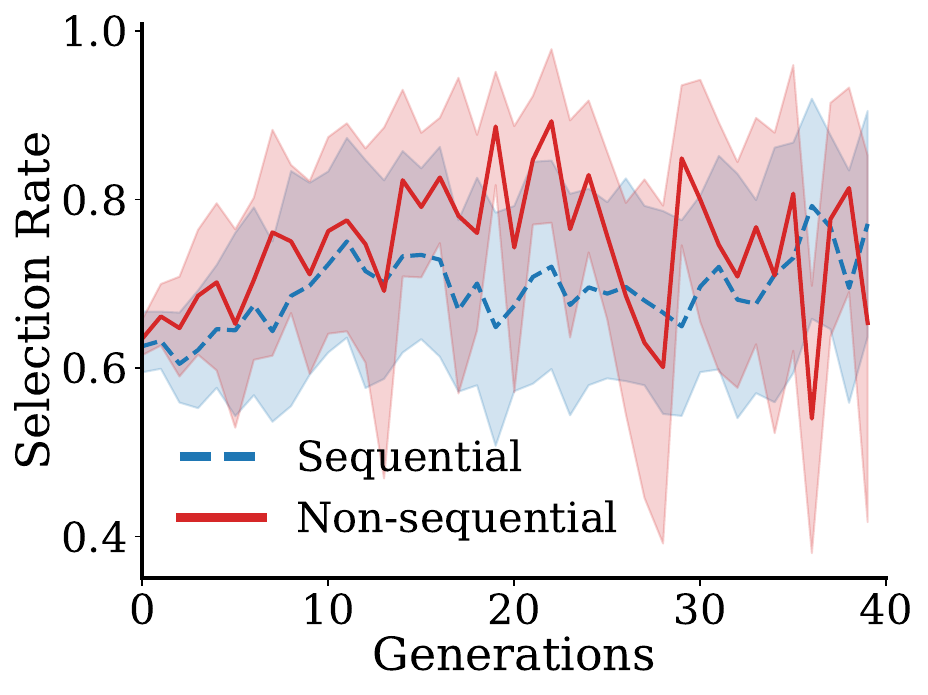} &
\includegraphics[width=0.3\textwidth]{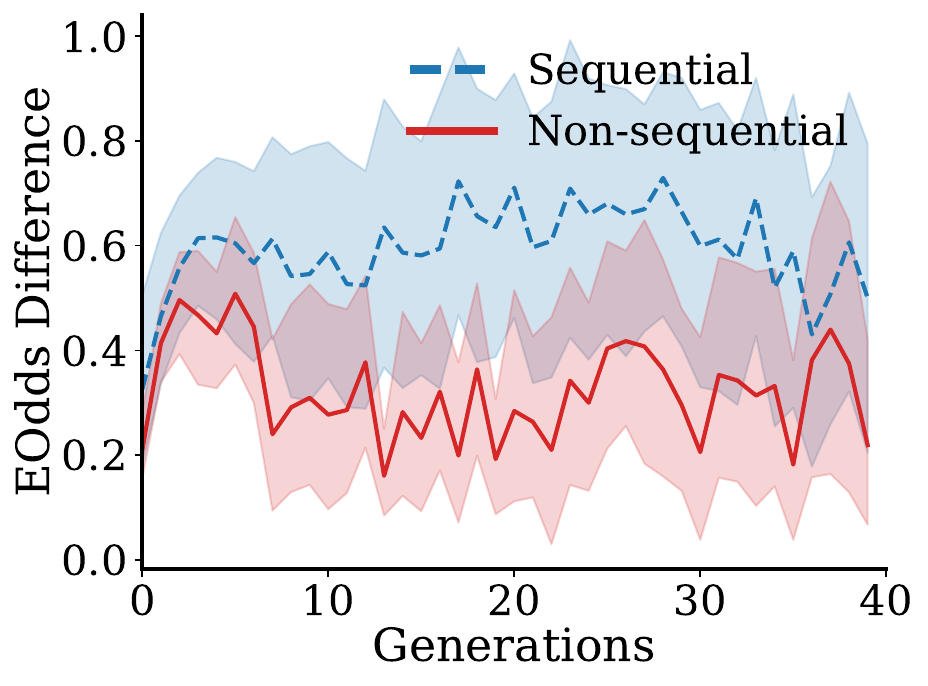}
\\
\end{tabular}
\caption{\texttt{ColoredSVHN} results for sequential versus non-sequential classifiers in \sgsc and \sgnsc. \textit{Top:} accuracy and accuracy difference between groups. \textit{Bottom:} demographic parity difference, selection rate, and equalized odds difference. }
\label{fig:abl_noseq_svhn}
\end{figure}

The non-sequential classifiers are more sensitive to changes in the distribution, as seen in the accuracy over generations and selection rate graphs in \Cref{fig:abl_noseq_cmnist} and \Cref{fig:abl_noseq_svhn} for \texttt{ColoredMNIST} and \texttt{ColoredSVHN}, respectively. 
This is consistent with the intuition that the sequential classifiers are `adapting' their mapping of $\mathcal{X}\rightarrow\mathcal{L}$ to changes in $\mathcal{X}$ caused by model collapse. 
This allows the sequential classifiers to experience more generations of utility compared to the non-sequential classifiers, as seen in their higher accuracies. 
As model collapse causes strong imbalance towards the beneficial class (as determined by $A_L$), the non-sequential classifiers eventually only predict the positive label, decreasing DP and EOdds unfairness as both groups receive the same predictions and error rates (in \texttt{ColoredMNIST}, EOdds difference drops to 0 as error rates from only giving the positive prediction are identical due to group and class balance). Meanwhile, the sequential classifiers for \texttt{ColoredMNIST} and\texttt{ColoredSVHN} instead evolve to only give a beneficial prediction to a majoritized sample, increasing unfairnesses (see \Cref{ssec:results_gen}).

Note that the achievement of higher fairness in the non-sequential classifier case indicates higher fairness with respect to the original distribution. This may be undesirable in some cases, particularly those applicable to algorithmic reparation, which specifically notes that equality of of model outputs to base rates in a dataset does not guarantee equity. This is especially true if the dataset is collected with any biases, including compounding Intersectional biases which these experiments do not inform upon~\cite{Davis2021reparation}.

\section{Model Collapse in Generators}\label{app:gen_loss}
\begin{figure}
\centering
\begin{tabular}{cc}
    \includegraphics[width=0.4\textwidth]{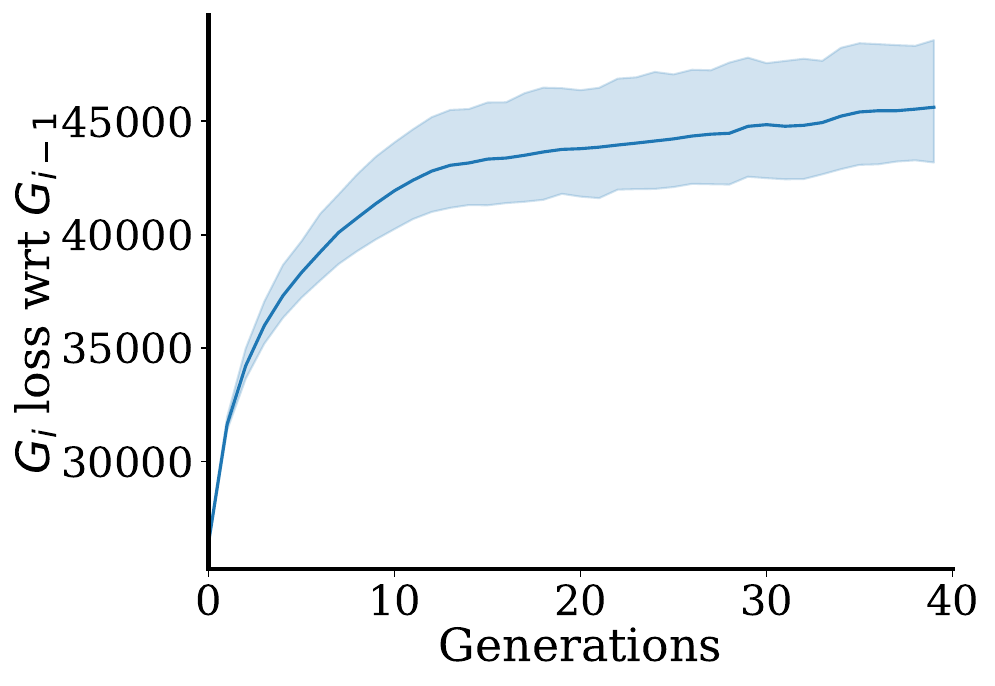} &
    \includegraphics[width=0.4\textwidth]{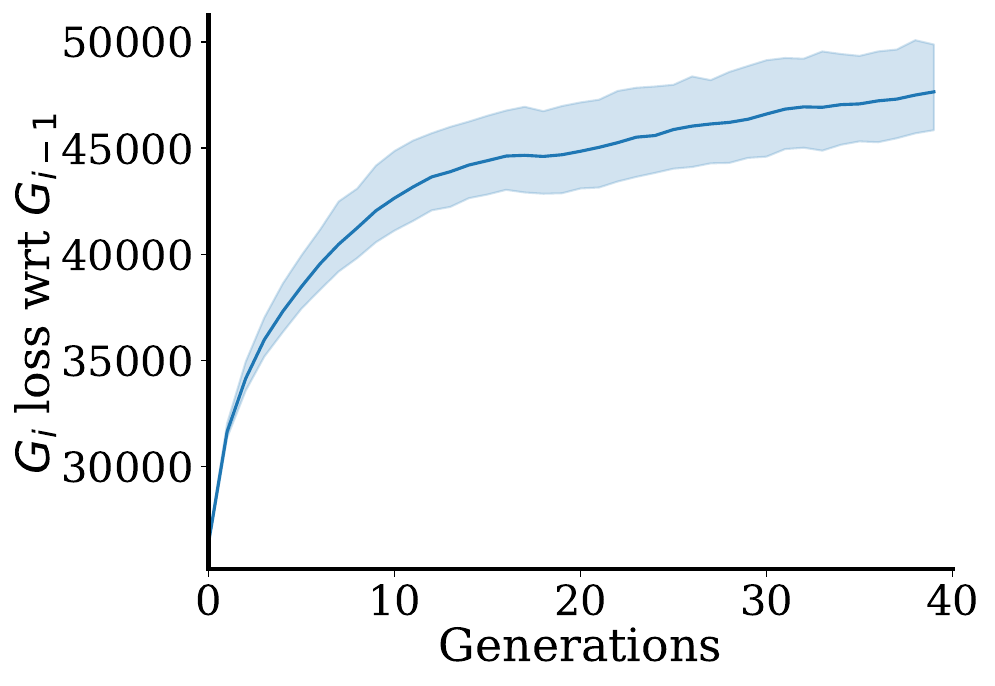}
    \\
    \includegraphics[width=0.4\textwidth]{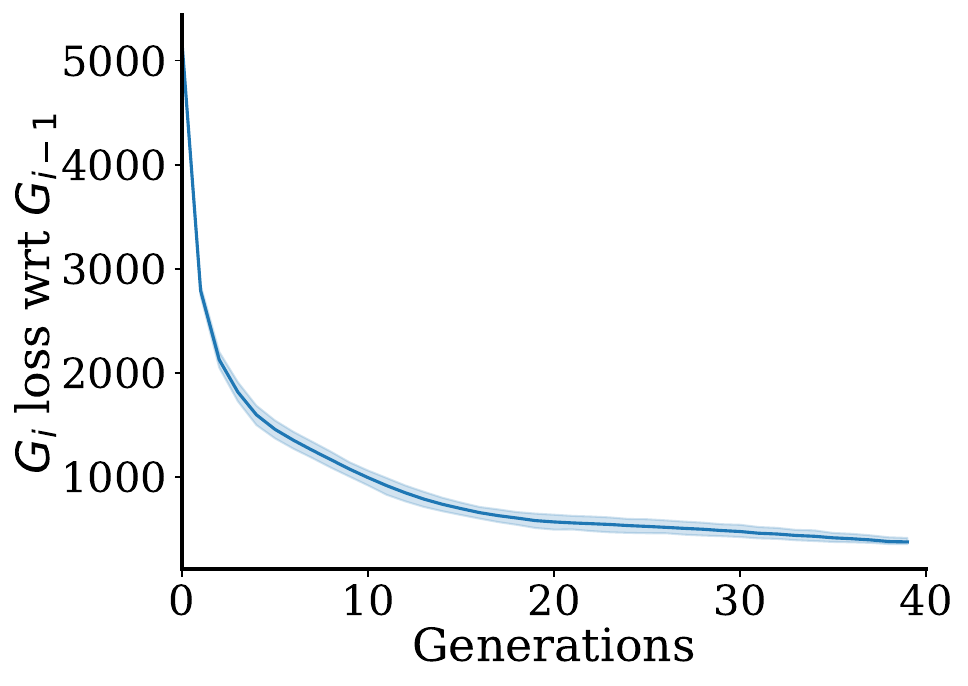} &
    \includegraphics[width=0.4\textwidth]{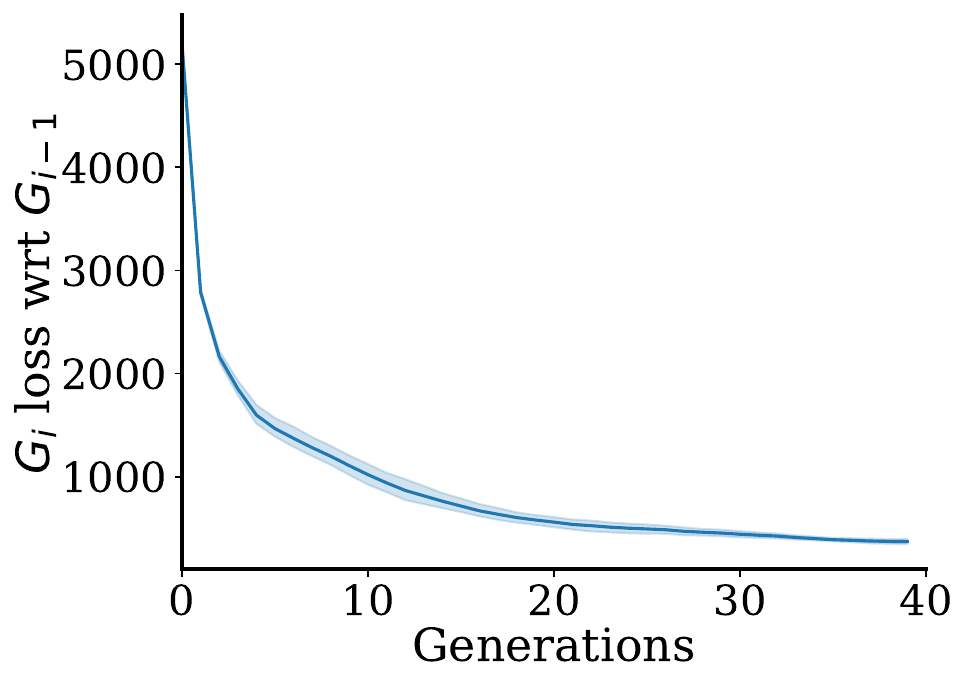}
    \\
    \includegraphics[width=0.4\textwidth]{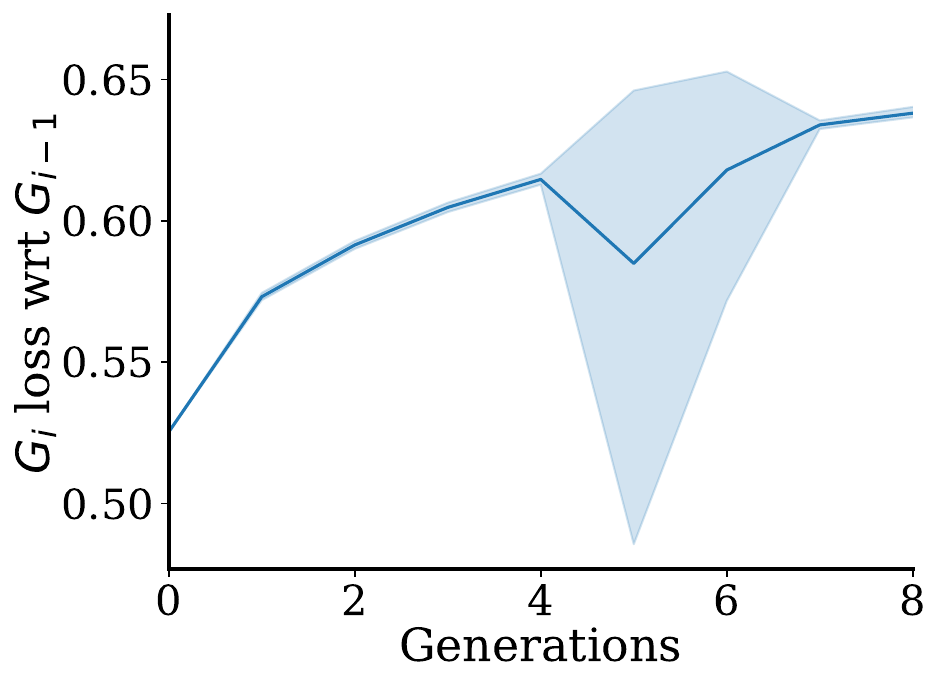} &
    \includegraphics[width=0.4\textwidth]{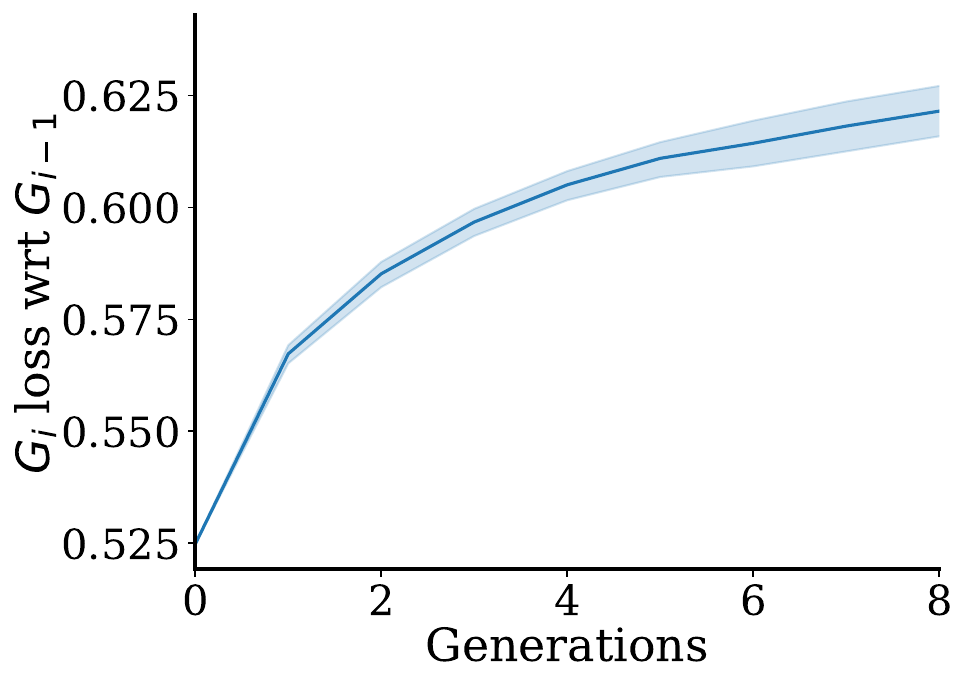} \\
    \includegraphics[width=0.4\textwidth]{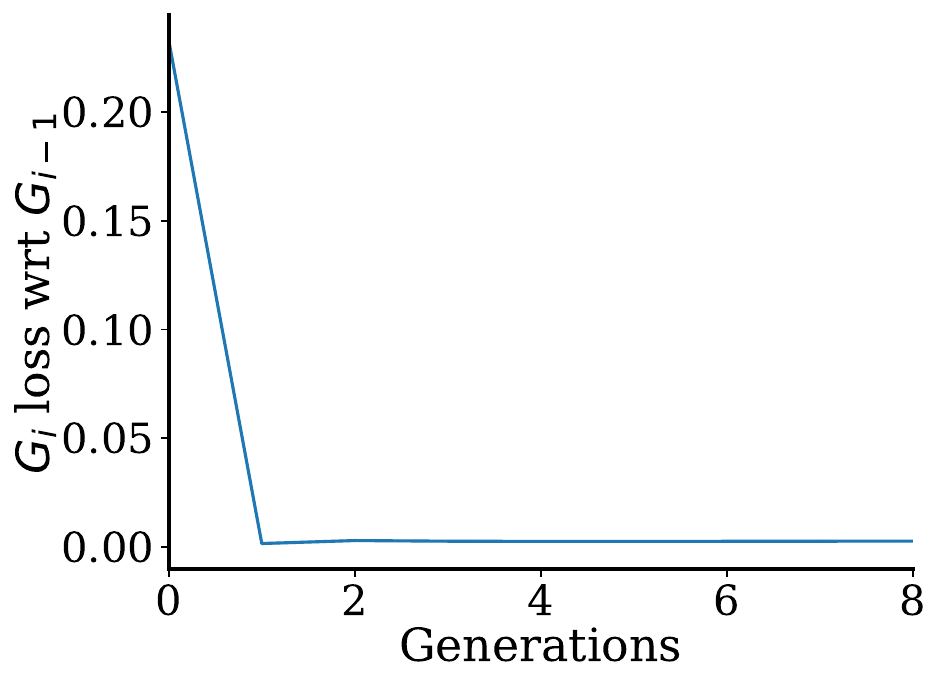} &
    \includegraphics[width=0.4\textwidth]{gen_losses/ff_mc_gl.pdf}
\end{tabular}
\caption{\textit{Top row}: \texttt{ColoredMNIST}, \textit{Second row:} \texttt{ColoredSVHN}, \textit{Third row:} \texttt{CelebA}, \textit{Bottom row:} \texttt{FairFace}. \textit{Left}: model collapse losses, \textit{Right:} model collapse with \genalgname losses.}
\label{fig:gen_losses}
\end{figure}

We show the losses with respect to the parent generator loss ($\mathcal{L}(G_i, G_{i-1})$) over generations as they undergo model collapse and while subject to \genalgname (which causes a minor adjustment). Intuition would suggest that the distribution collapses to be increasingly easy-to-learn, such that successive generators inherit simplified versions (due to finite sampling of their parents) of the problem and so perform better. We observe this effect with the smoothly decreasing loss curves of \texttt{ColoredSVHN} and \texttt{FairFace} in \Cref{fig:gen_losses}. 

However, for both \texttt{ColoredMNIST} and \texttt{CelebA}, we see the exact opposite curve. The child generators are faced with an increasingly hard-to-learn distribution. We hypothesize that this may be due to one of two causes. 1) We do not perform hyper-parameter tuning for the generators at each generation, and perhaps \texttt{ColoredMNIST} and \texttt{CelebA} experience hyperparameter instability. 2) Perhaps this is simply a quirk of model collapse, finite sampling of heavy-tailed distributions may lead to enough bias and noise to significantly complicate the learning task. We propose to investigate the stability of model collapse in future work.

\begin{figure}
    \centering
    \begin{tabular}{cc}
         \includegraphics[height=5cm]{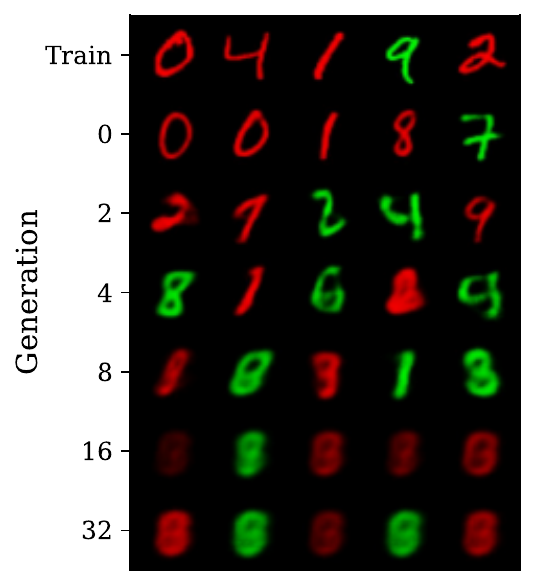}&  
         \includegraphics[height=5cm]{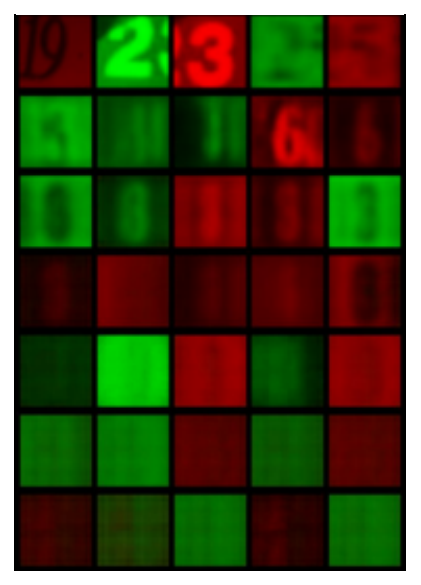}\\
         \includegraphics[width=.45\textwidth]{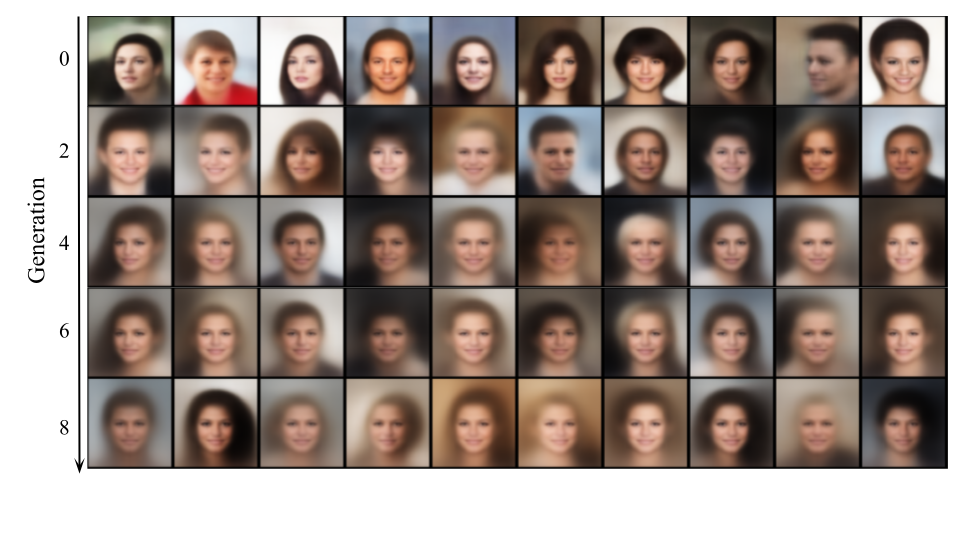}&  
         \includegraphics[width=.45\textwidth]{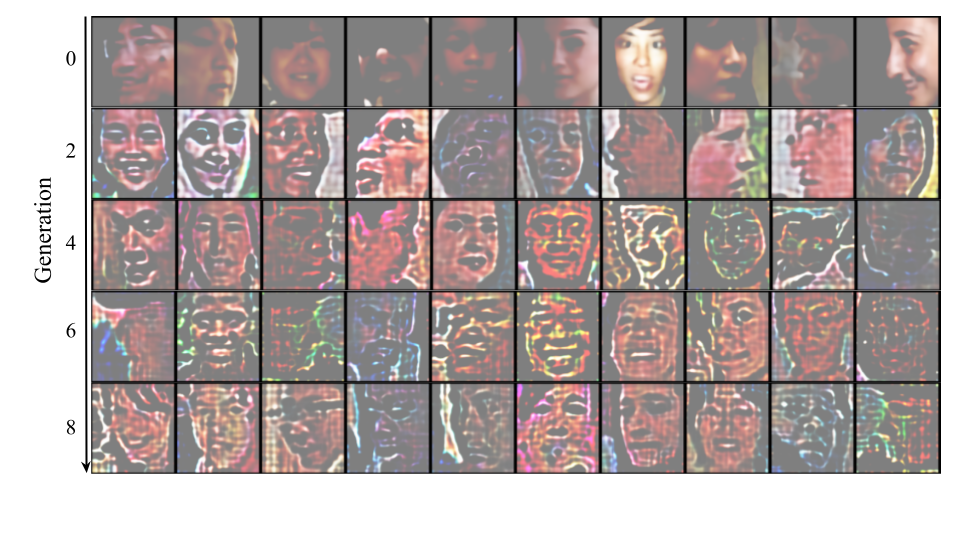}\\
    \end{tabular}
    \caption{Samples from generators undergoing model collapse in \sgsc for \texttt{ColoredMNIST} (\textit{top left}), \texttt{ColoredSVHN} (\textit{top right}), \texttt{CelebA} (\textit{bottom left}), and \texttt{FairFace} (\textit{bottom right}).}
    \label{fig:mc_images}
\end{figure}

We also provide some examples generated by generators undergoing model collapse in \Cref{fig:mc_images}. 

\section{Additional Results}
\subsection{\seqc Results}
\begin{figure}[ht]
\centering
\begin{tabular}{ccc}
\includegraphics[width=0.31\textwidth]{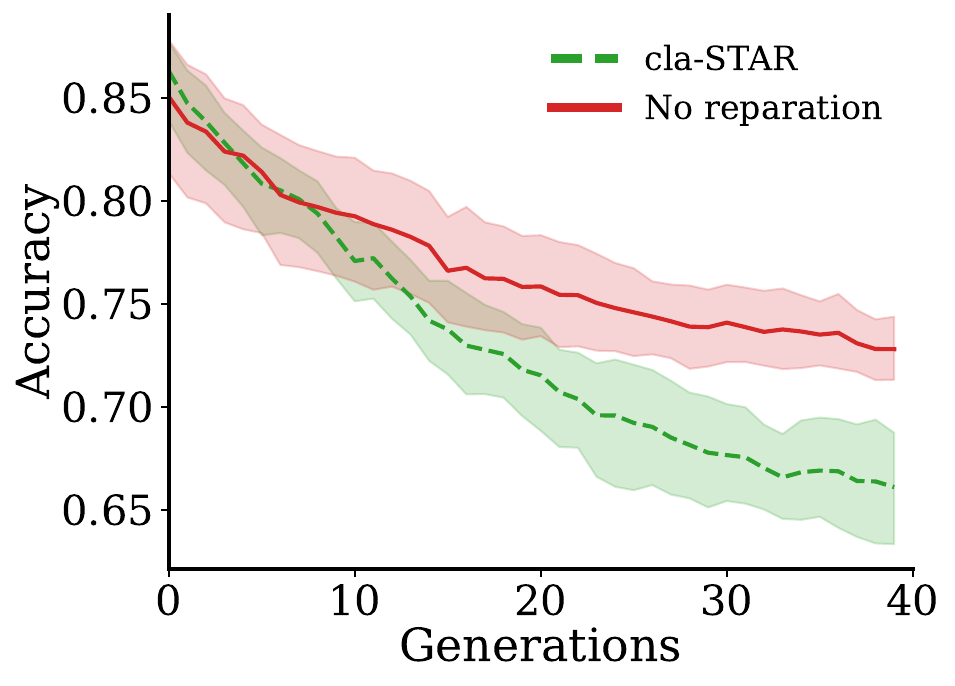} &
\includegraphics[width=0.31\textwidth]{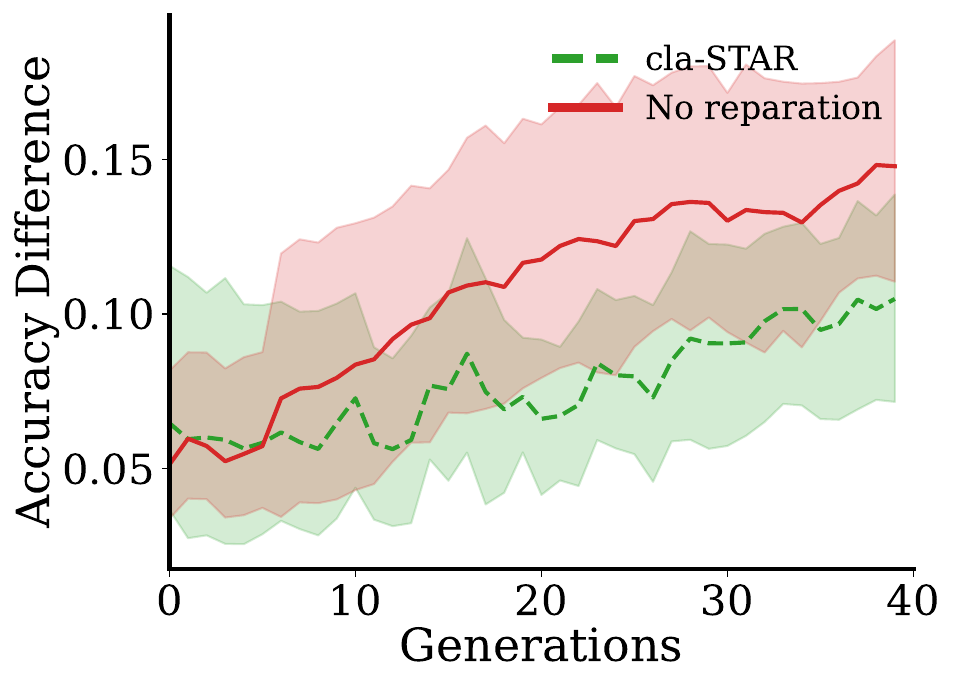} &
\includegraphics[width=0.31\textwidth]{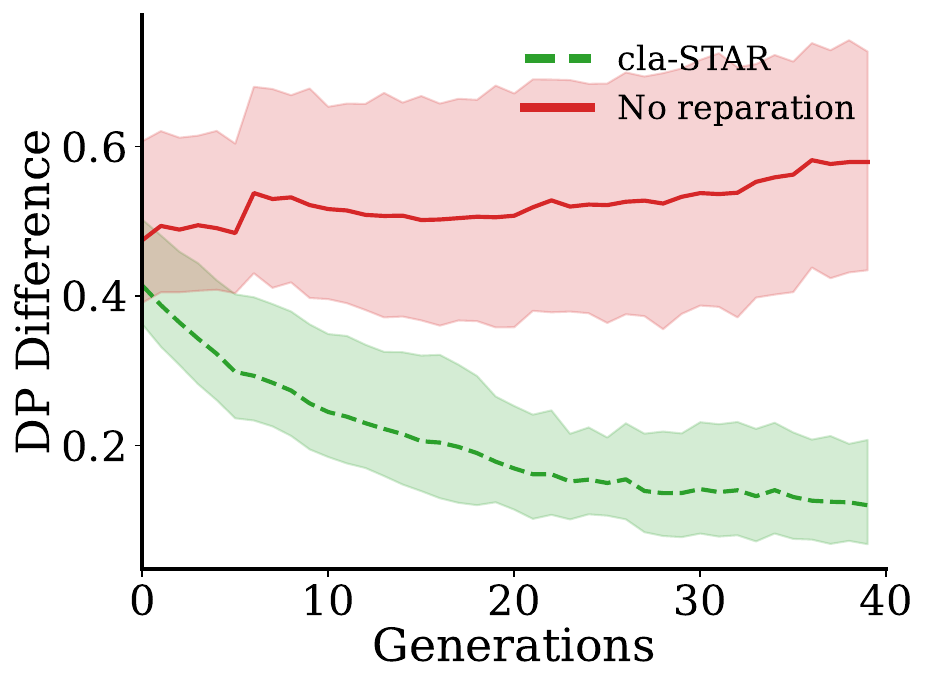} \\
\end{tabular}
\begin{tabular}{ccc}
\includegraphics[width=0.31\textwidth]{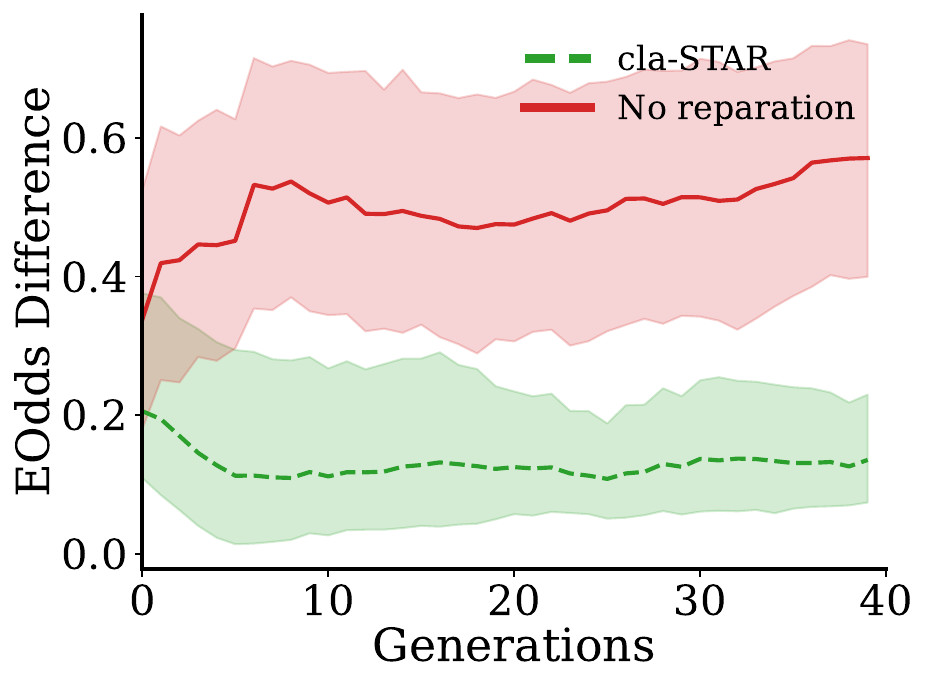}
\includegraphics[width=0.31\textwidth]{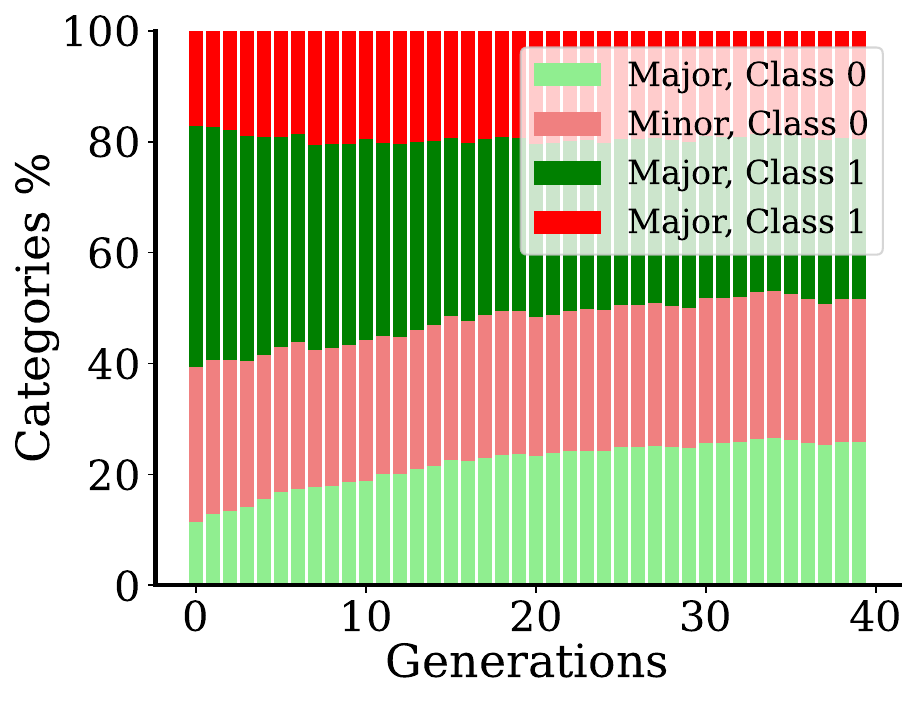} &
\includegraphics[width=0.31\textwidth]{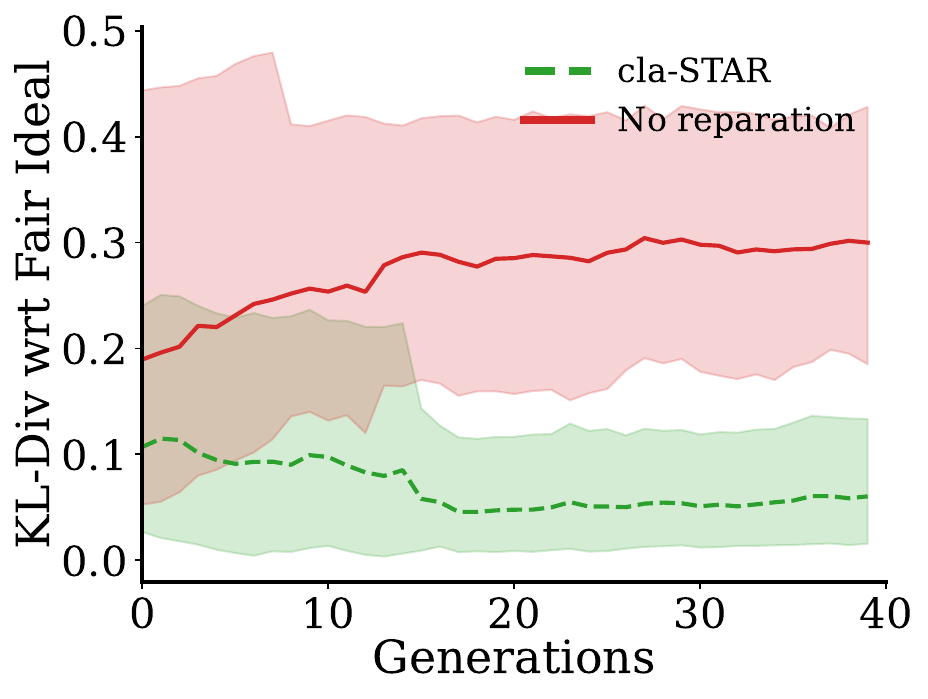}\\
\end{tabular}
\caption{\texttt{ColoredSVHN} results for \seqc on the evaluation set. \textit{Top:} accuracy, accuracy difference, and demographic parity difference. We observe lower fairness differences with \claalgname, with a cost of more inaccuracy.
\textit{Bottom:} equalized odds difference, the \distrname created during \claalgname, and the KL-divergence between \claalgname fairness ideal and classifier \distrname. The KL-Divergence decreases with \claalgname as the batches become more evenly balanced across group and class. }
\label{fig:SVHN_nomc}
\end{figure}

\begin{figure}[ht]
\centering
\begin{tabular}{ccc}
\includegraphics[width=0.31\textwidth]{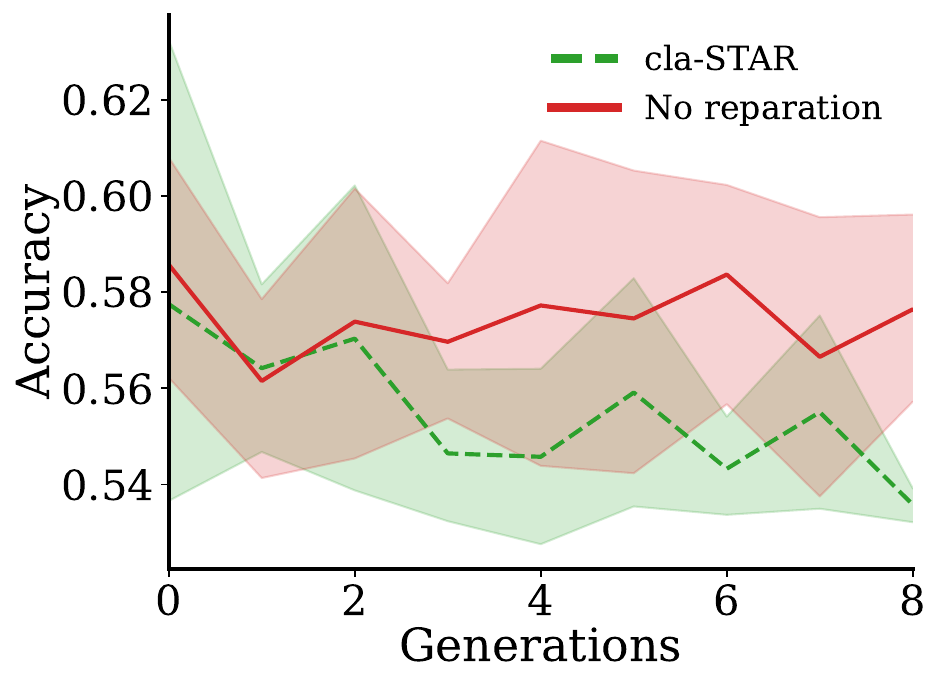} &
\includegraphics[width=0.31\textwidth]{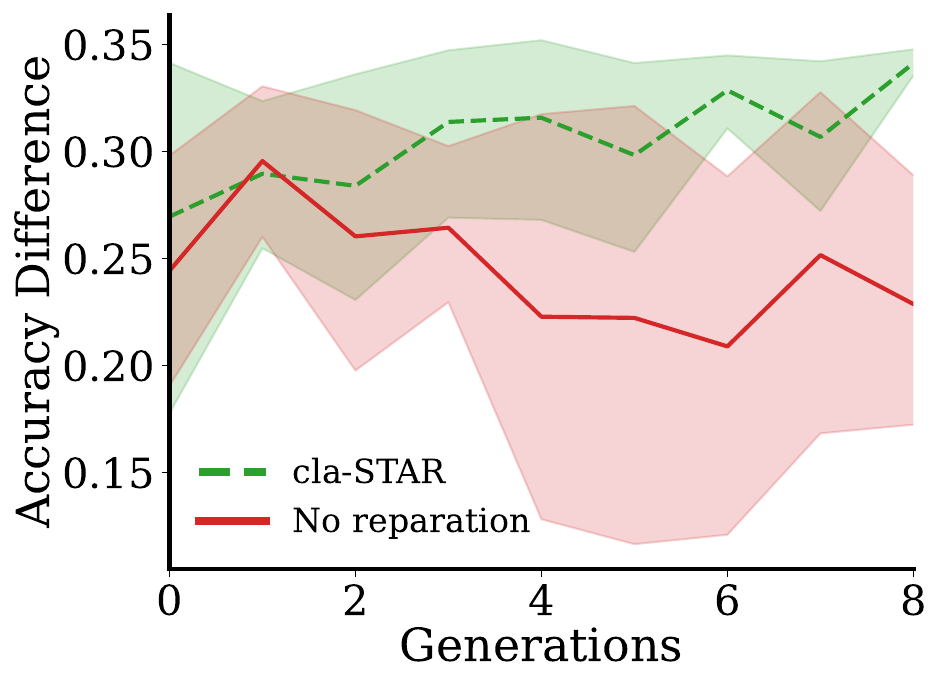} &
\includegraphics[width=0.31\textwidth]{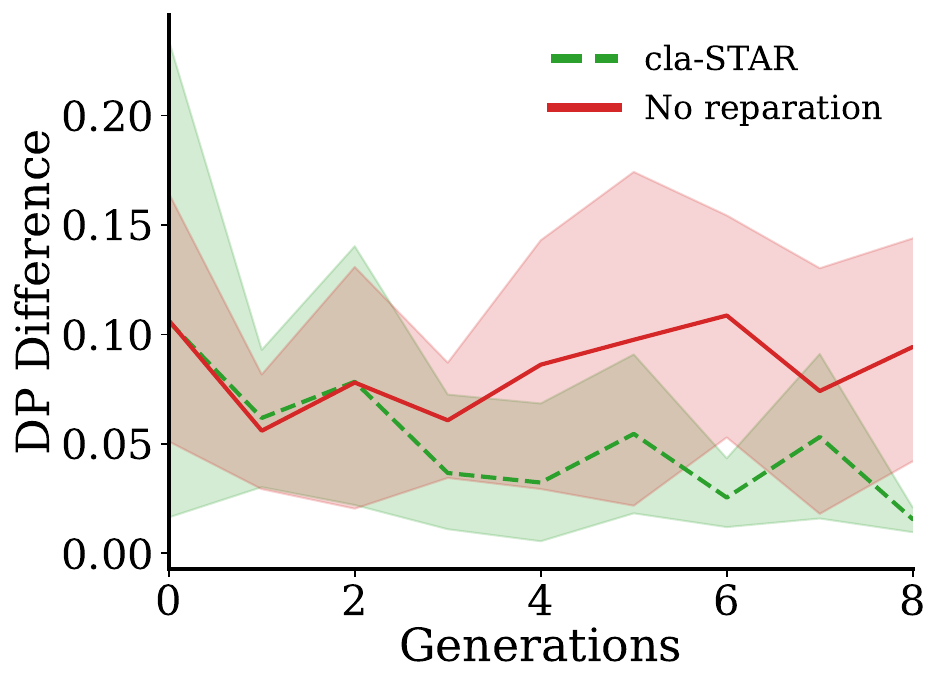} \\
\end{tabular}
\begin{tabular}{cc}
\includegraphics[width=0.31\textwidth]{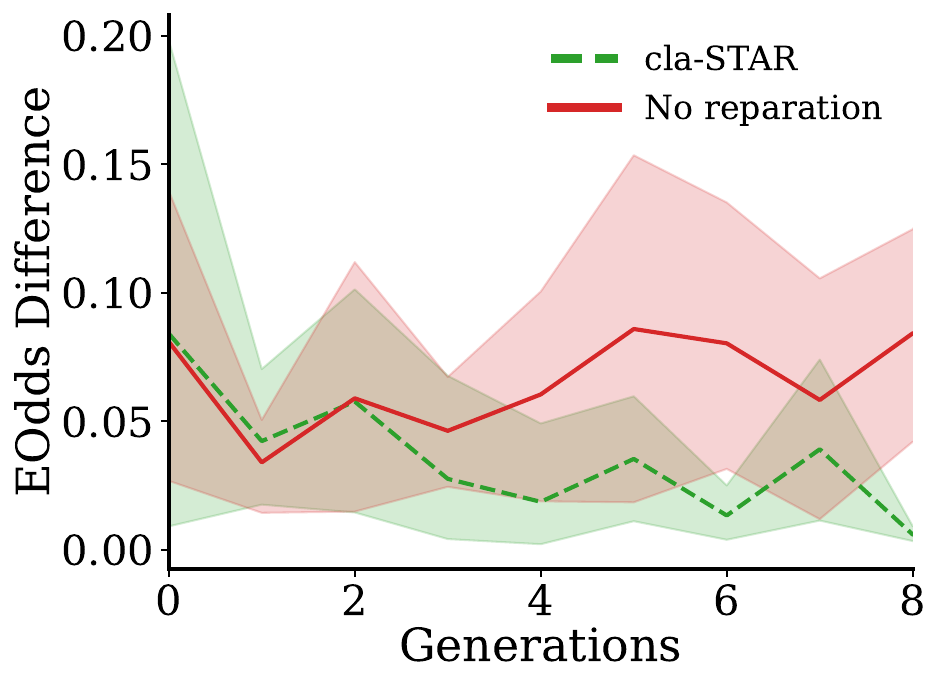}
\includegraphics[width=0.31\textwidth]{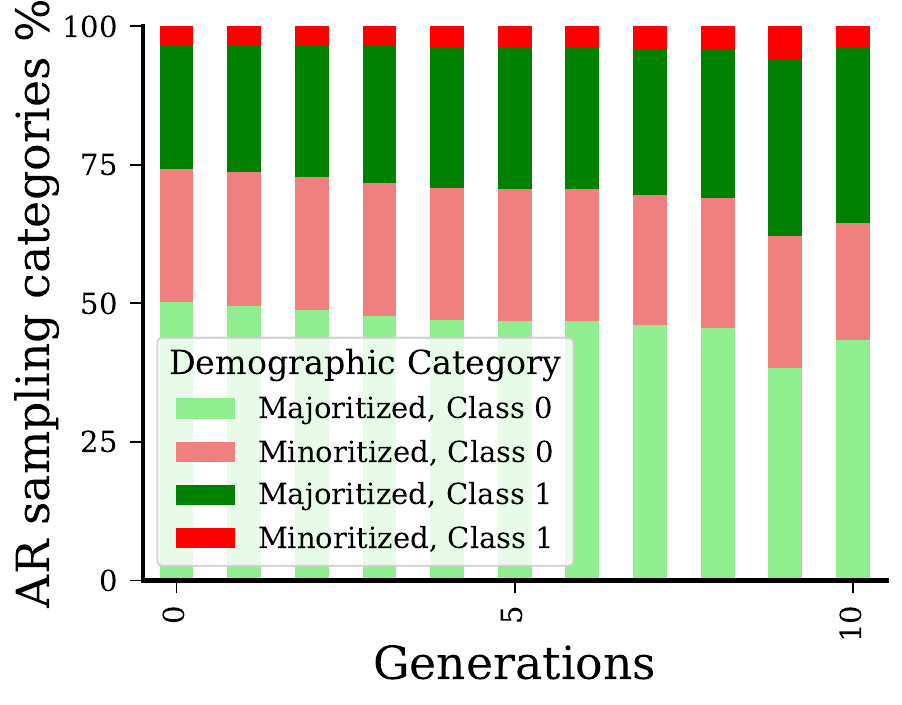} \\
\end{tabular}
\caption{\texttt{CelebA} results for \seqc on the evaluation set. \textit{Top:} accuracy, accuracy difference, and demographic parity difference. 
Better fairness (lower fairness difference) and higher accuracy is achieved with \claalgname. \textit{Bottom:} equalized odds difference and the \distrname created during \claalgname. The batches become more evenly balanced across group and class.}
\label{fig:celeba_nomc}
\end{figure}

\begin{figure}[ht]
\centering
\begin{tabular}{ccc}
\includegraphics[width=0.31\textwidth]{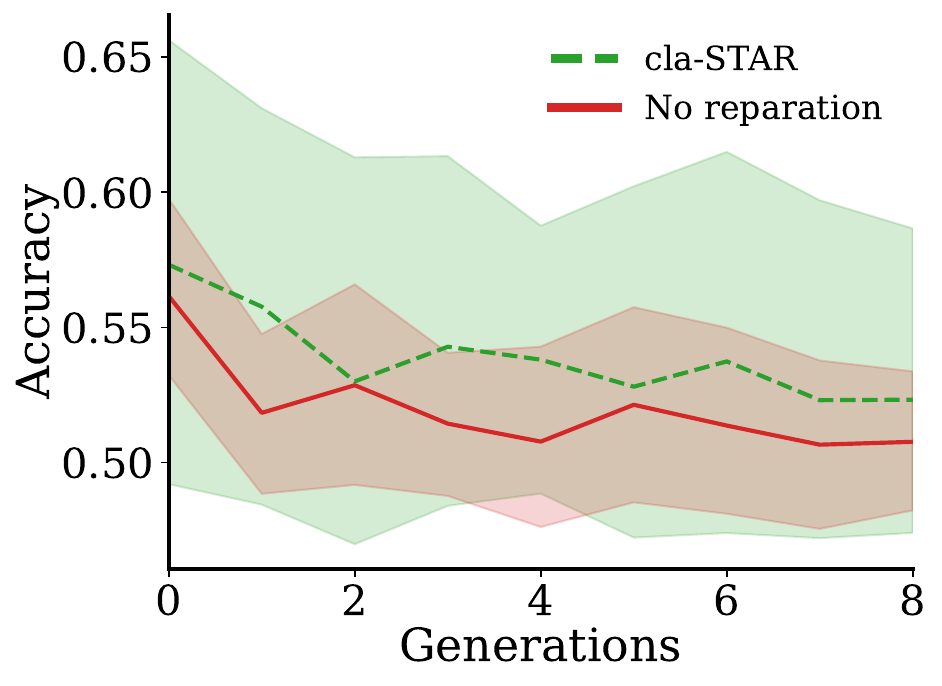} &
\includegraphics[width=0.31\textwidth]{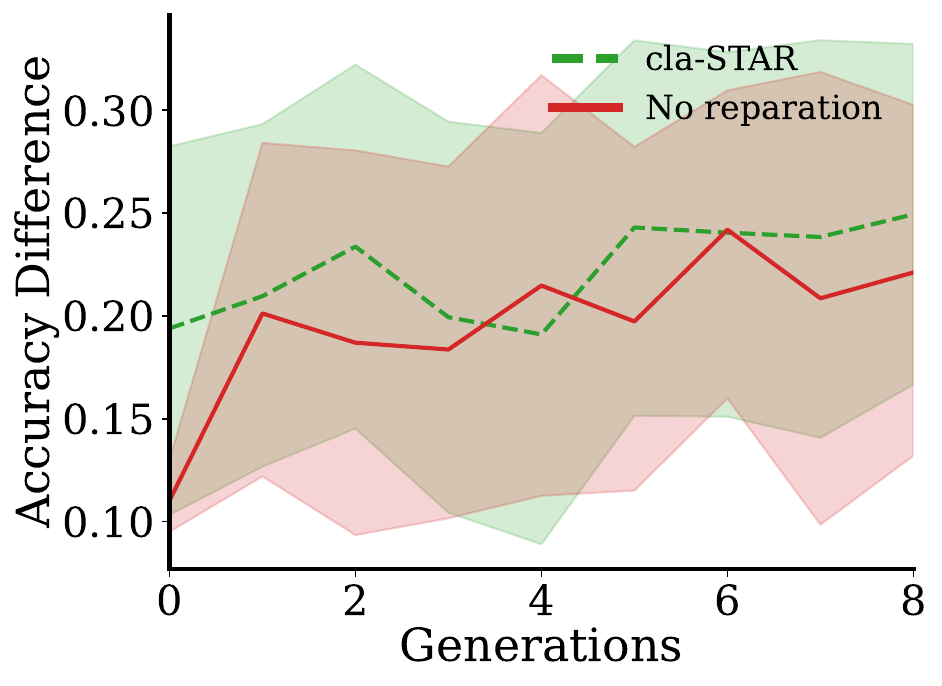} &
\includegraphics[width=0.31\textwidth]{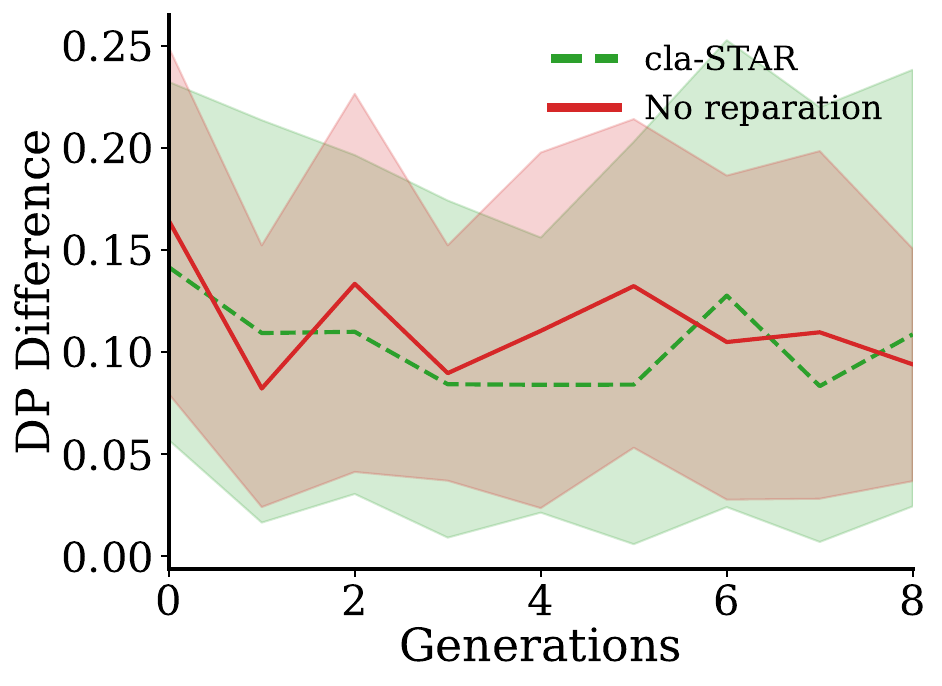} \\
\end{tabular}
\begin{tabular}{cc}
\includegraphics[width=0.31\textwidth]{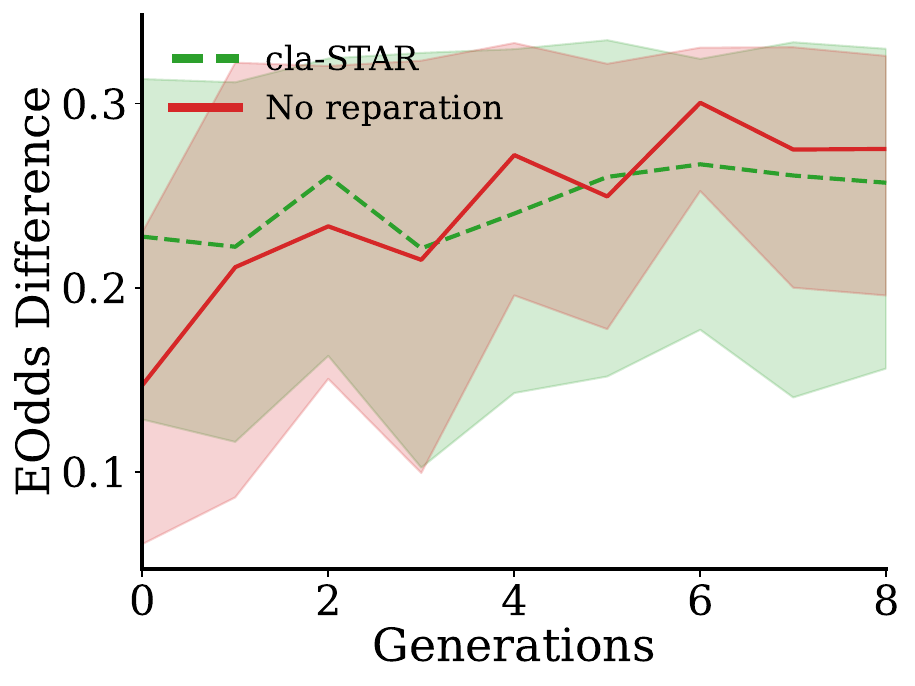}
\includegraphics[width=0.31\textwidth]{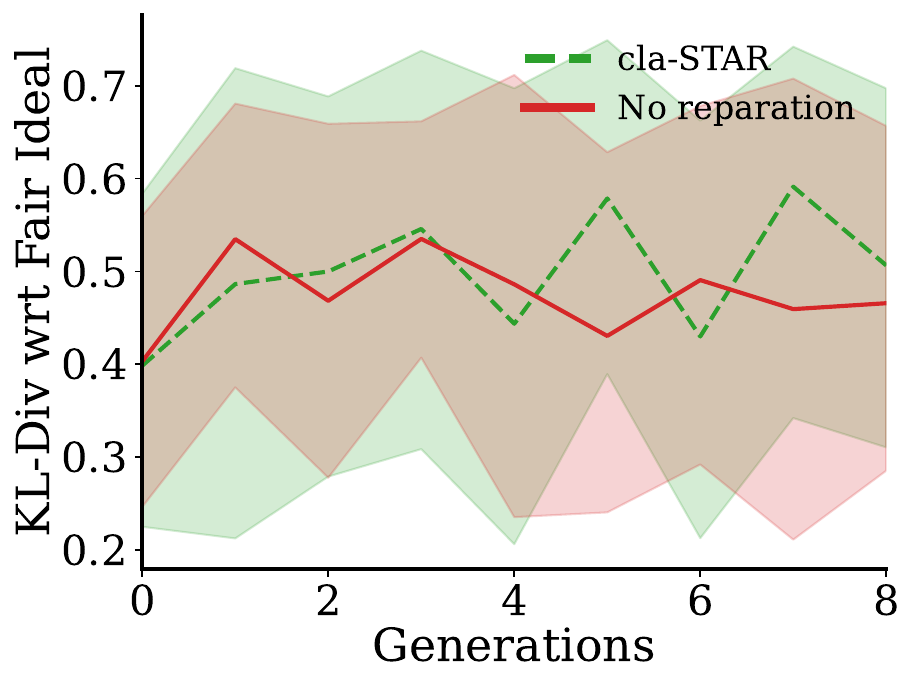}\\
\end{tabular}
\caption{\texttt{FairFace} results for \seqc on the evaluation set. \textit{Top:} accuracy, accuracy difference, and demographic parity difference. We observe lower fairness differences with \claalgname, with a cost of more inaccuracy.
\textit{Bottom:} equalized odds difference, and the KL-divergence between \claalgname fairness ideal and classifier \distrname. We do not report the \distrname formed by \claalgname as there are 28 categories; instead, refer to the \distrname of the classifiers in \Cref{ssec:nomc_strata}.}
\label{fig:FairFace_nomc}
\end{figure}

We provide full suites of figures for \texttt{ColoredMNIST}, \texttt{ColoredSVHN}, \texttt{CelebA}, and \texttt{FairFace} on \seqc. See Figures ~\ref{fig:ColoredMNIST_nomc}, ~\ref{fig:SVHN_nomc}, ~\ref{fig:celeba_nomc} and ~\ref{fig:FairFace_nomc}, respectively.

\subsubsection{\claalgname Batch Balances}\label{ssec:nomc_strata}
\begin{figure}[ht]
    \centering
    \begin{tabular}{ccc}
       \includegraphics[width=.3\textwidth]{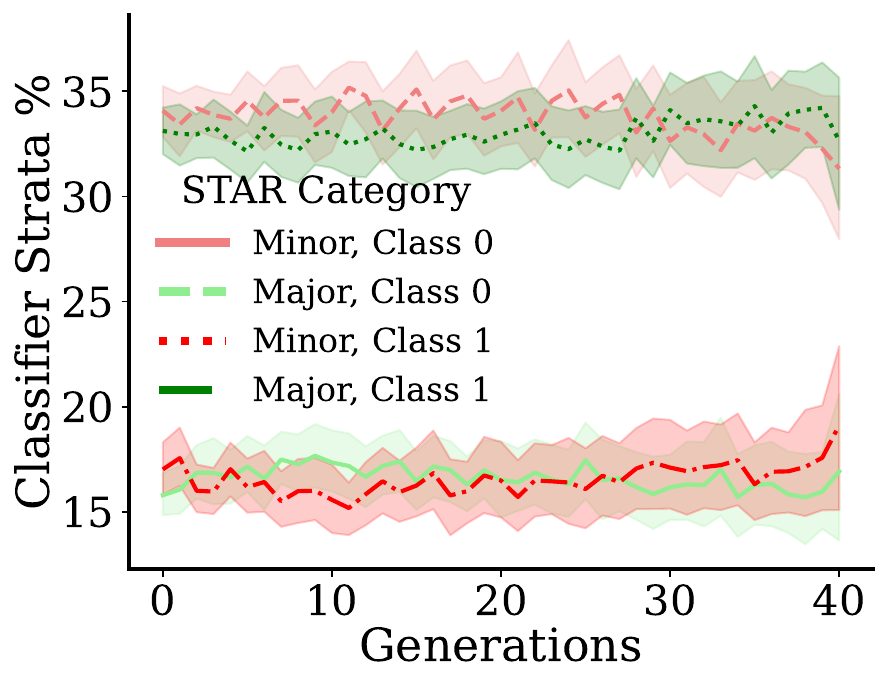}  & \includegraphics[width=.3\textwidth]{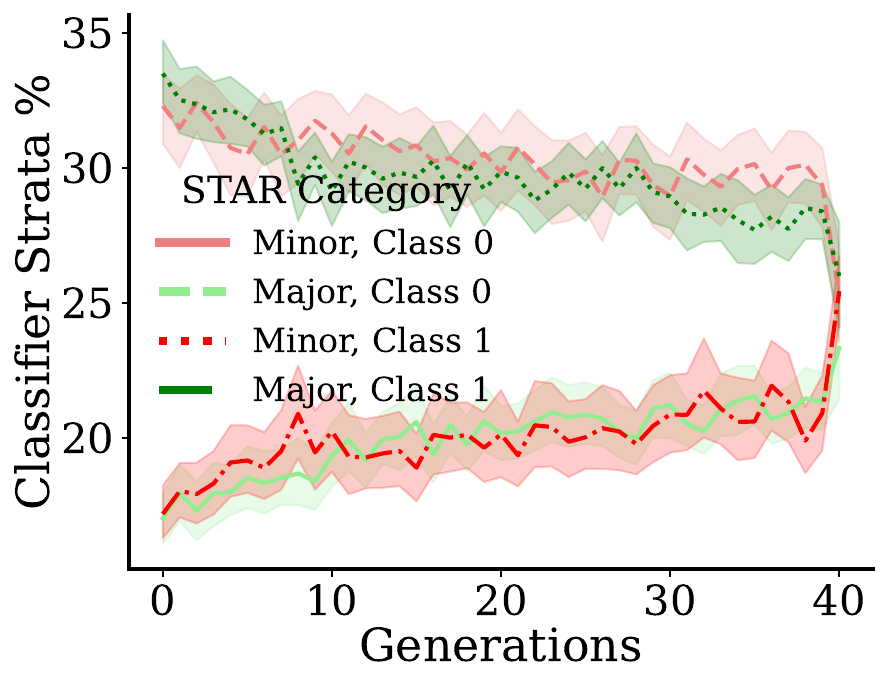} &
       \includegraphics[width=.3\textwidth]{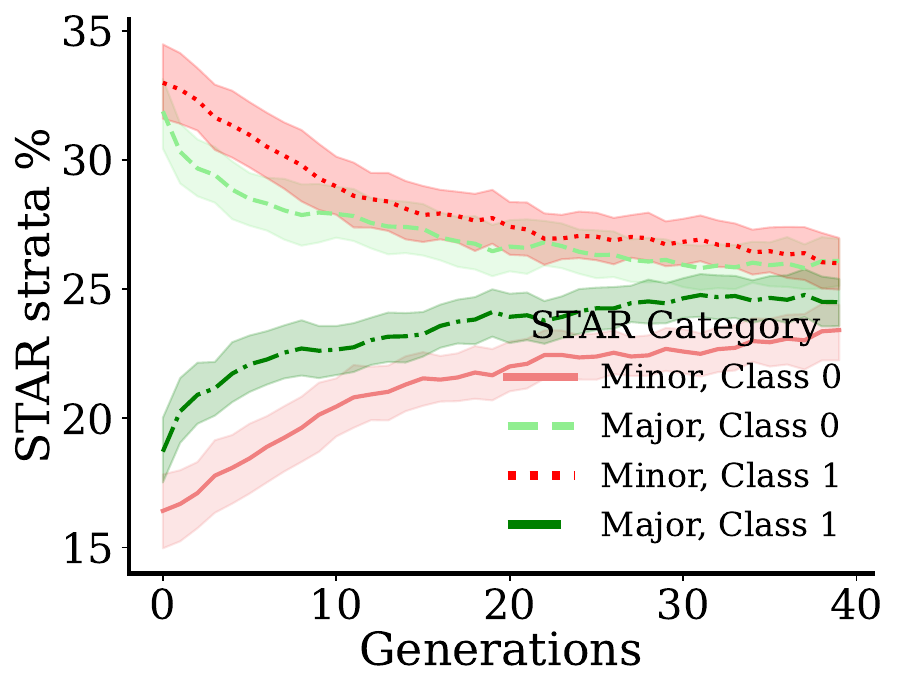} \\
       \includegraphics[width=.3\textwidth]{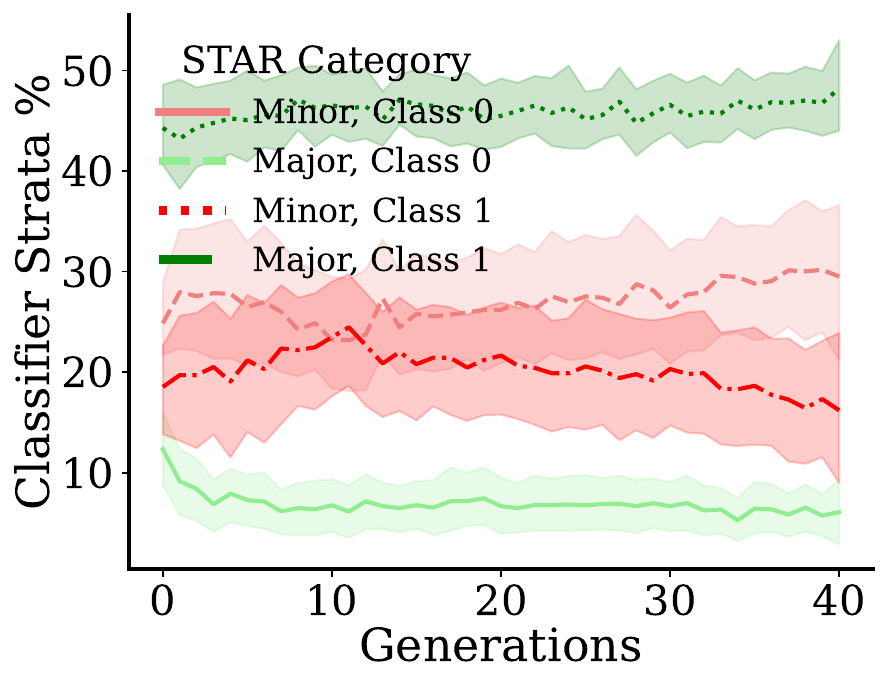} & \includegraphics[width=.3\textwidth]{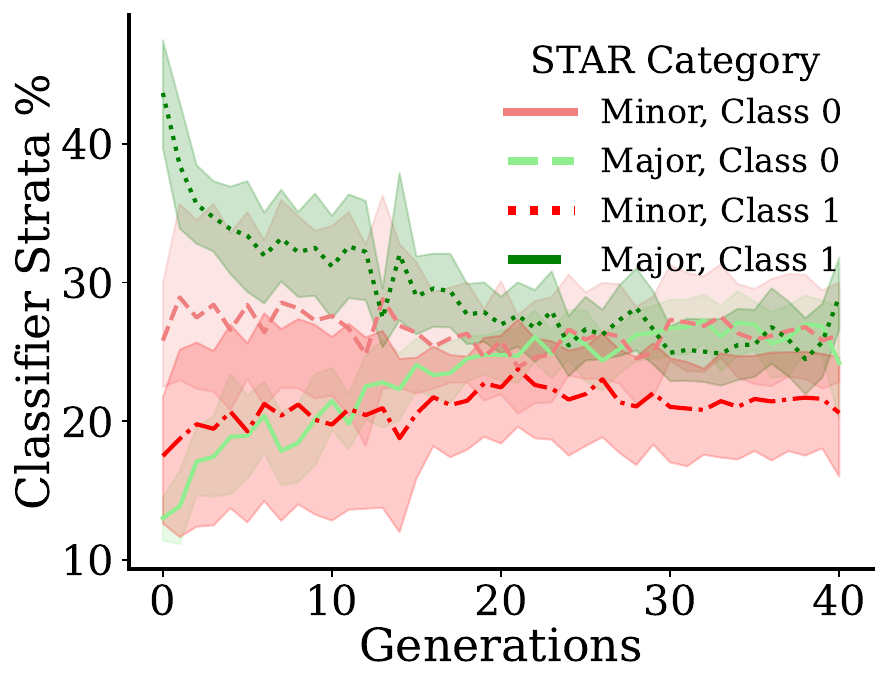} & \includegraphics[width=.3\textwidth]{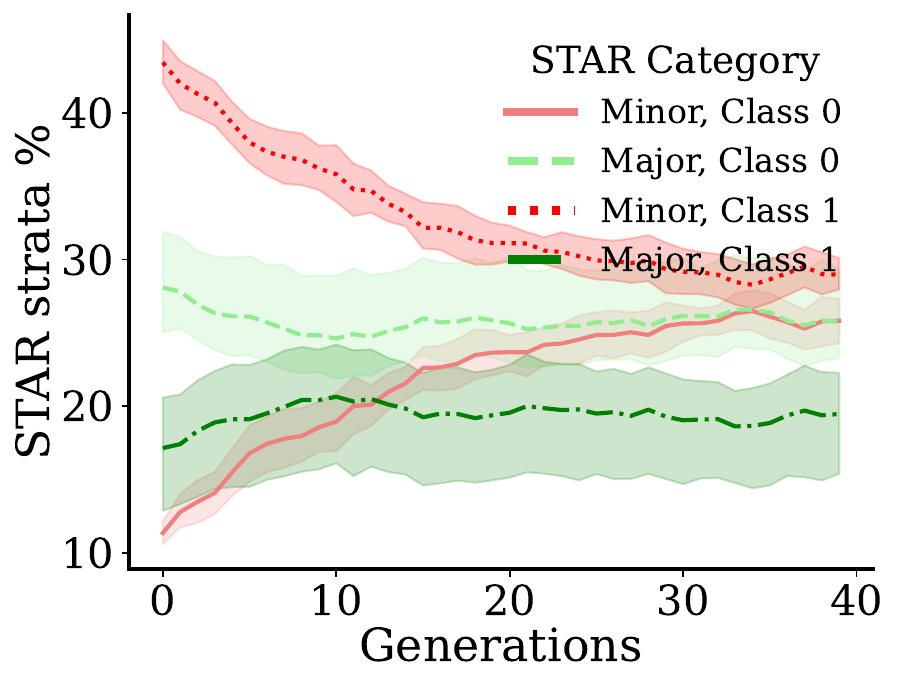} \\
       \includegraphics[width=.3\textwidth]{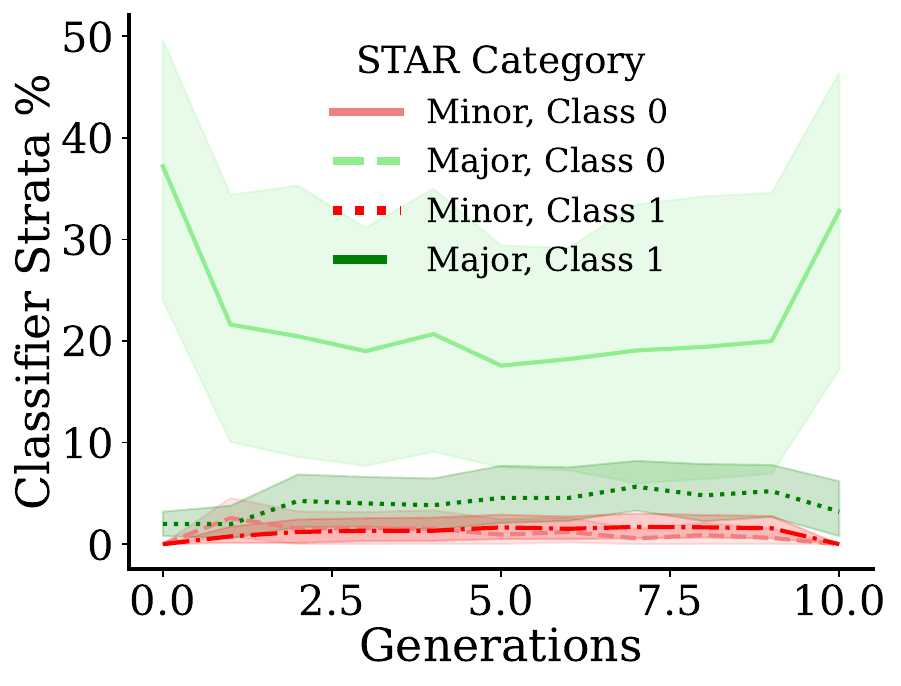} & \includegraphics[width=.3\textwidth]{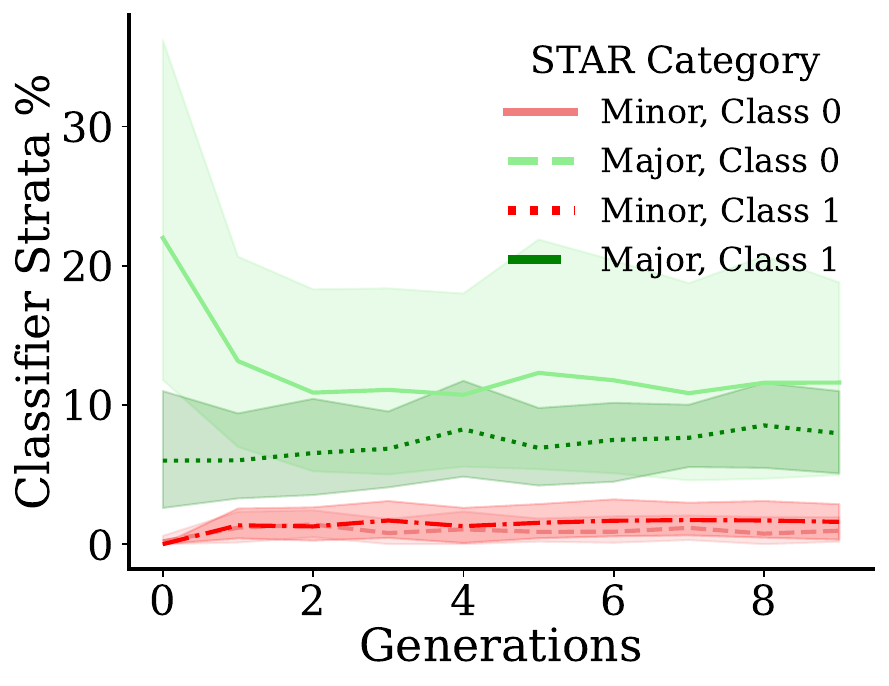} & \includegraphics[width=.3\textwidth]{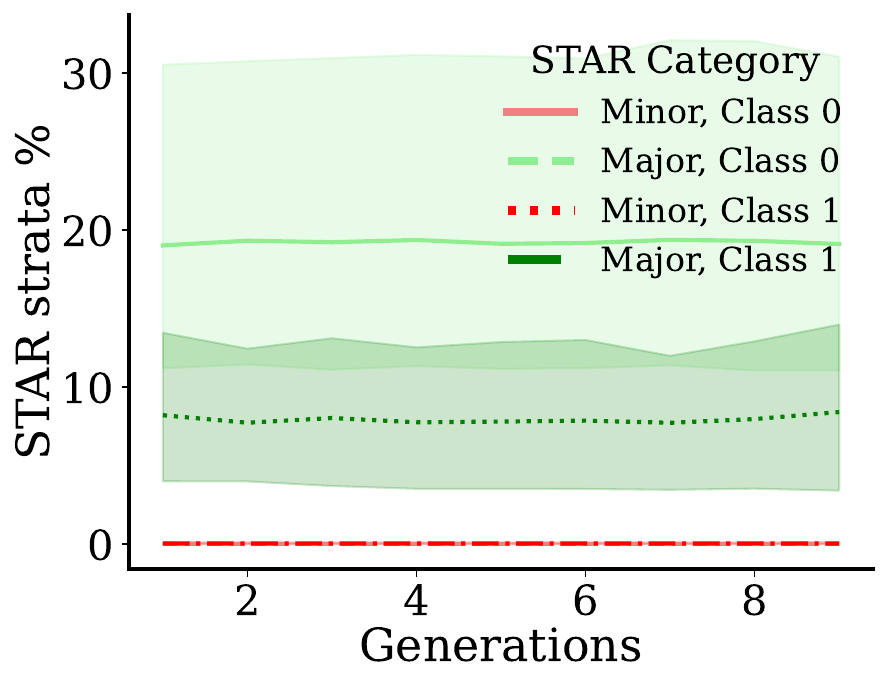} \\
    \end{tabular}
    \caption{\distrname balances for datasets in \seqc. \textit{Left:} The \distrname of classifiers without reparation. \textit{Center:} The \distrname resulting from classifiers with \claalgname. \textit{Right:} The \distrname used to train classifiers with \claalgname. \textit{Top:} \texttt{ColoredMNIST}. \textit{Second row:} \texttt{ColoredSVHN}. \textit{Bottom:} \texttt{FairFace}, instead of showing all 28 categories, we choose the two categories per label that are most frequently the largest and smallest portion of the batch across all generations.}
    \label{fig:nomc_stratas}
\end{figure}

We report the composition of the batches used when training the classifiers with and without \claalgname in the \seqc setting in \Cref{fig:nomc_stratas}. These figures show the \distrname \claalgname uses to train classifiers, the resulting classifier \distrname, and the \distrname of classifiers trained without any reparation. Usually, the \claalgname \distrname are the most balanced, followed by the \distrname of classifiers that received reparation. In \texttt{FairFace}, the batches are mostly older white males, and sometimes younger white non-males; the least populated categories are usually younger people from the Indian, South East Asian, and Hispanic/Latino races. 

\subsection{\sgsc Results}
\begin{figure}
\centering
\begin{tabular}{cccc}
\includegraphics[width=0.22\textwidth]{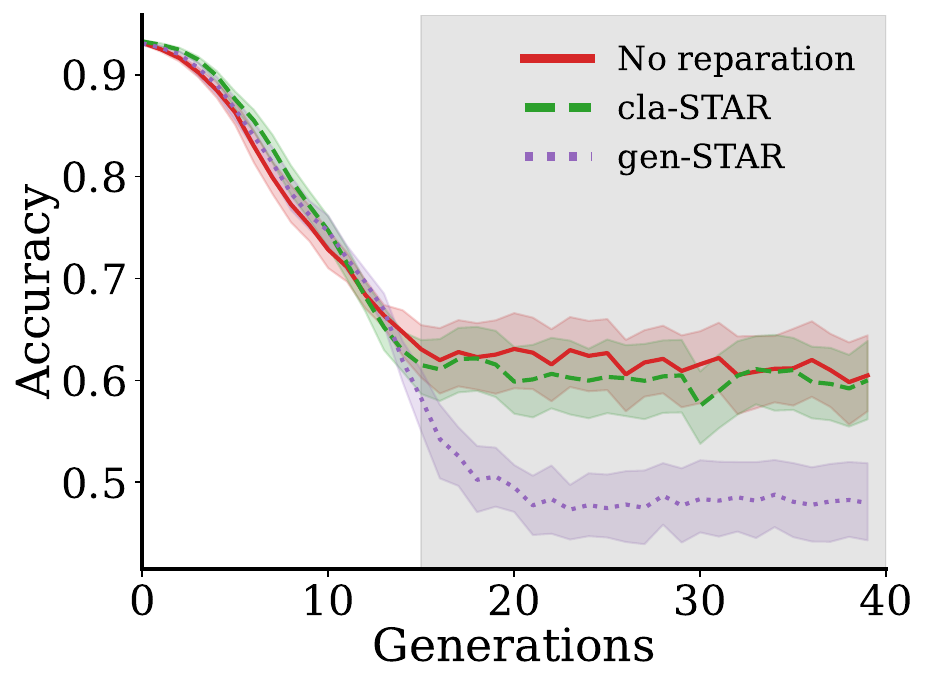} &
\includegraphics[width=0.22\textwidth]{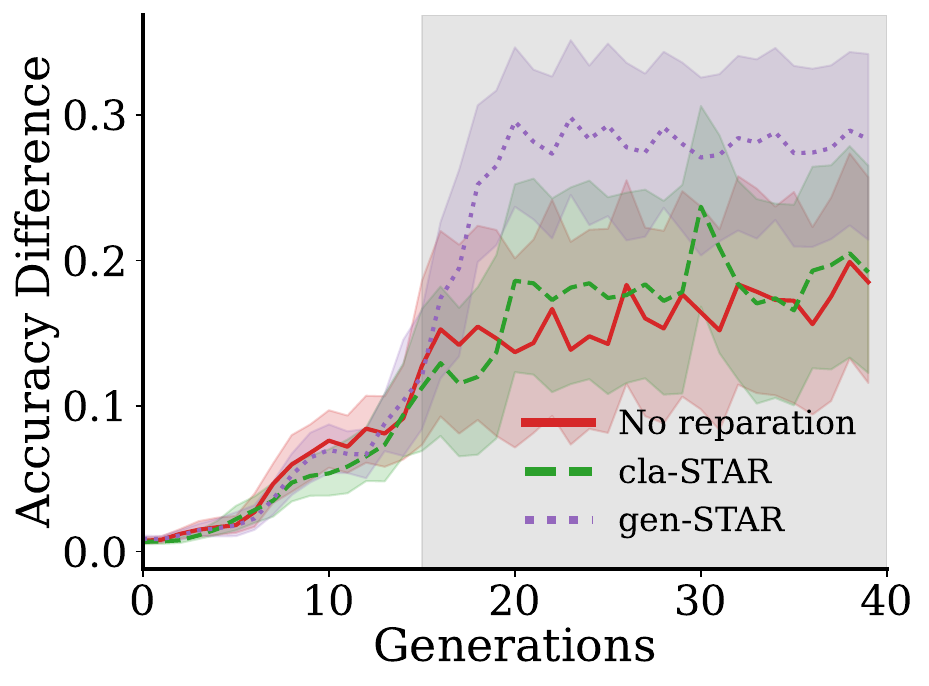} &
\includegraphics[width=0.22\textwidth]{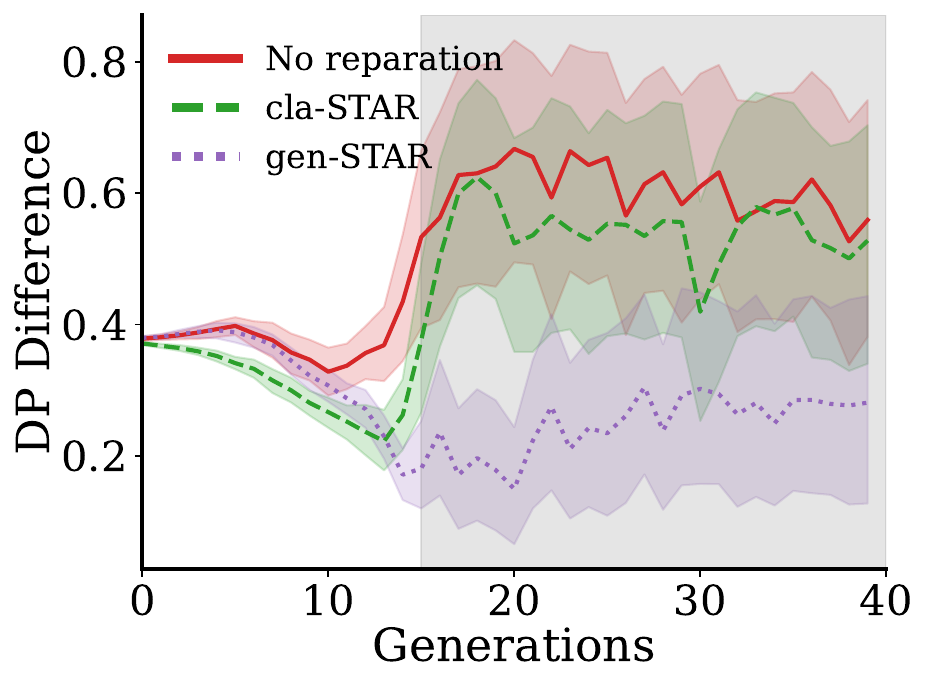} &
\includegraphics[width=0.22\textwidth]{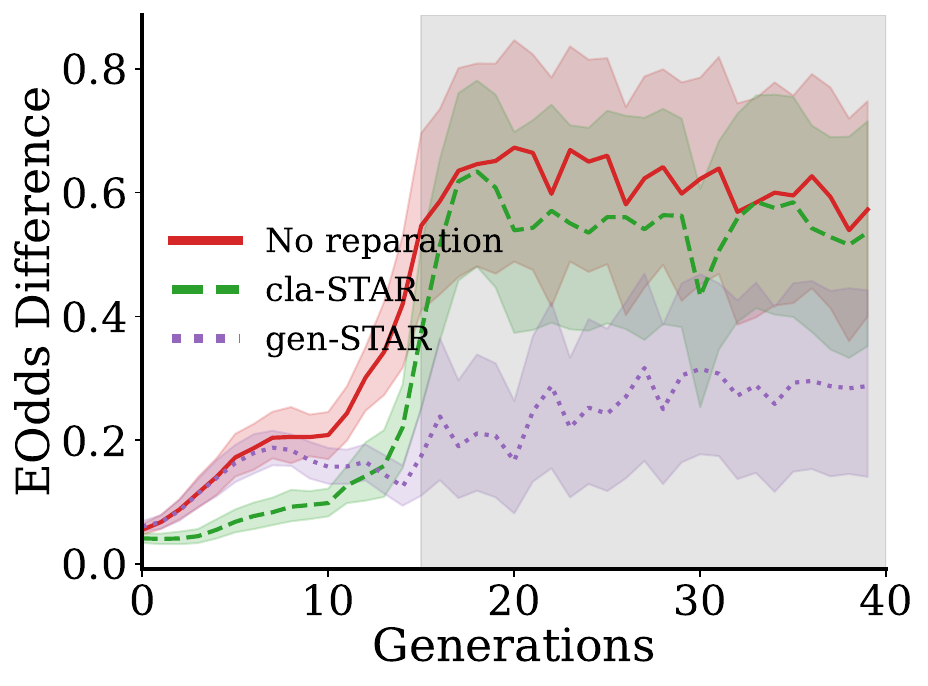} \\
\end{tabular}
\begin{tabular}{cccc}
\includegraphics[width=0.22\textwidth]{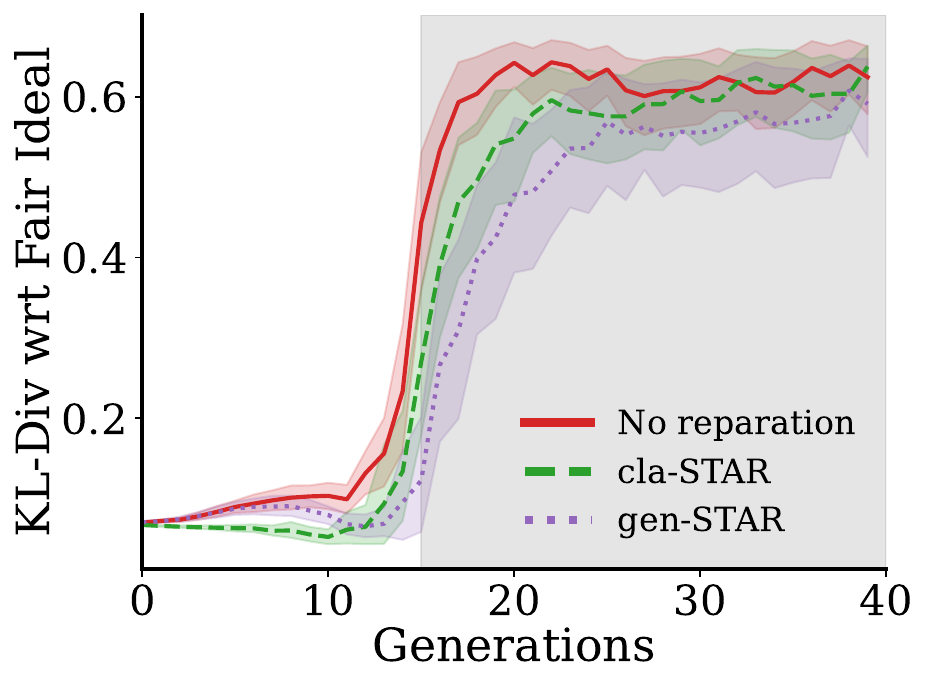} &
\includegraphics[width=0.22\textwidth]{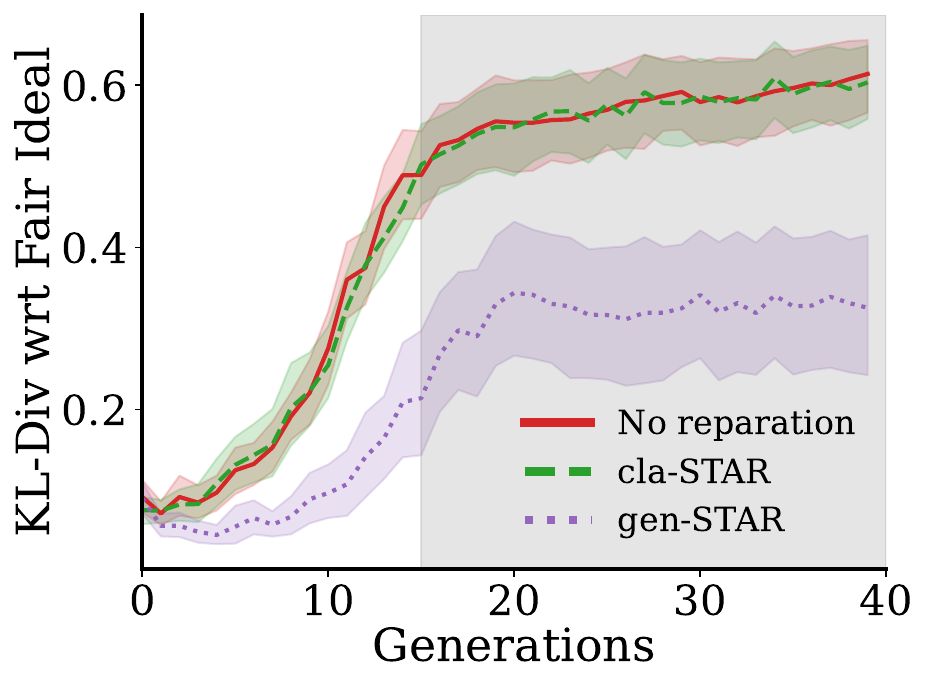} &
\includegraphics[width=0.22\textwidth]{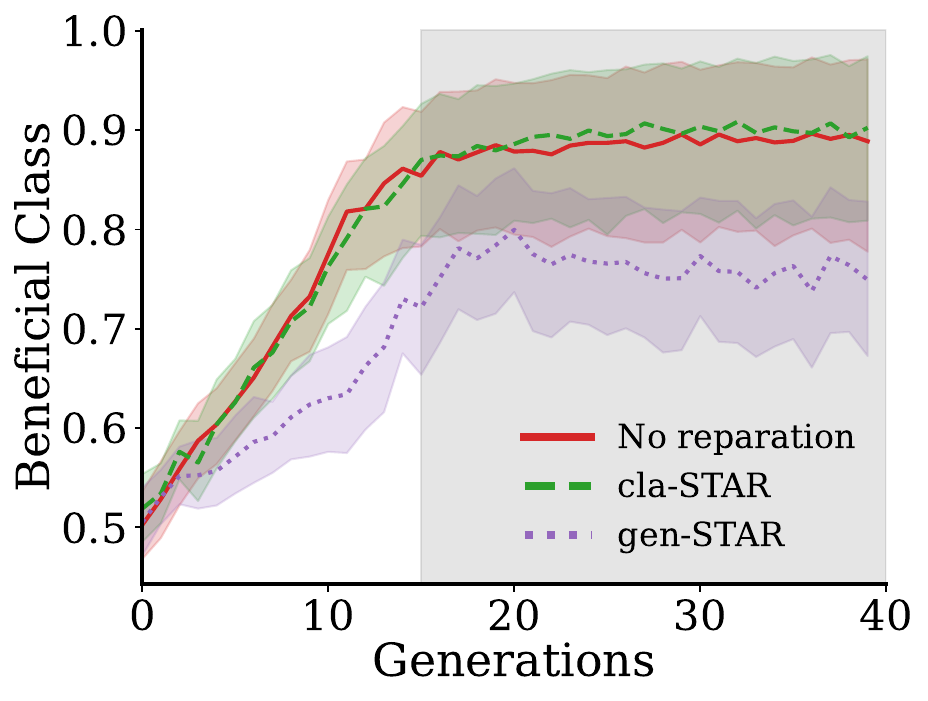} &
\includegraphics[width=0.22\textwidth]{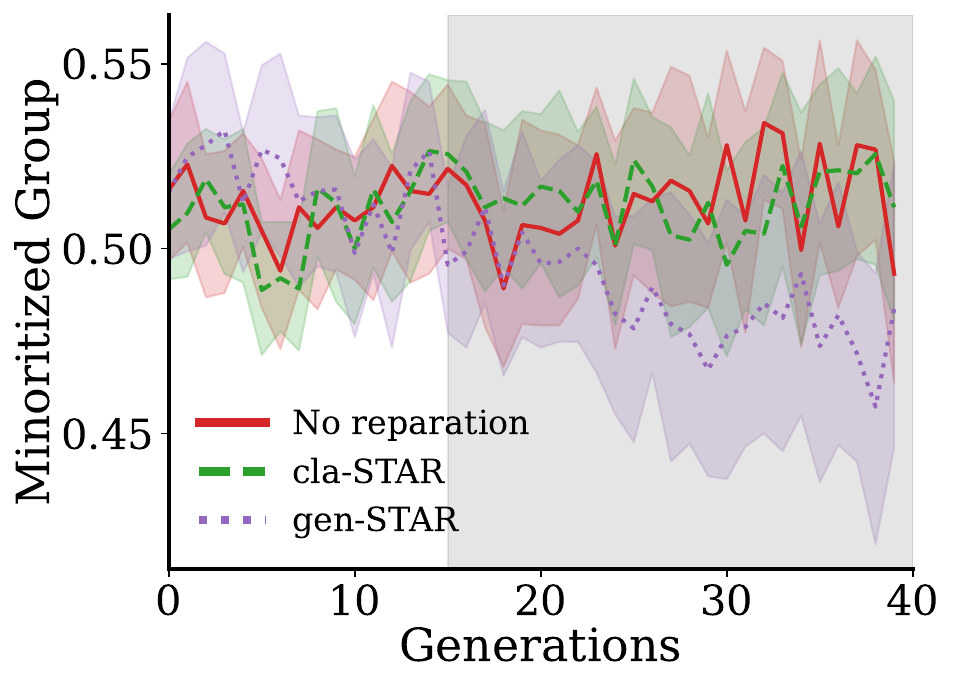} \\
\end{tabular}
\caption{\texttt{ColoredMNIST} results for \sgsc. \textit{Top:} shows accuracy, accuracy difference, demographic parity difference, and equalized odds difference. For the latter three, lower values are better. \textit{Bottom:} KL-Divergence between fairness ideal and classifiers, and between fairness ideal and generator \distrname, the class balance, and group balance. Shading shows collapsed generations.  
We observe that \genalgname leads to better representation and fairness, though with a cost to the accuracy metrics. Recall that for \texttt{ColoredMNIST}, we find metric tension between EOdds, accuracy difference, and the \algname fairness ideal.} 
\label{fig:ColoredMNIST_mc}
\end{figure}

\begin{figure}
\centering
\begin{tabular}{cccc}
\includegraphics[width=0.22\textwidth]{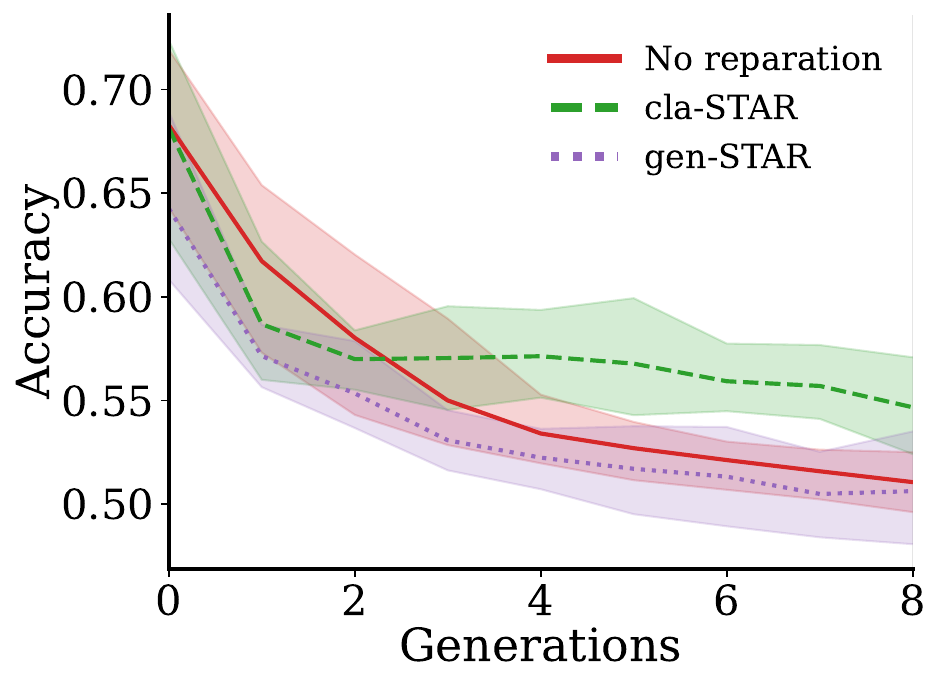} &
\includegraphics[width=0.22\textwidth]{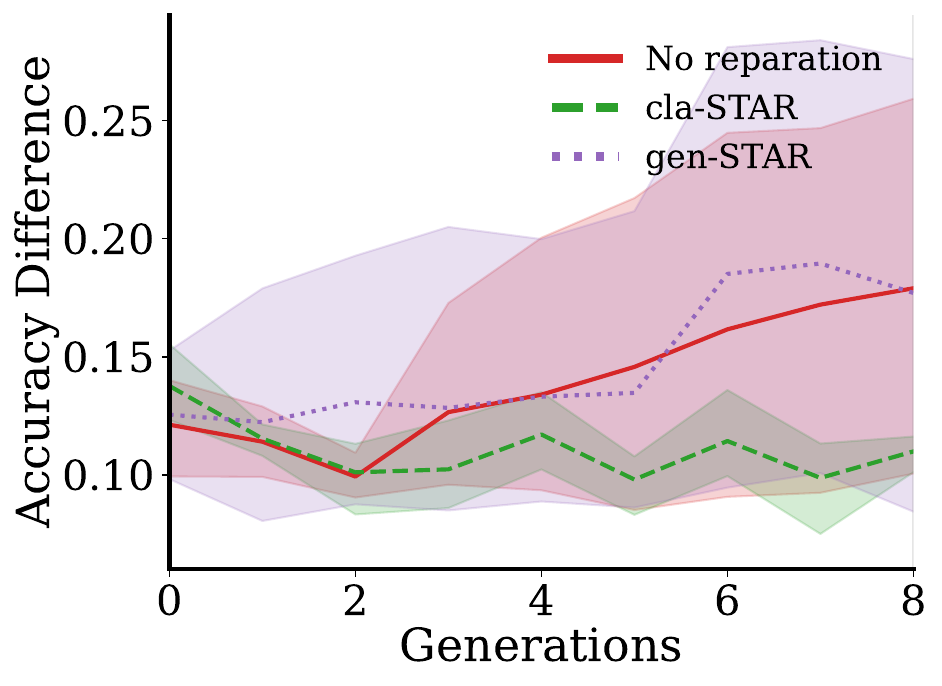} &
\includegraphics[width=0.22\textwidth]{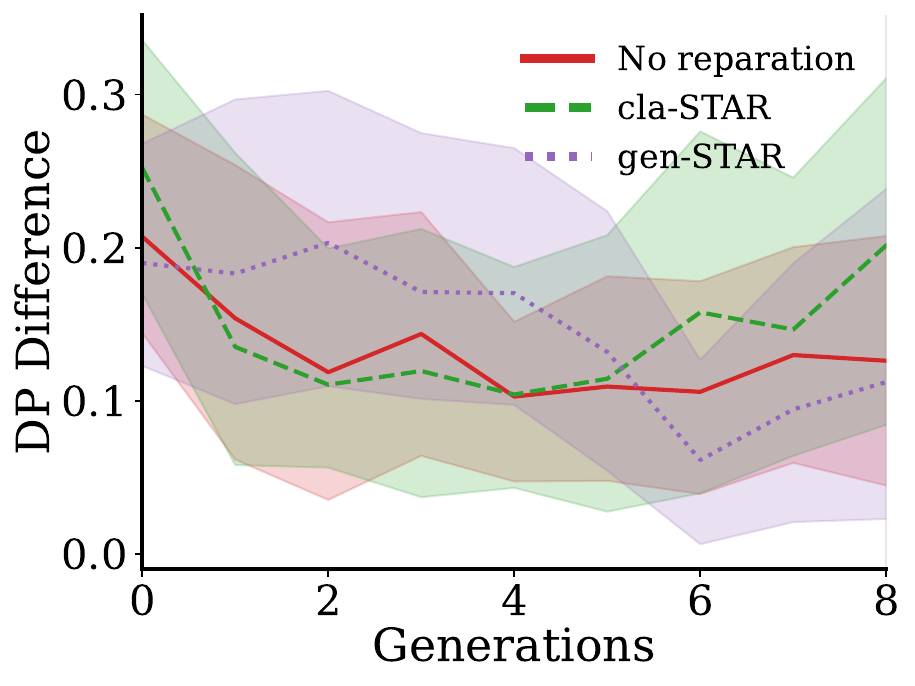} &
\includegraphics[width=0.22\textwidth]{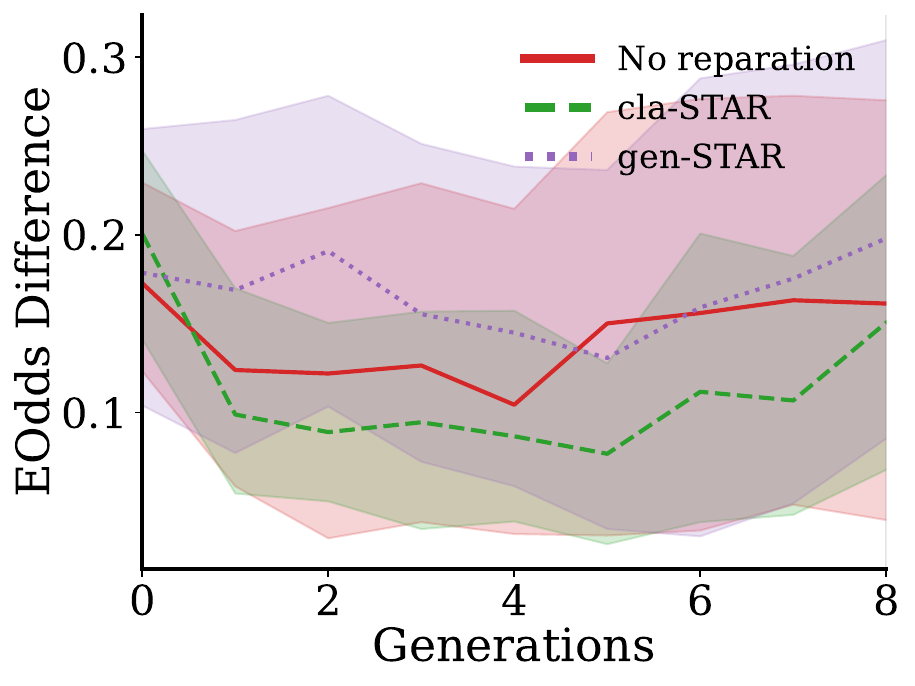} \\
\end{tabular}
\begin{tabular}{ccc}
\includegraphics[width=0.22\textwidth]{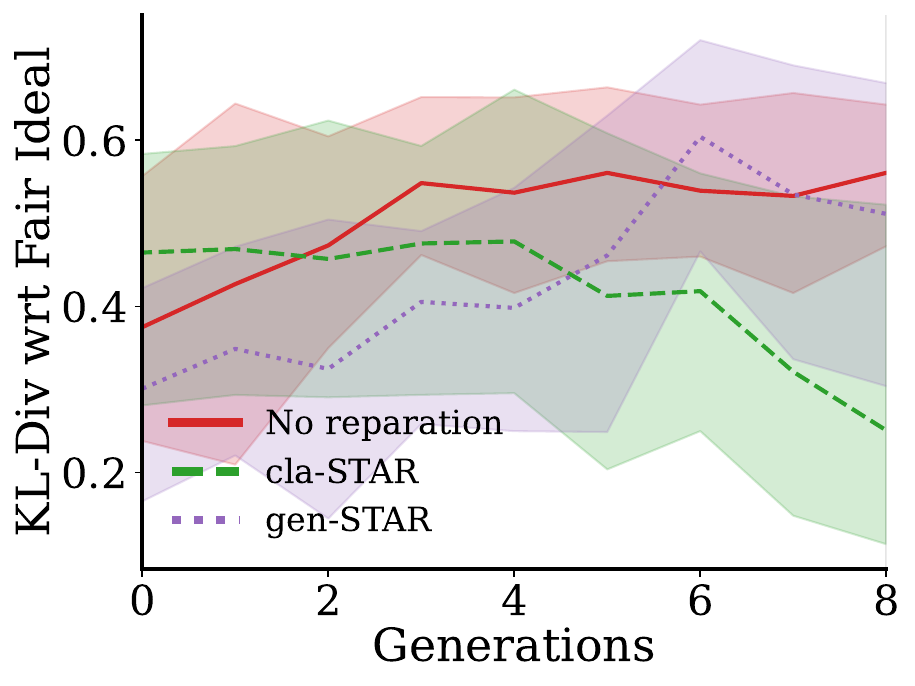} &
\includegraphics[width=0.22\textwidth]{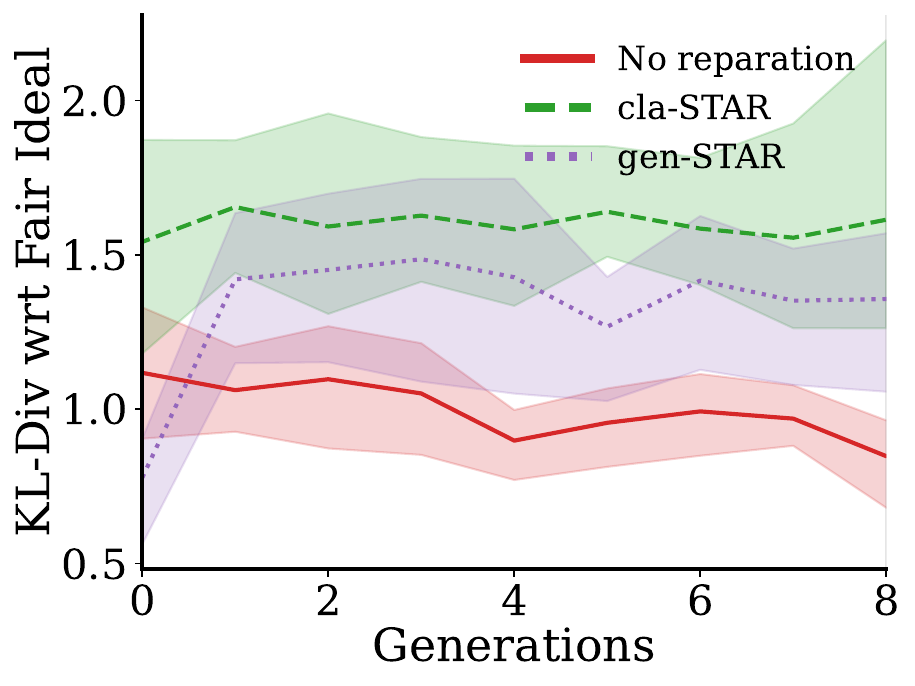} &
\includegraphics[width=0.22\textwidth]{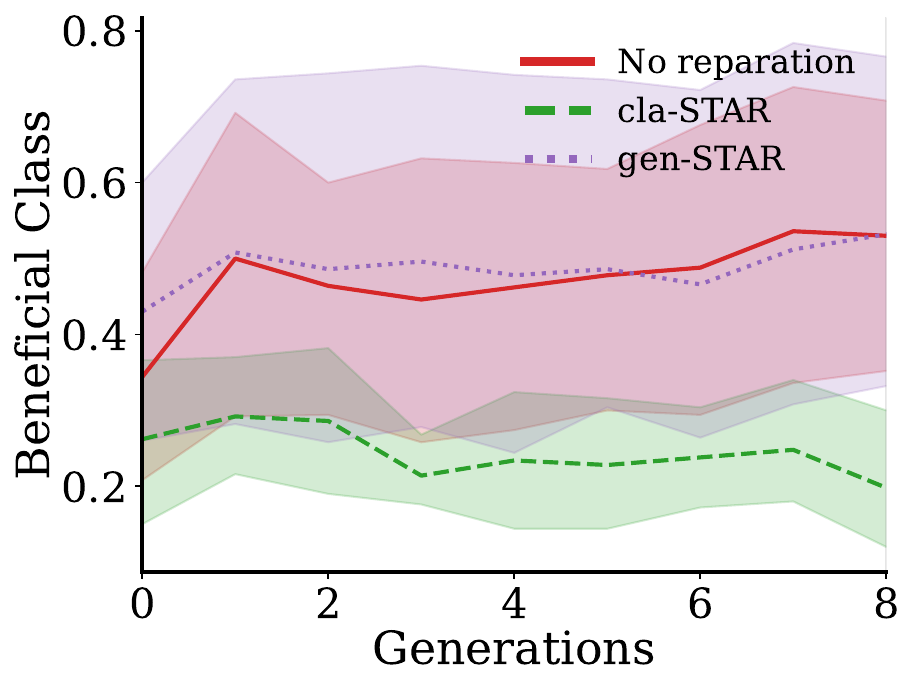} \\
\end{tabular}
\begin{tabular}{cc}
\includegraphics[width=0.22\textwidth]{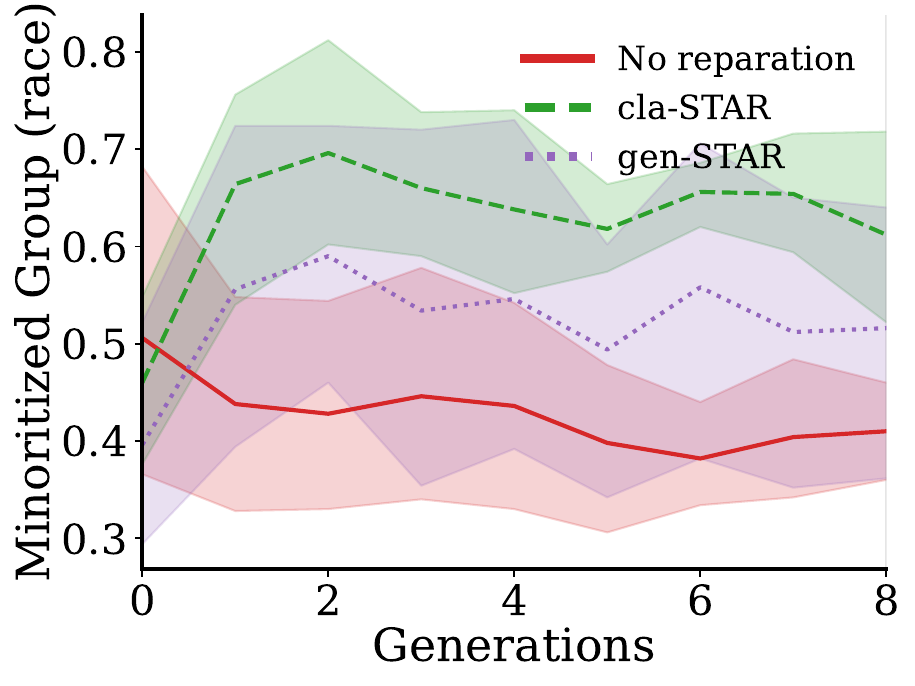} &
\includegraphics[width=0.22\textwidth]{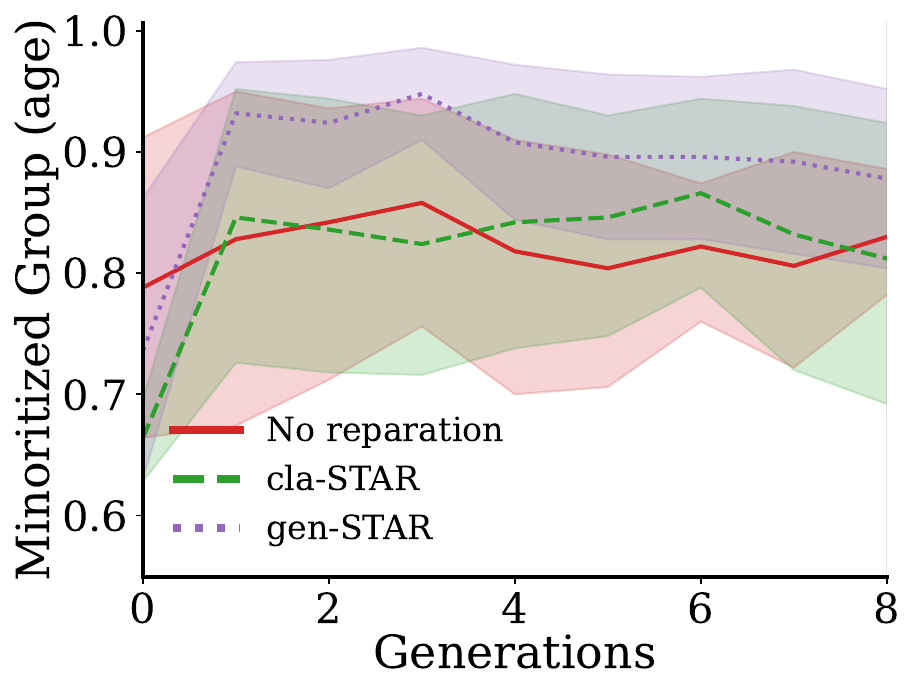}\\
\end{tabular}
\caption{\texttt{FairFace} results for \sgsc. \textit{Top:} shows accuracy, accuracy difference, demographic parity difference, and equalized odds difference. For the latter three, lower values are better. \textit{Center:} KL-Divergence between fairness ideal and classifiers, and between fairness ideal and generator \distrname, and the class balance. \textit{Bottom:} shows the group balance for sensitive attributes race and age by reporting the plurality race and age at each generation. Shading shows collapsed generations.
For this dataset, the annotator for race ($A_{S_1}$) as roughly 45\% utility on all groups aside from `white,' where it is 80\% accurate. This is despite the near-perfect balance between racial groups. This effects our annotations for race, which, alongside with a similarly biased lineage of generators leads to a lack of samples from non-white races. This thwarts \algname, resulting in similar unfairnesses as results without reparation. } 
\label{fig:FairFace_mc}
\end{figure}

We provide full suites of figures for \texttt{ColoredMNIST}, \texttt{ColoredSVHN}, \texttt{CelebA}, and \texttt{FairFace} on \sgsc. See Figures ~\ref{fig:ColoredMNIST_mc}, ~\ref{fig:SVHN_mc}, ~\ref{fig:celeba_mc} and ~\ref{fig:FairFace_mc}, respectively.

\subsubsection{\algname Batch Balances}\label{ssec:mc_strata}
\begin{figure}[ht]
    \centering
    \begin{tabular}{ccc}
       \includegraphics[width=.3\textwidth]{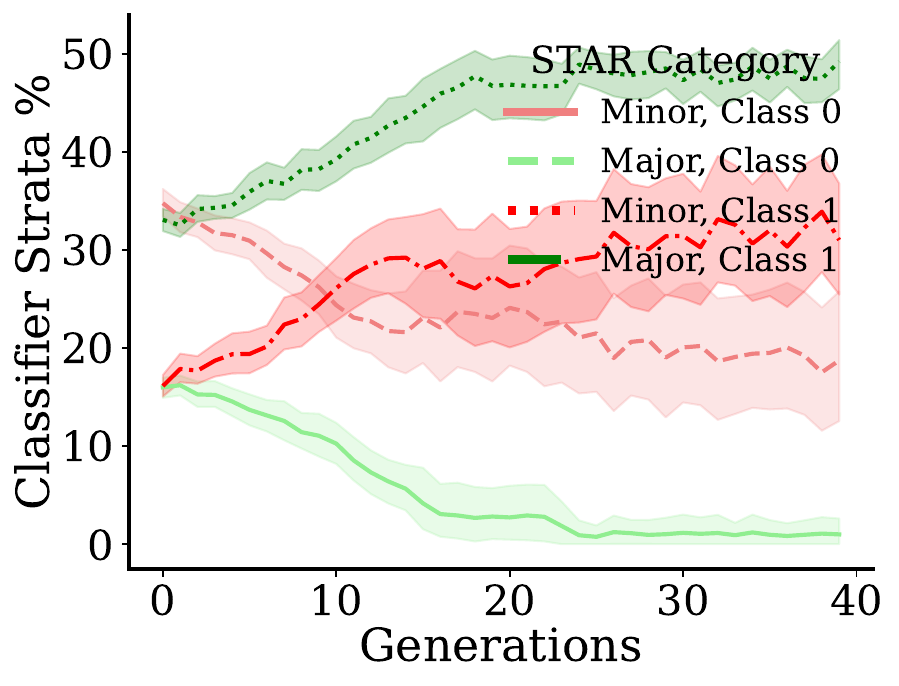}  & \includegraphics[width=.3\textwidth]{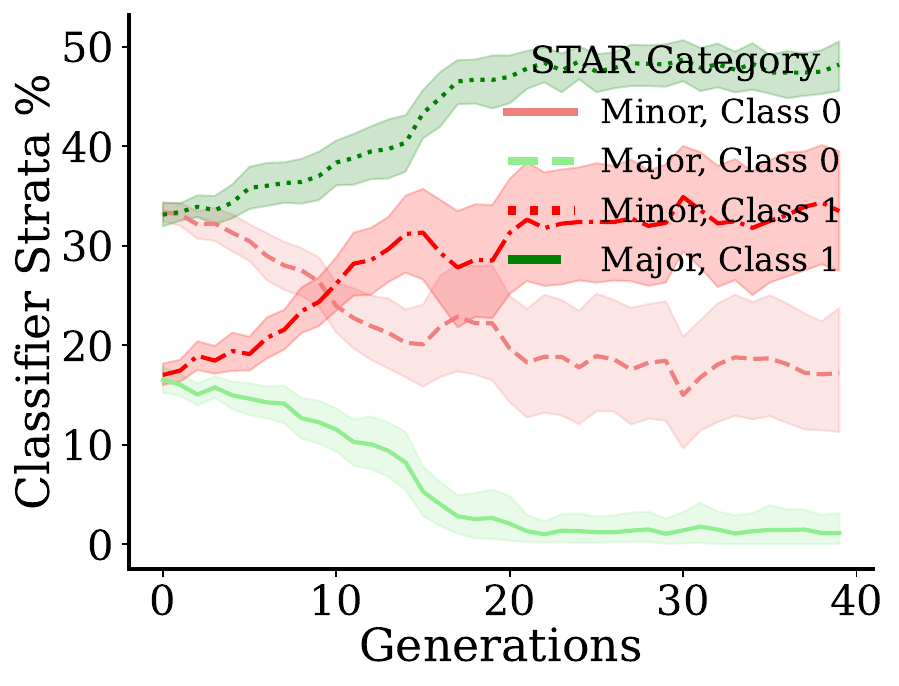} &
       \includegraphics[width=.3\textwidth]{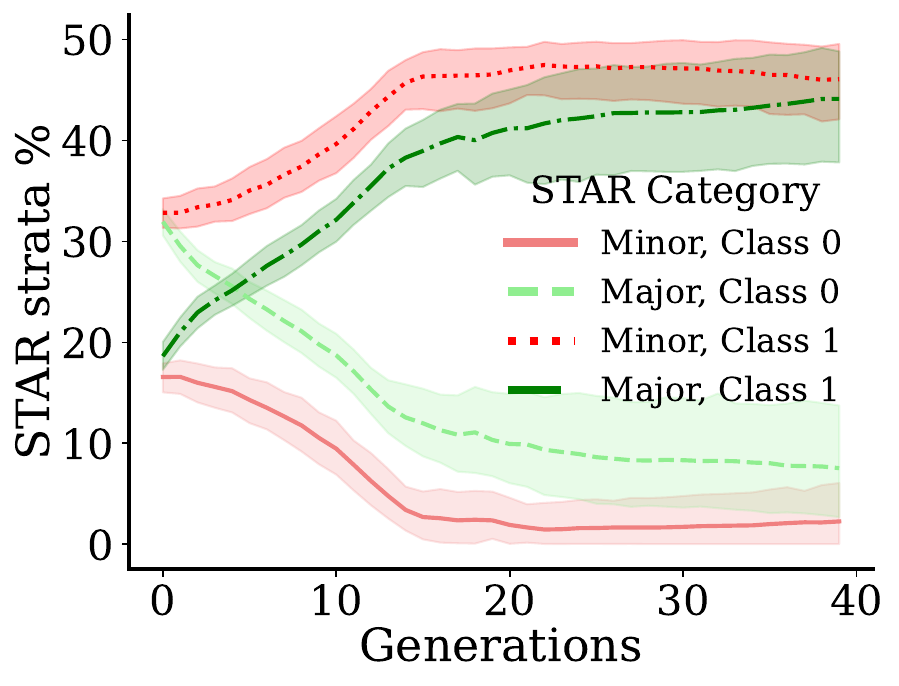} \\
       \includegraphics[width=.3\textwidth]{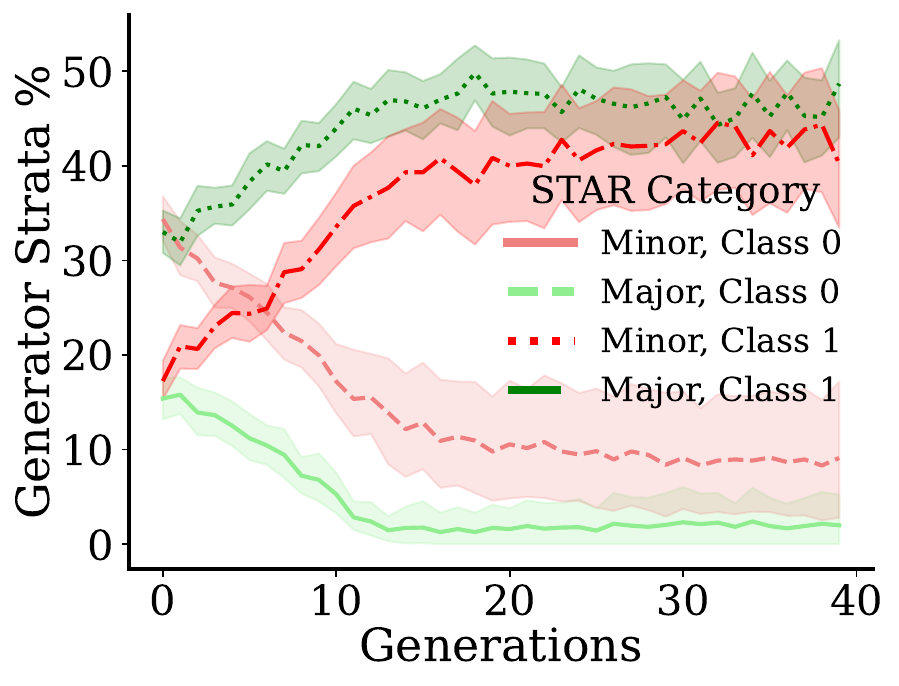}  & \includegraphics[width=.3\textwidth]{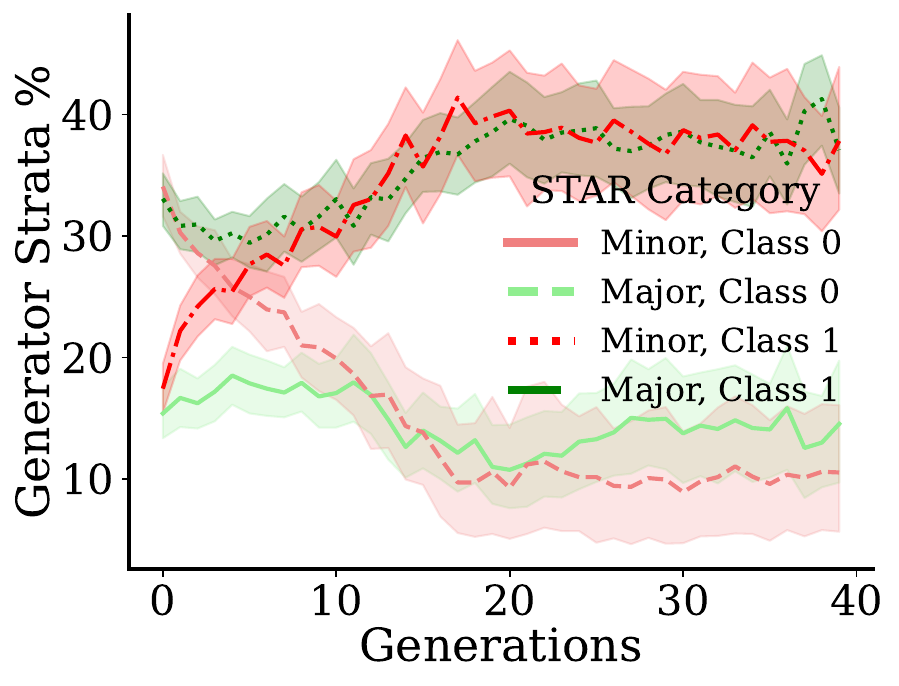} &
       \includegraphics[width=.3\textwidth]{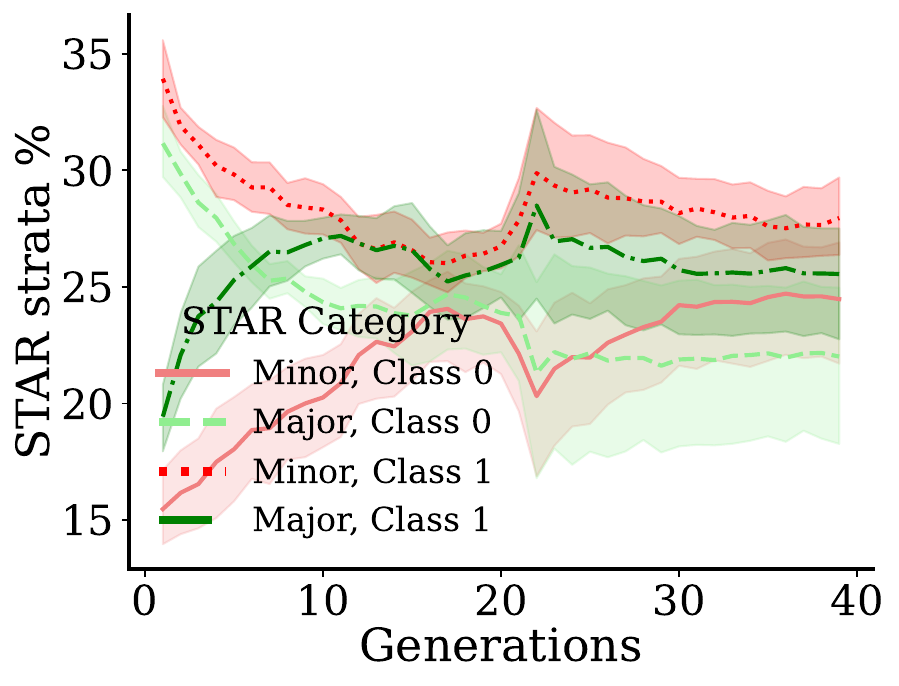} \\
    \end{tabular}
    \caption{\texttt{ColoredMNIST} \distrname balances for datasets in the \sgsc setting. \textit{Top:} \distrname resulting from classifiers without reparation, \distrname resulting from classifiers with \claalgname, and the \distrname used to train classifiers with \claalgname. 
    \textit{Bottom:} \distrname resulting from generators without reparation, \distrname resulting from generators with \genalgname, and the \distrname used to train generators with \genalgname. We see that \genalgname accomplishes a more balanced \distrname than \claalgname, but ultimately both are unable to get perfect balance due to model collapse casuing class imbalance.}
    \label{fig:cmnist_mc_stratas}
\end{figure}

\begin{figure}[ht]
    \centering
    \begin{tabular}{ccc}
       \includegraphics[width=.3\textwidth]{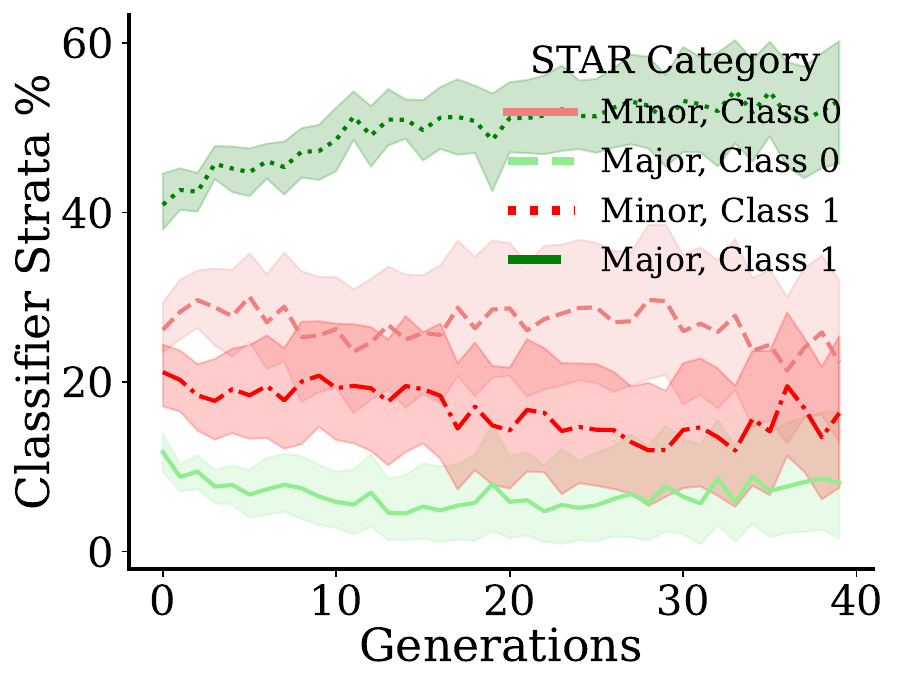}  & \includegraphics[width=.3\textwidth]{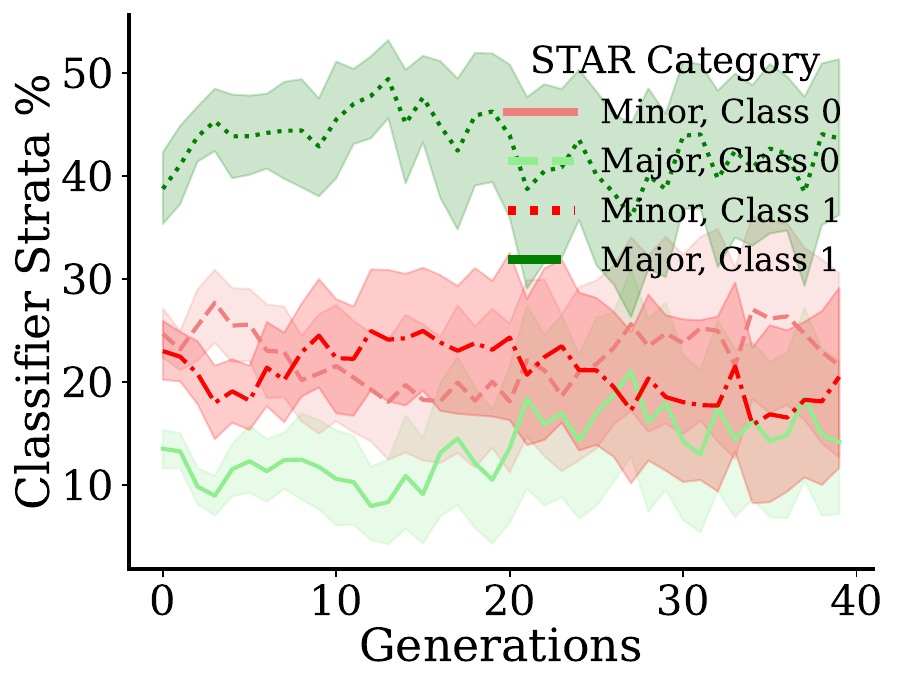} &
       \includegraphics[width=.3\textwidth]{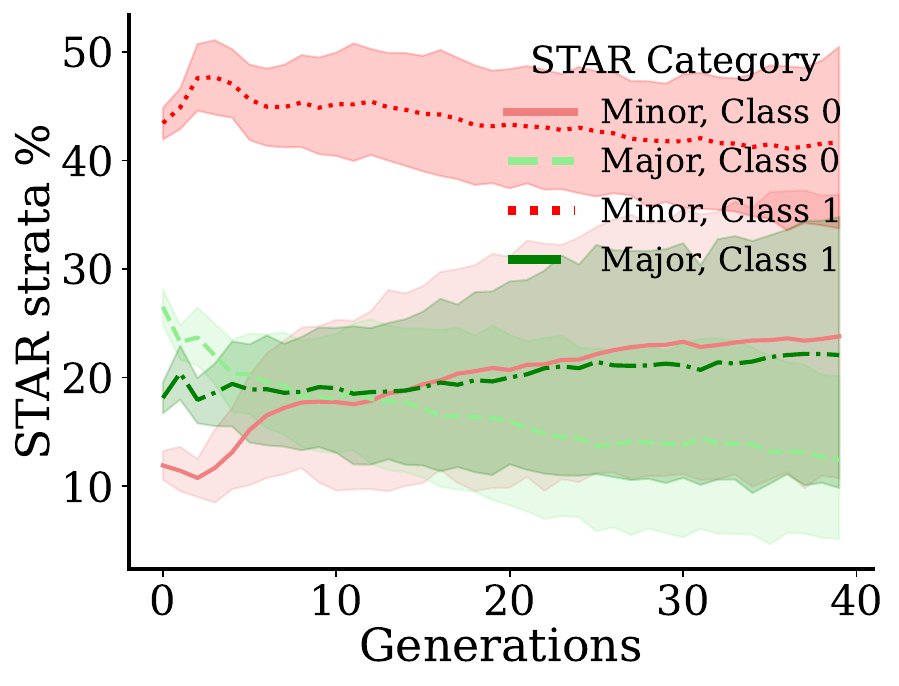} \\
       \includegraphics[width=.3\textwidth]{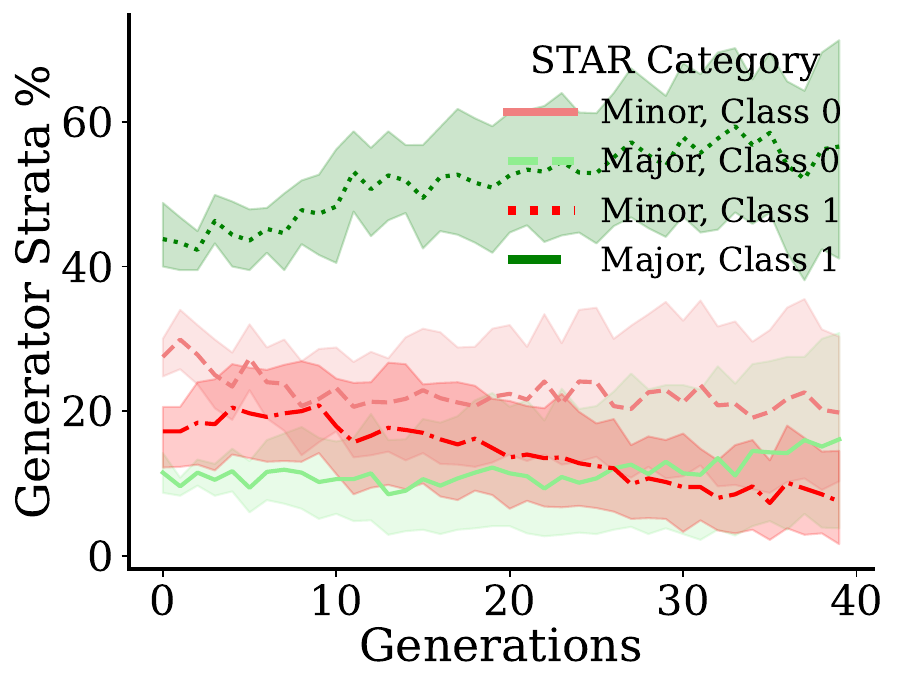}  & \includegraphics[width=.3\textwidth]{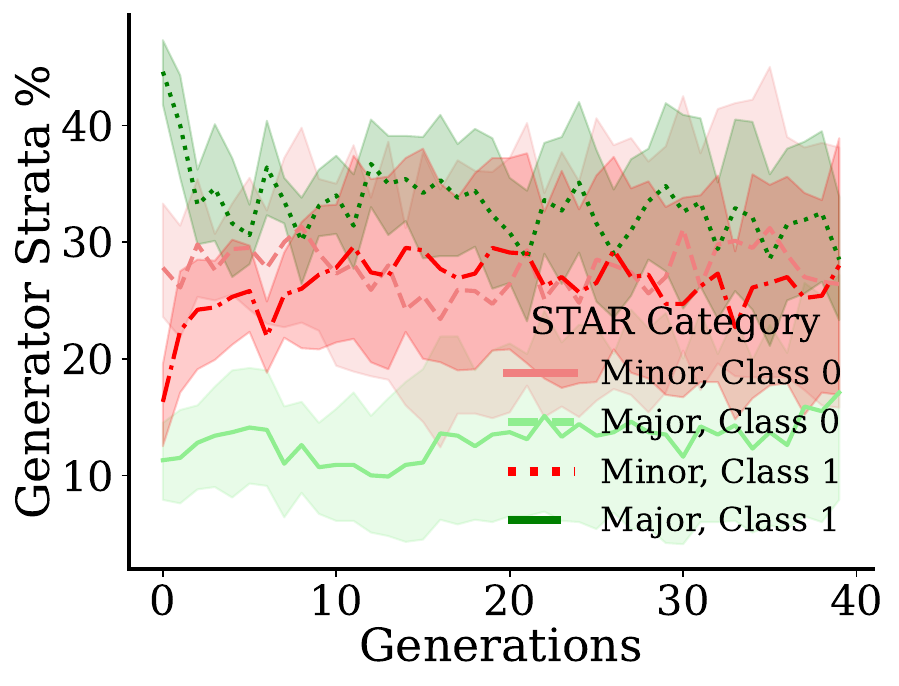} &
       \includegraphics[width=.3\textwidth]{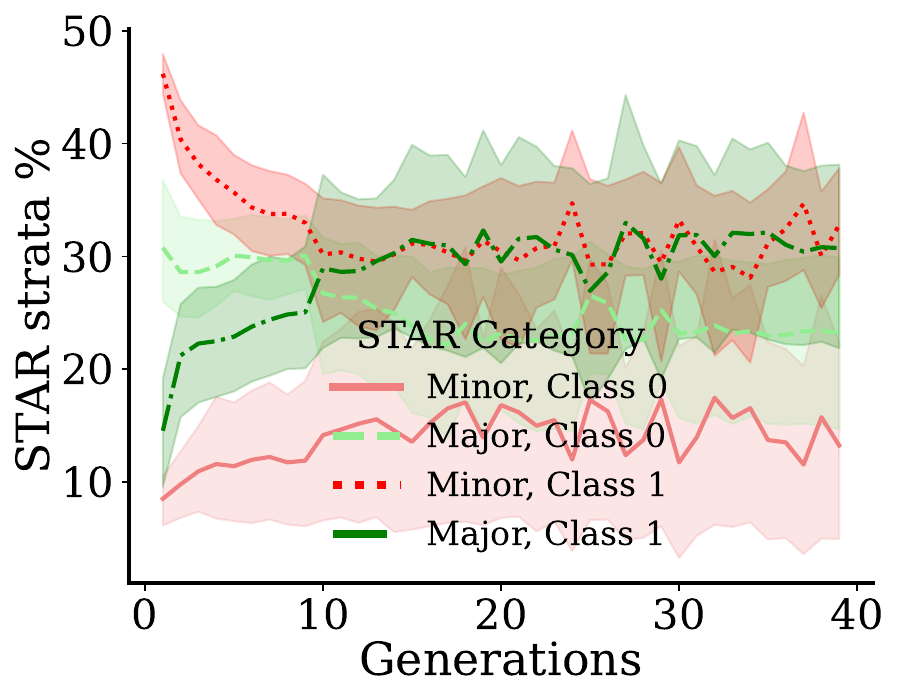} \\
    \end{tabular}
    \caption{\texttt{ColoredSVHN} \distrname balances for datasets in the \sgsc setting. \textit{Top:} \distrname resulting from classifiers without reparation, \distrname resulting from classifiers with \claalgname, and the \distrname used to train classifiers with \claalgname. 
    \textit{Bottom:} \distrname resulting from generators without reparation, \distrname resulting from generators with \genalgname, and the \distrname used to train generators with \genalgname. We can see that \genalgname results in slightly more balanced representation than does \claalgname. }
    \label{fig:svhn_mc_stratas}
\end{figure}

\begin{figure}[ht]
    \centering
    \begin{tabular}{ccc}
       \includegraphics[width=.3\textwidth]{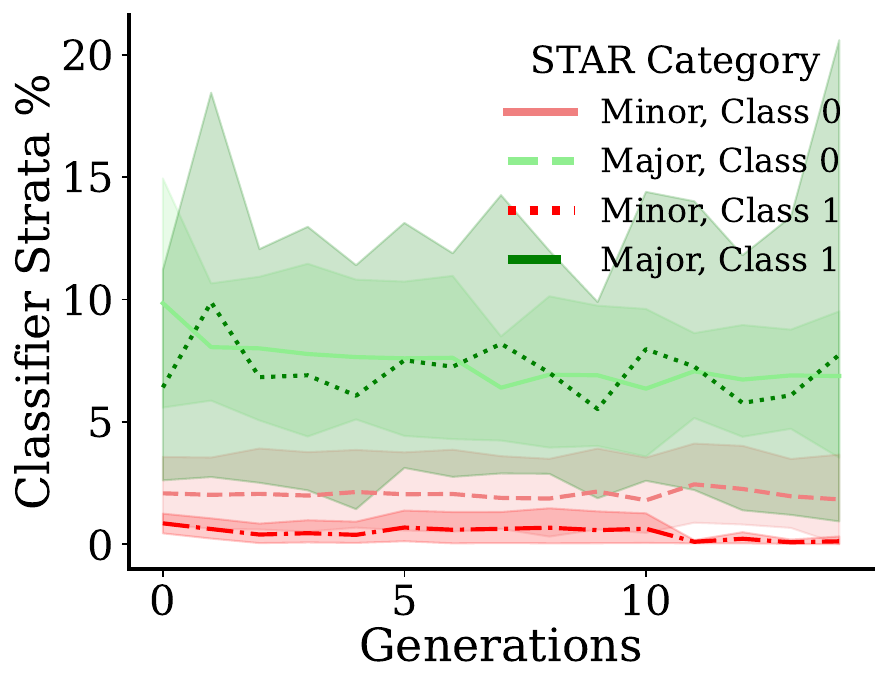}  & \includegraphics[width=.3\textwidth]{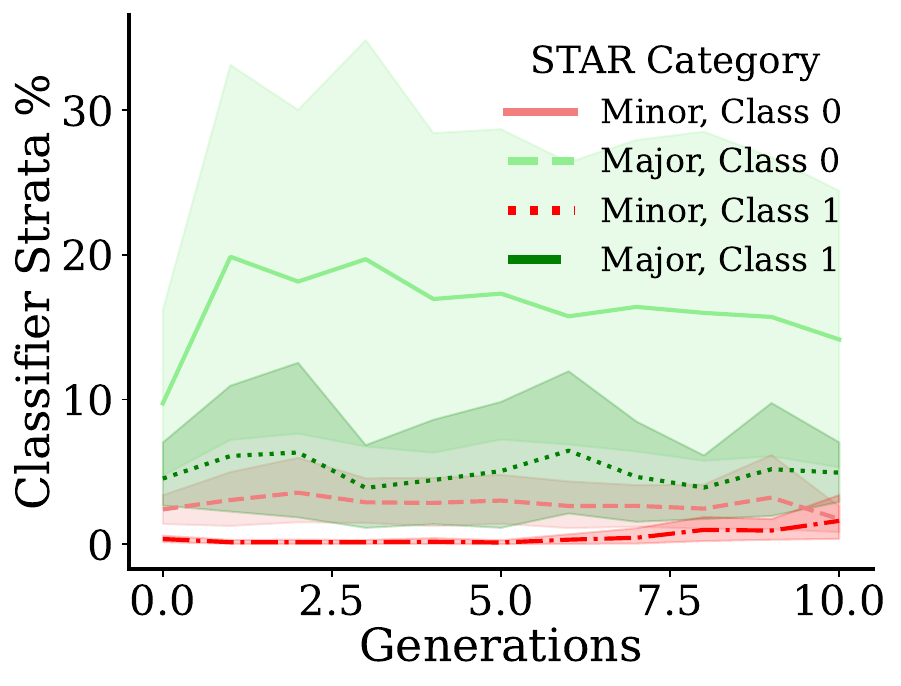} &
       \includegraphics[width=.3\textwidth]{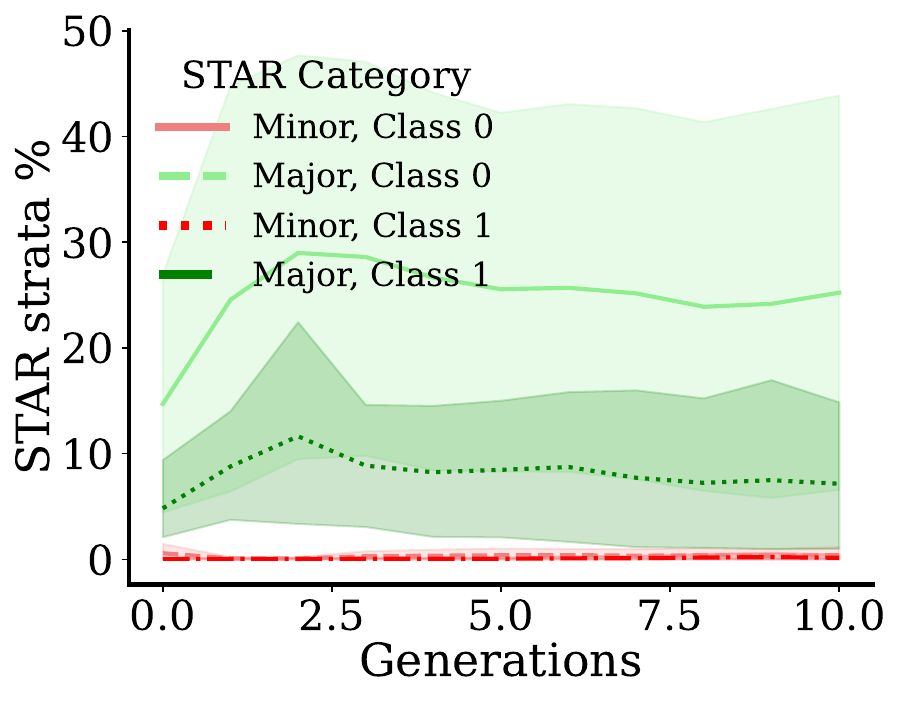} \\
       \includegraphics[width=.3\textwidth]{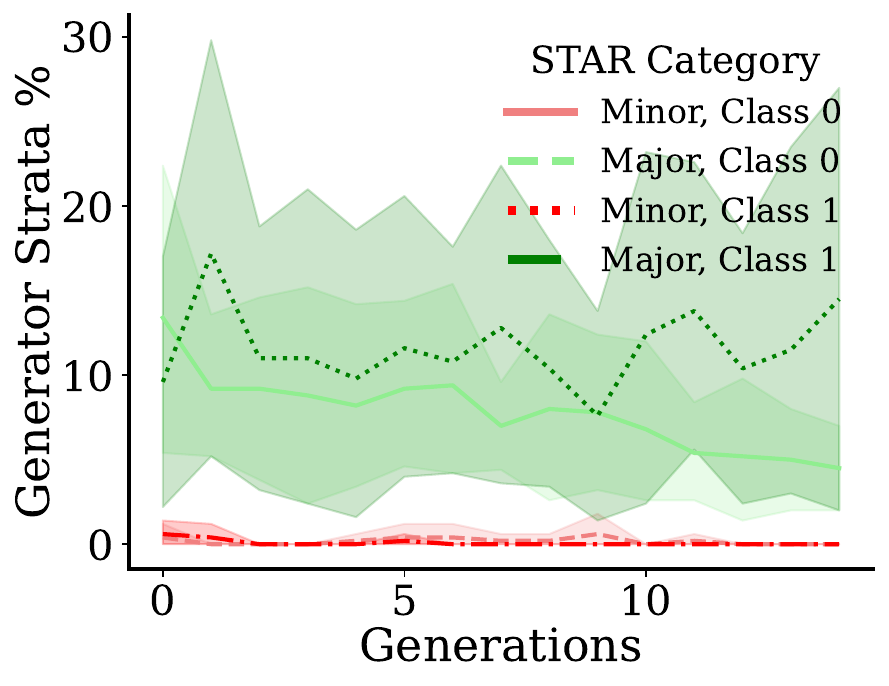}  & \includegraphics[width=.3\textwidth]{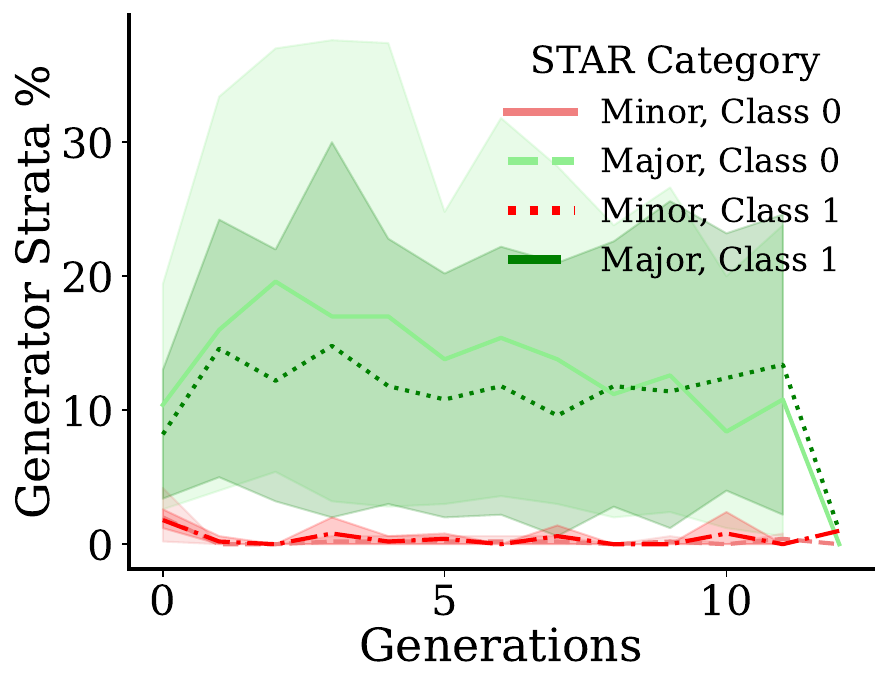} &
       \includegraphics[width=.3\textwidth]{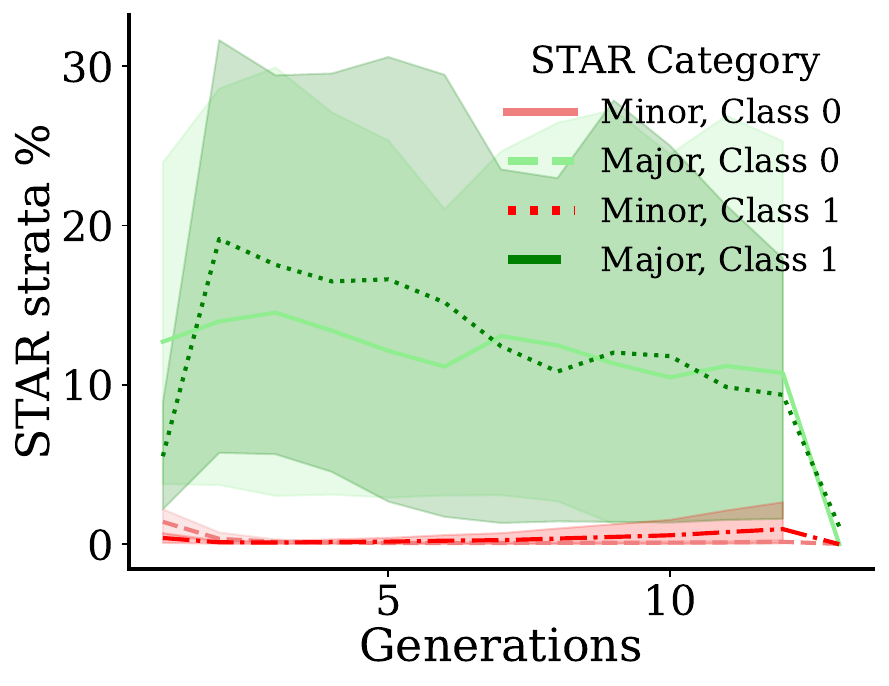} \\
    \end{tabular}
    \caption{\texttt{FairFace} \distrname balances for datasets in the \sgsc setting. \textit{Top:} \distrname resulting from classifiers without reparation, \distrname resulting from classifiers with \claalgname, and the \distrname used to train classifiers with \claalgname. 
    \textit{Bottom:} \distrname resulting from generators without reparation, \distrname resulting from generators with \genalgname, and the \distrname used to train generators with \genalgname. Neither AR simulation is able to achieve balance due to large racial disparities.}
    \label{fig:FairFace_mc_stratas}
\end{figure}

We compare the composition of the batches used when training the classifiers with and without \algname in the \sgsc setting. For \texttt{ColoredMNIST}, see \Cref{fig:cmnist_mc_stratas}, \texttt{ColoredSVHN}, see \Cref{fig:svhn_mc_stratas}, and \texttt{FairFace}, see \Cref{fig:FairFace_mc_stratas}.
These figures show the \distrname \algname uses to train models, the resulting model \distrname, and the \distrname of models trained without  reparation. 
In \texttt{FairFace}, the batches are mostly older white males, and sometimes younger white non-males; the least populated categories are usually people from the Indian, Middle Eastern, South East Asian, Black, and Hispanic/Latino races. 

\section{Relative Performances of MIDS}\label{app:rel_mids}
If the model trainer is unaware of MIDS occurring over time, they may see only the relative performances (\textit{i.e.,} performance of generation $i$ measured \textit{w.r.t.} generation $i-1$) of each generation compared to its prior generation. In this case, when each generation of models is trained to have relatively high performance, it may look as though the models are performing well, though not when compared to the original data distribution. This may lead to overstating the model's performance, which for the FML metrics results in fairwashing the model due to inadequate validation and testing~\citep{aivodji2019fairwashing}. For \texttt{ColoredMNIST} and \texttt{ColoredSVHN}, we report results on the testing set for the `actual' results (classifiers measured against the testing set) and for the relative results (classifiers measured against the previous generation's classifier predictions on the testing set inputs). We choose not to present these two graphs on the same plots to prevent confusion as they measure two different properties.

\begin{figure}[ht]
\centering
\begin{tabular}{cc}
\includegraphics[width=0.4\textwidth]{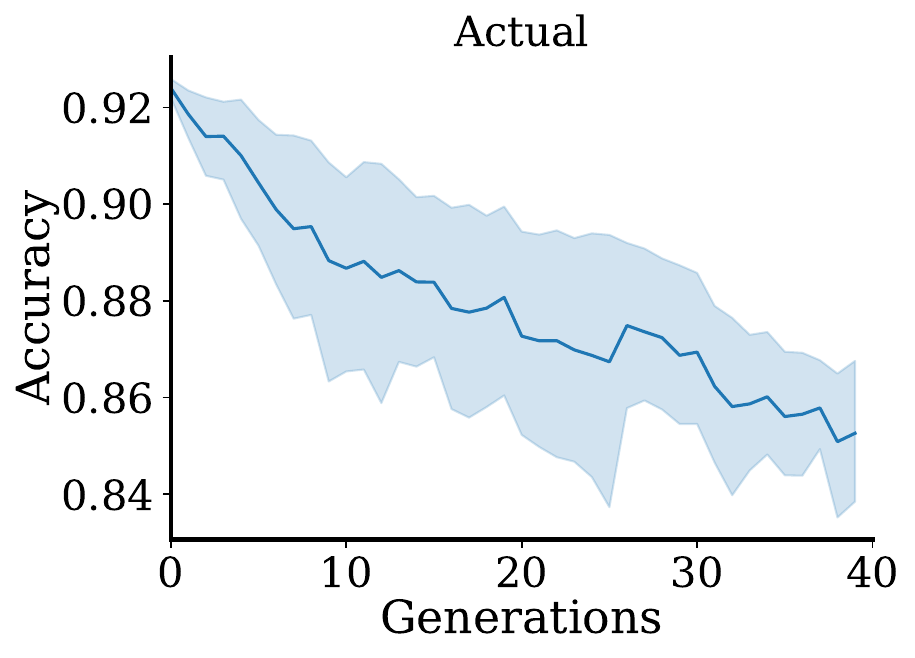} &
\includegraphics[width=0.4\textwidth]{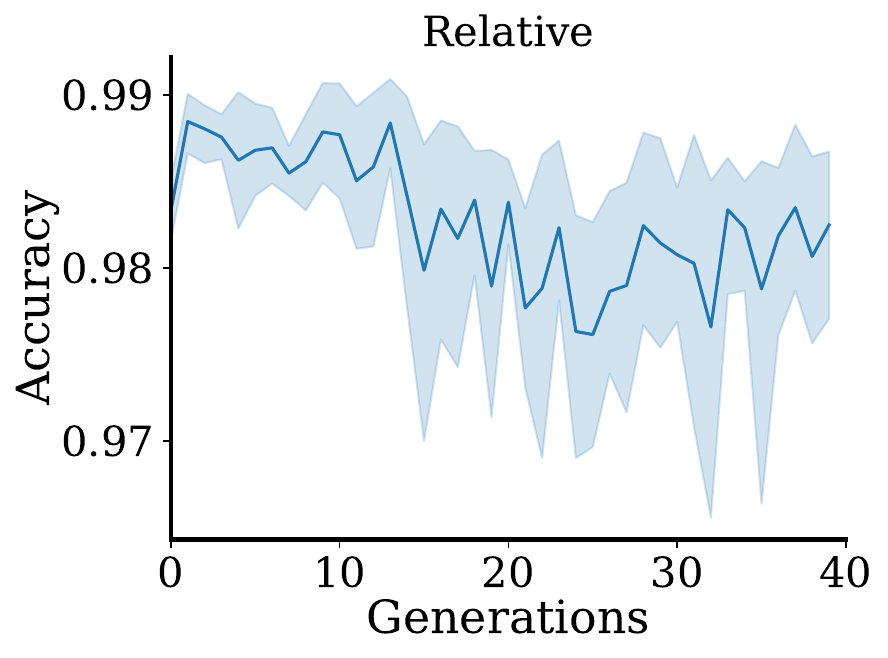} 
\\
\includegraphics[width=0.4\textwidth]{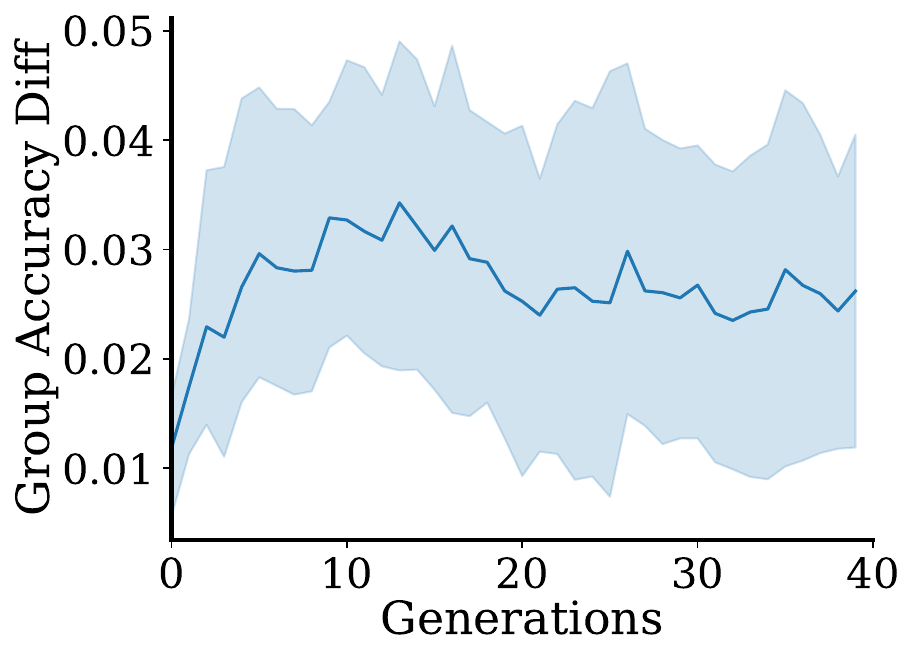} &
\includegraphics[width=0.4\textwidth]{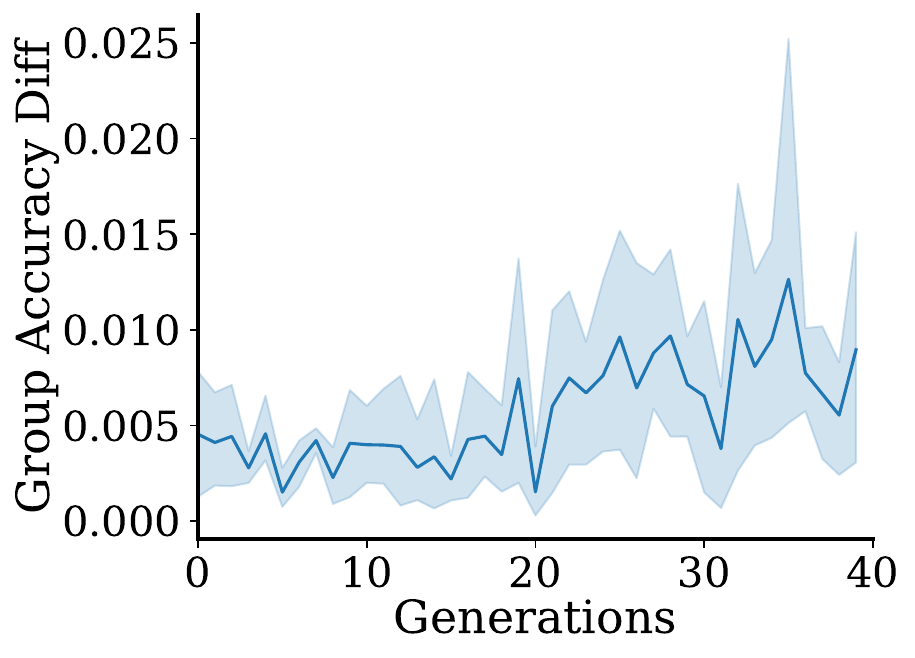} 
\\
\includegraphics[width=0.4\textwidth]{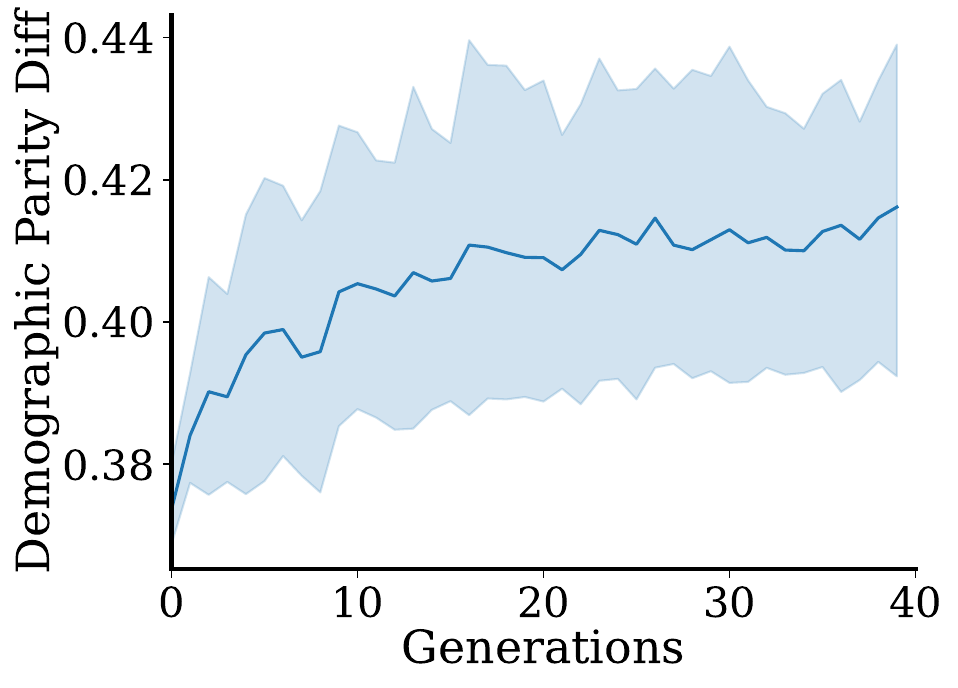} &
\includegraphics[width=0.4\textwidth]{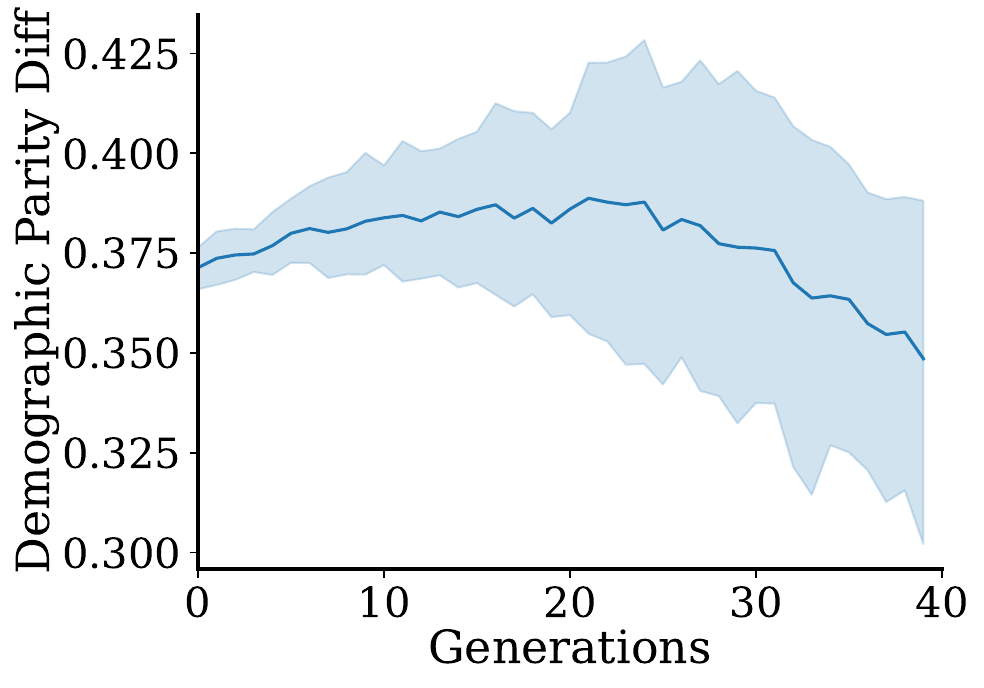} 
\\
\includegraphics[width=0.4\textwidth]{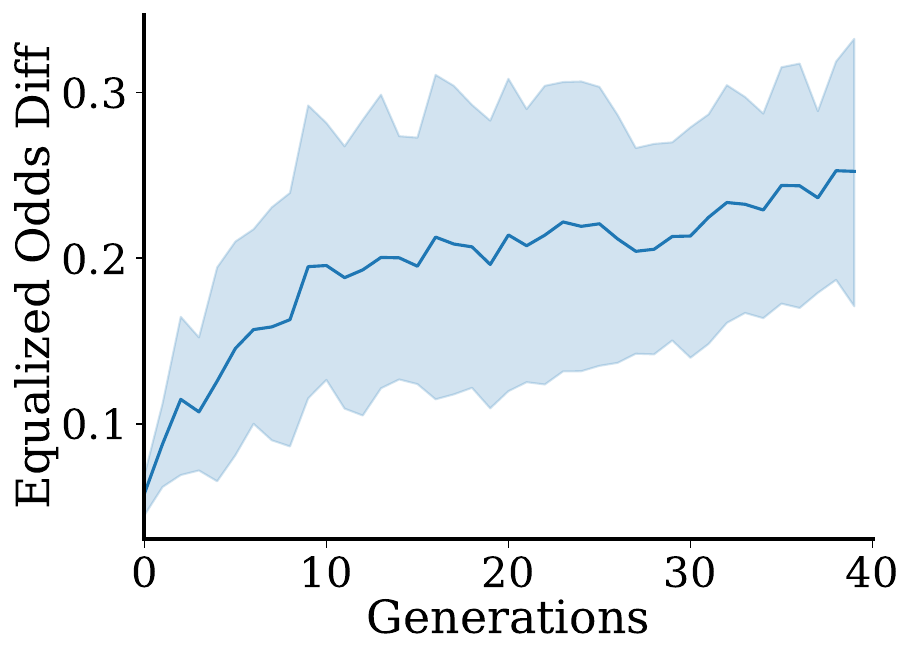} &
\includegraphics[width=0.4\textwidth]{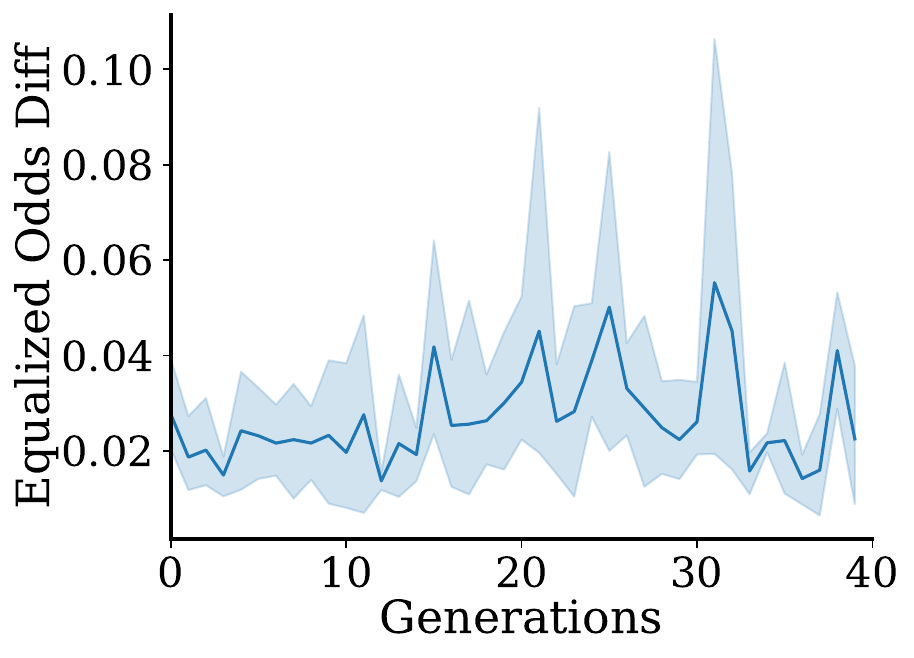} 
\\
\end{tabular}
\caption{Accuracy, demographic parity difference, and equalized odds difference in \seqc on \texttt{ColoredMNIST}. Higher accuracy is better, but for the FML metrics higher difference is worse. \textit{Left:} Performances on the test set. \textit{Right:} Relative performances between models. The model quality of accuracy and equalized odds in the relative performances is far higher than the actual results. In equalized odds, this shows that even if small unfairnesses were tolerated over while training each classifier, the result over time accrues high unfairness compared to the original testing set.}
\label{fig:rel_cmnist_nomc}
\end{figure}

\begin{figure}[ht]
\centering
\begin{tabular}{cc}
\includegraphics[width=0.4\textwidth]{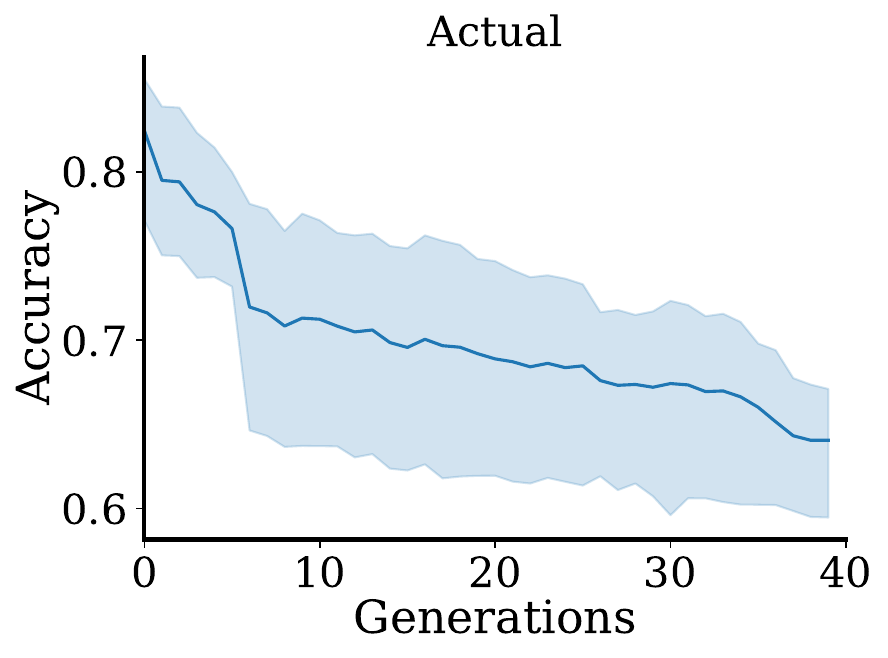} &
\includegraphics[width=0.4\textwidth]{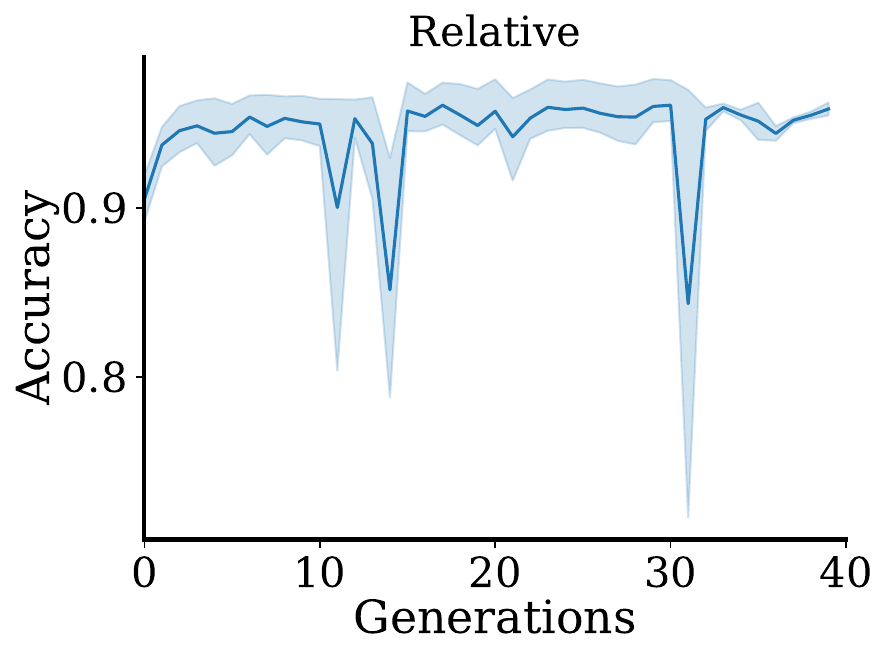} 
\\
\includegraphics[width=0.4\textwidth]{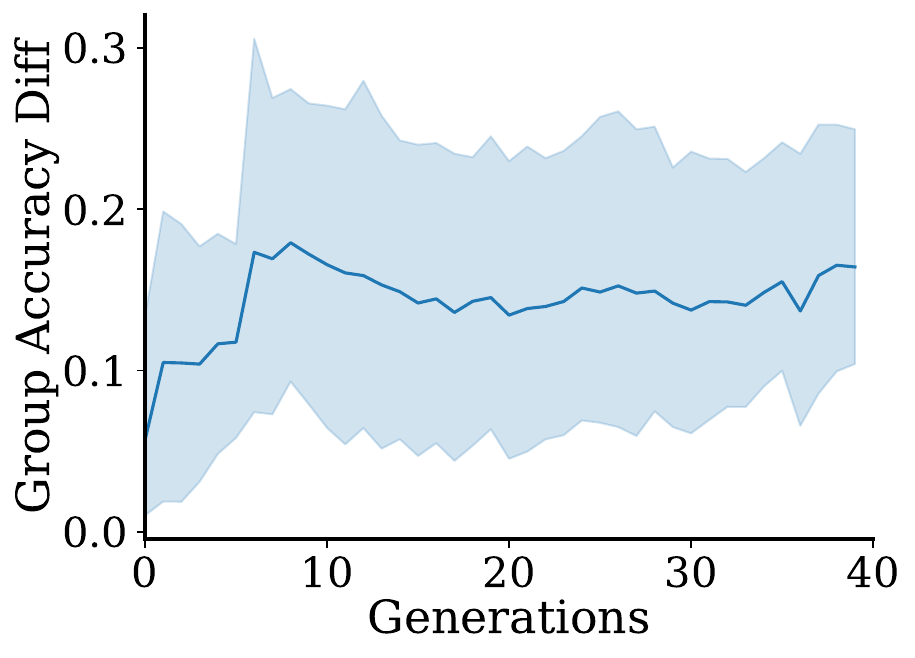} &
\includegraphics[width=0.4\textwidth]{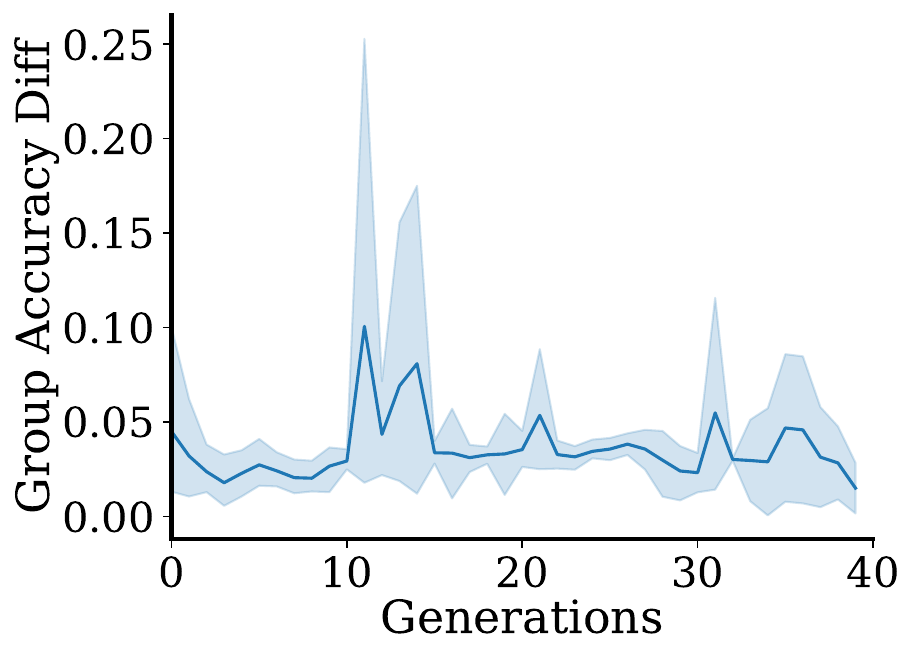} 
\\
\includegraphics[width=0.4\textwidth]{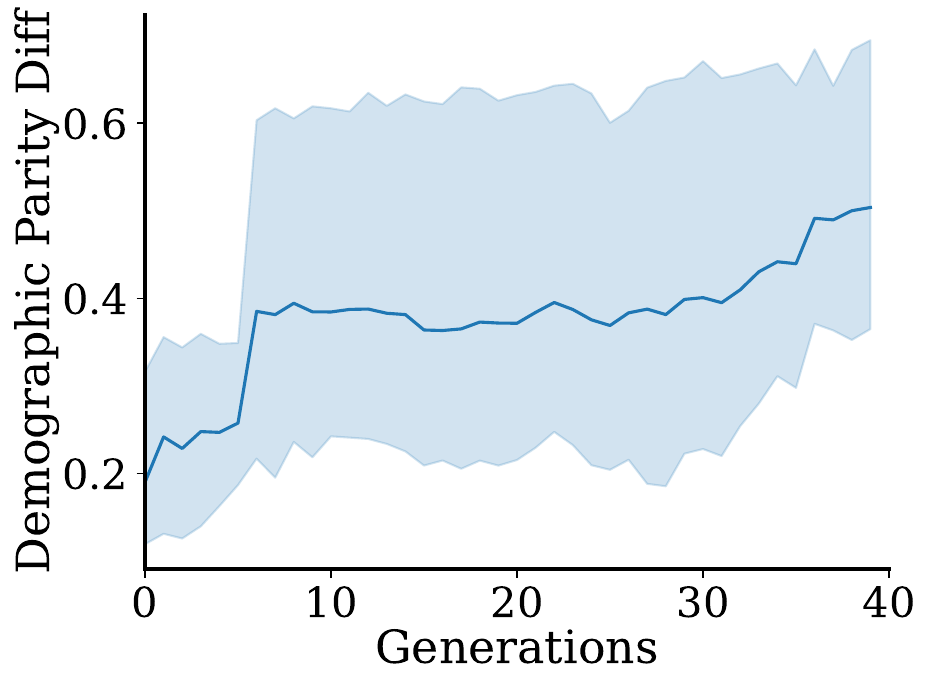} &
\includegraphics[width=0.4\textwidth]{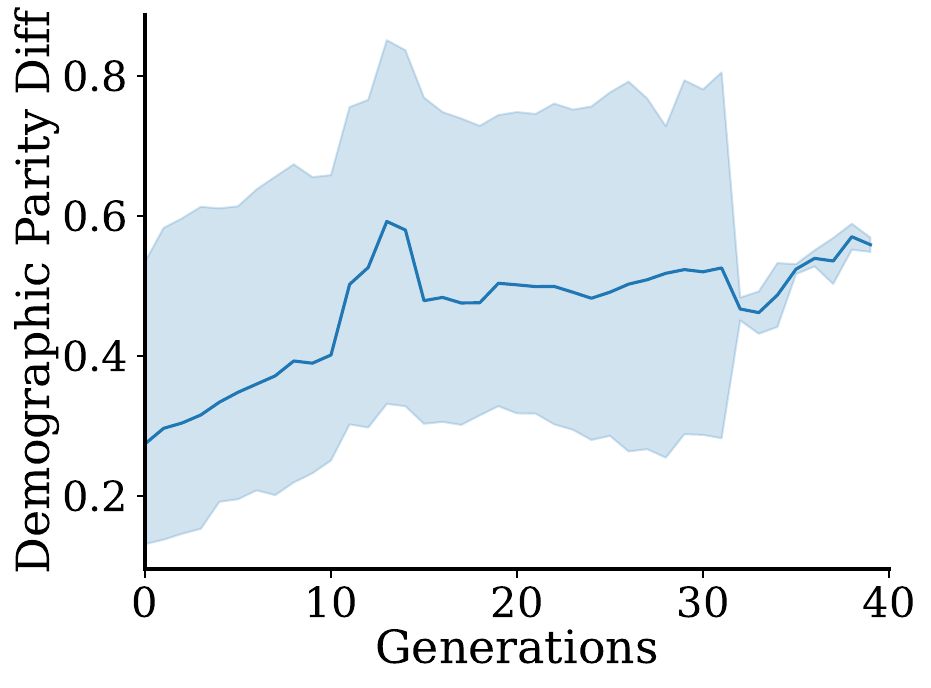} 
\\
\includegraphics[width=0.4\textwidth]{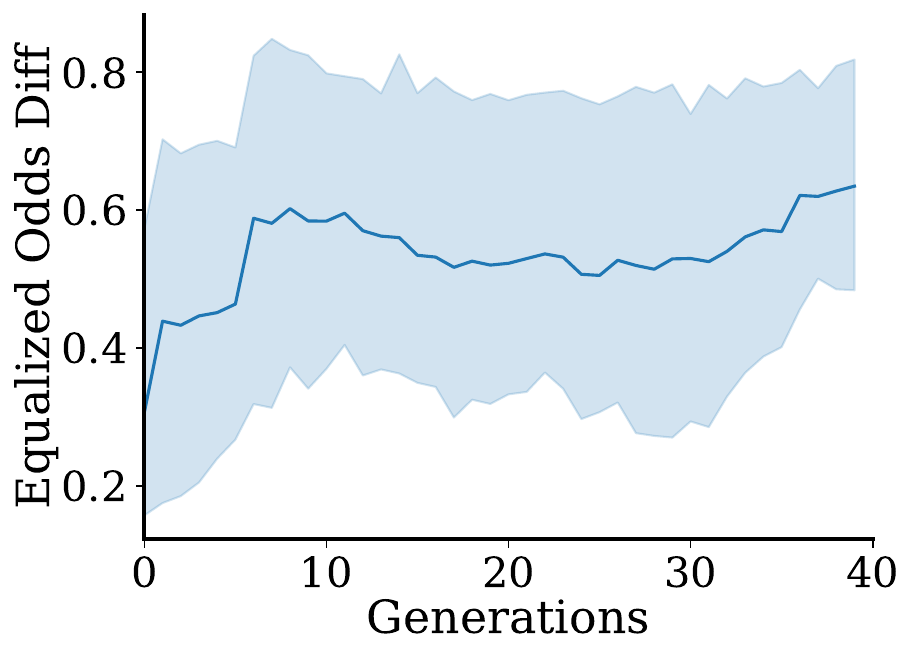} &
\includegraphics[width=0.4\textwidth]{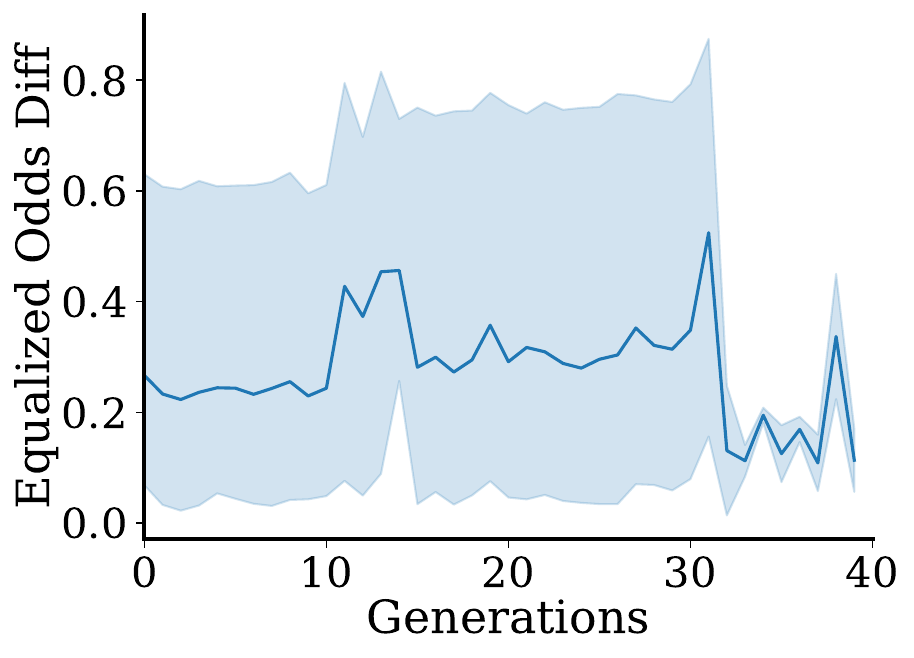} 
\\
\end{tabular}
\caption{Accuracy, demographic parity difference, and equalized odds difference in \seqc on \texttt{ColoredSVHN}. Higher accuracy is better, but for the FML metrics higher difference is worse. \textit{Left:} Results on the test set. \textit{Right:} Relative performances between models. }
\label{fig:rel_svhn_nomc}
\end{figure}

\begin{figure}[ht]
\centering
\begin{tabular}{cc}
\includegraphics[width=0.35\textwidth]{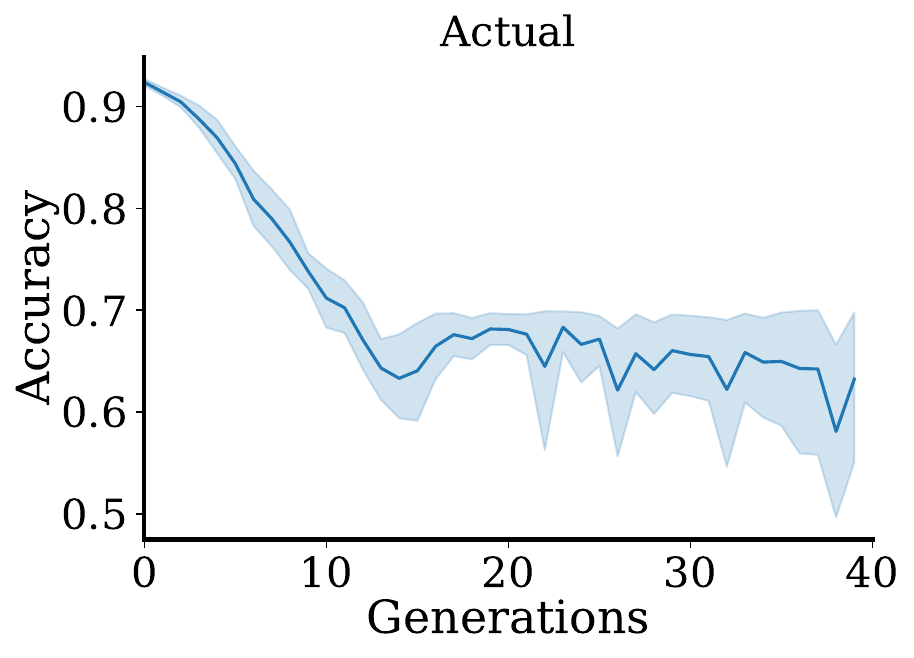} &
\includegraphics[width=0.35\textwidth]{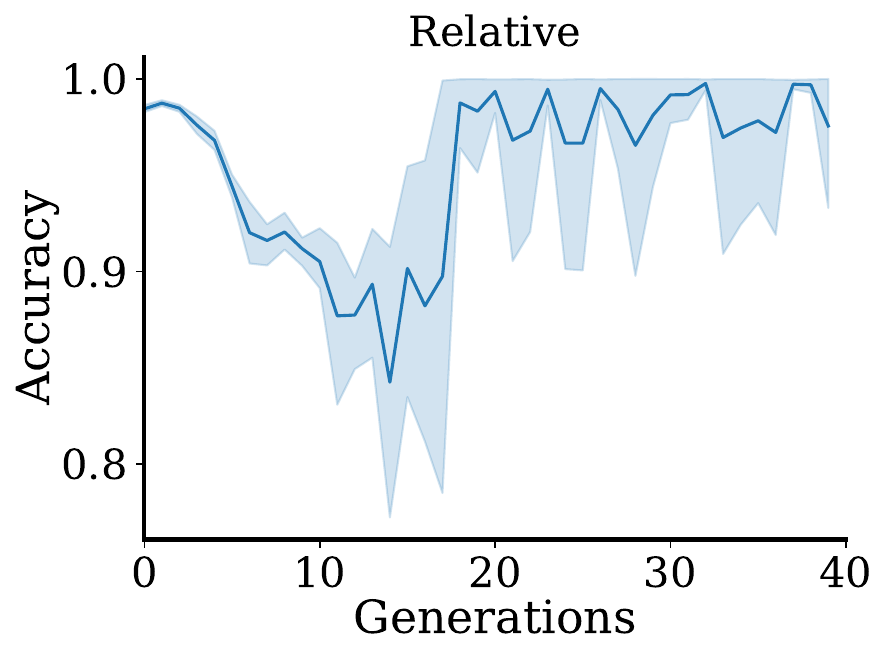} 
\\
\includegraphics[width=0.35\textwidth]{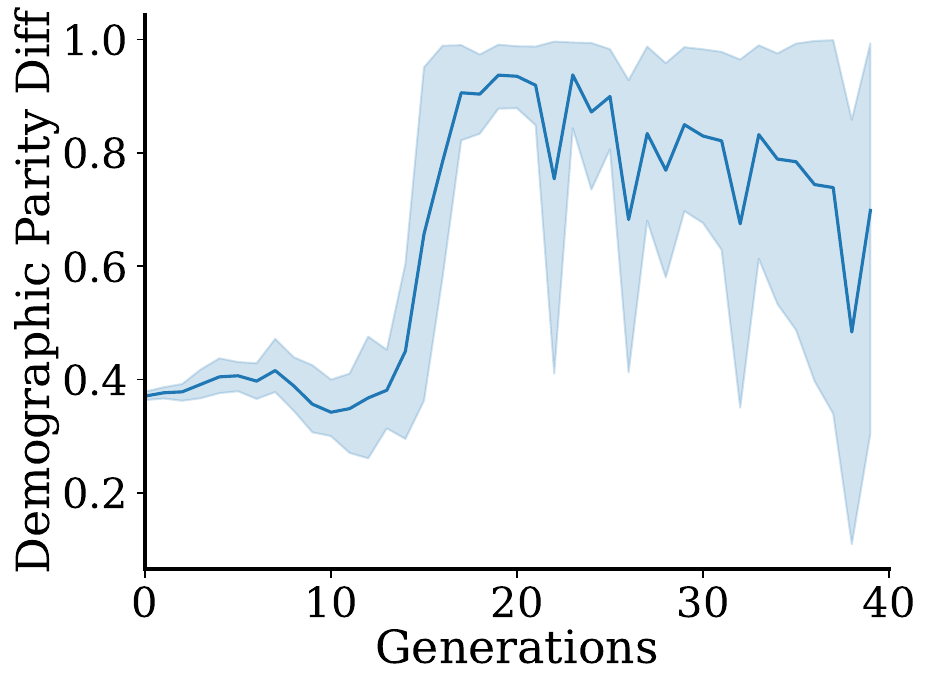} &
\includegraphics[width=0.35\textwidth]{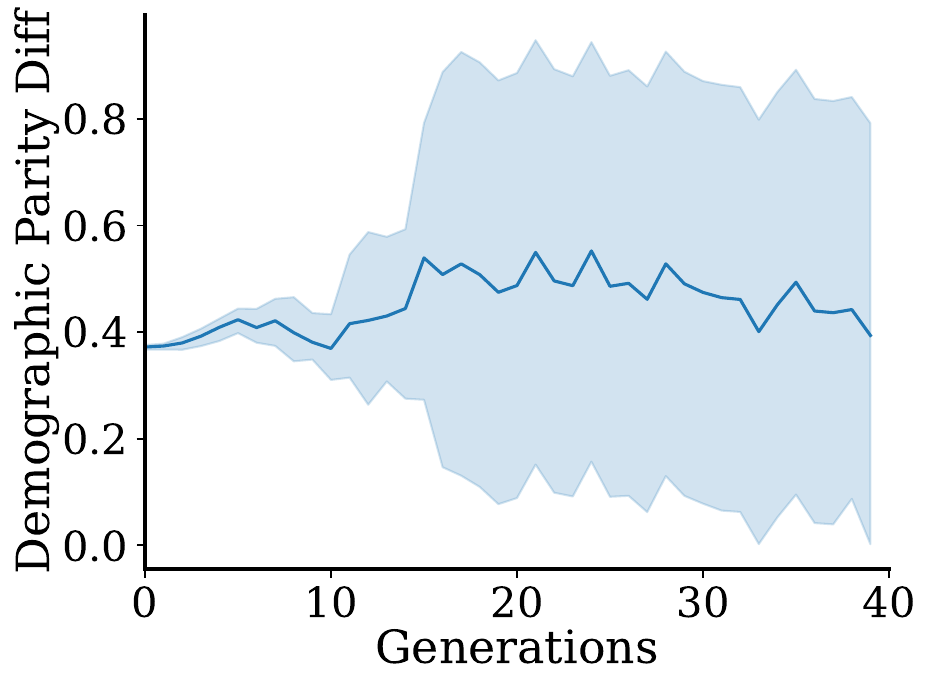} 
\\
\includegraphics[width=0.35\textwidth]{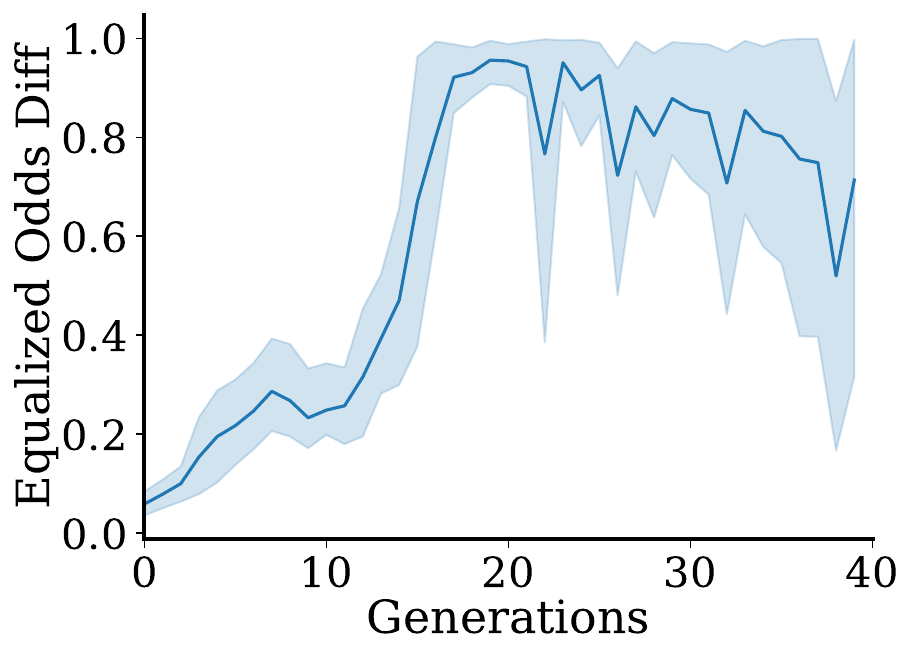} &
\includegraphics[width=0.35\textwidth]{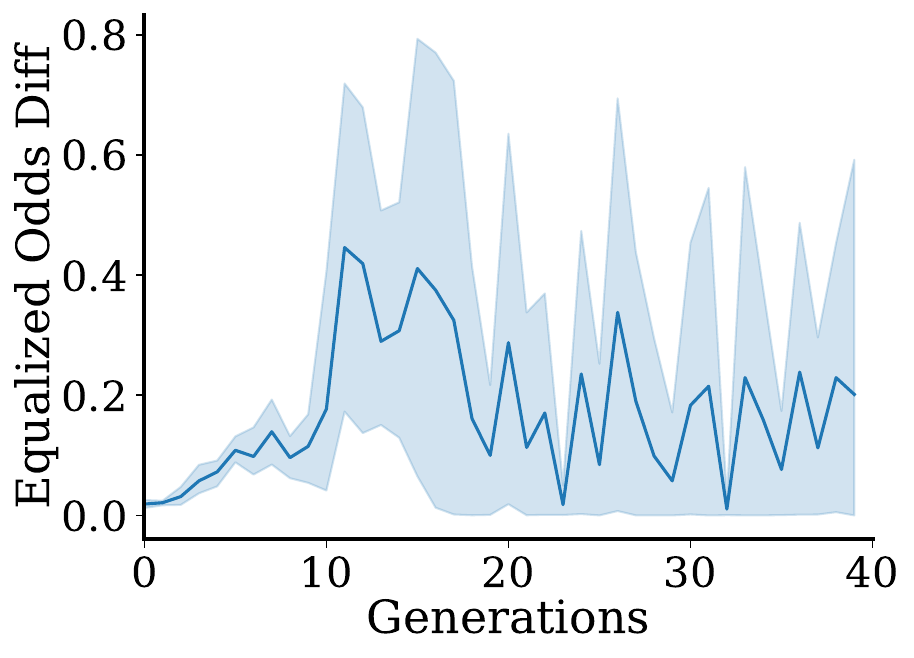} 
\\
\includegraphics[width=0.35\textwidth]{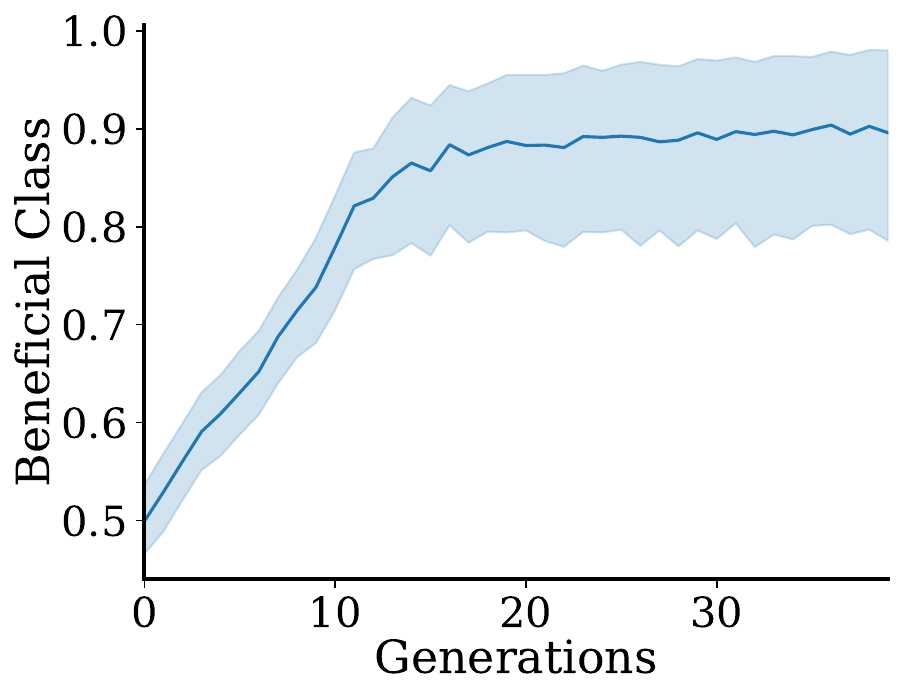} &
\includegraphics[width=0.35\textwidth]{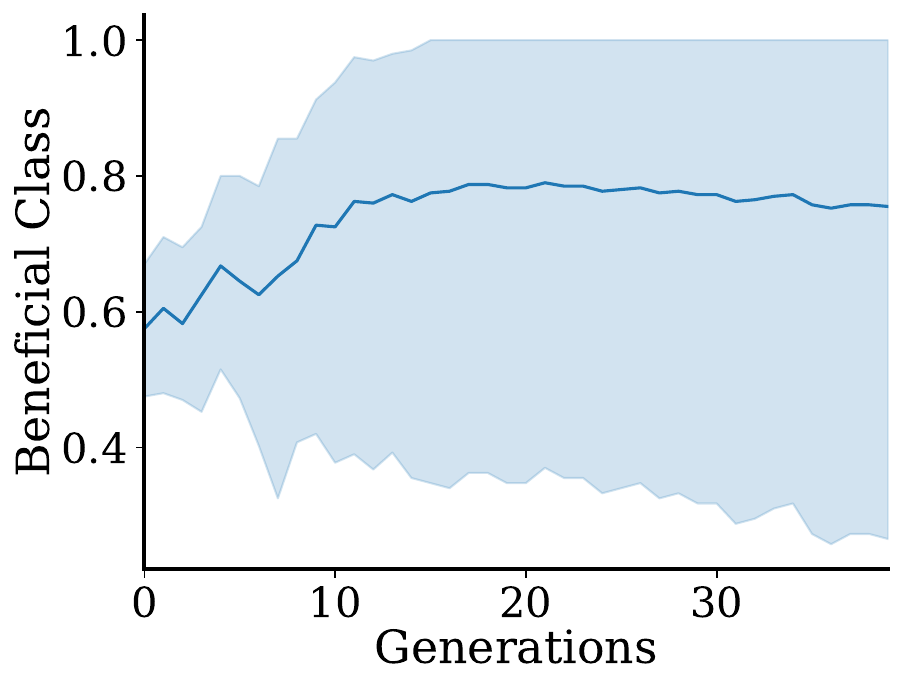} 
\\
\includegraphics[width=0.35\textwidth]{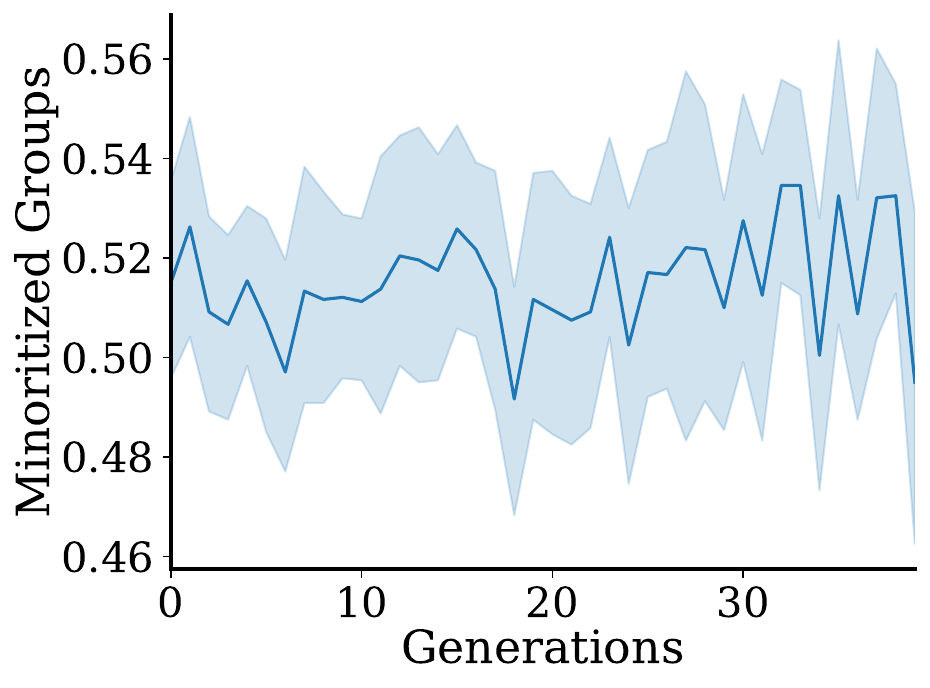} &
\includegraphics[width=0.35\textwidth]{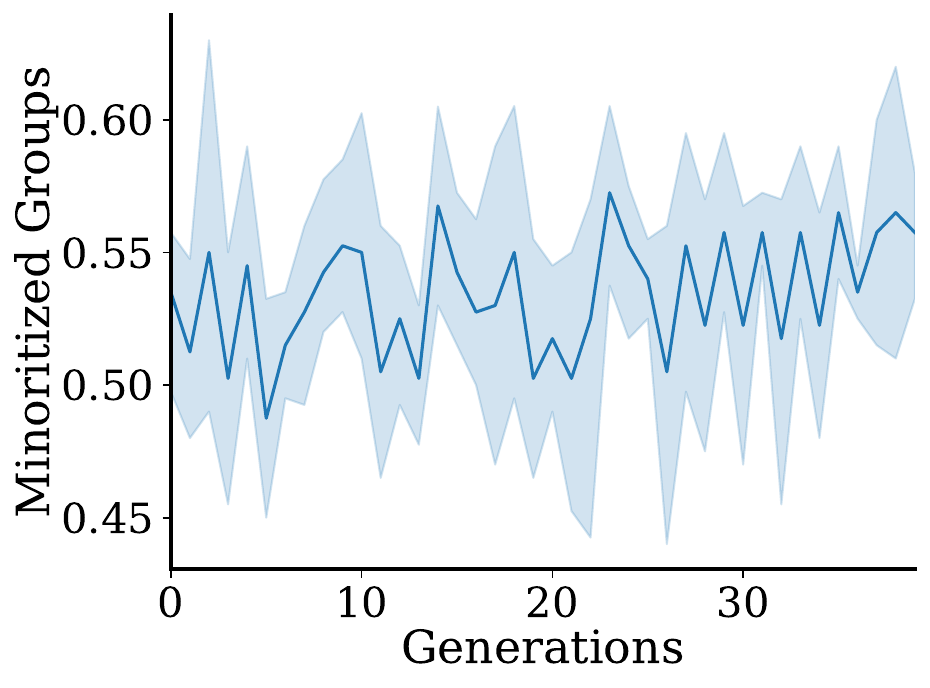} 
\\
\end{tabular}
\caption{Accuracy, demographic parity difference, equalized odds difference, and rates of the beneficial class and minoritized group in \sgsc on \texttt{ColoredMNIST}. Higher accuracy is better, but for the FML metrics higher difference is worse. \textit{Left:} Results on the test set. \textit{Right:} Relative performances between models.}
\label{fig:rel_cmnist_mc}
\end{figure}

\begin{figure}[ht]
\centering
\begin{tabular}{cc}
\includegraphics[width=0.35\textwidth]{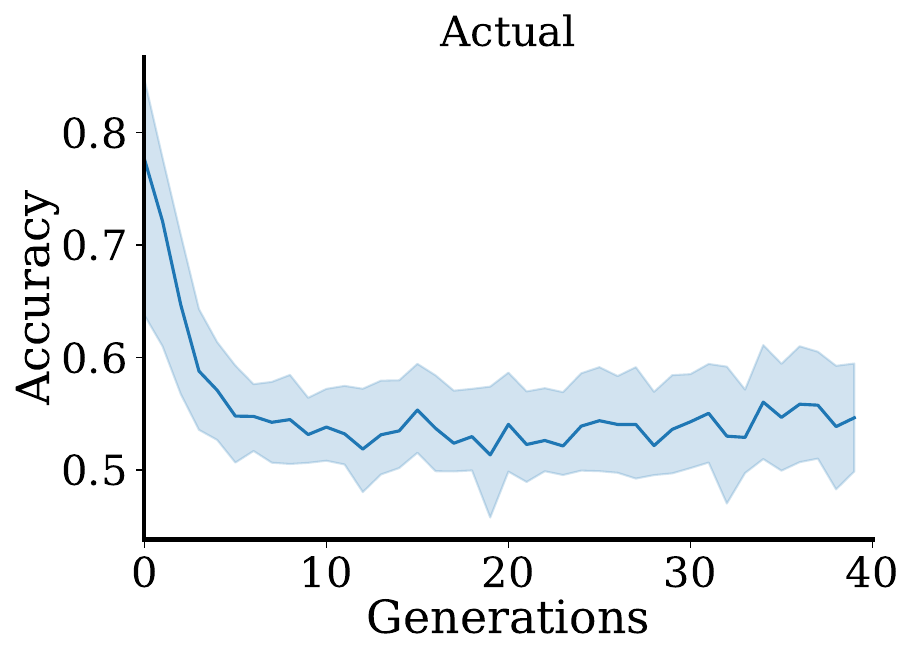} &
\includegraphics[width=0.35\textwidth]{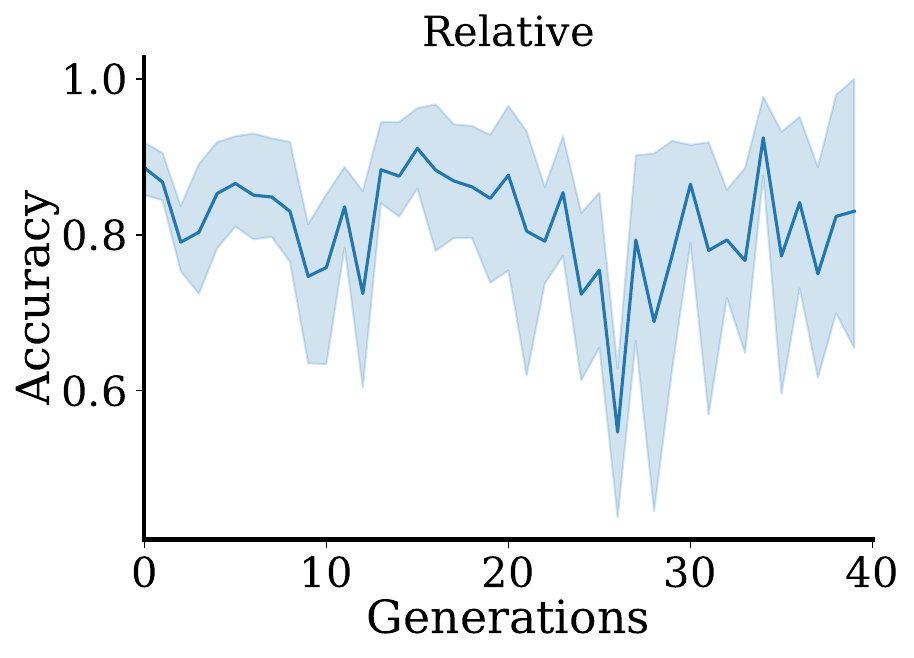} 
\\
\includegraphics[width=0.35\textwidth]{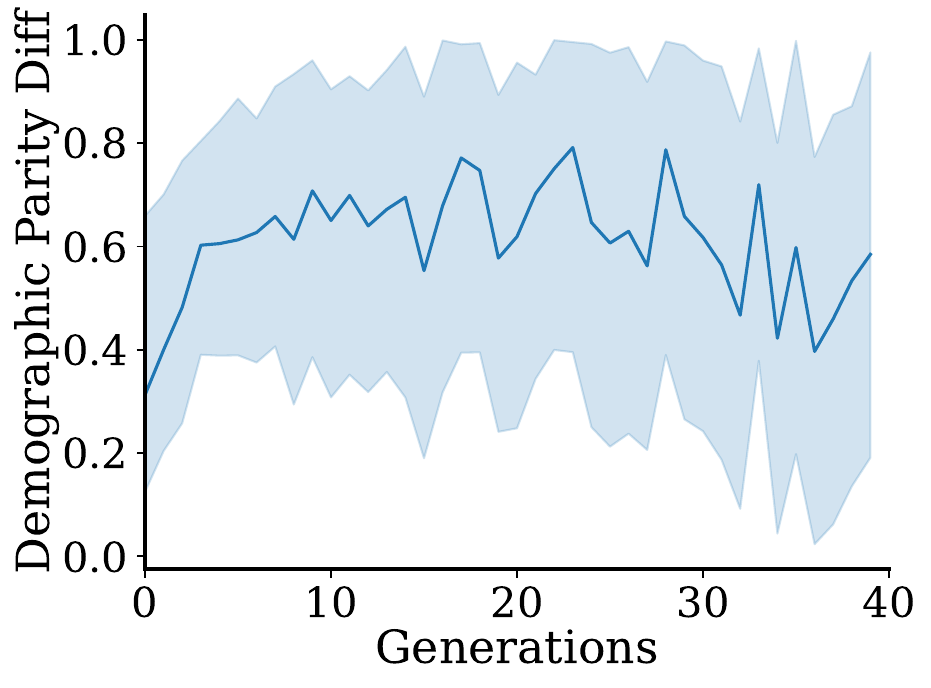} &
\includegraphics[width=0.35\textwidth]{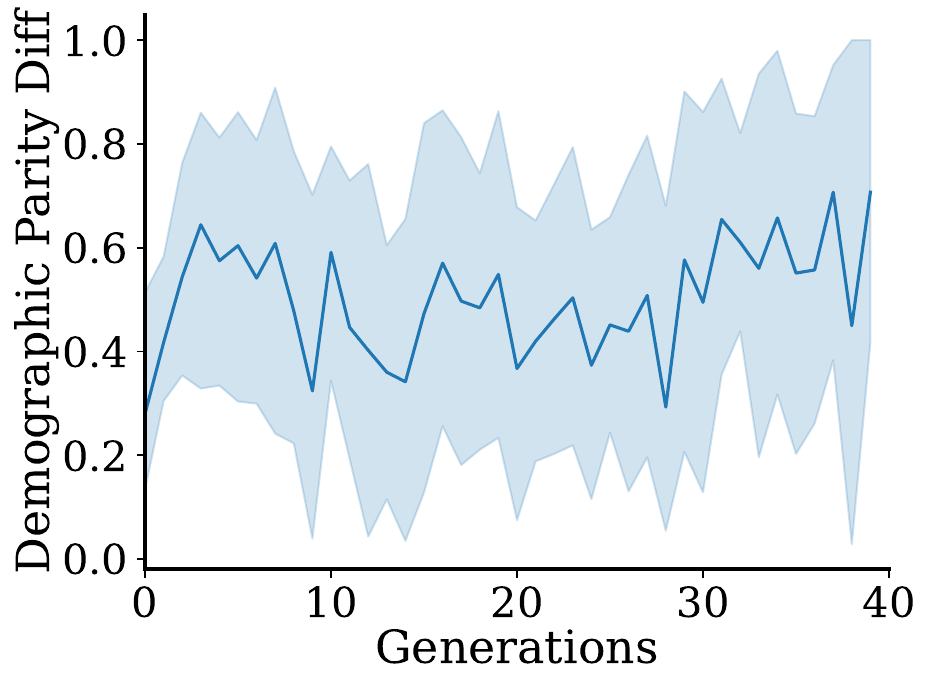} 
\\
\includegraphics[width=0.35\textwidth]{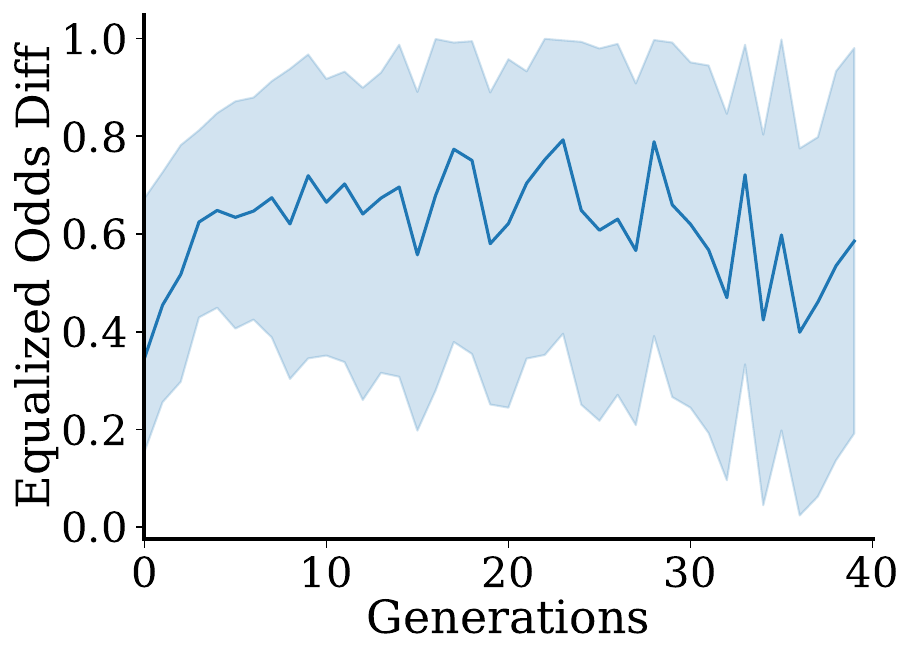} &
\includegraphics[width=0.35\textwidth]{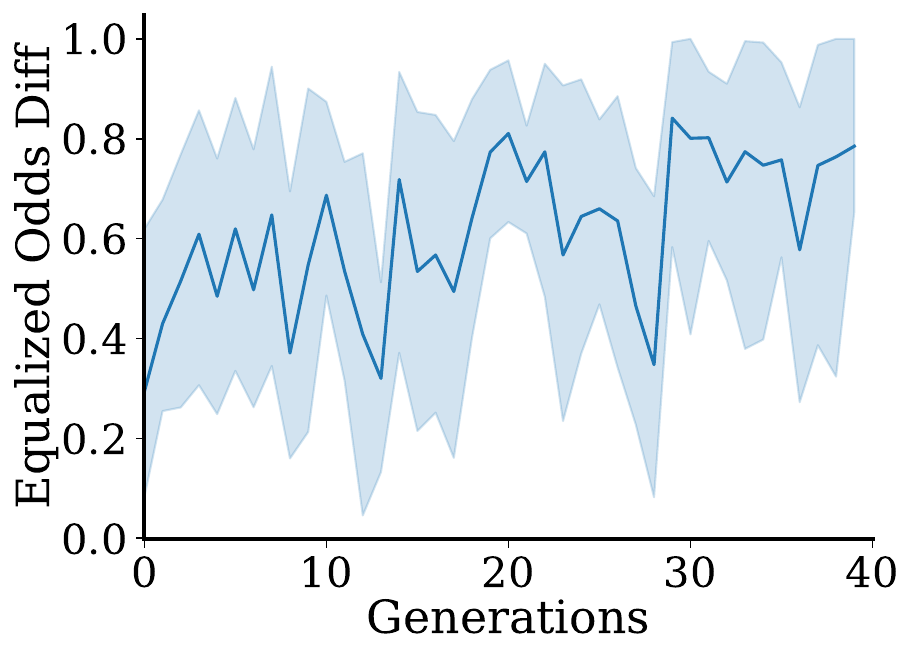} 
\\
\includegraphics[width=0.35\textwidth]{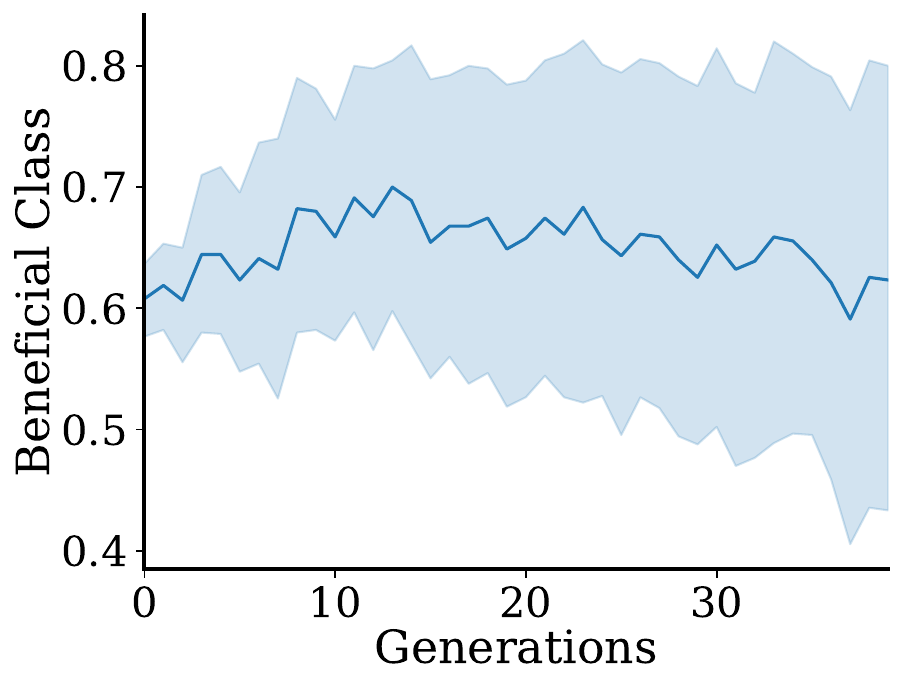} &
\includegraphics[width=0.35\textwidth]{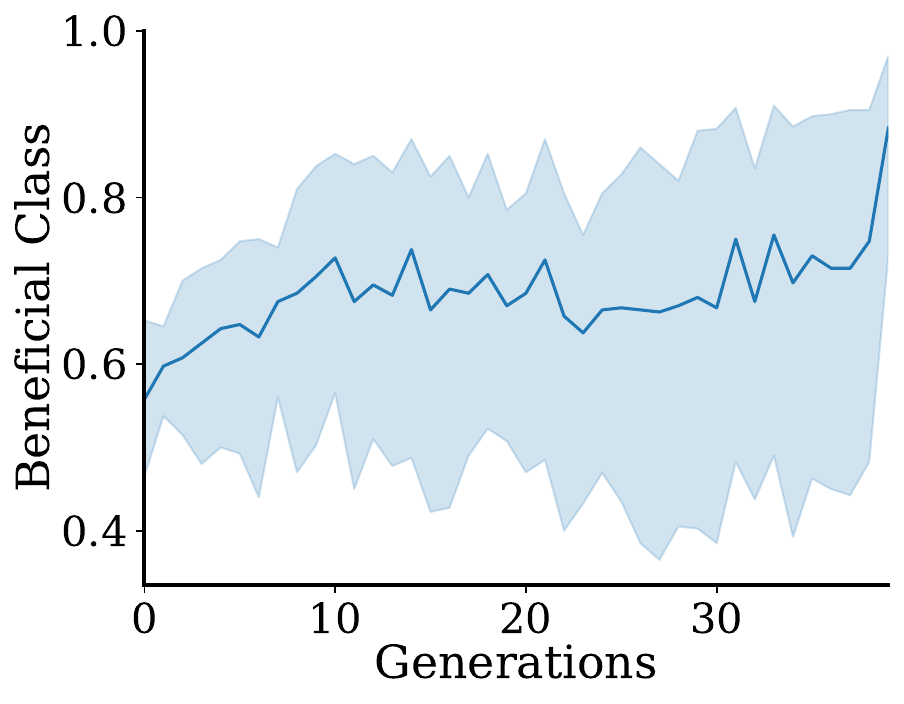} 
\\
\includegraphics[width=0.35\textwidth]{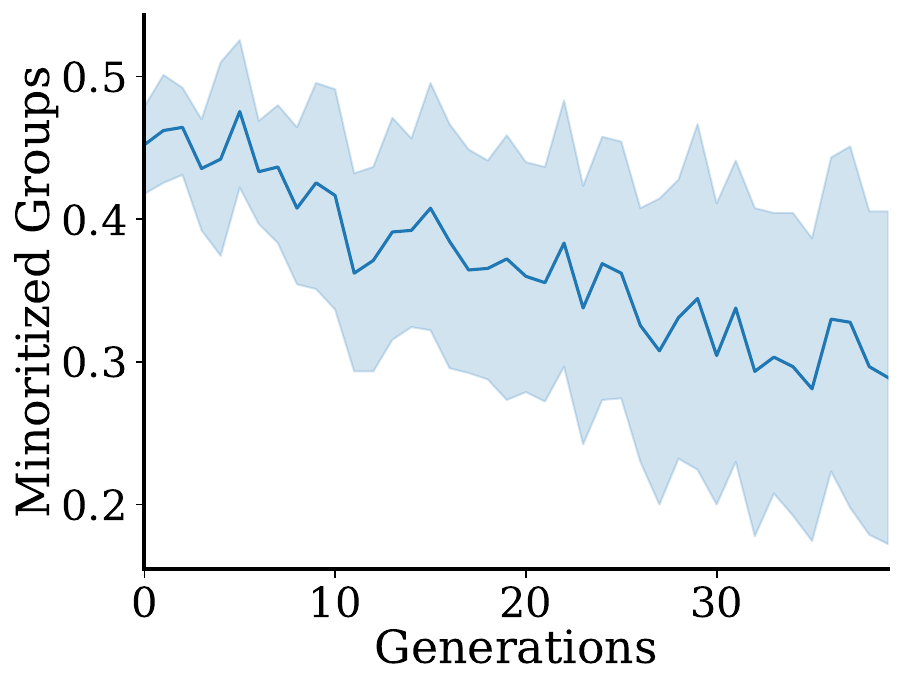} &
\includegraphics[width=0.35\textwidth]{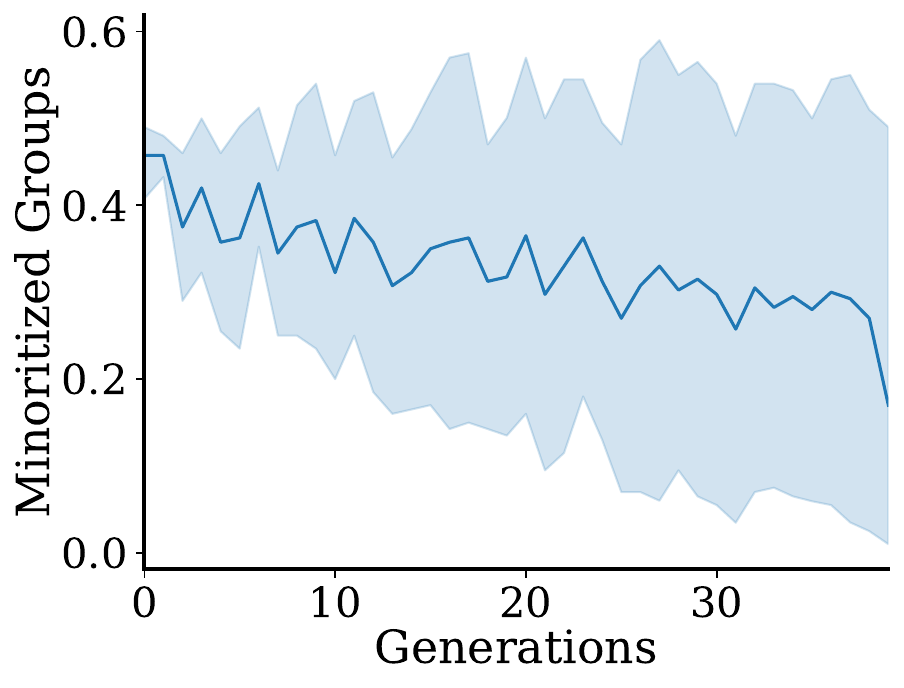} 
\\
\end{tabular}
\caption{Accuracy, demographic parity difference, equalized odds difference, and rates of the beneficial class and minoritized group in \sgsc on \texttt{ColoredSVHN}. Higher accuracy is better, but for the FML metrics higher difference is worse. \textit{Left:} Results on the test set. \textit{Right:} Relative performances between models. }
\label{fig:rel_svhn_mc}
\end{figure}

For reference, \texttt{ColoredMNIST} plots are in \Cref{fig:rel_cmnist_nomc} for \seqc and \Cref{fig:rel_cmnist_mc} for \sgsc. \texttt{ColoredSVHN} plots are in \Cref{fig:rel_svhn_nomc} for \seqc and \Cref{fig:rel_svhn_mc} for \sgsc. 

These results also demonstrate how even when training each new classifier with a small tolerance for unfairness can accrue to high unfairness. For example, consider the relative equalized odds results in \Cref{fig:rel_cmnist_nomc} which on average stay below 0.06 for each generation accrue to over 0.2.  

\Cref{fig:rel_cmnist_mc} shows the point of model collapse in the \sgsc setting in \texttt{ColoredMNIST} (collapse by generation 15) can be seen in the relative accuracy plot and in the increase in variance in the other relative plots. Similarly as found in the \seqc plots discussed above, low relative equalized odds difference and a relatively balanced minoritized group under-report the actual unfairness and imbalance.

\section{Co-Occuring MIDS}\label{app:fancy}
\begin{figure}[ht]
\centering
\begin{tabular}{cccc}
\includegraphics[width=0.23\textwidth]{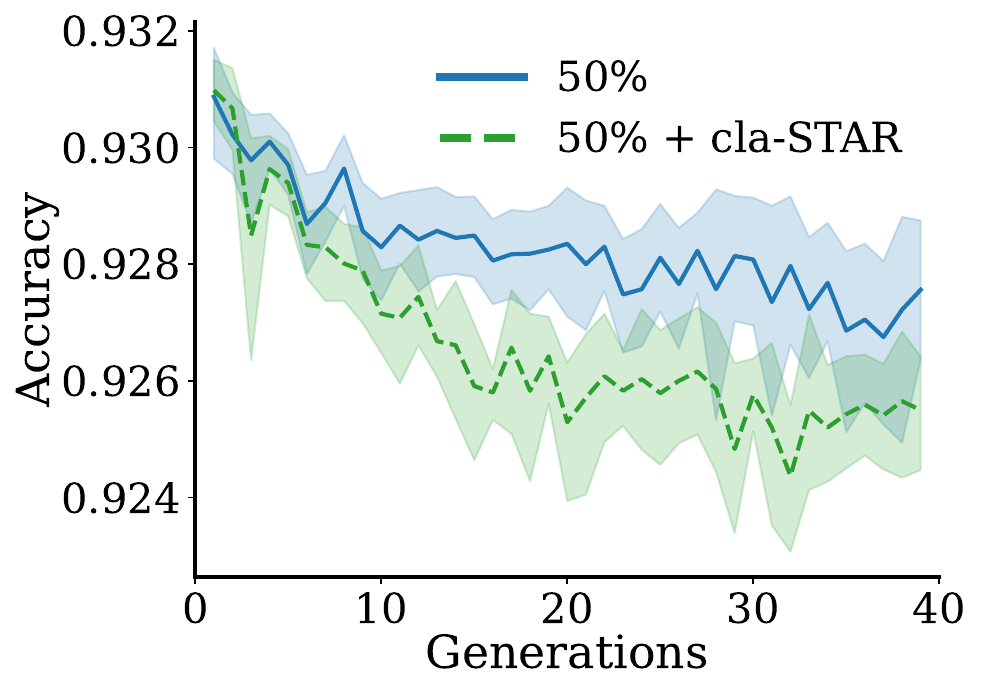} &
\includegraphics[width=0.23\textwidth]{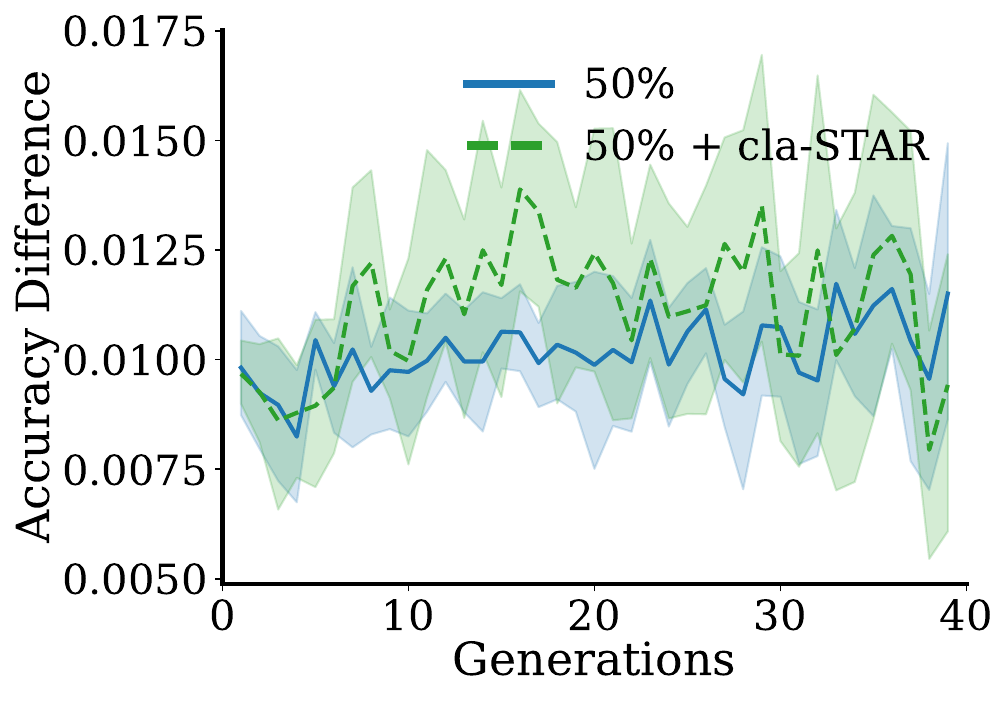} &
\includegraphics[width=0.23\textwidth]{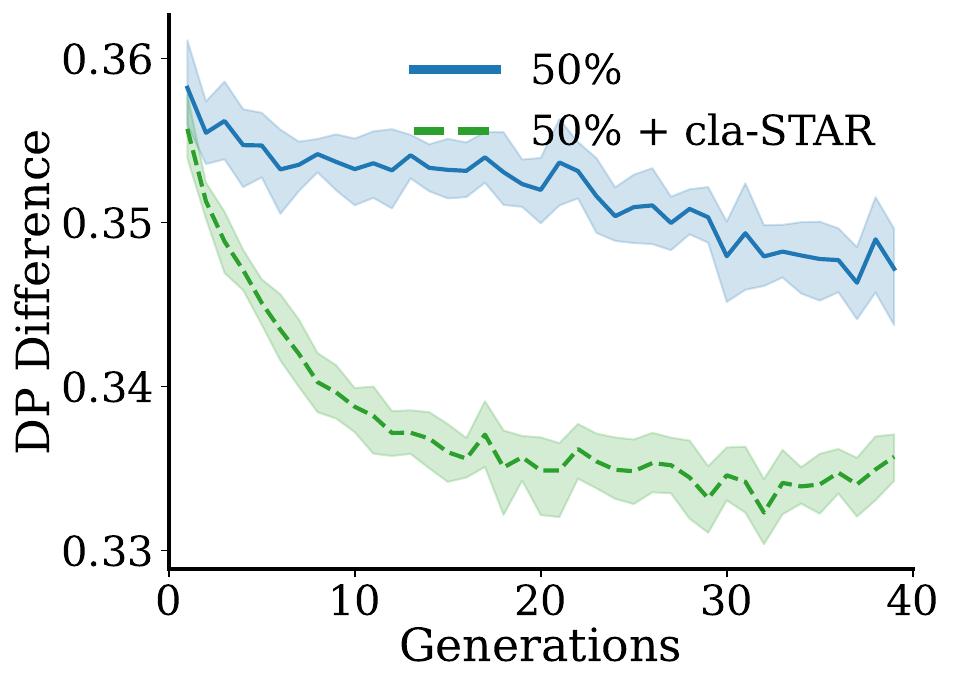} &
\includegraphics[width=0.23\textwidth]{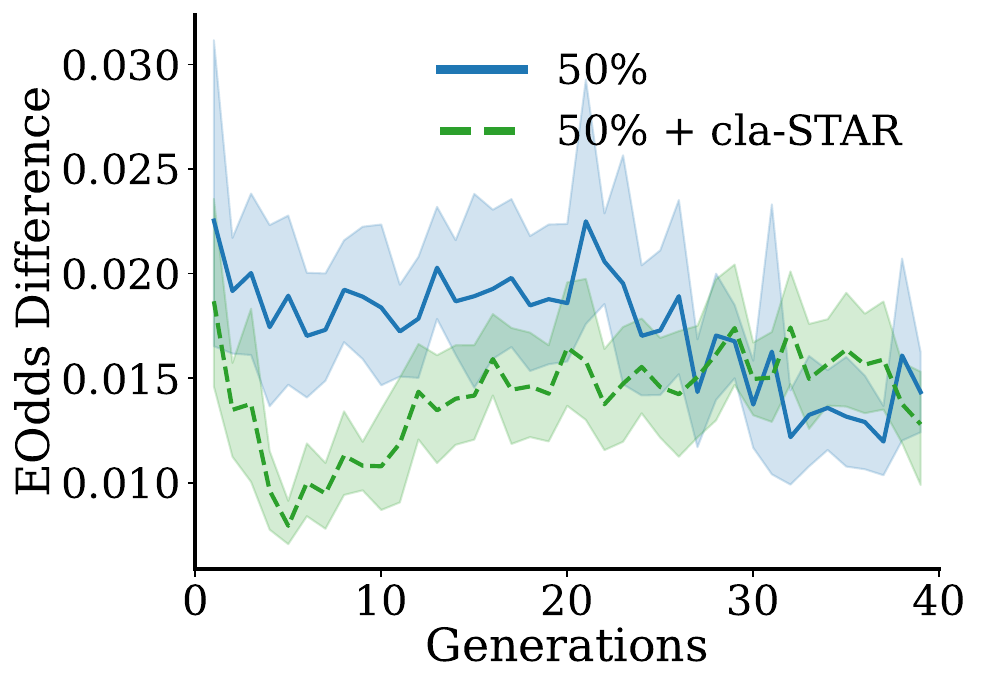}
\end{tabular}
\caption{\texttt{ColoredMNIST} results for \seqc when training with half synthetic and half non-synthetic data. Plots show accuracy, accuracy difference, demographic parity difference, and equalized odds difference. Non-synthetic data is sampled according to the \distrname of the prior classifier to model disparity amplification. \claalgname leads to more DP and EOdds fairness even while disparity amplification and performative prediction occur. }
\label{fig:nomc_50}
\end{figure}

\begin{figure}[ht]
    \centering
    \begin{tabular}{cccc}
       \includegraphics[width=.22\textwidth]{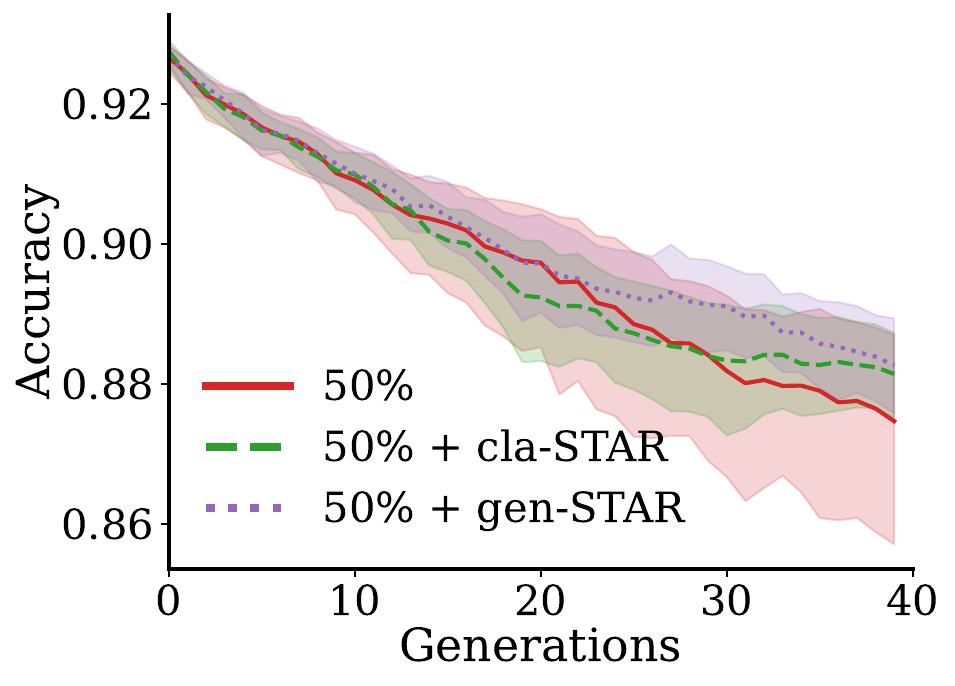}  &  \includegraphics[width=.22\textwidth]{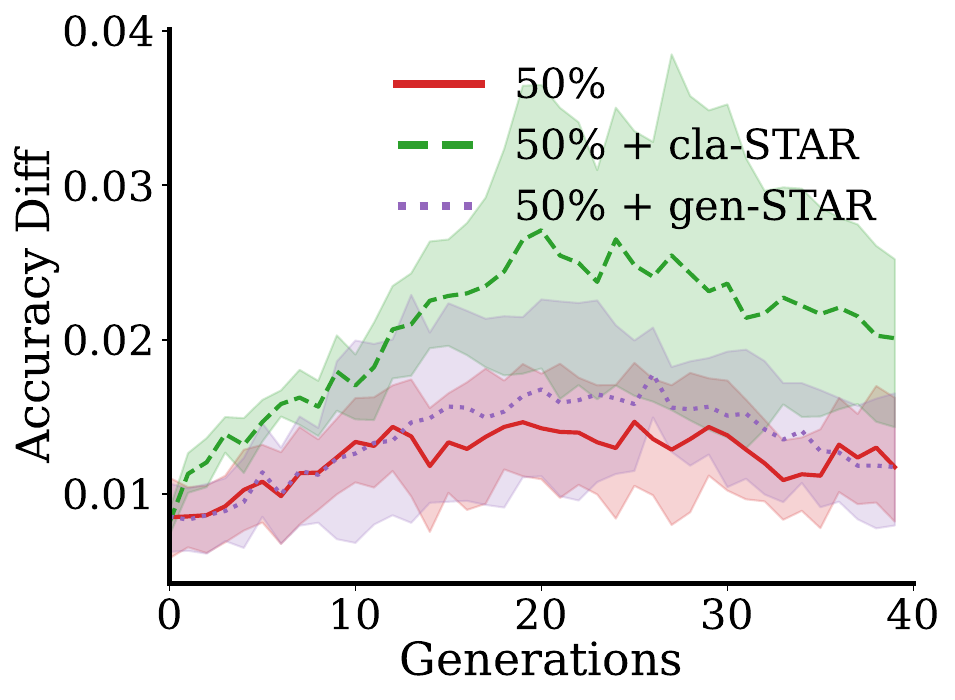} &
       \includegraphics[width=.22\textwidth]{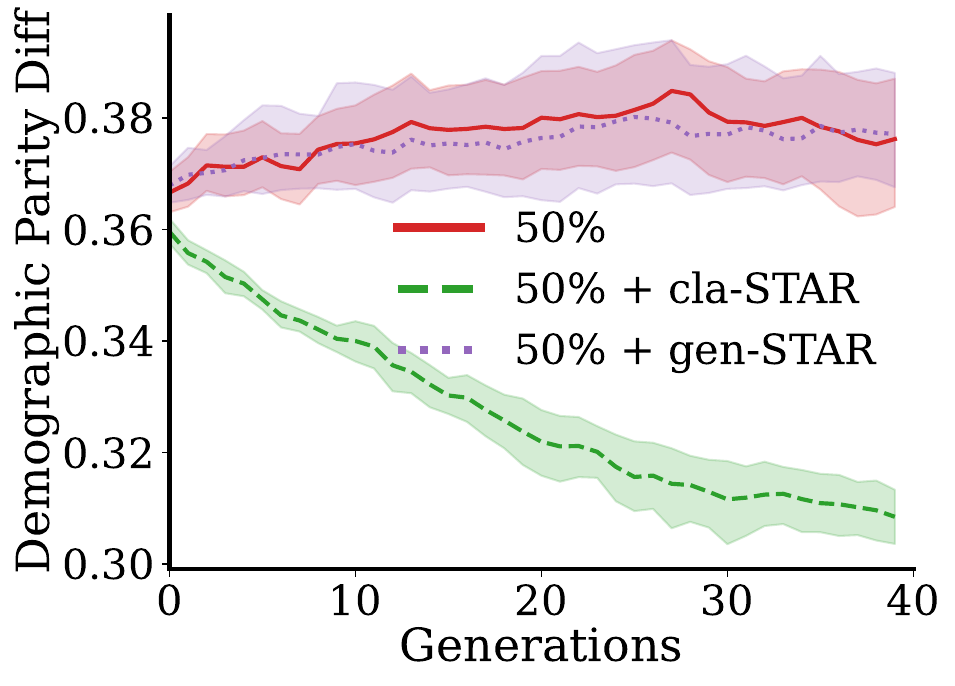} &
       \includegraphics[width=.22\textwidth]{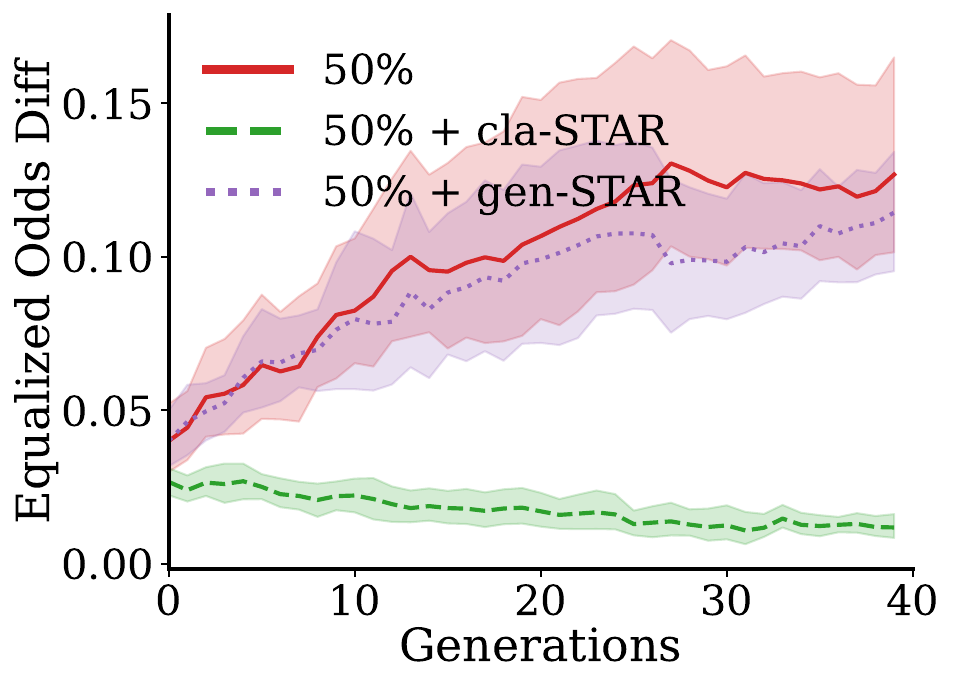} \\
        \includegraphics[width=.22\textwidth]{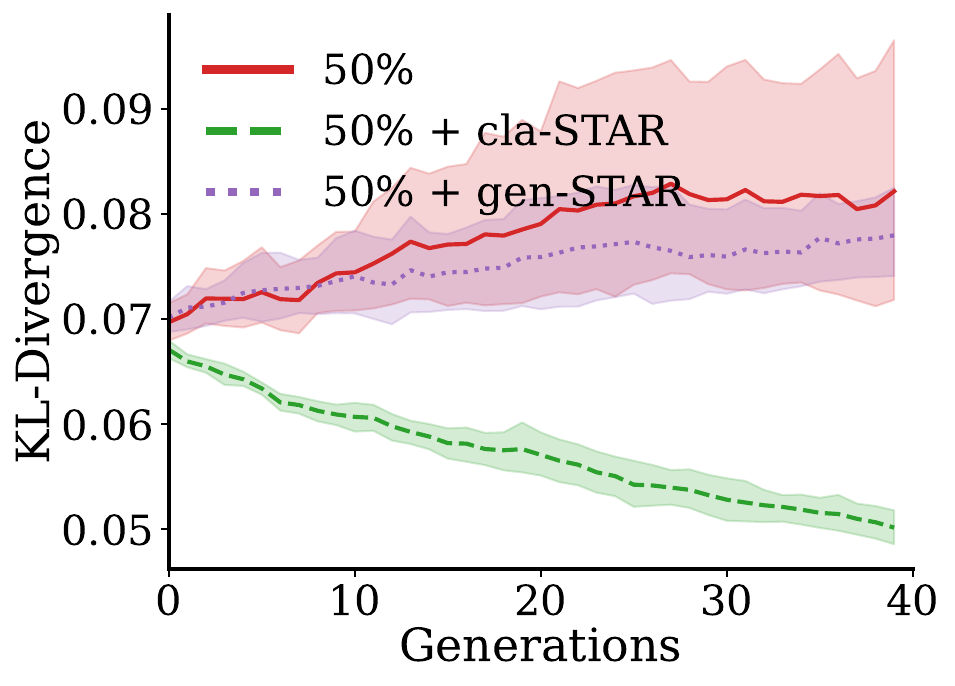} & 
        \includegraphics[width=.22\textwidth]{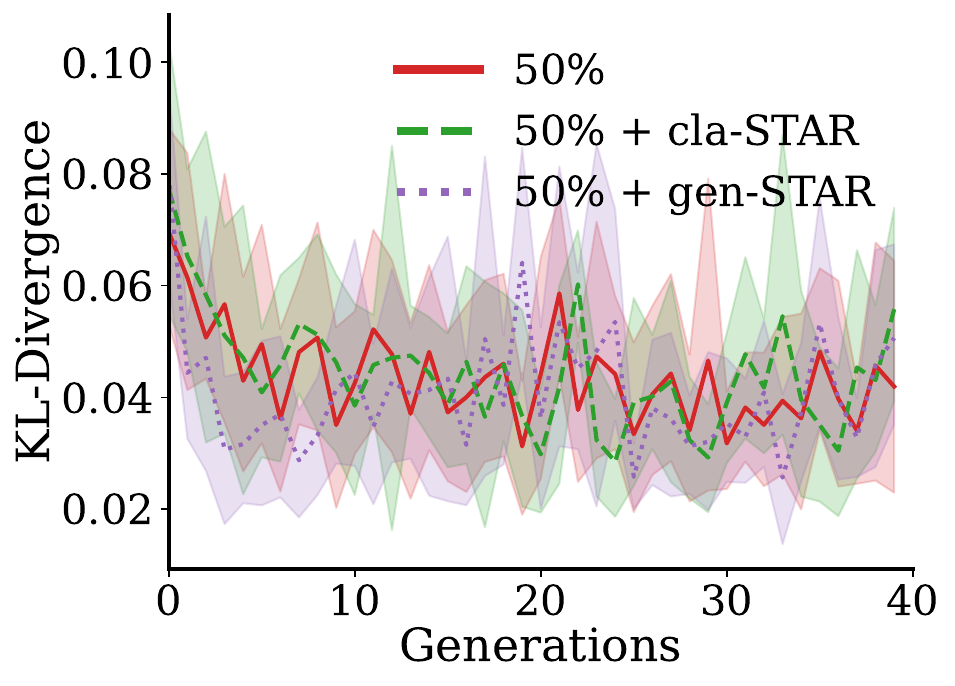} &
        \includegraphics[width=.22\textwidth]{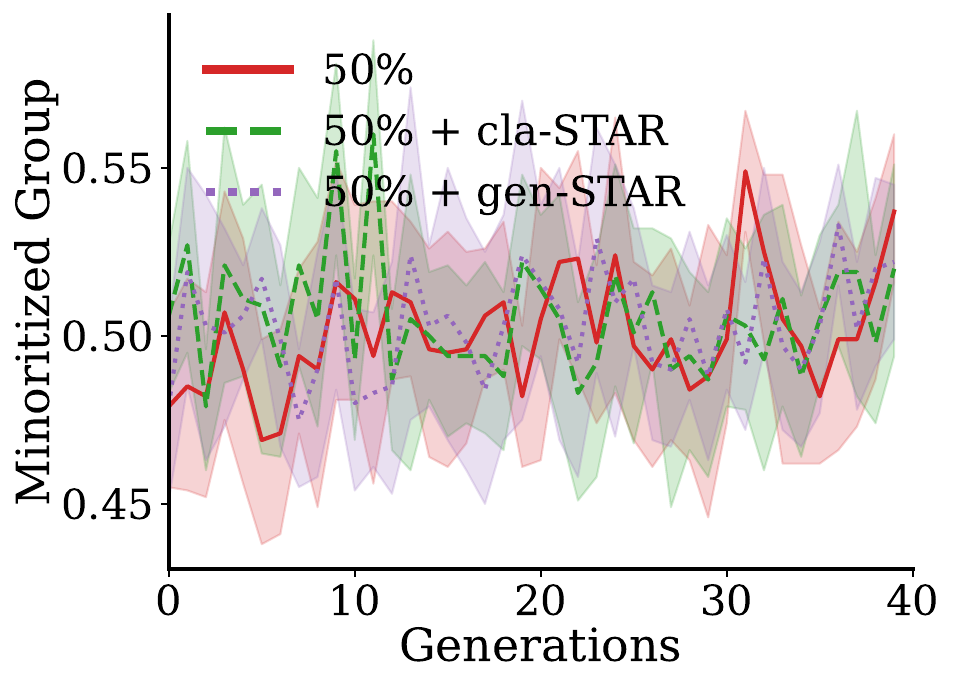} &
        \includegraphics[width=.22\textwidth]{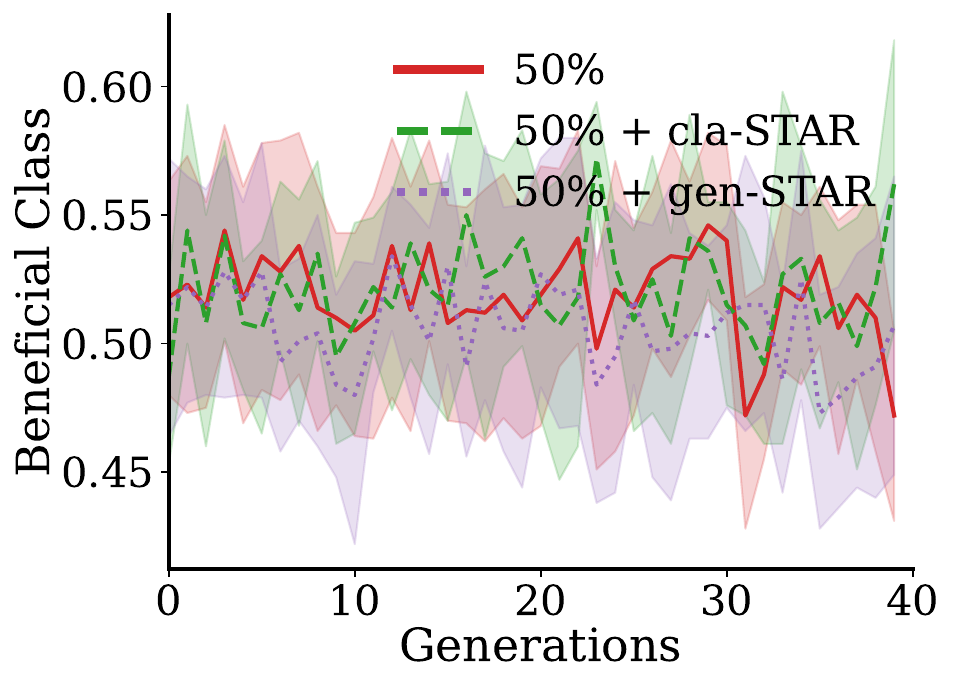} \\
    \end{tabular}
    \caption{Training on a mixture of synthetic and non-synthetic data, where performative prediction, model collapse, and disparity amplification can co-occur, for \texttt{ColoredMNIST} on \sgsc. \textit{Top:} accuracy, accuracy difference, demographic parity difference, and equalized odds difference. \textit{Bottom:} KL-Divergence between the \algname fairness ideal and the \distrname of classifiers and generators, the group balance, and the label balance.}
    \label{fig:mc_50}
\end{figure}

\begin{figure}[ht]
    \centering
    \begin{tabular}{ccc}
        \includegraphics[width=.3\textwidth]{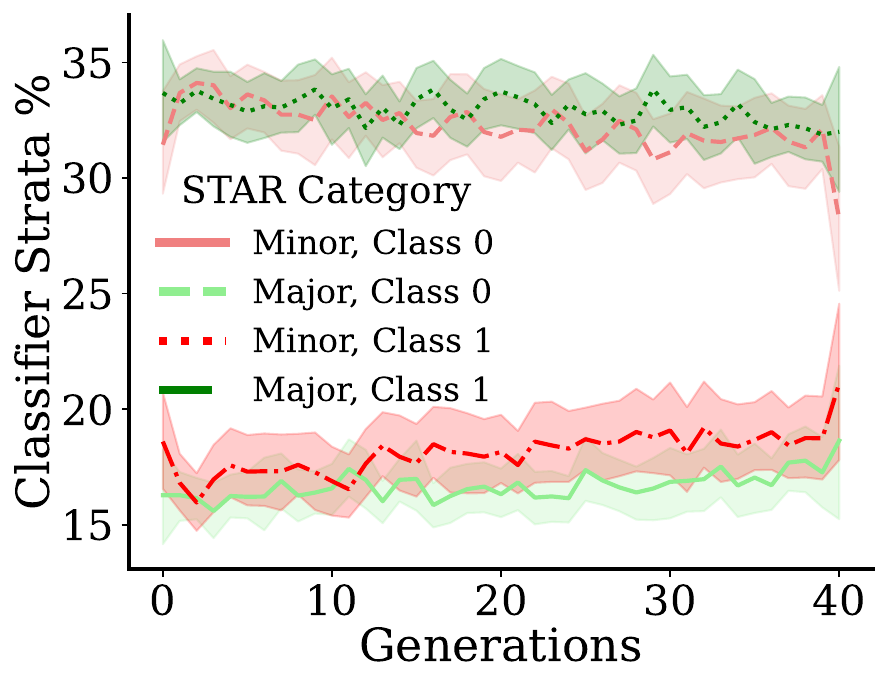} & 
        \includegraphics[width=.3\textwidth]{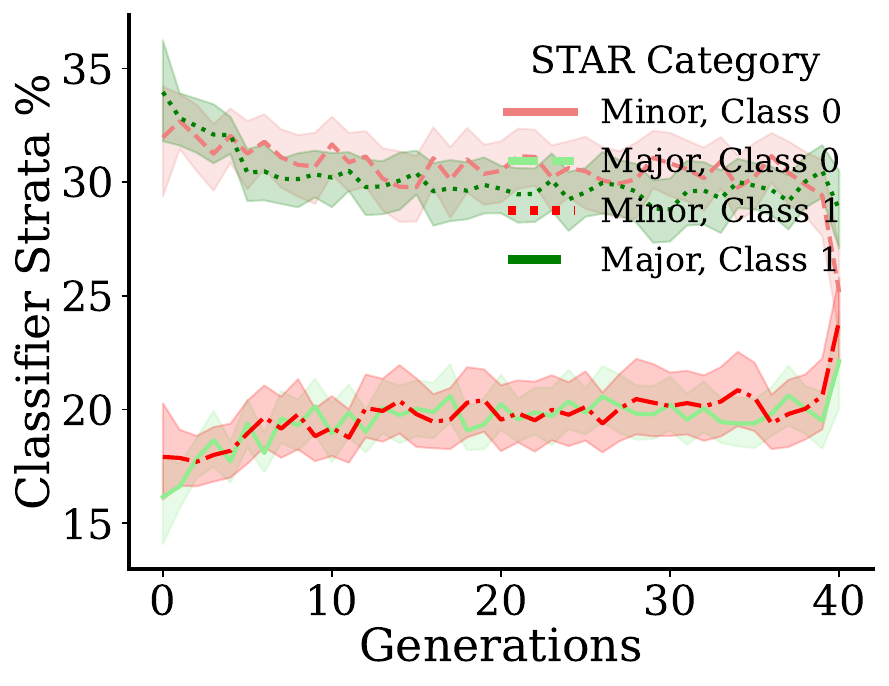} &
        \includegraphics[width=.3\textwidth]{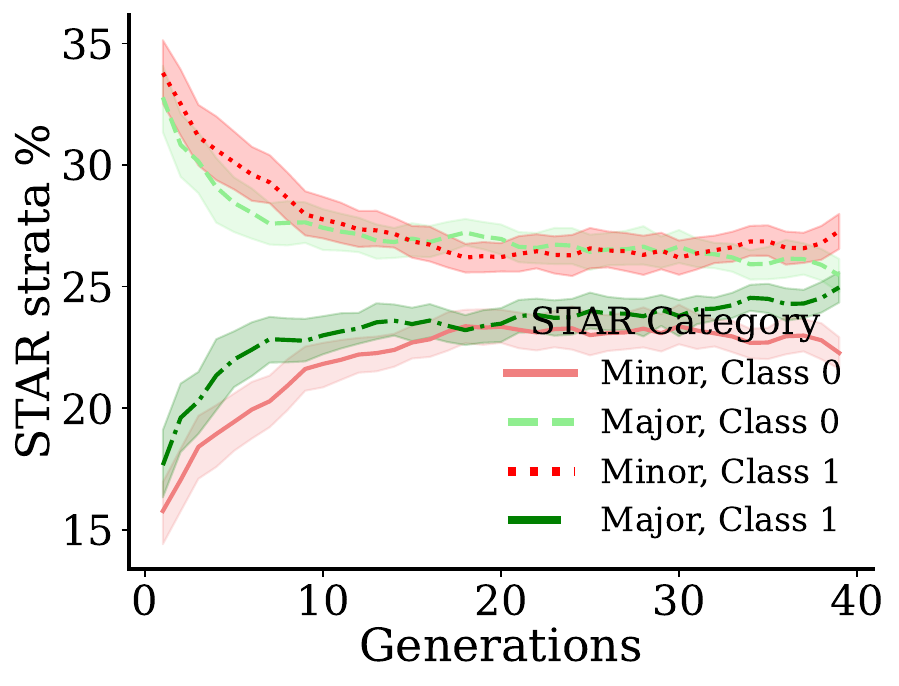} \\
        \includegraphics[width=.3\textwidth]{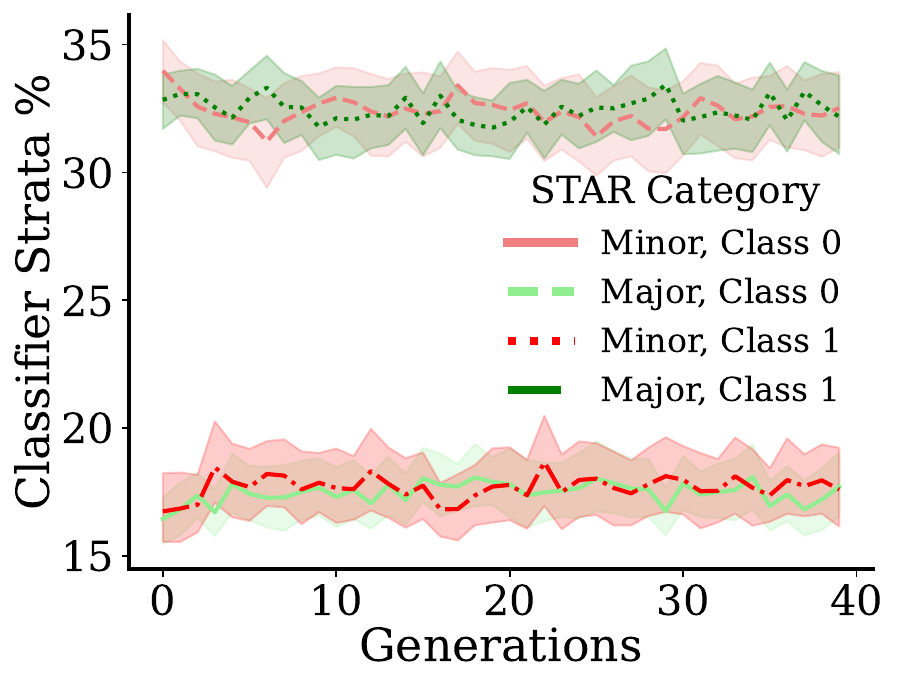} & 
        \includegraphics[width=.3\textwidth]{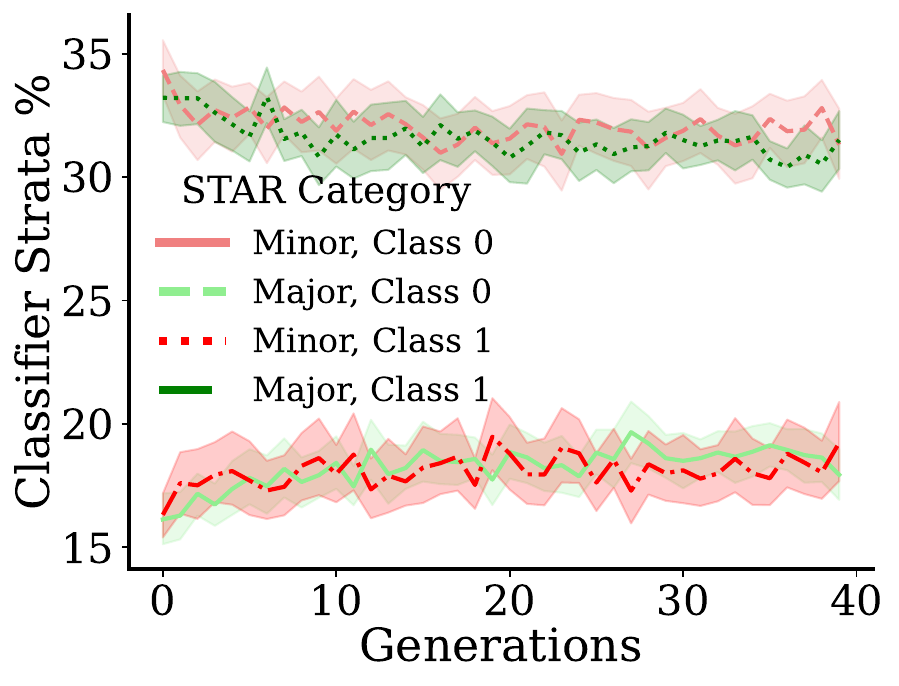} &
        \includegraphics[width=.3\textwidth]{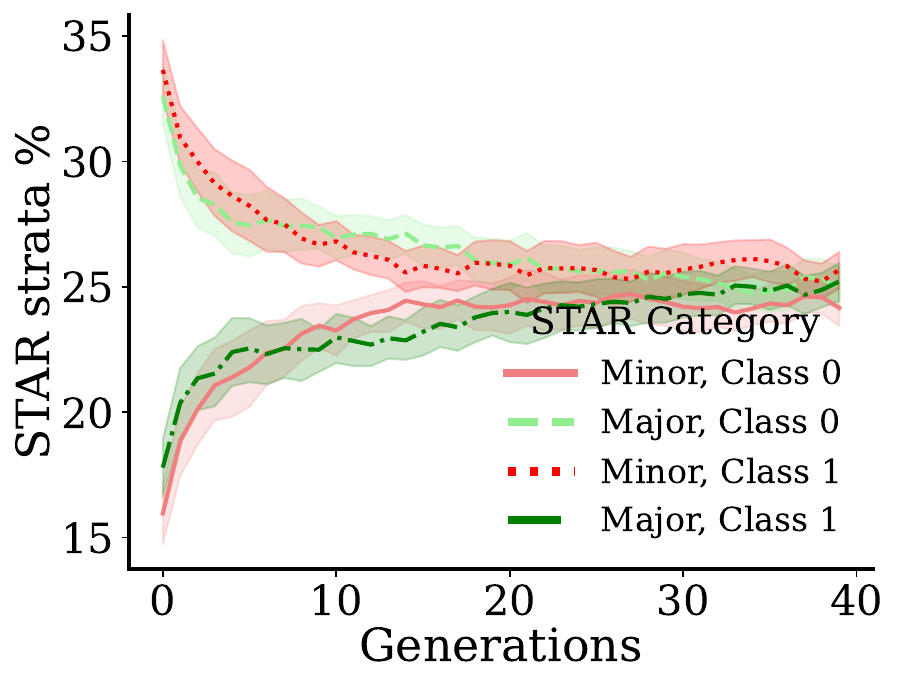} \\
        \includegraphics[width=.3\textwidth]{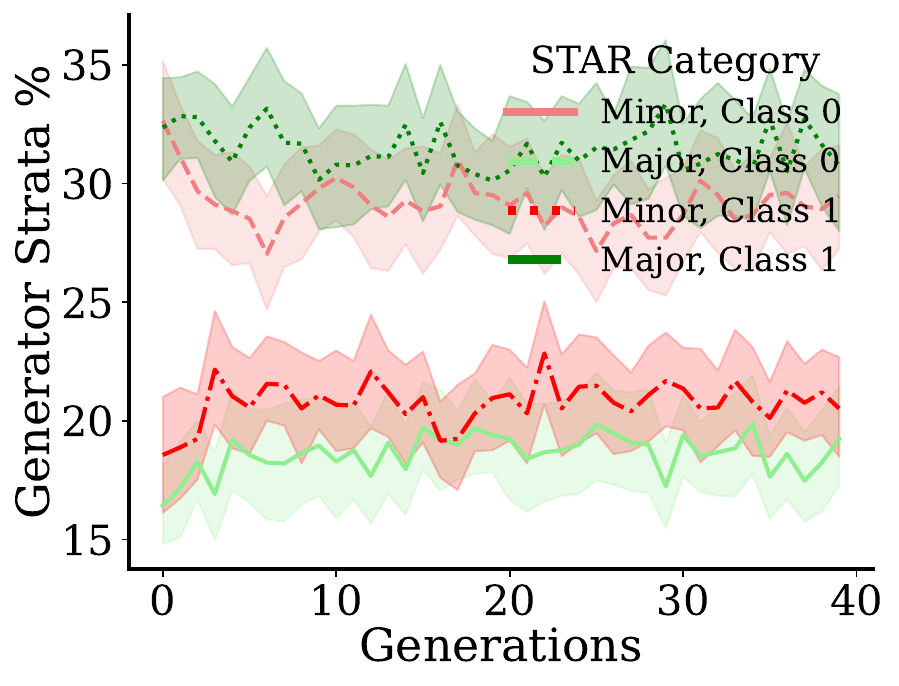} & 
        \includegraphics[width=.3\textwidth]{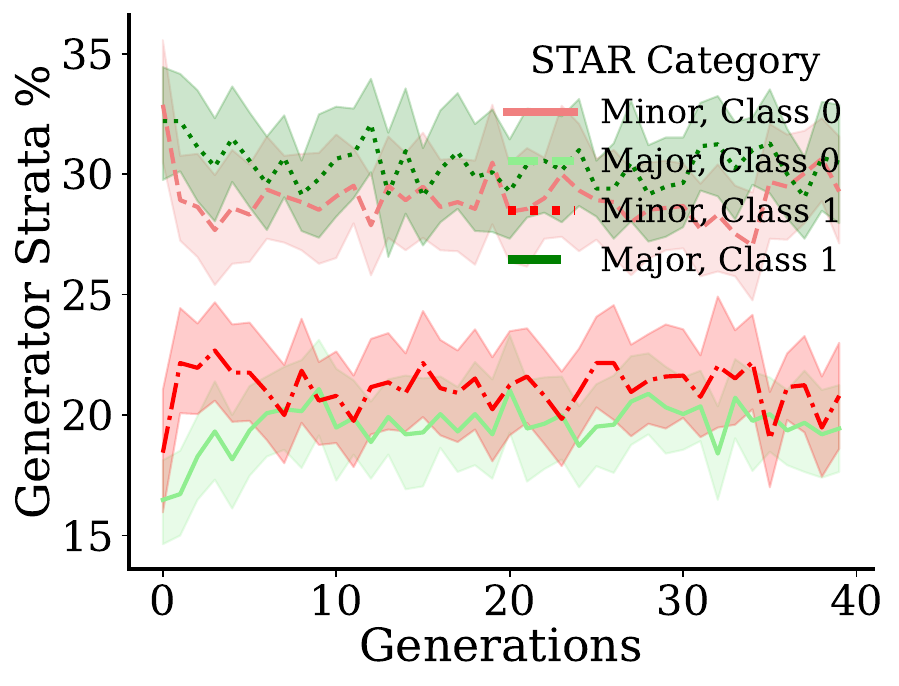} &
        \includegraphics[width=.3\textwidth]{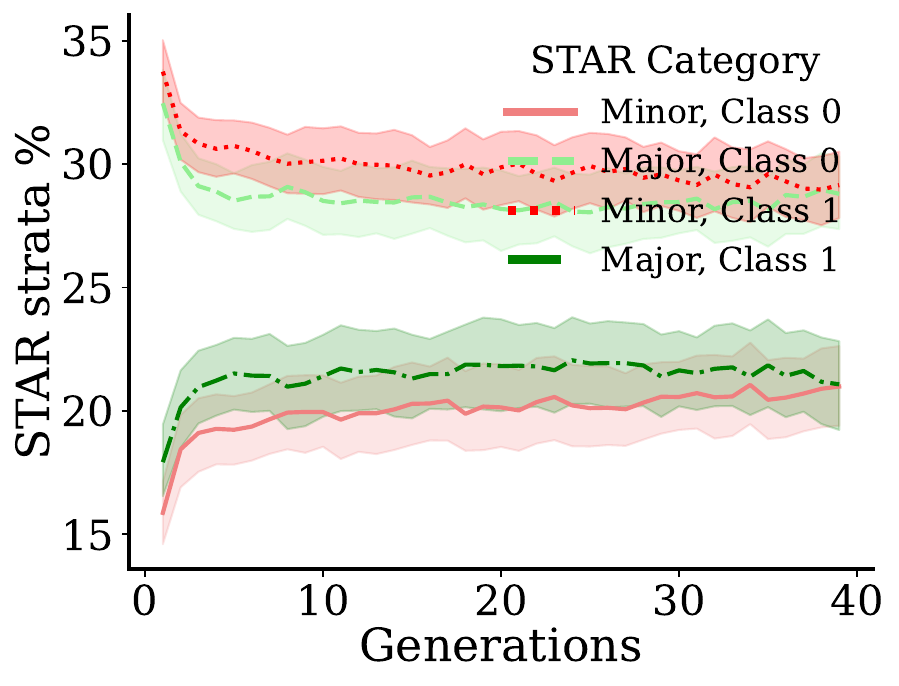} \\
    \end{tabular}
    \caption{\textit{Left:} The \distrname of classifiers without reparation. \textit{Center:} The \distrname resulting from models with \algname. \textit{Right:} The \distrname used to train models with \algname. 
    \textit{Top:} \texttt{ColoredMNIST} in \seqc with a mixture of synthetic and non-synthetic data. 
    \textit{Second row:} \texttt{ColoredMNIST} in \sgsc with a mixture of synthetic and non-synthetic data, reporting \distrname of the classifiers and using \claalgname. 
    \textit{Bottom:} \texttt{ColoredMNIST} in \sgsc with a mixture of synthetic and non-synthetic data, reporting \distrname of the generators and using \genalgname. }
    \label{fig:50_stratas}
\end{figure}

In this section we present two sets of experiments to showcase how disparity amplification can co-occur with performative prediction, and also with model collapse. We evaluate these experiments for \texttt{ColoredMNIST}, where \Cref{fig:nomc_50} shows the \seqc case, \Cref{fig:mc_50} shows the \sgsc case, and \Cref{fig:50_stratas} shows the \distrname for models trained in both settings.

For the \seqc setting, we train each classifier in the lineage from a 50/50 mixture of data from $G_0$ and from the original training set. In the \sgsc setting, the generators are trained from this data mixture, though the downstream classifiers are trained entirely from their corresponding generator's synthetic outputs. The inclusion of human-generated data moderates the degree of model collapse to showcase other effects. 

For disparity amplification to co-occur, we use stratified sampling on the original training set portion of the data mixture, where the strata are determined by the classifier's label distribution over the groups. Note that this is not disparity amplification as discussed in \citet{Hashimoto2018fairness}, which is due to performance failures, but instead due to label and group representation. This approximates the effects of the classifiers on the human-generated data distribution, and shows how this effects feeds into the other MIDS.
We conduct additional experiments to showcase the effects of AR at the classifiers or generators, in isolation from and in combination with the disparity amplification sampling strategy.

Note that at the limit where there is no synthetic data, we recover the promising technical question of how to create a biased sampling mechanism that begets fairness in a downstream model trained from a generator.

\end{document}